\documentclass[preprint,12pt]{elsarticle}

\usepackage{amssymb}
\usepackage{mathbbol}
\usepackage{amsmath,amsthm,amsfonts,amscd,amsmath,amsfonts} % Some packages to write mathematics.
\usepackage{upgreek}
\usepackage{mdframed}
\usepackage{multicol}
\usepackage{graphicx}
\usepackage{nomencl}
\usepackage[nottoc]{tocbibind}
\usepackage{setspace}
\usepackage{subfig}
\usepackage{verbatim}   	% Allows quoting source with commands.
\usepackage{makeidx}    	% Package to make an index.
\usepackage{color}
\usepackage{xcolor}
\usepackage{float}
\usepackage{enumerate}
\usepackage{booktabs,tabularx}
\usepackage{url}	
\usepackage{booktabs,longtable,array}
\usepackage{tcolorbox}
\usepackage{xcolor}
\usepackage{multirow}
\usepackage[linesnumbered,ruled,vlined]{algorithm2e}
\usepackage{ulem}
\usepackage[pdftex,bookmarks]{hyperref}
\usepackage{xurl}
\usepackage{multirow}
\usepackage{bbm, dsfont}
\usepackage[percent]{overpic}
\usepackage[table]{xcolor}
\usepackage{booktabs}

\newcommand{\blue}[1]{\textcolor{black}{#1}}
\colorlet{blue}{black} % Render blue revision marks in black

\usepackage[margin=1.2in]{geometry} % Adjust margins here

\usepackage{etoolbox} % noindent after the sections/subsections
\makeatletter
\patchcmd{\@startsection}
  {\@afterindenttrue}
  {\@afterindentfalse}
  {}{}
\makeatother

\journal{Neural Networks}

\begin{document}

%========================================================================
% First Page
%========================================================================
\begin{frontmatter}



\title{
\large
\textbf{Evidence-Guided Neural Architecture Selection under Uncertainty for Subject-Specific Blood Glucose Forecasting}
}


\author[inst1]{Md Azharul Islam}
\author[inst2]{Dwyer Deighan}
\author[inst1]{Tarunraj Singh}
\author[inst1]{Danial Faghihi\corref{cor1}}

\affiliation[inst1]{organization={Department of Mechanical and Aerospace Engineering, \\University at Buffalo},%Department and Organization
            %addressline={Address One}, 
            city={Buffalo},
            %postcode={00000}, 
            state={NY},
            country={USA}}
            
\affiliation[inst2]{organization={
Computational and Data-Enabled Sciences, \\
University at Buffalo},%Department and Organization
            %addressline={Address One}, 
            city={Buffalo},
            %postcode={00000}, 
            state={NY},
            country={USA}}

\cortext[cor1]{Corresponding Author, \texttt{danialfa@buffalo.edu} (D. Faghihi)}            

%========================================================================
% Abstract
%========================================================================
\begin{abstract}
% Objective: Statement of the problem -- 1 sentence
Neural architecture selection for time-series forecasting is challenging under limited, noisy, and heterogeneous data, particularly when relative predictive performance alone does not establish reliability for the intended task.
%
% Methods: How you are going to attack this problem -- 2 sentences
\blue{
We propose \textit{EVIDENT} (EVidence-based IDEntification of Neural
archiTectures), an ordered architecture selection framework that seeks
a balance between representational capacity, support from available
data, and task-specific forecasting adequacy.
In this framework, Bayesian model plausibility prioritizes evaluated candidates within
capacity levels, ordered by increasing architectural complexity, while validation under predictive uncertainty governs acceptance, progression, and stopping.
The procedure advances to a higher capacity level when no shortlisted
candidate meets the prescribed forecasting requirement and stops
at the first level yielding an accepted candidate.
}
% Summary of results -- 2 sentences
\blue{
We evaluate EVIDENT for glucose forecasting in in silico subjects
and individuals with type~1 diabetes using recorded patient data,
with Bayesian temporal convolutional networks as predictors.
A synthetic electrocardiogram study extends the evaluation to
Bayesian long short-term memory networks.
The results show that high model plausibility does not necessarily
imply forecasting acceptance, supporting the distinction between
model ranking and task-specific validation. Integration with
adaptive neural architecture search further demonstrates how
different candidate exploration methods can operate within
the same selection procedure.
%
% Conclusion: "take home" message summarizing the study -- 1 sentence
EVIDENT makes probabilistic forecasting adequacy an explicit
decision criterion for architecture exploration and stopping,
with acceptance conditional on the evaluated candidates,
available data, and prescribed forecasting requirements.
}
\end{abstract}


\end{frontmatter}

%%\linenumbers

% \clearpage
%\tableofcontents

%\newpage
%========================================================================
% Document body
%========================================================================
\section{Introduction}
\label{sec:introduction}
%------------------------------------------------------------------------------------------------
% 1. Identify the problem and say why it is important
%------------------------------------------------------------------------------------------------

%-- architecture selection openings
Neural network architecture and hyperparameter selection remains a central challenge in time-series learning \cite{ZHANG, LUCAS, HU}, particularly in regimes with limited, noisy, and heterogeneous data. This challenge becomes more pronounced when the objective extends beyond pointwise accuracy to reliable forecasting for consequential decision making in physical and biomedical systems. In such settings, architecture design must balance temporal expressivity, effective parameter learning from available data, and generalization to unseen conditions.
In practice, however, architecture choice often relies on trial-and-error tuning or \blue{selection based on held-out predictive performance}. While these approaches can yield accurate models for a given data split, \blue{relative performance comparisons alone do not establish whether a candidate satisfies the forecasting requirements of the intended task. This motivates explicit consideration of both model support from the available data and task-specific acceptance under predictive uncertainty.}
%%-- diabetes as an example
Subject-specific blood glucose forecasting in
\blue{individuals with type 1 diabetes provides a motivating application}
for this problem.
The goal is to predict future glucose trajectories from continuous
glucose monitoring (CGM) data together with recorded meal intake and
insulin delivery, \blue{with potential applications in patient-specific
treatment planning and insulin-delivery control}.
This setting is particularly challenging due to complex and delayed
glucose-insulin dynamics, strong inter- and intra-patient variability,
and wearable sensor noise.
\blue{A central challenge is to use retrospective data to identify
architectures that support forecasting for individual subjects.
Strong performance on historical records alone does not establish
forecasting adequacy for a particular individual under subsequent
conditions. Architecture selection must therefore consider both
the support provided by the available data and the adequacy of
individual forecasts under predictive uncertainty.}

%------------------------------------------------------------------------------------------------
% 2. The current state of the field and current limitation(s) in the field
%------------------------------------------------------------------------------------------------
\blue{Neural architecture search and related design strategies explore candidate configurations using different search mechanisms and performance estimates \cite{elsken2019neural, wang2018hybrid, ghosh2022designing, optuna_2019, ALSABRI}. 
More fundamentally, model validation is typically driven by empirical performance assessment, often under large data assumptions \cite{yaseen2023quantification, twomey1997validation, arzani2023interpreting, samek2021explaining, zhong2022explainable}.
For consequential forecasting, relative model ranking and
task-specific acceptance address complementary questions.
An architecture that outperforms other evaluated candidates may
still require further assessment to determine whether its predictions
meet the requirements of the intended task.
Aggregate prediction errors provide useful performance summaries,
while assessment across individual subjects or forecast windows
and consideration of predictive uncertainty provide additional
information about forecasting adequacy.
These considerations motivate a model-construction procedure
that explicitly coordinates model ranking with task-specific acceptance.
}

%------------------------------------------------------------------------------------------------
% 3. How do you plan on filling this hole with a novel approach
%------------------------------------------------------------------------------------------------
\blue{To address this need, we propose EVIDENT (EVidence-based
IDEntification of Neural archiTectures), an ordered architecture selection
procedure that seeks a balance between representational capacity,
support from the available data, and forecasting adequacy.
EVIDENT coordinates Bayesian training
\cite{li2025federated, jantre2021layer, mackay1995probable,
islam2024predicting, sevilla2024bayesian},
evidence-based model comparison
\cite{immer2021scalable, tan2022toward},
and validation under uncertainty
\cite{singh2024framework, oden2017predictive}
through distinct ranking, acceptance, and stopping decisions.
Ordered capacity levels structure exploration from less to more
expressive architectures. Insufficient capacity may restrict the
temporal behavior a model can represent, whereas additional flexibility
may not be adequately supported by the available data.
The underlying idea is to decouple \textit{model ranking}
from \textit{model validation}, such that Bayesian evidence assesses
relative support for evaluated architectures under the assumed prior
and likelihood, while task-specific validation assesses forecasting
reliability under predictive uncertainty. To this end, evidence determines which
candidates are shortlisted within each capacity level, and validation
determines whether to accept them, stop, or proceed to the next level.
If no shortlisted candidate is accepted across the prescribed levels,
the procedure reports that outcome.
%
%The methodological contribution lies in a framework for coordinating model support, predictive uncertainty, and acceptance requirements to guide architecture selection and stopping decisions for temporal forecasting.
} 
\blue{Moreover, EVIDENT accommodates existing architecture search algorithms within each capacity level, using approximate evidence to prioritize candidates for subsequent validation. Its ranking acceptance structure is modular, such that the inference method, search implementation, and validation metric can be adapted to the size and context of the problem. More generally, Bayesian evidence could be replaced by a suitable information criterion for deterministically trained models while preserving the separation between ranking and task-specific acceptance; such variants are outside the present evaluation.}

\blue{
In this paper, we first examine the behavior of approximate Bayesian evidence across temporal convolutional network (TCN) architectures and temporal forecasting configurations using synthetic glucose trajectories.
We then demonstrate EVIDENT through three complementary case
studies. The \textit{in silico glucose study} provides a controlled
evaluation across subjects and population-training folds.
The \textit{recorded patient glucose study} extends the evaluation
to measured patient records from the OhioT1DM dataset
\cite{marling2020ohiot1dm}. The \textit{synthetic electrocardiogram (ECG) study}
examines transfer to another signal and network family through
electrocardiogram forecasting with Bayesian long short-term memory
(LSTM) networks.
The recorded-patient glucose and synthetic ECG studies also
incorporate neural architecture search (NAS) within capacity levels,
demonstrating compatibility with adaptive candidate exploration.
}

%------------------------------------------------------------------------------------------------
% 4. The final paragraph: summary of sections
%------------------------------------------------------------------------------------------------
The remainder of the paper is organized as follows. 
\blue{Section~\ref{sec:related_work} reviews related approaches to architecture search, Bayesian model selection, and probabilistic forecasting.}
Section~\ref{sec:methods} develops the proposed methodology, including the glucose forecasting formulation, the feasible TCN architecture space, Bayesian learning, approximate evidence-based architecture ranking, and the EVIDENT algorithmic framework.
\blue{Section~\ref{sec:results} presents preliminary analyses of Bayesian plausibility for TCN forecasting, followed by the in silico glucose, recorded-patient glucose, and synthetic ECG studies.}
Section~\ref{sec:discussion} discusses the broader implications, limitations, and extensions of the approach, and Section~\ref{sec:conclusions} provides concluding remarks.

\vspace{1in}

%%%%%%%%%%%%%%%%%%%%%%%%%%%%%%%%%%%%%%%%%%%%%%%%%%%%%%%%%%%%%%%%
{
\color{blue}
\begin{center}
\section*{Nomenclature Table}
\end{center}

\begingroup
\small
\renewcommand{\arraystretch}{1.1}
\renewcommand{\theHtable}{nomenclature}

\begin{longtable}{@{}p{0.32\textwidth}p{0.64\textwidth}@{}}
\toprule
\textbf{Symbol} & \textbf{Definition} \\
\midrule
\endfirsthead

\toprule
\textbf{Symbol} & \textbf{Definition} \\
\midrule
\endhead

\midrule
\endfoot

\bottomrule
\endlastfoot

\multicolumn{2}{@{}l}{\textbf{Forecasting and TCN architecture}} \\

$G^{(k)},M^{(k)},I^{(k)}$ &
Glucose, meal-intake, and insulin-delivery inputs at time index $k$. \\

$k,j$ &
Discrete time or forecast-origin index; temporal-block index
(sample index in the ECG equations). \\

$d_G,d_M,d_I$ &
Input-history lengths for glucose, meals, and insulin. \\

$H,s$ &
Forecast-horizon length; stride used to extract training windows. \\

$\mathcal{T}_{w,\xi}$ &
TCN forecasting operator with trainable parameters $w$
and architecture $\xi$. \\

$\overline{\mathcal{T}}_{w,\xi}$ &
Posterior predictive mean of the TCN forecast. \\

$w,W$ &
Vector of trainable network parameters; number of its components. \\

$\xi,\Xi$ &
Architecture configuration; feasible architecture space. \\

$N$ &
Number of temporal blocks in each TCN encoder and decoder. \\

$f,b$ &
Convolutional filter size; encoder dilation base. \\

$c_j,\tilde{c}_j$ &
Numbers of channels in encoder and decoder block $j$. \\

$\delta_j$ &
Encoder dilation factor, $\delta_j=b^{j-1}$;
decoder dilation is unity. \\

$\mathrm{RF},\tau_{\min}$ &
Receptive-field extent; minimum required temporal context. \\

\midrule
\multicolumn{2}{@{}l}{
\textbf{Bayesian learning and architecture selection}} \\

$\mathcal{D}_i,\mathcal{D},\mathcal{D}_{\mathrm{val}}$ &
Patient-specific training data; population training data;
held-out validation data.\\

$N_{D_i},N_D,N_{\mathrm{patient}}$ &
Training-window count for patient $i$; total training-window
count; number of population-training patients. \\

$\sigma_{\mathrm{pr}},\sigma_{\mathrm{noise}}$ &
Prior standard deviation of network parameters;
likelihood-noise standard deviation. \\

$\sigma,\sigma_{\mathrm{MAP}}$ &
Inference hyperparameters
$(\sigma_{\mathrm{pr}},\sigma_{\mathrm{noise}})$
and their MAP estimate. \\

$\pi_{\mathrm{pr}},\pi_{\mathrm{like}},\pi_{\mathrm{post}}$ &
Prior, likelihood, and posterior distributions, respectively. \\

$\pi_{\mathrm{evid}}(\mathcal{D}\mid\sigma,\xi)$ &
Evidence conditional on inference hyperparameters
and architecture. \\

$\pi_{\mathrm{evid}}(\mathcal{D}\mid\xi)$ &
Architecture-level evidence after marginalizing
the inference hyperparameters. \\

$q_{\eta}(w),\eta$ &
Variational approximation to the weight posterior;
its distribution parameters. \\

$\eta^{\mathrm{opt}},w_{\mathrm{MAP}}$ &
Optimized variational parameters; MAP weight estimate,
approximated by the variational posterior mean. \\

$P$ (Appendix B) &
Hessian of the negative log-posterior with respect
to network parameters. \\

% $P$ (Ohio capacity table) &
% Network parameter count, denoted by $W$ elsewhere. \\

$\mathcal{M},\mathcal{M}_{\mathrm{initial}}$ &
Architecture comparison set; initial candidate architecture pool. \\

$\mathcal{M}^{(l)},\widehat{\mathcal{M}}^{(l)}$ &
Candidate architectures in level $l$;
evidence-selected validation shortlist in that level. \\

$M$ (architecture comparison) &
Number of architectures in the comparison set. \\

$l,L$ &
Capacity-level index; total number of capacity levels. \\

$\rho_m$ &
Posterior plausibility of architecture $\xi_m$,
normalized over the evaluated comparison set. \\

\midrule
\multicolumn{2}{@{}l}{\textbf{Validation}} \\

$h$ &
Forecast-step index within a validation window,
$h=1,\ldots,H$. \\

$N_{\mathrm{val}}$ &
Number of scored validation windows in a held-out fold. \\

$\sigma_{\mathrm{data}}^2,\sigma_{i,h}^2$ &
Measurement-noise variance and total predictive variance
at forecast step $h$ in validation window $i$, respectively. \\

$\mathbb{d}$ &
Task-specific probabilistic validation metric. \\

$Tol$ &
Prescribed validation tolerance. \\

\midrule
\multicolumn{2}{@{}l}{\textbf{ECG forecasting and LSTM}} \\

$E,\widehat{E},B$ &
Measured ECG voltage; forecast ECG voltage;
respiratory-belt signal. \\

$\tau_R,RR_{\mathrm{last}}$ &
Elapsed time since the latest detected R peak;
last available detected RR interval. \\

$\mathcal{L}_{w,\xi}$ &
Bayesian LSTM forecasting operator. \\

$N_{LSTM},h,d$ (LSTM) &
Number of recurrent layers; hidden units per layer;
input feature dimension. \\

$\boldsymbol{x}^{(j)},\boldsymbol{h}^{(j)},\boldsymbol{c}^{(j)}$ &
LSTM input vector, hidden state, and cell state at sample $j$. \\

$\boldsymbol{i}^{(j)},\boldsymbol{f}^{(j)},\boldsymbol{o}^{(j)}$ &
LSTM input, forget, and output gates. \\

$\widetilde{\boldsymbol{c}}^{(j)}$ &
Candidate LSTM cell state. \\

$\boldsymbol{W}_q,\boldsymbol{U}_q,\boldsymbol{b}_q$ &
Input and recurrent weight matrices and bias vector
for $q\in\{i,f,o,c\}$. \\

\midrule
\multicolumn{2}{@{}l}{\textbf{Abbreviations}} \\

EVIDENT &
EVidence-based IDEntification of Neural archiTectures. \\

BG, CGM &
Blood glucose; Continuous glucose monitoring. \\

T1D, ECG &
Type 1 diabetes; Electrocardiogram. \\

TCN, LSTM &
Temporal convolutional network; Long short-term memory. \\

ELBO, KL &
Evidence lower bound; Kullback--Leibler divergence. \\

MAP, NLPD &
Maximum a posteriori; Negative log predictive density. \\

RMSE, CV &
Root mean-squared error; Coefficient of variation. \\

Conv1D, ReLU, AvgPool &
One-dimensional convolution; Rectified linear unit;
Average pooling. \\

\end{longtable}

\addtocounter{table}{-1}
\endgroup
}

%========================================================
% Methods
%========================================================
{
\color{blue}
\section{Related Work}\label{sec:related_work}

\paragraph{Architecture search}
Neural architecture search (NAS) combines candidate generation with
performance estimation. ENAS uses reinforcement learning and shared
weights \cite{pham2018efficient}, whereas DARTS optimizes a continuous
relaxation of operation choices \cite{hanxiao2019darts}.
Bayesian optimization models configuration performance
\cite{snoek2012practical}; BOHB combines model based proposals with
adaptive training budgets \cite{falkner2018bohb}.
AutoAMS extends automated design to attention-based multimodal graph
learning \cite{ALSABRI}, while NAS-Bench-101 provides precomputed
evaluations for reproducible comparisons \cite{ying2019bench}.
\blue{These approaches address candidate architecture exploration, performance
estimation, and computational allocation in different ways.
Search mechanisms can be paired with different objectives, including
accuracy or probabilistic scores. EVIDENT complements compatible
search methods by specifying how evidence-based ranking within
capacity levels and task-specific probabilistic validation govern
candidate acceptance, progression, and stopping.
To this end, while these search methods propose candidates, EVIDENT determines how
evaluated candidates are prioritized and assessed for acceptance.}

\paragraph{Bayesian model selection}
MacKay's evidence framework balances data fit against prior dependent
complexity \cite{mackay1992practical}; scalable Laplace approximations
extend marginal-likelihood optimization to modern networks
\cite{immer2021scalable}.
Evidence assesses model support under the assumed prior and
likelihood, but its ranking need not coincide with predictive
generalization performance \cite{lotfi2022bayesian}.
Earlier frameworks for physical and neural surrogate model selection
combine complexity ordering, Bayesian plausibility, and validation
under uncertainty \cite{farrell2015selection,singh2024framework, tan2022toward, oden2017predictive}.
EVIDENT builds on these principles to seek a balance between
representational capacity, support from available data, and forecasting
adequacy in temporal neural predictors. Its forecasting-specific
formulation combines candidate exploration within capacity levels
with window-based probabilistic validation. The present study examines this
formulation across glucose and synthetic ECG forecasting using
Bayesian TCNs and LSTMs.

\paragraph{Probabilistic and personalized forecasting}
Deep time-series forecasting research addresses complementary
questions of temporal representation, predictive uncertainty, and
personalization. N-BEATS, Autoformer, and TimesNet introduce different
representations of temporal structure
\cite{oreshkin2019n,wu2021autoformer,wu2022timesnet}.
LSTNN, a lightweight spatio-temporal network for traffic forecasting,
separates seasonal and trend components through sampling based
time-series decomposition and indicates that compact architectures
can remain competitive with more complex models \cite{wang2025lightweight}.
DeepAR models predictive distributions \cite{salinas2020deepar},
while Multi-Granularity Autoformer incorporates a quantile loss for
probabilistic long-term power load forecasting \cite{yang2025multi}.
For blood glucose forecasting in type~1 diabetes, a systematic review
of 26 studies examines the datasets, input signals, prediction horizons,
architectures, and evaluation metrics of deep learning predictors \cite{calzavara2026systematic}. The paper identifies physiological plausibility and interpretability as open concerns for clinical use. Within this literature, GluNet uses dilated convolutions to predict
a categorical distribution over quantized future glucose changes
\cite{li2020glunet}, and dilated recurrent networks (DRNN) set their depth
and dilation rates by comparing configurations on validation RMSE
\cite{zhu2020dilated}. Gluformer estimates predictive distributions
of blood glucose \cite{sergazinov2023gluformer}, while multitask
learning transfers population information to personalized predictors
\cite{daniels2021multitask}.
Sun et al. \cite{sun2026future} distill future meal and insulin
information available during training into a predictor using
historical inputs.
These developments provide forecasting models and training
strategies that can inform candidate design.
Multimodal architectures from other domains, such as GP-M3Net for
Alzheimer's disease classification \cite{zhang2026gpm3net} and MCD4SR
for sequential recommendation \cite{zhang2026mcd4sr}, are examples of
candidate designs that future extensions of EVIDENT could consider for
combining glucose, meal, and insulin inputs of unequal reliability.
EVIDENT addresses the complementary question of how relative model
support and task-specific probabilistic acceptance jointly govern
architecture selection and validation. An extended methodological
comparison is provided in Table~\ref{tab:related_work_comparison}.
}

%========================================================
% Methods
%========================================================
\section{Methods}\label{sec:methods}

\blue{
We present the methodology using glucose forecasting with Bayesian
TCNs, which underlies the preliminary synthetic glucose analysis
and the in silico and recorded patient glucose studies in this paper.
The subsequent Bayesian learning, evidence-based ranking, and
task-specific validation procedure is formulated in this setting
but also applies to other forecasting tasks and network families.
The Bayesian LSTM predictor and task-specific implementation for
synthetic ECG forecasting are described in
Section~\ref{sec:ecg_results}.
}

%++++++++++++++++++++++++++++++++++++++++++++++++++++++++++++++++++++++++
\subsection{Glucose forecasting and feasible architecture space}
\begin{figure}[!ht]
    \centering
    \includegraphics[trim={2.in 0in 2.in 0in}, clip, width=1\linewidth]{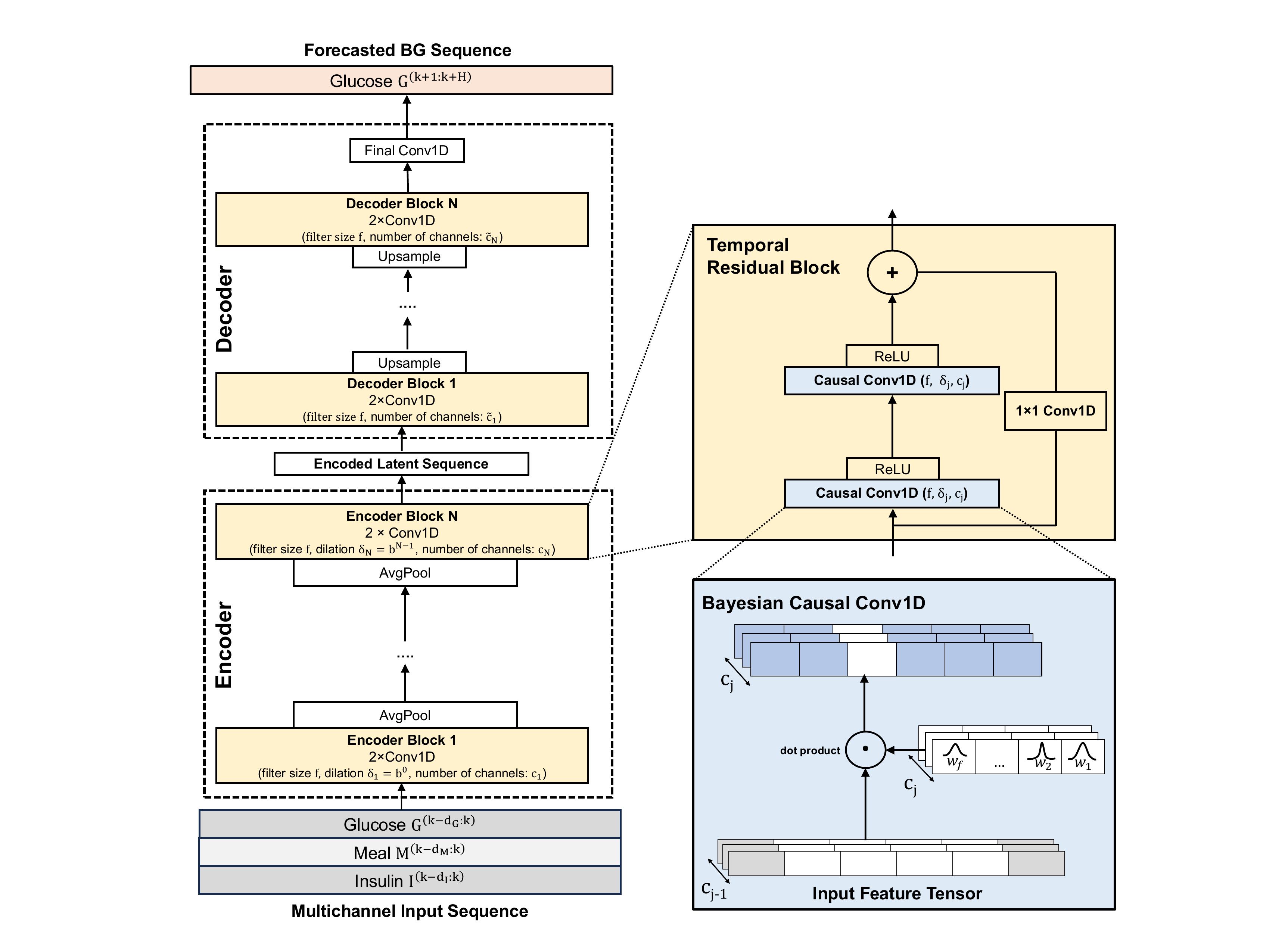}
    \vspace{-0.4in}
\caption{
(Left Panel) Bayesian temporal convolutional network (TCN) architecture used for multi-step blood-glucose (BG) forecasting 
${G}^{(k+1:k+H)} = \mathcal{T}_{w,\xi}\!(G^{(k-d_G:k)},\, M^{(k-d_M:k)},\, I^{(k-d_I:k)})$. 
The encoder contains $N$ temporal residual blocks with $c_j$ channels, filter size $f$, and dilation $\delta_j=b^{j-1}$; average-pooling layers between successive encoder blocks progressively reduce the temporal resolution and produce an encoded latent sequence. The decoder consists of $N$ residual convolutional blocks with $\tilde{c}_j$ channels, unit-dilation causal convolutions, and upsampling layers between successive decoder blocks. 
(Top-right Panel) Temporal residual block consisting of two Bayesian causal Conv1D layers with ReLU activations and a residual skip connection; a $1\times1$ Conv1D is used on the skip path when the input and output channel dimensions differ. 
(Bottom-right Panel) The Bayesian parameterization of a causal Conv1D layer, where the convolutional weights are represented by probability distribution determined by Bayesian training.
}
    \label{fig:arch}
\end{figure}

We consider multi-step blood-glucose (BG) forecasting from multichannel time series consisting of glucose, meal intake, and insulin delivery sequences. At discrete time index $k$, the model maps a window of recent observations to a future glucose trajectory over a prediction horizon of length $H$. Let \(G^{(k-d_G:k)} \in \mathbb{R}^{d_G}\), \(M^{(k-d_M:k)} \in \mathbb{R}^{d_M}\), and \(I^{(k-d_I:k)} \in \mathbb{R}^{d_I}\) denote the corresponding input histories of glucose, meal intake, and insulin delivery, respectively. The forecasting model is posed as a nonlinear operator
$
\mathcal{T}_{w, \xi} : \mathbb{R}^{d_G} \times \mathbb{R}^{d_M} \times \mathbb{R}^{d_I} \rightarrow \mathbb{R}^{H},
$
where \(\xi\) is the architecture parameters and \(w \in \mathbb{R}^{W}\) denotes all trainable parameters.
The predictor is then written as
\begin{equation}
{G}^{(k+1:k+H)}
=
\mathcal{T}_{w,\xi}\!\left(
G^{(k-d_G:k)},\,
M^{(k-d_M:k)},\,
I^{(k-d_I:k)}
\right),
\label{eq:tcn_operator}
\end{equation}
where \({G}^{(k+1:k+H)} \in \mathbb{R}^{H}\) denotes the predicted future BG trajectory.
%
%% --- explaning the TCN architecture
In this work, $\mathcal{T}_{w,\xi}$ is implemented using an encoder--decoder temporal convolutional network (TCN) \cite{bai2018empirical, zaidi2021multi}, as illustrated in Figure~\ref{fig:arch}. The input is represented as a multichannel sequence with three channels corresponding to $G^{(k-d_G:k)}$, $M^{(k-d_M:k)}$, and $I^{(k-d_I:k)}$. 
The encoder consists of $N$ stacked temporal blocks.
Each encoder block $j$ maps an input feature tensor to an output feature tensor with $c_j$ channels and is implemented as a temporal residual block composed of two Bayesian dilated causal 1D convolutional layers (Conv1D) with filter size $f$, followed by ReLU nonlinearities and a residual skip connection. When the input and output channel dimensions differ, a point-wise $1 \times 1$ Conv1D is applied on the skip path to linearly project the input channels before element-wise addition.
This residual-block structure is inspired by ResNets \cite{he2016deep} and improves training of deeper TCNs by facilitating gradient flow across blocks.
The convolutional layers use causal padding so that the temporal dimension is preserved within each block.
The dilation factor in encoder block $j$ is defined as $\delta_j = b^{j-1}$, where $b$ is the dilation base. 
Dilation determines the spacing between adjacent filter operations and when $\delta_j=1$, the convolution uses consecutive time samples, while for $\delta_j>1$, the filter skips intermediate samples and therefore covers a wider temporal range without increasing the filter size. The geometric growth of $\delta_j$ with depth $j$ enlarges the receptive field and enables the network to capture long-range temporal dependencies while preserving causality. Following each encoder block, except the last, an average pooling layer (AvgPool) reduces the temporal resolution, yielding a compact latent representation.
The decoder maps the latent sequence back to the forecast resolution through $N$ decoder blocks with number of channels $\tilde{c}_1,\ldots,\tilde{c}_N$. Each decoder block follows the same residual structure, consisting of two causal Conv1D layers with filter size $f$, ReLU activations, and a residual connection with a $1 \times 1$ projection when the block's input and the second convolution's output have different channel dimensions, so that the two tensors can be summed element-wise in the residual connection. In contrast to the encoder, the decoder convolutions use unit dilation (i.e., $\delta_j=1$ for all decoder blocks $j=1,\ldots,N$). Between successive decoder blocks, upsampling layers increase the temporal resolution.
Finally, a Bayesian Conv1D layer maps the decoder feature tensor to the predicted BG trajectory ${G}^{(k+1:k+H)}$.

Within this architecture family, the receptive field is a derived quantity determined by \(\xi\), indicating set of entries of the input that affect a specific entry of the output. For the temporal blocks in Figure~\ref{fig:arch}, each of which contains two causal dilated convolutions, the resulting receptive field is
\begin{equation}
\mathrm{RF}
=
1+2(f-1)\sum_{j=0}^{N-1} b^j
=
1+2(f-1)\frac{b^N-1}{b-1}.
\label{eq:rf_closed}
\end{equation}
The feasible architecture space is defined by the finite combinations of discrete architecture parameters (number of blocks $N$, dilation base $b$, filter size $f$, number of channels in encoder $\{c_j\}_{j=1}^N$ and decoder $\{\tilde{c}_j\}_{j=1}^N$, stride $s$, input lengths $\{d_G, d_M, d_I\}$, and output horizon $H$), subject to structural and temporal constraints,
\begin{equation}
\Xi
=
\left\{
\xi :
\mathrm{RF}\geq \tau_{\min},
\;
f \geq b
\right\},
\label{eq:feasible_space}
\end{equation}
where \(\tau_{\min}\) denotes the minimum temporal extent required to capture relevant glucose dynamics (e.g., delayed effects of meals and insulin), and \(f \geq b\) enforces a no-holes condition for dilated convolutions such that every input in the sequence contributes to the output. Additional problem-specific constraints may be imposed based on the desired prediction horizon and computational budget. Despite these restrictions, the combinatorial structure of \(\xi\) induces a large candidate architecture space, motivating the need for a principled architecture discovery strategy.

%++++++++++++++++++++++++++++++++++++++++++++++++++++++++++++++++++++++++
\subsection{Bayesian learning of TCN predictors}\label{sec:tcn_bayes}
We pose the training of the TCN predictor in a Bayesian setting by placing a probability distribution over the trainable parameters, i.e., convolutional weights \(w\). The Bayesian training is motivated by the limited and heterogeneous nature of the data, the need for quantifying uncertainty in the forecasts, and the requirement that architecture selection be robust to overconfident data fits.
In our particular problem, training is performed using population-level data obtained from multichannel time series across multiple type 1 diabetes patients. The TCN is trained using a sliding-window procedure with stride \(s\), applied over the full dataset. 
The retrospective data associated with patient $i$ is written as
\begin{equation}
\mathcal{D}_i
=
\left\{
\left(
G_{\mathrm{data}, i}^{(k-d_G:k)},\,
M_{\mathrm{data}, i}^{(k-d_M:k)},\,
I_{\mathrm{data}, i}^{(k-d_I:k)},\,
G_{\mathrm{data}, i}^{(k+1:k+H)}
\right)
\right\}_{k=1}^{N_{D_i}}.
\end{equation}
Here, \(G_{\mathrm{data}, i}\) denotes the observed glucose trajectory from continuous glucose monitoring (CGM), \(M_{\mathrm{data}, i}\) denotes the recorded meal-related input, and \(I_{\mathrm{data}, i}\) denotes the recorded insulin delivery.
Each index \(k\) corresponds to a sliding temporal window extracted from the trajectory of patient \(i\), and \(N_{D_i}\) denotes the total number of admissible input-output windows obtained from that patient.
The complete population-based training dataset constructed from \(N_{\mathrm{patient}}\) patients is then given by
\begin{equation}
\mathcal{D}
=
\bigcup_{i=1}^{N_{\mathrm{patient}}}
\mathcal{D}_i .
\end{equation}
For a given architecture \(\xi\), Bayesian training is defined by a prior distribution on the weights and a likelihood function for the observed trajectories. We use a Gaussian prior
$
\pi_{\mathrm{pr}}(w | \sigma_{\mathrm{pr}},\xi)
=
\mathcal{N}\!\left(0,\sigma_{\mathrm{pr}}^2 I_{W}\right),
$
and assume additive Gaussian noise model with variance \(\sigma_{\mathrm{noise}}^2\), leading to log-likelihood being proportional to
\begin{equation}
\log \pi_{\mathrm{like}}(\mathcal{D} \mid w,\sigma_{\mathrm{noise}},\xi)
\;\propto\;
-\frac{1}{2\sigma_{\mathrm{noise}}^2}
\sum_{k=1}^{N_D}
\left\|
G_{\mathrm{data}}^{(k+1:k+H)}
-
\mathcal{T}_{w,\xi}\!\big(G^{(k-d_G:k)}, M^{(k-d_M:k)}, I^{(k-d_I:k)}\big)
\right\|_2^2,
\end{equation}
where \(\sigma=(\sigma_{\mathrm{pr}},\sigma_{\mathrm{noise}})\) is called inference hyperparameters and 
\(N_D\) denotes the number of training input--output windows
over all patients.
Under these assumptions, the posterior distribution over the  trainable parameters is given by
\begin{equation}
\label{eq:bayes_w}
\pi_{\mathrm{post}}(w\mid \mathcal{D},\sigma,\xi)
=
\frac{\pi_{\mathrm{like}}(\mathcal{D}\mid w,\sigma_{\mathrm{noise}},\xi)\,\pi_{\mathrm{pr}}(w\mid \sigma_{\mathrm{pr}},\xi)}
{\pi_{\mathrm{evid}}(\mathcal{D}\mid \sigma_{\mathrm{pr}},\sigma_{\mathrm{noise}},\xi)}.
\end{equation}
\blue{Sampling-based posterior inference can be computationally expensive
for TCNs with high-dimensional parameters.}
 We thus employ variational inference with a tractable variational distribution \(q_{\eta}(w)\), parameterized by \(\eta\), chosen here as a mean-field Gaussian. The optimal variational distribution is obtained by minimizing the Kullback--Leibler divergence to the true posterior, equivalently by maximizing the evidence lower bound (ELBO):
\begin{equation}
\eta^{\mathrm{opt}}
=
\arg\min_{\eta}
\left[
\mathrm{KL}\!\left(
q_{\eta}(w)\,\|\,\pi_{\mathrm{pr}}(w \mid \sigma_{\mathrm{pr}},\xi)
\right)
-
\mathbb{E}_{q_{\eta}(w)}
\left[
\log \pi_{\mathrm{like}}(\mathcal{D} \mid w,\sigma_{\mathrm{noise}},\xi)
\right]
\right].
\end{equation}
This yields an approximate posterior over the TCN weight parameters.
Given the posterior distribution over \(w\) and fixed \(\sigma\) and \(\xi\), the predictive distribution for a new input sequence is
\begin{align}
\pi\!\left(
G^{*(k+1:k+H)}
\,\middle|\,
G^{(k-d_G:k)},\, M^{(k-d_M:k)},\, I^{(k-d_I:k)},\, \mathcal{D},\, \sigma,\, \xi
\right) = 
\nonumber \\
\int
\pi\!\left(
G^{*(k+1:k+H)}
\,\middle|\,
G^{(k-d_G:k)},\, M^{(k-d_M:k)},\, I^{(k-d_I:k)},\, w,\, \sigma,\, \xi
\right)
\pi_{\mathrm{post}}(w\mid \mathcal{D},\sigma,\xi)\,dw,
\label{eq:posterior_predictive}
\end{align}
which integrates the TCN predictions over the learned weight posterior. In practice, this distribution is approximated via Monte Carlo sampling from \(q_{\eta^{\mathrm{opt}}}(w)\), yielding predictive mean trajectories and corresponding uncertainty.

%++++++++++++++++++++++++++++++++++++++++++++++++++++++++++++++++++++++++
\subsection{Evidence as an architecture selection criterion}
\label{sec:plausibility}

To move from single-model Bayesian training to architecture discovery, we use the evidence distribution in \eqref{eq:bayes_w} as the scoring function for selecting the inference hyperparameters \(\sigma=(\sigma_{\mathrm{pr}},\sigma_{\mathrm{noise}})\) and for comparing candidate architectures $\xi$. The evidence assesses relative model support by balancing data fit
and prior-dependent complexity after marginalizing over the
network parameters \cite{mackay1995probable, singh2024framework, immer2021scalable}.
For fixed \(\sigma\) and \(\xi\), the evidence is defined as
$
\pi_{\mathrm{evid}}(\mathcal{D} | \sigma,\xi)
=
\int
\pi_{\mathrm{like}}(\mathcal{D} | w,\sigma_{\mathrm{noise}},\xi)\,
\pi_{\mathrm{pr}}(w | \sigma_{\mathrm{pr}},\xi)\,dw.
$
This quantity evaluates how well the architecture \(\xi\), together with \(\sigma\), explains the training data after marginalizing over parameter uncertainty. The posterior over the inference hyperparameters is given by
\begin{equation}
\pi_{\mathrm{post}}(\sigma\mid \mathcal{D},\xi)
=
\frac{
\pi_{\mathrm{evid}}(\mathcal{D}\mid \sigma,\xi)\,
\pi_{\mathrm{pr}}(\sigma\mid \xi)
}{
\pi_{\mathrm{evid}}(\mathcal{D}\mid \xi)
}.
\label{eq:sigma_post}
\end{equation}
In this work, \blue{the posterior in \eqref{eq:sigma_post} is
approximated by a Gaussian centered at its maximum a posteriori
estimate \(\sigma_{\mathrm{MAP}}\), using a Laplace approximation.}
The architecture-level evidence is obtained by marginalizing over
\(\sigma\),
$
\pi_{\mathrm{evid}}(\mathcal{D}| \xi)
=
\int
\pi_{\mathrm{evid}}(\mathcal{D} | \sigma,\xi)\,
\pi_{\mathrm{pr}}(\sigma | \xi)\,d\sigma,
$
\blue{with the integral approximated using a Laplace expansion around
\(\sigma_{\mathrm{MAP}}\). Details of these approximations are provided
in ~\ref{app:evidence}, following \cite{singh2024framework}.}
\blue{Let \(\mathcal M=\{\xi_m\}_{m=1}^{M}\) denote the finite set of architectures being compared. While architecture probabilities are conditional on this comparison set, we suppress this conditioning in the notation for simplicity.} The posterior plausibility of architecture \(\xi_m\) is defined as
\begin{equation}
\rho_m
=
\pi_{\mathrm{post}}(\xi_m\mid \mathcal{D})
=
\frac{
\pi_{\mathrm{evid}}(\mathcal{D}\mid \xi_m)\,
\pi_{\mathrm{pr}}(\xi_m)
}{
\sum_{j=1}^{M}
\pi_{\mathrm{evid}}(\mathcal{D}\mid \xi_j)\,
\pi_{\mathrm{pr}}(\xi_j)
},
\qquad m=1,\ldots,M.
\label{eq:rho_m}
\end{equation}
Under a uniform prior over architectures, \(\rho_m\) reduces to normalized evidence.
The plausibility weights \(\{\rho_m\}_{m=1}^{M}\) can be used either to select the most plausible architecture 
\blue{
$\xi_{\mathrm{plausible}}=\xi_{m^\star}$ with 
$m^\star=\arg\max_{m\in\{1,\ldots,M\}}\rho_m$ } or to form a plausibility-weighted ensemble predictor:
\begin{eqnarray}
\pi\!\left(
G^{* (k+1:k+H)}
\,\middle|\,
G^{(k-d:k)},\,M^{(k-d:k)},\,I^{(k-d:k)},\,\mathcal{D}
\right) =
\nonumber \\
\sum_{m=1}^{M}
\rho_m\,
\pi\!\left(
G^{* (k+1:k+H)}
\,\middle|\,
G^{(k-d:k)},\,M^{(k-d:k)},\,I^{(k-d:k)},\,\mathcal{D},\,\xi_m
\right).
\label{eq:bma}
\end{eqnarray}

%++++++++++++++++++++++++++++++++++++++++++++++++++++++++++++++++++++++++
\subsection{EVIDENT: Evidence-based identification
of neural architectures}\label{sec:evident}
\blue{EVIDENT is an iterative architecture discovery strategy that
coordinates Bayesian training, evidence-based ranking, and validation
under predictive uncertainty to balance representational capacity,
support from available data, and reliable task-specific forecasting performance.
It explores ordered capacity levels, prioritizes evaluated candidates
within each level by approximate Bayesian evidence, and validates
shortlisted candidates. Acceptance determines whether the procedure
stops or advances to the next capacity level.} 
\blue{Architectures with insufficient capacity may
fail to capture relevant temporal dynamics, whereas more flexible
architectures permit a broader range of behaviors that the available
data may not adequately constrain. In the Bayesian evidence
perspective \cite{mackay1995probable}, additional flexibility is
supported when its contribution to explaining the observations
compensates for the associated complexity, as determined by the
prior and likelihood. EVIDENT combines this evidence-based comparison
with ordered capacity exploration and validation under predictive
uncertainty. It therefore seeks architectures whose flexibility is
supported by the training data and whose forecasts satisfy the
task-specific requirements.}
An algorithmic summary of EVIDENT is provided in Figure~\ref{fig:algorithm}, \blue{and its components are described below.}

\paragraph{Candidate pool and capacity levels}
Let \(\mathcal{M}_{\mathrm{initial}}=\{\xi_m\}_{m=1}^{M}\) denote the finite, but potentially large, candidate architecture set induced by the feasible architecture space. EVIDENT partitions this set into ordered capacity levels,
$
\mathcal{M}_{\mathrm{initial}}
=
\bigcup_{l=1}^{L}\mathcal{M}^{(l)},
\;
\mathcal{M}^{(l)}\cap \mathcal{M}^{(r)}=\varnothing
\ \text{for}\ l\neq r,
$
where the levels are arranged from lower to higher model capacity according to a prescribed complexity measure, such as the number of trainable parameters. This construction avoids committing apriori to either overly small architectures, which may lack sufficient expressive power, or highly over-parameterized architectures, whose parameters cannot be reliably informed by the available data. Instead, EVIDENT searches progressively over increasingly expressive architecture classes guided by posterior plausibility.

%

%---------------
\begin{figure}[H]
\centering
\fbox{
\begin{minipage}{0.92\linewidth}
\textbf{EVIDENT: EVidence-based IDEntification of Neural archiTectures}

\vspace{0.3em}
\textbf{Input:} Initial architecture pool \(\mathcal{M}_{\mathrm{initial}}=\{\xi_m\}_{m=1}^M\), 
training data \(\mathcal{D}\), 
validation data \(\mathcal{D}_{\mathrm{val}}\), 
tolerance \(Tol\).

\textbf{Output:} \blue{Architectures selected through evidence-based
ranking and task-specific validation, or no accepted candidate
within the evaluated pool.}

\vspace{0.3em}
\textbf{Procedure:}
\begin{enumerate}
\item Partition \(\mathcal{M}_{\mathrm{initial}}\) into \blue{ordered capacity levels} \(\{\mathcal{M}^{(l)}\}_{l=1}^L\).
\item For \(l=1,\dots,L\):
\begin{enumerate}
    \item \blue{Explore candidate architectures within
\(\mathcal{M}^{(l)}\), perform Bayesian training on \(\mathcal{D}\),
and 
rank evaluated candidates by posterior plausibility
(equivalently, by approximate evidence under a uniform architecture prior).
Retain the top-ranked candidates in
\(\widehat{\mathcal{M}}^{(l)} \subseteq \mathcal{M}^{(l)}\).}

    \item Validate \(\xi \in \widehat{\mathcal{M}}^{(l)}\) on \(\mathcal{D}_{\mathrm{val}}\):
\[
\mathbb{d}\!\left(\mathcal{D}_{\mathrm{val}},\,\mathcal{T}_{\xi}\right)\leq Tol.
\]

    \item \blue{If at least one shortlisted candidate satisfies the
validation requirements, return the accepted architecture(s)
and terminate; otherwise, proceed to the next capacity level.}
\end{enumerate}

\item \blue{If no shortlisted candidate is accepted after all
capacity levels have been explored, report that no architecture
was accepted under the prescribed search and validation protocol.}

\end{enumerate}
\end{minipage}
}
\caption{EVIDENT workflow. Architectures are explored from lower to higher capacity, ranked by posterior plausibility, and validated using a task-specific probabilistic metric \(\mathbb{d}\) (e.g., NLPD). 
\blue{Within-level exploration may use enumeration or adaptive search. Only shortlisted candidates undergo validation, and the procedure stops at the first level yielding an accepted candidate.
Validation follows the prescribed forecasting protocol,
including subject-specific adaptation where applicable.}
}
\label{fig:algorithm}
\end{figure}
%---------------

\paragraph{Level-wise architecture ranking}
\blue{Within each capacity level \(l\), candidate architectures may be evaluated
exhaustively or explored using an architecture search algorithm.
Each evaluated candidate undergoes Bayesian training and evidence
estimation as described above. Candidates are ranked by posterior
plausibility, while under a uniform architecture prior within the comparison
set, this ranking is equivalent to ranking by approximate log-evidence.
The shortlist
\(\widehat{\mathcal{M}}^{(l)} \subseteq \mathcal{M}^{(l)}\)
contains the highest-ranked evaluated candidate or several candidates
with comparable posterior plausibility. Only shortlisted candidates
advance to task-specific validation.
Thus, the search algorithm determines which candidates are explored within a level, while validation determines acceptance and progression between levels. The search space, algorithm, and within-level termination settings are specified for each experiment.
In the results provided in this paper, 
the in silico glucose study enumerates candidates within each
explored capacity level, whereas the recorded patient glucose and
synthetic ECG studies use Optuna \cite{optuna_2019} for within-level
candidate exploration.
Plausibility values in \eqref{eq:rho_m} are normalized over the evaluated candidates in the current level, thus evidence from unvisited levels is not required.
}

\paragraph{Task-specific validation under uncertainty}
This stage is the central step in EVIDENT as 
\blue{it assesses whether shortlisted predictors
meet the prescribed forecasting requirements under predictive
uncertainty. Its design follows the intended forecasting scenario.
In the glucose studies, population-trained models are adapted to
individuals excluded from population training and validated on a
subsequent period excluded from adaptation. The in silico glucose
study repeats this assessment across population-training folds.
Validation used for acceptance is part of architecture selection.}
Acceptance is determined using one or more probabilistic validation criteria together with a prescribed tolerance $Tol$.
\blue{Probabilistic forecasting is assessed using predictive scoring rules, complemented by diagnostics of accuracy and calibration \cite{gneiting2007probabilistic, sergazinov2025improving, maupin2018validation}. These evaluation principles do not, by themselves, determine a universal numerical acceptance threshold. In EVIDENT, the tolerance is a task-specific input whose interpretation depends on the scoring convention, forecast horizon, and data scaling. Architecture acceptance and the stopping capacity level are therefore conditional on this requirement. 
Changing the tolerance can alter candidate acceptance and,
consequently, the stopping capacity level. Sensitivity analyses
assess how these decisions depend on the prescribed requirement.
In an operational application, the tolerance would need
to be justified against application-specific forecasting
requirements and assessed separately from the data used for
architecture selection.
}

In this work, The primary validation metric is the negative log predictive
density (NLPD),
\blue{evaluated separately for each
forecast window and averaged over its \(H\) forecast steps:}
\begin{equation}
\mathrm{NLPD}_i
=
\frac{1}{H}\sum_{h=1}^{H}
\left[
\frac{
\left(
G_{\mathrm{data},i}^{(k+h)}
-
\left[
\overline{\mathcal{T}}_{w,\xi}\!\left(
G_{\mathrm{data},i}^{(k-d_G:k)},
M_{\mathrm{data},i}^{(k-d_M:k)},
I_{\mathrm{data},i}^{(k-d_I:k)}
\right)
\right]_h
\right)^2
}{2\sigma_{i,h}^{2}}
+
\frac{1}{2}\log \sigma_{i,h}^{2}
\right].
\label{eq:nlpd}
\end{equation}
\blue{Here, \(i\) indexes validation windows,
\(\overline{\mathcal{T}}_{w,\xi}\) denotes the posterior predictive
mean of the TCN forecast, and \([\cdot]_h\) denotes its \(h\)-th
forecast component. 
This Gaussian predictive score omits the additive constant
$\tfrac{1}{2}\log(2\pi)$ and all reported NLPD values and tolerances
use this convention.
The corresponding predictive variance is}
\begin{equation}
\sigma_{i,h}^{2}
=
\sigma_{\mathrm{data}}^{2}
+
\operatorname{Var}\!\left[
\left[
\mathcal{T}_{w,\xi}\!\left(
G_{\mathrm{data},i}^{(k-d_G:k)},
M_{\mathrm{data},i}^{(k-d_M:k)},
I_{\mathrm{data},i}^{(k-d_I:k)}
\right)
\right]_h
\right].
\label{eq:sigma_nlpd}
\end{equation}
\blue{The variance is taken over the network weight posterior
and \(\sigma_{\mathrm{data}}^{2}\) denotes observation noise variance such as CGM sensor errors.
A candidate is accepted only if
\(\mathrm{NLPD}_i \leq Tol\) for every required validation window
and subject. }

\paragraph{Iterative search and stopping criteria}
If at least one candidate in level \(l\) satisfies the validation criteria, the search terminates and EVIDENT returns the corresponding architecture \blue{satisfying the prescribed validation criterion}. Otherwise, the algorithm advances to the next level, \(l \leftarrow l+1\), and repeats the training, architecture ranking, and validation under uncertainty cycle. 
\blue{EVIDENT therefore returns an accepted candidate from the first capacity level whose evaluated shortlist satisfies the prescribed validation requirements.}
\blue{If no shortlisted candidate passes validation in the final
capacity level, EVIDENT reports that no architecture was accepted
under the prescribed search and validation protocol. In such cases, an optional decision to be made based depending computational budgets and other considerations is expanding the candidate architecture set.}
The resulting workflow favors plausible architectures supported by the training data and validated for the intended predictive task.

%========================================================
% Results
%========================================================
\section{Results}\label{sec:results}

\blue{This section examines EVIDENT through a preliminary analysis
and three complementary case studies. We first investigate how
approximate Bayesian evidence varies across TCN architectures and
temporal forecasting configurations using synthetic glucose data. We then apply EVIDENT in the in silico
glucose and recorded patient glucose studies using Bayesian TCNs,
followed by the synthetic ECG study using Bayesian LSTMs. These
studies examine evidence based ranking and task-specific acceptance
across different data sources, predictor families, and within-level
search implementations.}
All Bayesian TCN \blue{and LSTM} models were implemented in PyTorch
\cite{paszke2019pytorch}. \blue{The Bayesian TCNs and LSTMs were trained using
variational inference code adapted from the open-source implementations
described in \cite{shridhar2019comprehensive, deighan_2026_20044677}.
The recorded-patient glucose and synthetic ECG studies use Optuna
\cite{optuna_2019} for candidate exploration within capacity levels,
while retaining EVIDENT's ranking, validation, and
stopping rules. Within each level, Optuna's tree-structured Parzen estimator (TPE)
sampler~\cite{bergstra2011algorithms} proposes architectures from the level's
categorical candidate list to maximize the approximate log-evidence. }

%++++++++++++++++++++++++++++++++++++++++++++++++++++++++++++++++++++++++
\subsection{\blue{Preliminary analysis of Bayesian plausibility for TCN forecasting}}
\label{sec:plausibility_results}
We begin with a single-patient forecasting study to examine how Bayesian evidence responds to TCN architecture choices and whether it provides a consistent criterion for navigating the architecture space. This includes the analyses of how evidence varies with architectural complexity and its relationship to accuracy and uncertainty in predicted blood glucose trajectories. This experiment isolates the role of evidence as a selection metric before moving to population-level architecture discovery in the next section.
\begin{figure}[!ht]
    \centering
        \includegraphics[width=0.5\linewidth]{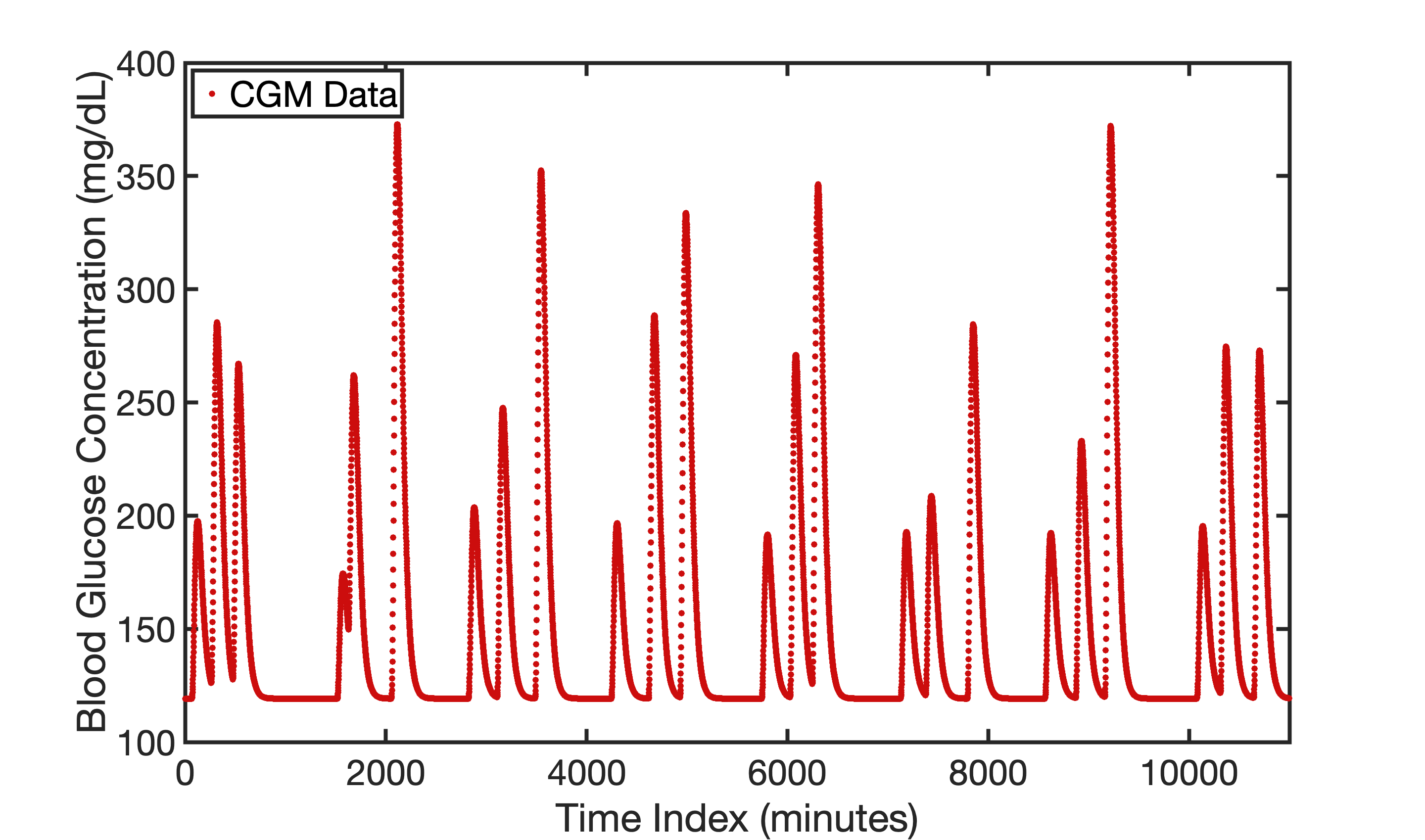}
        \\{\small (a)}\\
        \includegraphics[width=0.4\linewidth]{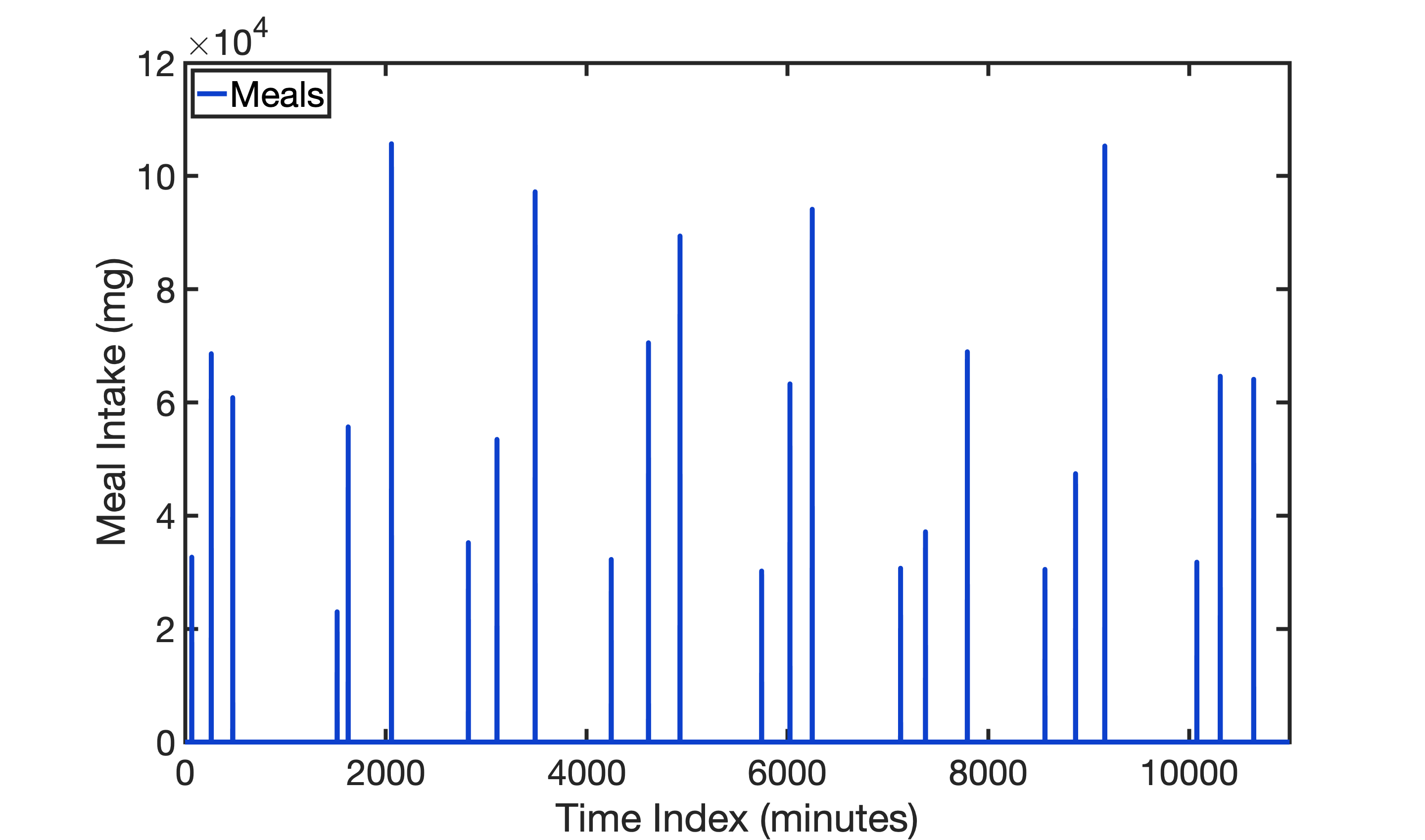}
        ~
        \includegraphics[width=0.4\linewidth]{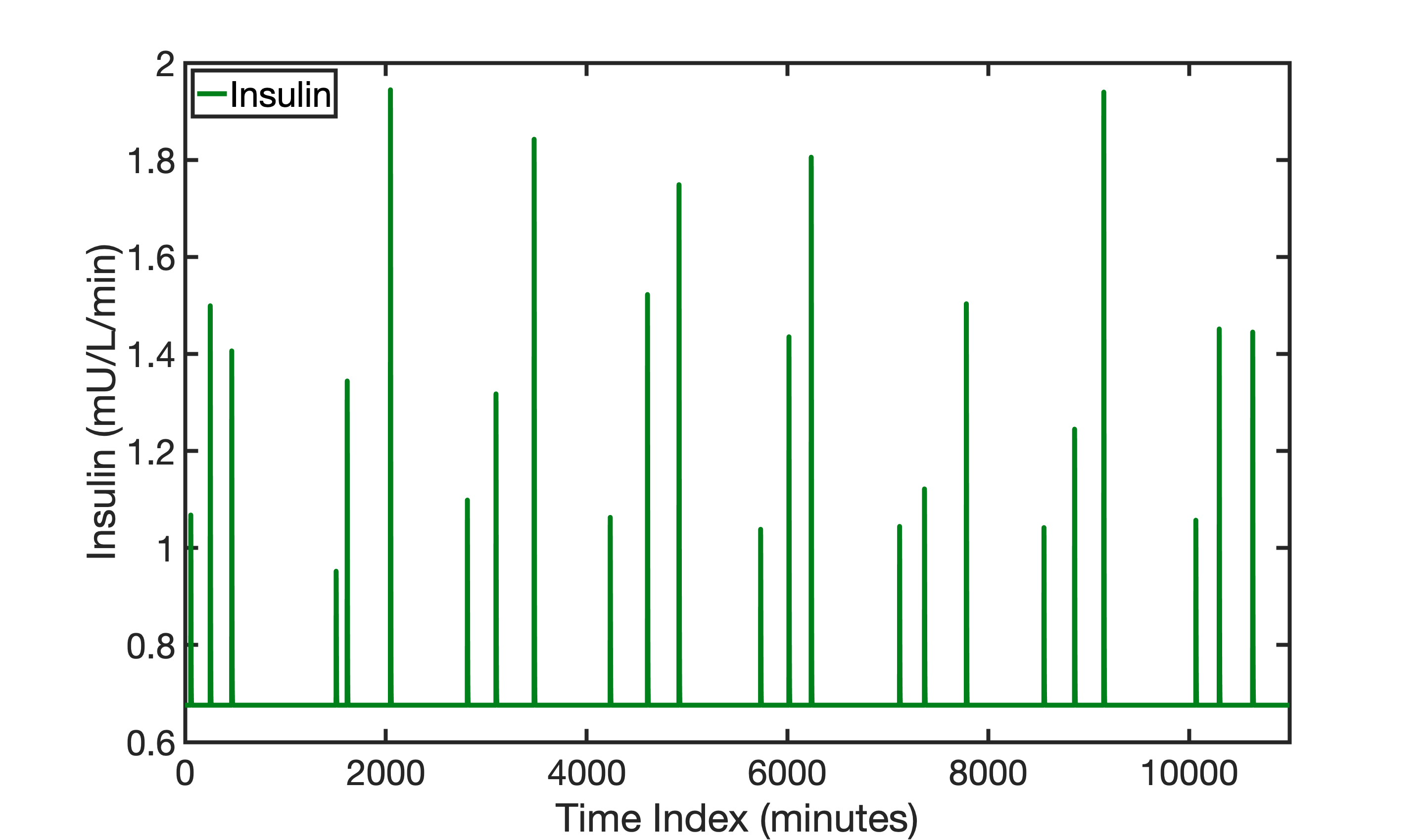}
        \\{\small (b) \hspace{2.6in} (c)}\\
    \caption{Single-patient training data used in the plausibility analysis: (a) blood glucose (CGM), (b) meal intake, and (c) insulin delivery, generated using the Bergman minimal model with Dalla Man meal absorption dynamics.}
    \label{fig:bergman}
\end{figure}
Figure~\ref{fig:bergman} shows the dataset generated from the Bergman minimal model with meal disturbances modeled via the Dalla Man absorption model \cite{bergman1981physiologic, dalla2006system}. To introduce variability, meal size and timing are randomly perturbed around nominal values, while insulin is administered using a deterministic basal–bolus protocol. Detailed model equations and parameter values are provided in the ~\ref{app:bergman}. 
In all TCNs, the input sequences for glucose $G$, meal $M$, and insulin $I$ share a common history window, i.e., $d_G = d_M = d_I$ in \eqref{eq:tcn_operator}.
All the upcoming results in this section were obtained using the same approximetly 90\% train and 10\% test split of the time series shown in Figure~\ref{fig:bergman}. 
The training set corresponds to \(0\text{--}10000\) minutes ($\approx 7$ days) and the test set corresponds to \(10000\text{--}11520\) minutes ($\approx 1$ day).

% -- Evidence vs # blocks/dilation
\begin{figure}[!ht]
\centering
\setlength{\tabcolsep}{4pt}
\renewcommand{\arraystretch}{1.0}

% Top row: architecture descriptors
\begin{tabular}{ccc}
\includegraphics[width=0.28\textwidth]{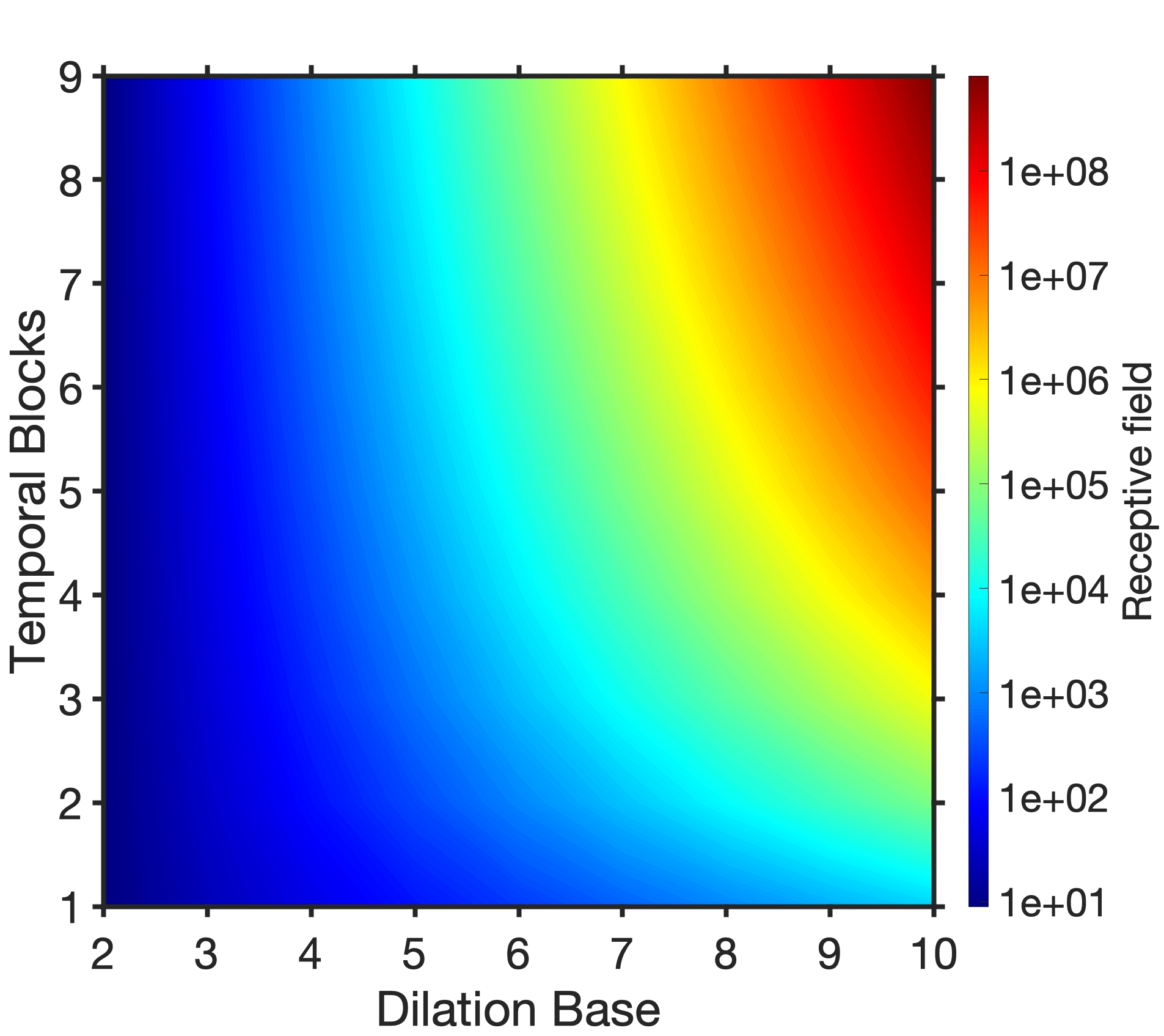} &
\includegraphics[width=0.28\textwidth]{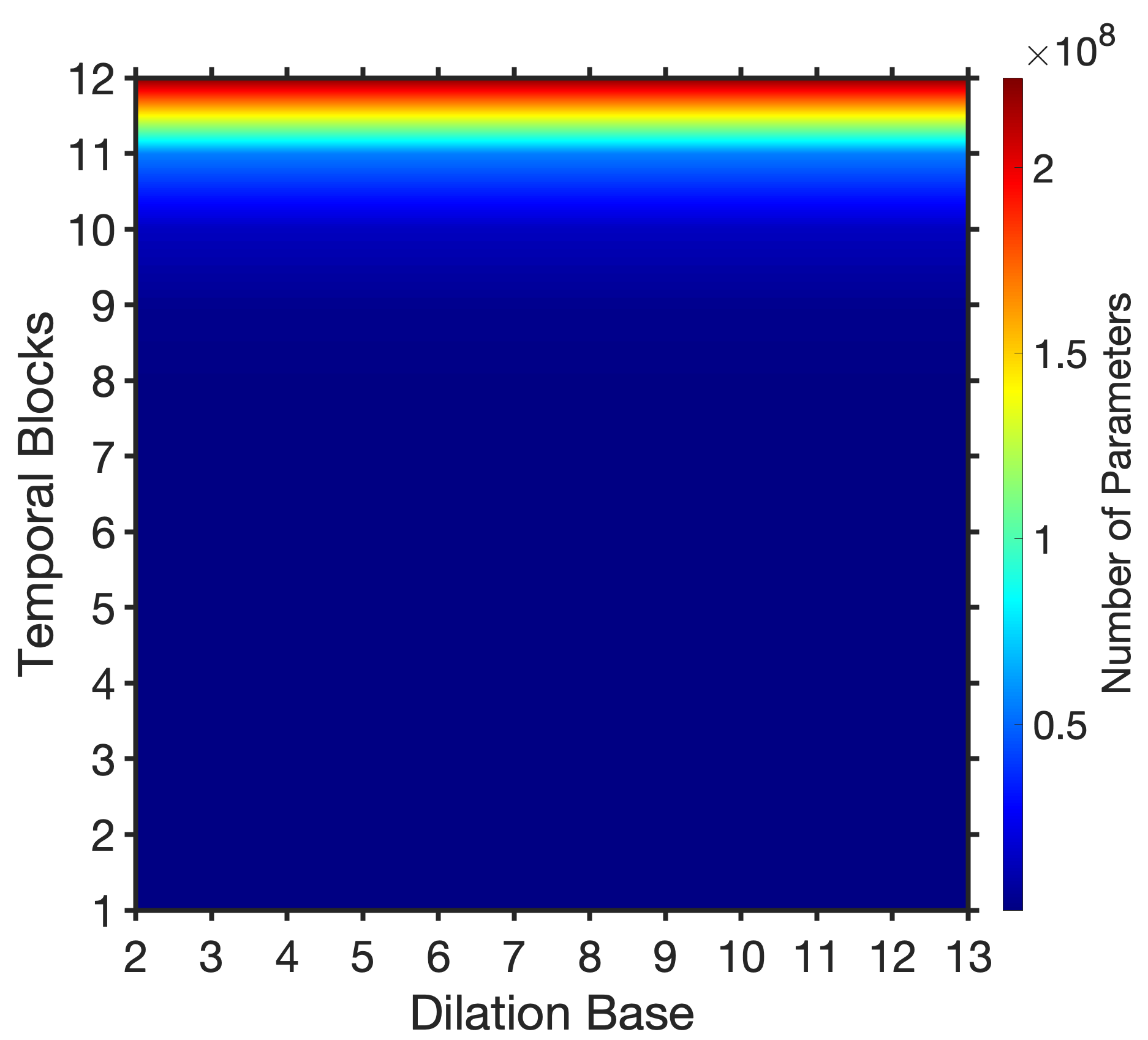} &
\includegraphics[width=0.45\textwidth]{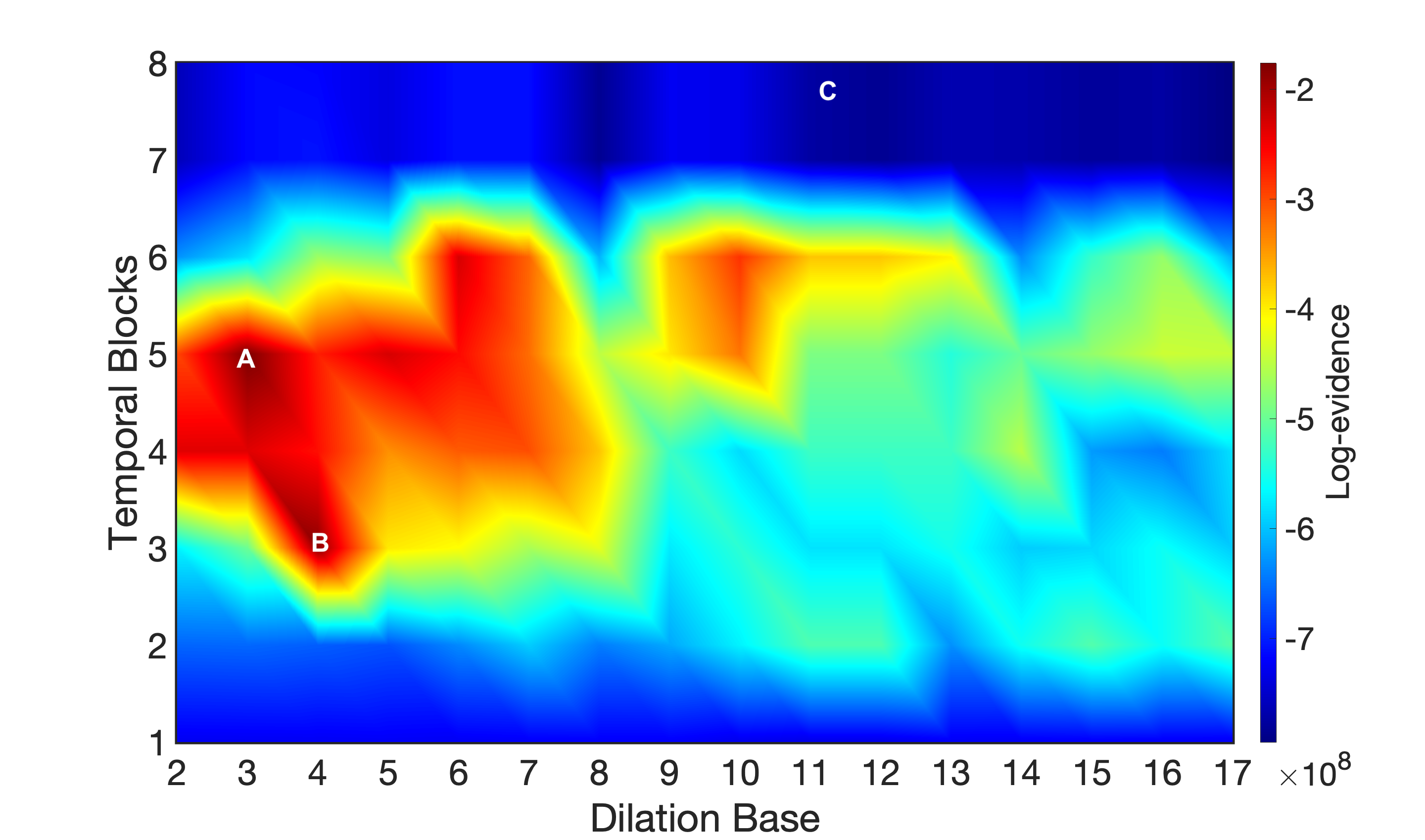}
\\
(a) & (b) & (c)
\end{tabular}
% Bottom row: representative predictions
\begin{tabular}{ccc}
\includegraphics[width=0.30\textwidth]{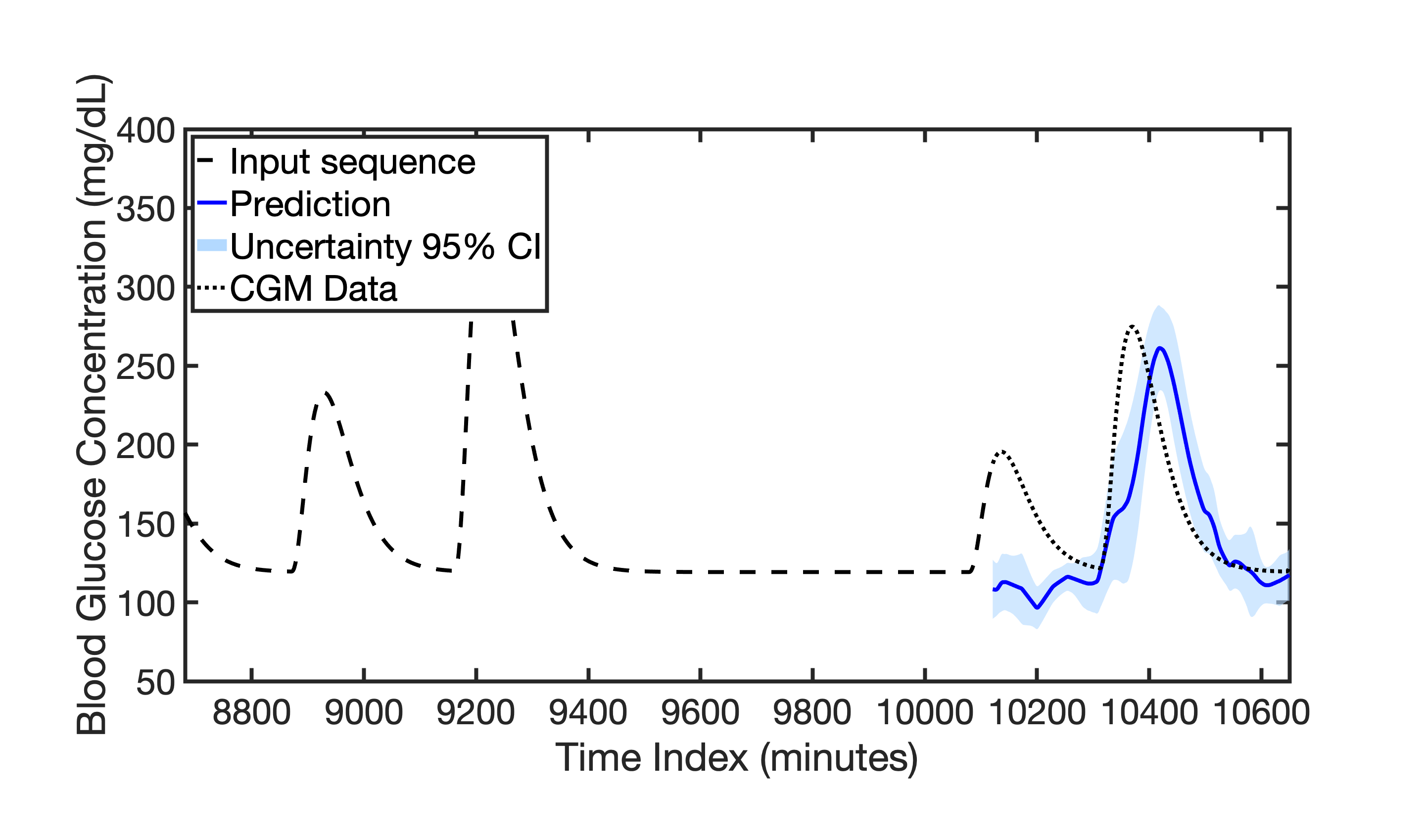} &
\includegraphics[width=0.30\textwidth]{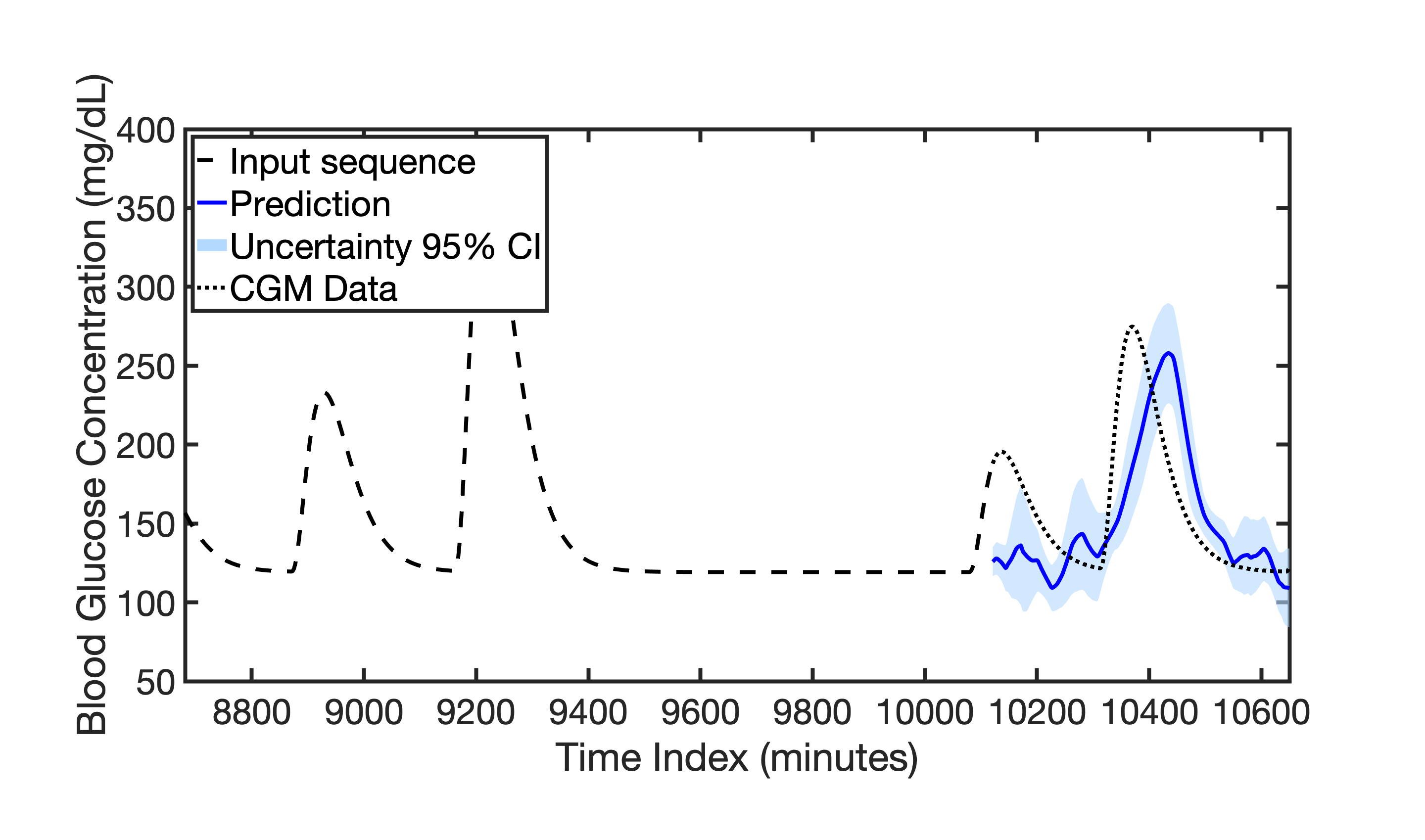} &
\includegraphics[width=0.30\textwidth]{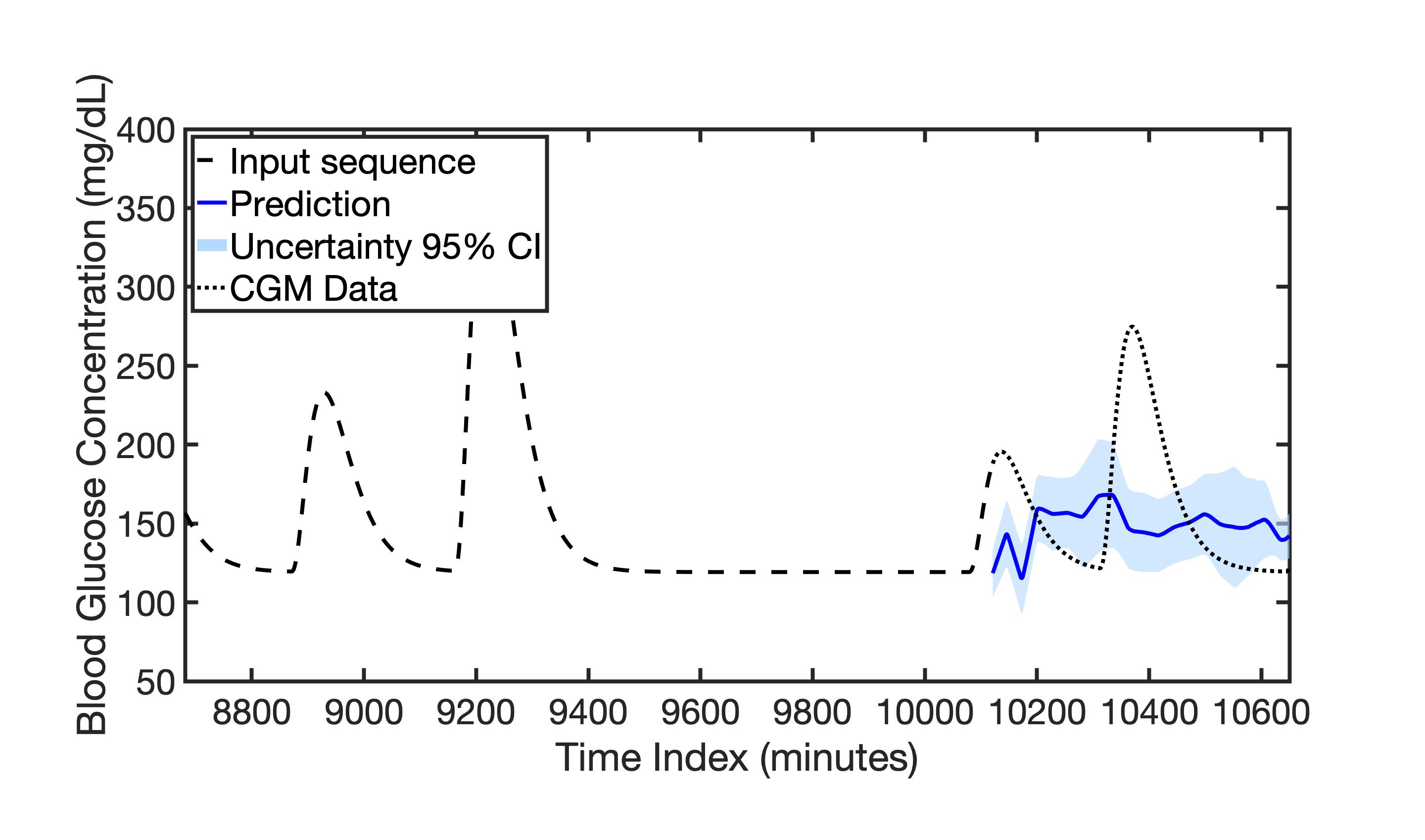} \\
(d) architecture A & (e) architecture B & (f) architecture C
\end{tabular}

\caption{
Evidence landscape and representative predictive behavior across the TCN architecture space.
(a,b) Receptive field and number of trainable parameters as a function of the number of temporal blocks $N$ and dilation base $b$.
(c) Log-evidence surface over $N$ and $b$, showing a highly non-uniform structure with high-evidence architectures concentrated in an intermediate region of the architecture space.
(d--f) Prediction horizons for representative architectures A, B, and C selected from distinct regions of the evidence landscape in (c). 
In all architectures filter size $f=5$, stride $s=3$, encoder/decoder channels $\{c_j\}=\{\tilde{c}_j\}=\{8\}$ repeated across blocks, input history $d_G=d_M=d_I=1400$~min, and prediction horizon $H=500$~min.
}
\label{fig:dil_evid_pred}
\end{figure}

We first examine how Bayesian evidence varies across TCN architectures with different receptive fields and model capacities. 
Figure~\ref{fig:dil_evid_pred}(a and b) show that both the number of temporal blocks \(N\) and the dilation base \(b\) increase the receptive field, while model size, measured by the number of trainable parameters, is primarily driven by \(N\). 
The evidence landscape of the 136 architectures in Figure~\ref{fig:dil_evid_pred}(c) is highly non-uniform, with high-evidence models concentrated in an intermediate region of the \((N,b)\) architecture space. This indicates that neither larger receptive fields nor larger model sizes alone improve model plausibility. Instead, evidence favors architectures that provide sufficient temporal context to capture meal and insulin dynamics while avoiding weakly informed parameters.
The forecast behavior of three representative architectures (A, B, and C) from different regions of the architecture space is shown in Figure~\ref{fig:dil_evid_pred}(d--f). Architectures A and B, located in high-evidence regions, produce accurate forecasts with tighter uncertainty bands. In contrast, architecture C, selected from a low-evidence region, yields less reliable predictions despite its larger number of temporal blocks. These results suggest that evidence is not merely measuring goodness-of-fit, but also identifying architectures with improved predictive reliability under uncertainty.
\blue{In these examples, higher evidence is associated with more
accurate forecasts and tighter predictive intervals, motivating
its use for candidate ranking alongside separate task-specific
validation.}

%

% -- Evidence vs stride/input/output
\begin{figure}[!ht]
\centering
\setlength{\tabcolsep}{4pt}
\renewcommand{\arraystretch}{1.0}
% ==================== TOP BLOCK: STRIDE ====================
\begin{tabular}{ccc}
\includegraphics[width=0.323\textwidth]{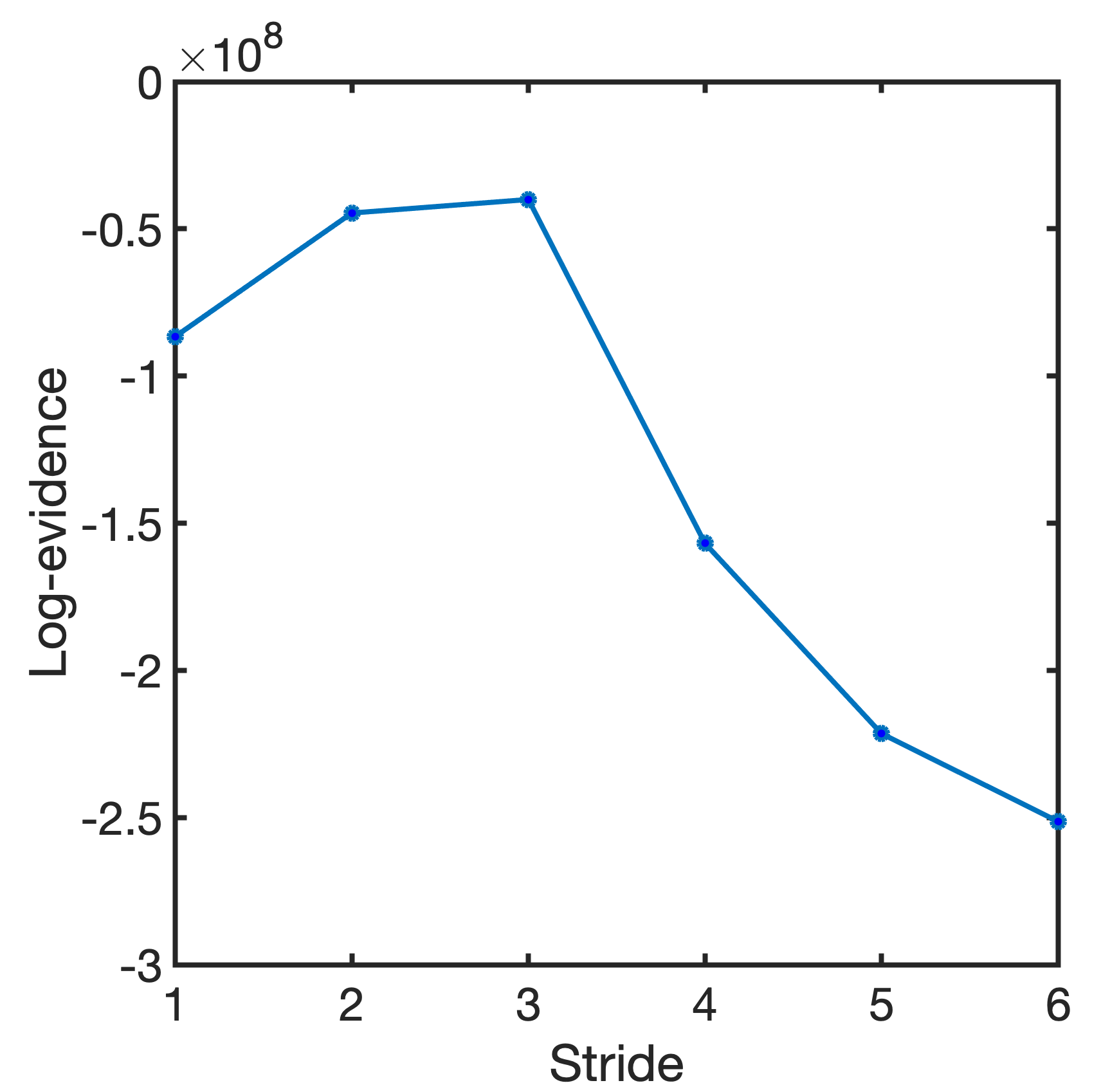} &
\includegraphics[width=0.31\textwidth]{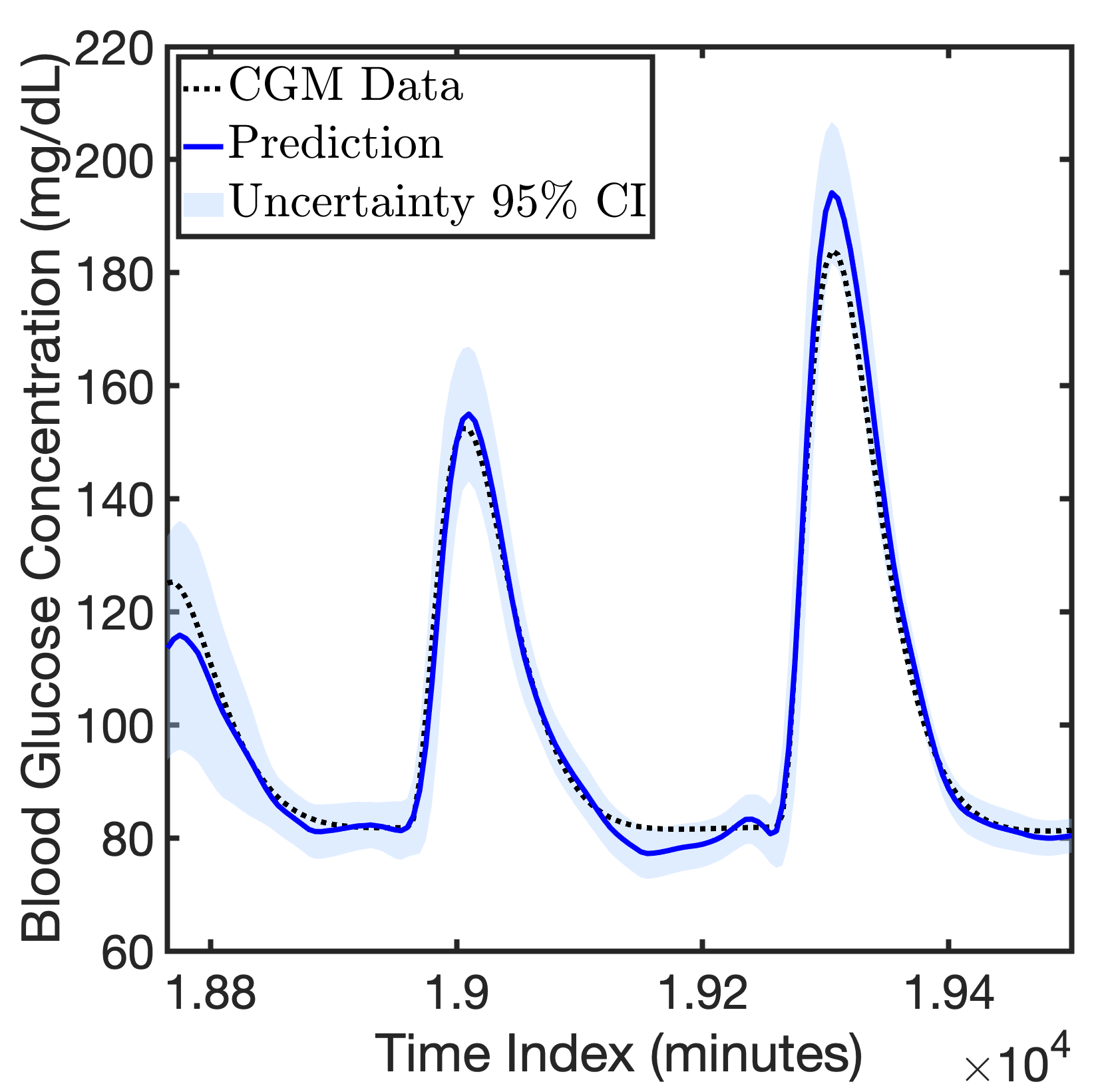} &
\includegraphics[width=0.31\textwidth]{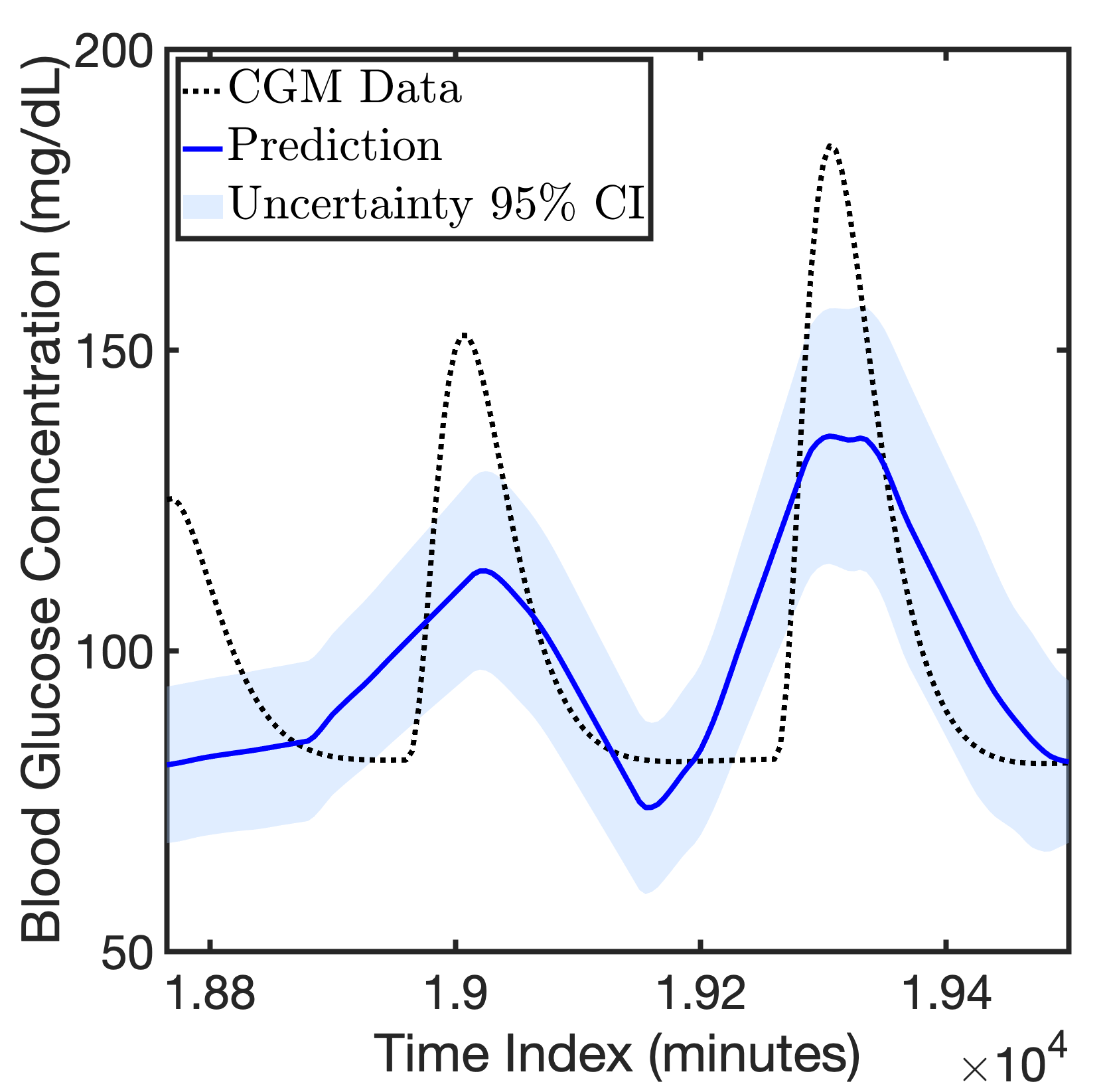} \\
(a) & (b) $s=3$ & (c) $s=6$
\end{tabular}
% ==================== BOTTOM BLOCK: INPUT-OUTPUT ====================
\begin{tabular}{c}
\includegraphics[width=0.78\textwidth]{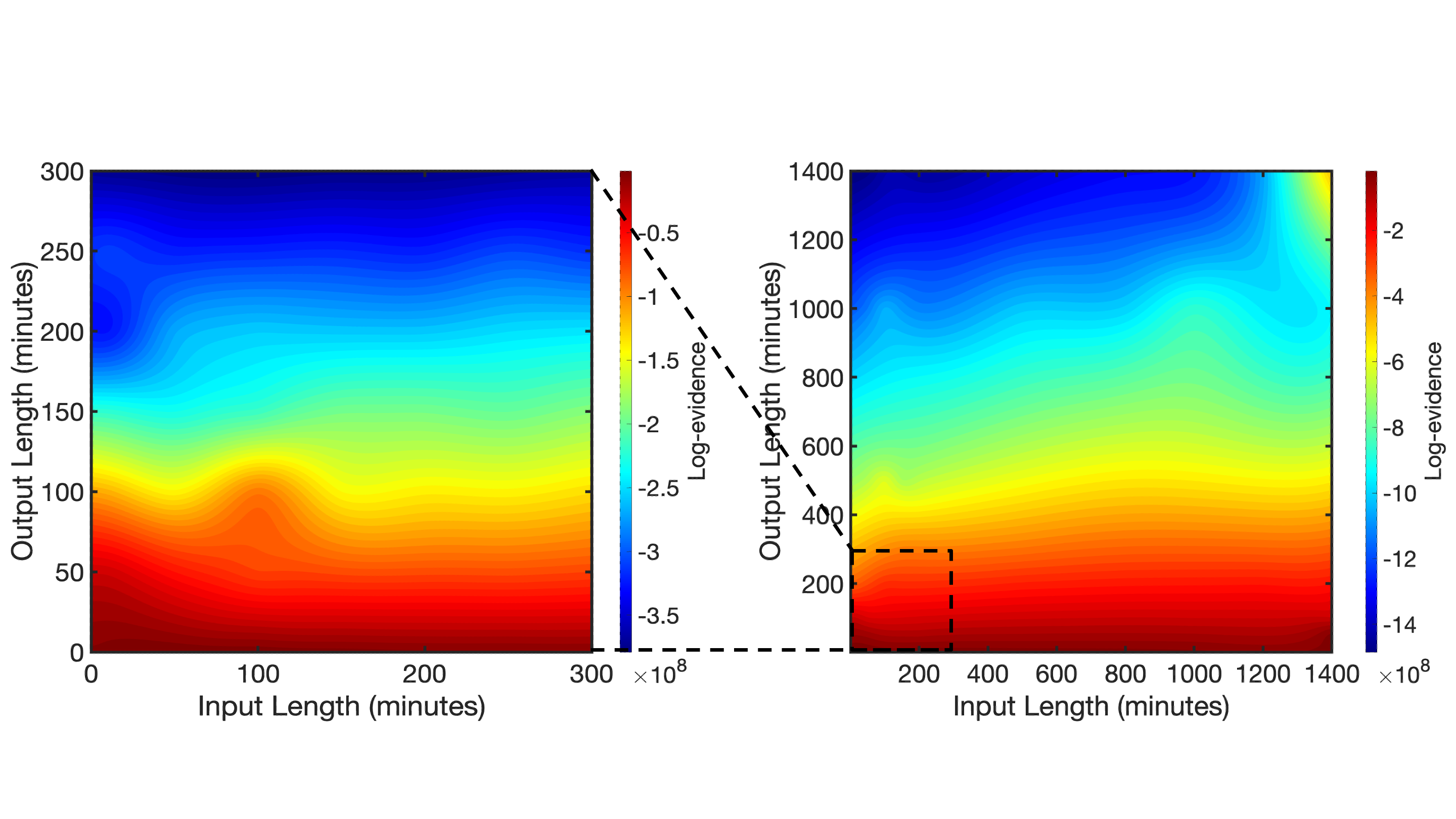} \\
(d)
\end{tabular}

\caption{ 
(a) Log-evidence as a function of the stride $s$ used to construct overlapping input--output windows for TCN training, showing that an intermediate stride provides the highest evidence. 
(b--c) Representative predictions for stride values $s=3$ and $s=6$. 
(d) Log-evidence surface over input history length $d_G=d_M=d_I$ and prediction horizon $H$; the coarse search over 1400 minutes localizes a high-evidence region, which is refined using a finer grid over the highlighted 300 minute region. 
For all panels, the TCN architecture is fixed at $N=4$ temporal blocks, filter size $f=5$, dilation base $b=3$, and encoder/decoder channels $\{c_j\}=\{\tilde{c}_j\}=\{12\}$. For (a--c), the input history and prediction horizon are fixed at $d_G=d_M=d_I=1400$~min and $H=1000$~min; for (d), the stride is fixed at $s=3$.
}
\label{fig:stride_inputoutput}
\end{figure}

In addition to architecture’s structural parameters, evidence also provides guidance for training and forecasting choices in TCN. Figure~\ref{fig:stride_inputoutput}(a-c) shows the effect of stride in constructing overlapping input–output windows and the evidence surface over input length $d$ and prediction horizon $H$. The evidence peaks at stride $s=3$, indicating a balance between highly redundant training windows at small stride and reduced sample efficiency at large stride. 
Figure~\ref{fig:stride_inputoutput}(d) shows that evidence decreases as the prediction horizon increases, indicating that longer-horizon forecasts are less supported by the available data. For a fixed horizon, the evidence surface also localizes a favorable range of input history lengths rather than increasing monotonically with longer input windows. To identify this region efficiently, we first evaluate a coarse grid over input and output lengths and then refine the high-evidence region using a finer search. 
\blue{These calculations examine evidence variation across both
TCN architectures and temporal forecasting configurations.}

%++++++++++++++++++++++++++++++++++++++++++++++++++++++++++++++++++++++++
%++++++++++++++++++++++++++++++++++++++++++++++++++++++++++++++++++++++++
\subsection{\blue{EVIDENT for glucose forecasting in silico subjects}}\label{sec:evident_results}

We now implement EVIDENT in a \blue{controlled in silico forecasting setting} involving inter-patient variability and data noise from CGM sensor. The objective is to leverage retrospective population data to identify TCN architectures that can reliably generalize to unseen individual T1D patients for predicting BG trajectory.

%++++++++++++++++++++++++++++++++++++++++++++++++++++++++++++++++++++++++
\subsubsection{Dataset, architecture space, and validation protocol}

\begin{figure}[!ht]
\centering
\includegraphics[width=0.6\linewidth]{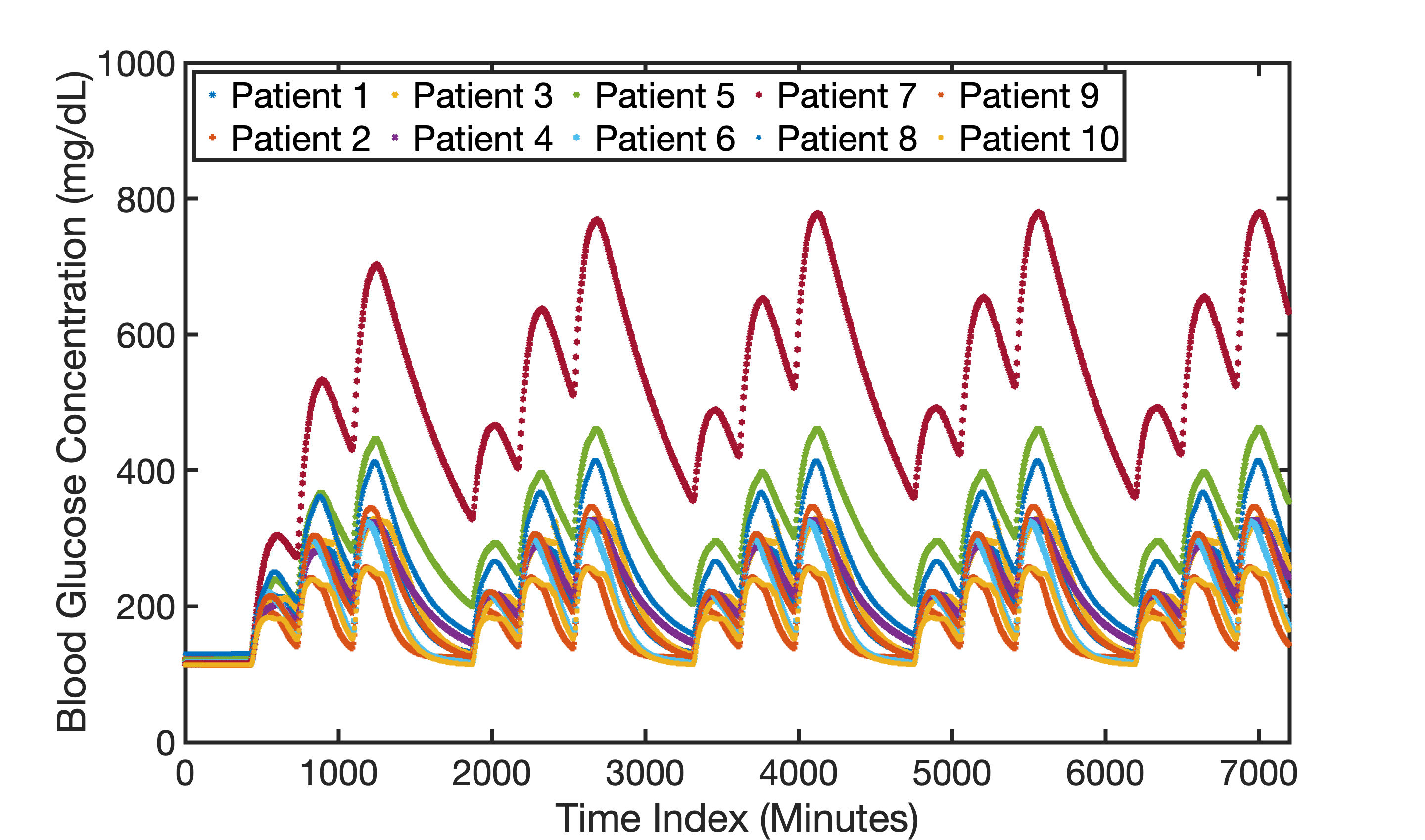}
\caption{Blood-glucose trajectories for 10 adult in silico subjects generated using the UVA/Padova T1D simulator \cite{man2014uva}.}
\label{fig:uva}
\end{figure}

% -- dataset
\paragraph{Population data and patient-wise folds}
Population-level type 1 diabetes (T1D) trajectories are generated using the UVA/Padova T1D simulator \cite{man2014uva}. The dataset consists of 10 adult in silico subjects, each providing continuous glucose monitoring (CGM) data together with the corresponding meal and insulin inputs, as shown in Figure~\ref{fig:uva}. To evaluate generalization under inter-patient variability, we use a 5-fold cross-validation protocol: in each fold, 8 patients are used for population training and the remaining 2 subjects are held out for patient-specific validation, as summarized in Table~\ref{tab:cv_folds}.

\begin{table}[H]
\footnotesize
\centering
\caption{Patient-wise 5-fold cross-validation splits for the 10 adult in silico patients.}
\label{tab:cv_folds}
\begin{tabular}{c l l}
\hline
\textbf{Fold} & \textbf{Training} & \textbf{Held out validation} \\
\hline
1 & Patients [1--8] & Patients [9, 10] \\
2 & Patients [1--5, 8--10] & Patients [6, 7] \\
3 & Patients [3--10] & Patients [1, 2] \\
4 & Patients [1, 2, 5--10] & Patients [3, 4] \\
5 & Patients [1--4, 6, 7, 9, 10] & Patients [5, 8] \\
\hline
\end{tabular}
\end{table}

%-- architecture pool
\paragraph{Candidate architecture pool}
The candidate TCN pool is constructed from the feasible architecture space in \eqref{eq:feasible_space}. To ensure adequate temporal context for glucose dynamics, architectures are constrained to satisfy
$
1400~\text{minutes} \le \mathrm{RF}(\xi) \le 2000~\text{minutes},
$
where the lower bound ensures access to at least one daily cycle and the upper bound avoids unnecessarily long temporal context. We also enforce the no-holes condition \(f\geq b\) for dilated convolutions. Additional implementation bounds on filter size, depth, and channel width are imposed to keep the search computationally tractable. These constraints yield a finite pool of 61 candidate TCN architectures. For EVIDENT, the candidates are partitioned into six ordered capacity levels according to the number of trainable parameters, as summarized in Table~\ref{tab:arch_level}. Full architecture specifications are provided in ~\ref{app:architecture_pool}.

%-- training specifics/validation protocol
\paragraph{Validation protocol}
The training and validation protocol is designed to emulate deployment to unseen T1D subjects with limited subject-specific data. In each cross-validation fold, candidate architectures are first trained on population data using input--output windows extracted from inter-meal intervals. The output horizon is set to 285 minutes, corresponding to the minimum inter-meal gap, and windows containing meal events within the prediction horizon are excluded.
For each held-out subject, the posterior distribution obtained from population training is subsequently used as the prior for subject-specific adaptation using four days of that subject’s data. Validation is then performed using rolling one-hour prediction windows  shifted to cover the full day 5, while excluding windows intersecting meal events and truncating predictions 10 minutes before meal onset to avoid scoring across unobserved disturbances. A candidate architecture is considered acceptable if its NLPD remains below the prescribed tolerance \(Tol\) across all one-hour validation windows for both held-out subjects.
\blue{For this methodological study, we use $Tol=14.5$ as an
empirically chosen reference setting for the rolling one-hour
forecasting screen, rather than a threshold derived from clinical
requirements or an external calibration procedure. Acceptance
requires every scored validation window for the required held-out
subjects to satisfy this criterion. The resulting acceptance
decisions are therefore conditional on this reference setting.
Table~\ref{tab:uva_tol_sensitivity} reports how the acceptance of
the selected predictor changes when the tolerance is varied while
its predictions and validation windows are held fixed.}

\begin{footnotesize}
\setlength{\tabcolsep}{5pt}
\renewcommand{\arraystretch}{1.3}
\begin{longtable}{p{1.8cm}p{1.8cm}p{2.0cm}p{2.0cm}p{1.6cm}p{1.8cm}p{2.0cm}}
\caption{Candidate TCN architectures organized into six capacity levels based on the number of trainable parameters. Each row reports the range of architecture hyperparameters used in level $l$: the number of temporal blocks $N$, filter size $f$, dilation base $b$, and starting number of encoder channels $c_1$. The dilation factor in encoder block $j$ is defined as $\delta_j = b^{j-1}$, and the number of encoder channels follows a doubling schedule $c_j = 2^{j-1} c_1$ for $j = 1, \dots, N$. The decoder channels $\{\tilde{c}_j\}_{j=1}^{N}$ mirrors the encoder channels in reverse order.}
\label{tab:arch_level}\\
\toprule
\textbf{Capacity level $l$} & \textbf{Parameter range} & \textbf{Number of architectures} & \textbf{Number of temporal blocks $N$} & \textbf{Filter size $f$} & \textbf{Dilation base $b$} & \textbf{Starting number of channels $c_1$} \\
\midrule
\endfirsthead
\toprule
\textbf{Capacity level $l$} & \textbf{Parameter range} & \textbf{Number of architectures} & \textbf{Number of temporal blocks $N$} & \textbf{Filter size $f$} & \textbf{Dilation base $b$} & \textbf{Starting number of channels $c_1$} \\
\midrule
\endhead
\endfoot
\bottomrule
\endlastfoot
1 & $\le 30$k      & 10 & $[3-4]$          & $[7- 13]$ & $[4- 9]$ & $[2, 4]$     \\
2 & $30$k--$100$k  &  9 & $[3- 6]$    & $[3- 13]$ & $[3- 9]$ & $[2, 8]$     \\
3 & $100$k--$400$k & 10 & $[3- 6]$    & $[3- 13]$ & $[2- 9]$ & $[2, 16]$    \\
4 & $400$k--$1$M   & 11 & $[3- 7]$ & $[3- 13]$ & $[2- 9]$ & $[2, 32]$    \\
5 & $1$M--$5$M     & 11 & $[3- 7]$ & $[3- 13]$ & $[2- 9]$ & $[4, 64]$    \\
6 & $\ge 5$M       & 10 & $[3- 7]$ & $[7- 13]$ & $[2-9]$ & $[8, 128]$   \\
\end{longtable}
\end{footnotesize}

%++++++++++++++++++++++++++++++++++++++++++++++++++++++++++++++++++++++++
\subsubsection{\blue{Architecture ranking and validation results}}

\begin{figure}[!ht]
\centering
%--- plausibility (Level 1)
\includegraphics[width=0.19\textwidth]{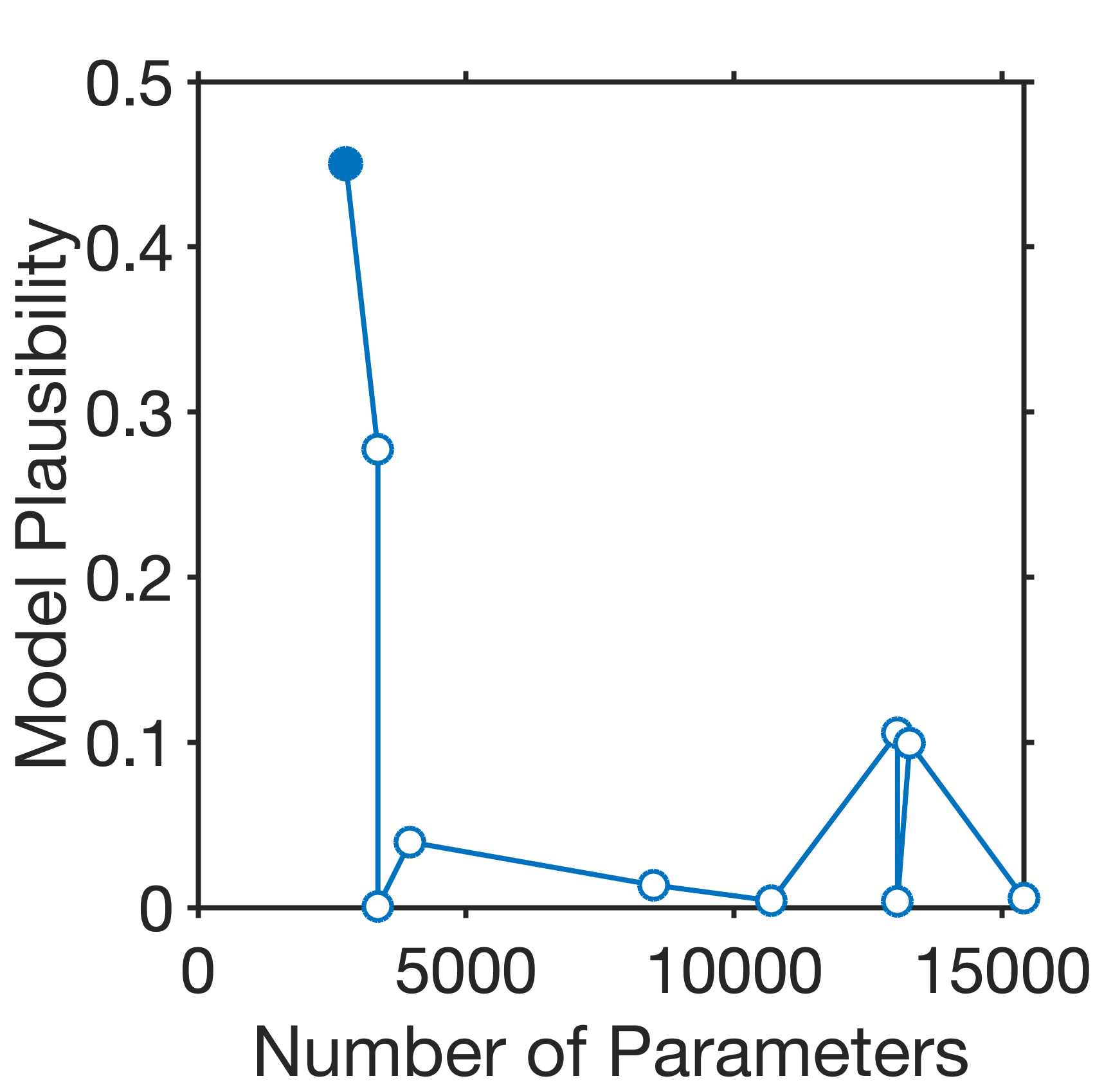}\hfill
\includegraphics[width=0.19\textwidth]{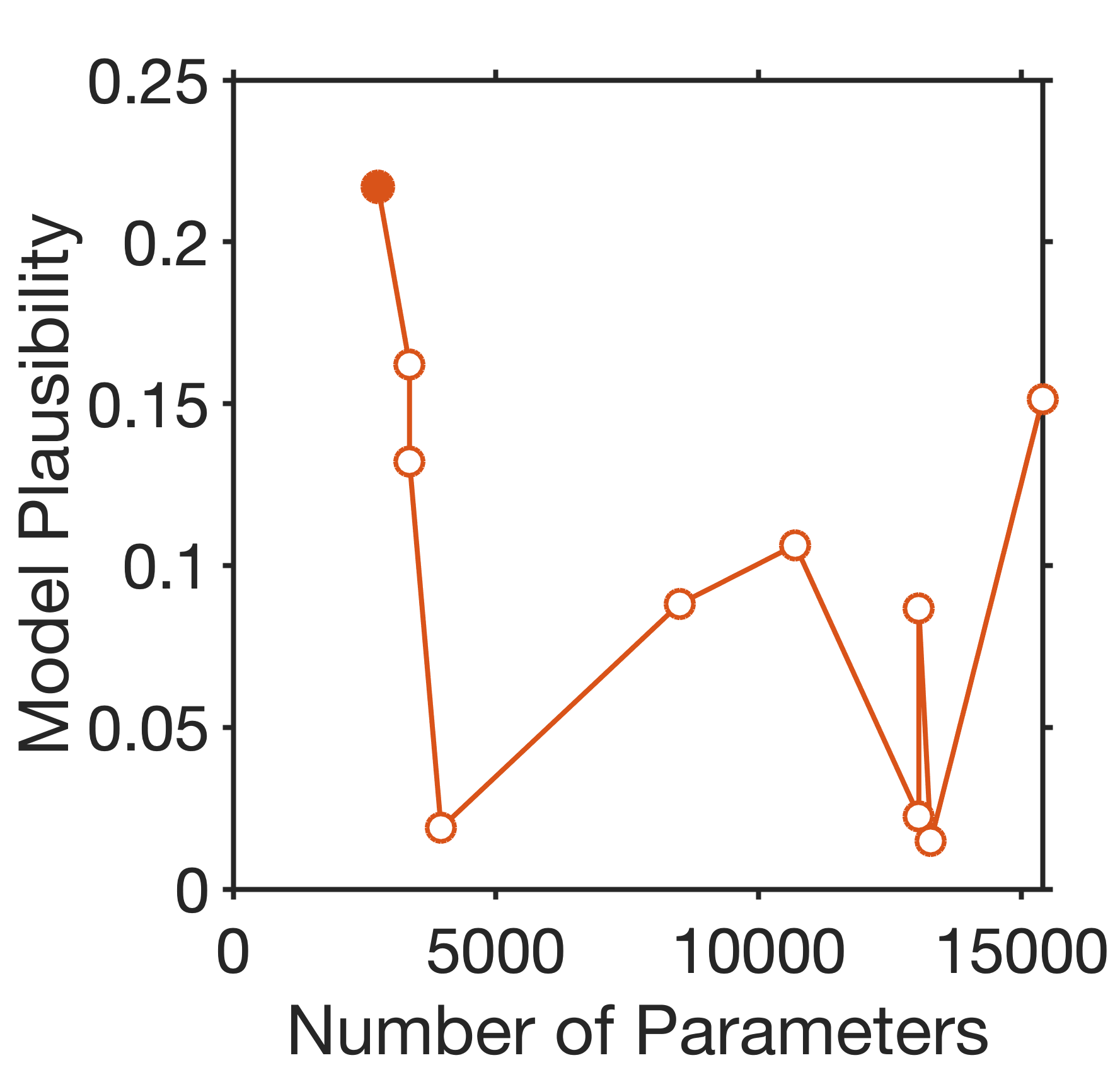}\hfill
\includegraphics[width=0.19\textwidth]{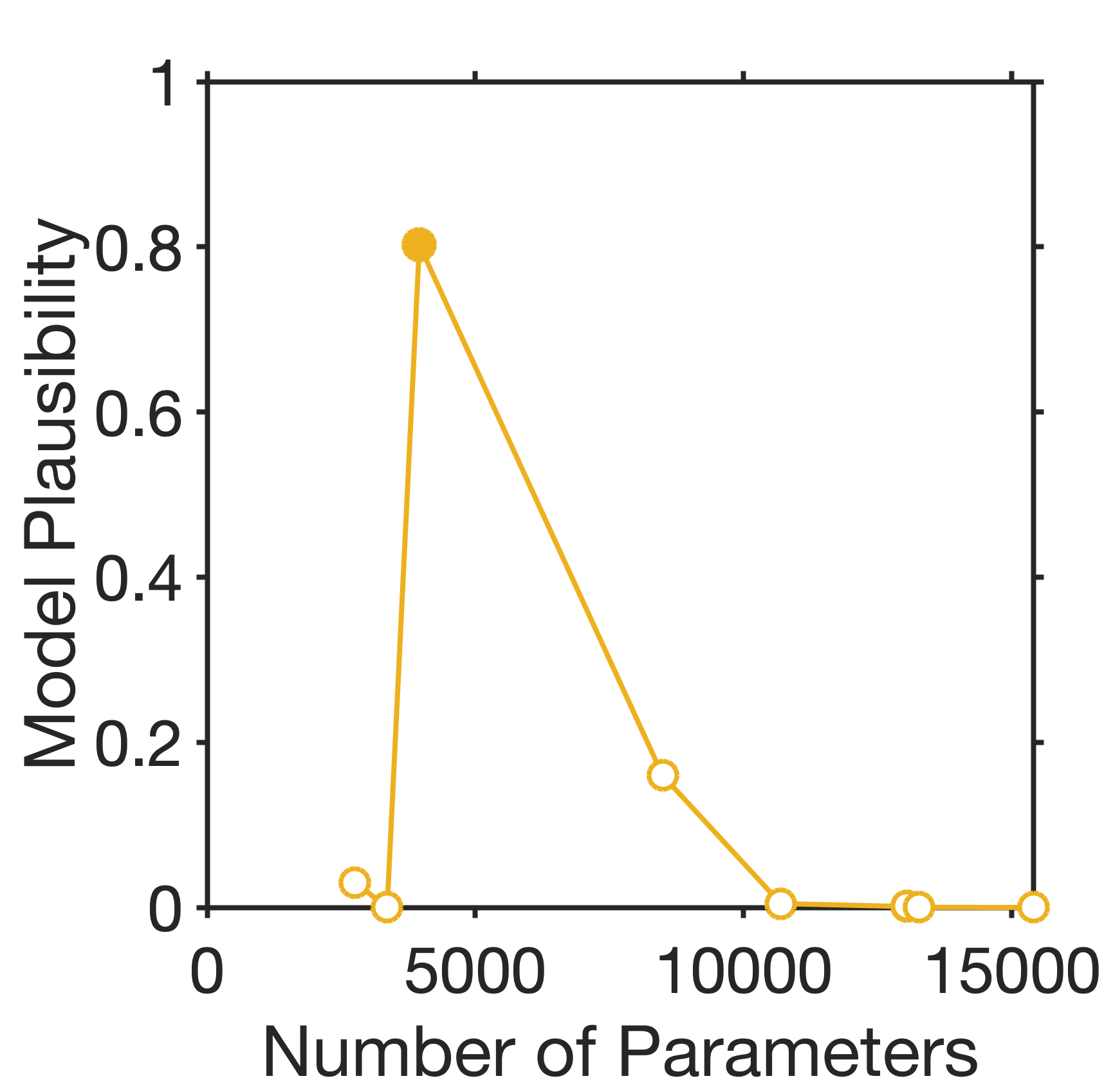}\hfill
\includegraphics[width=0.19\textwidth]{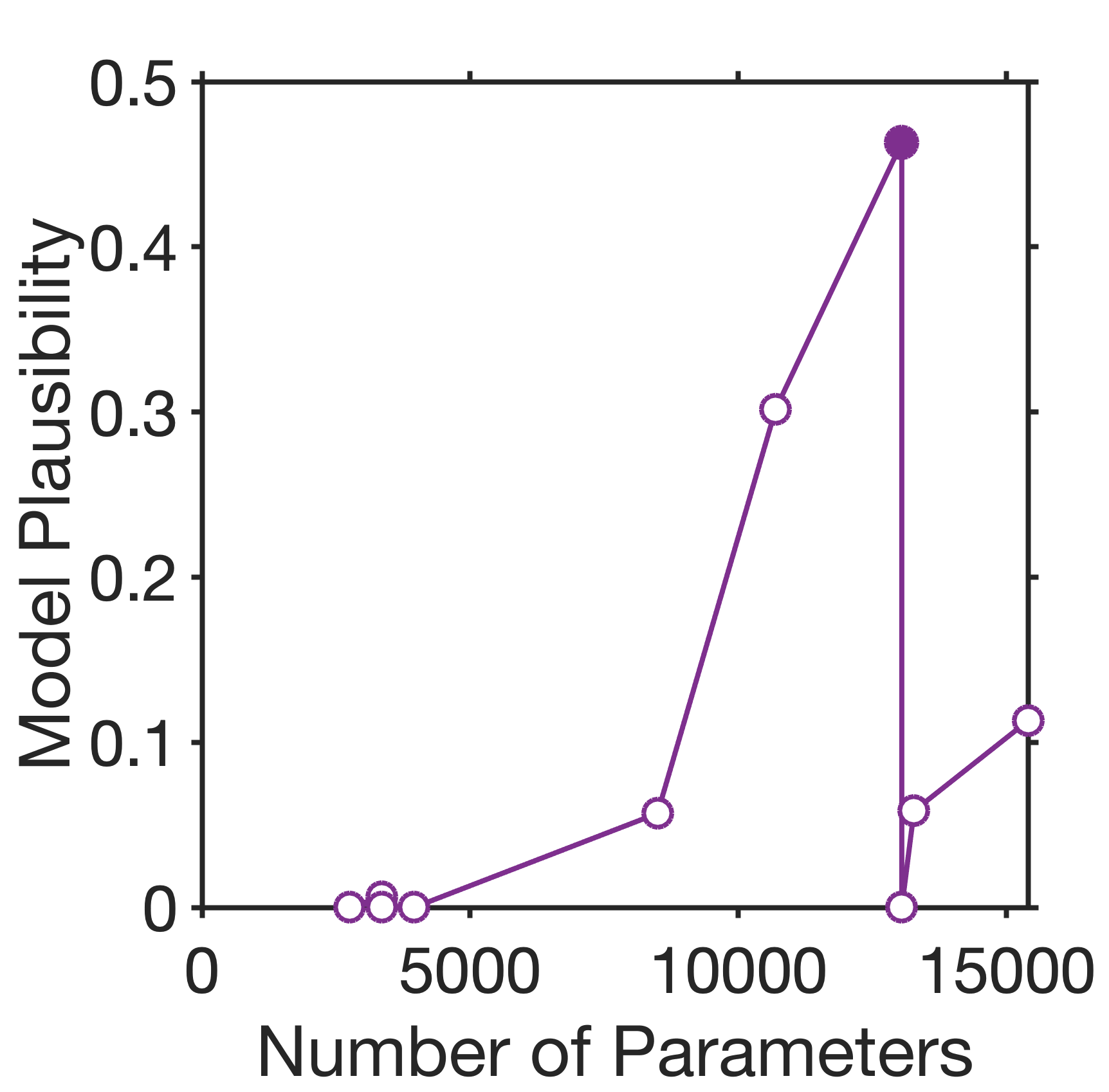}\hfill
\includegraphics[width=0.19\textwidth]{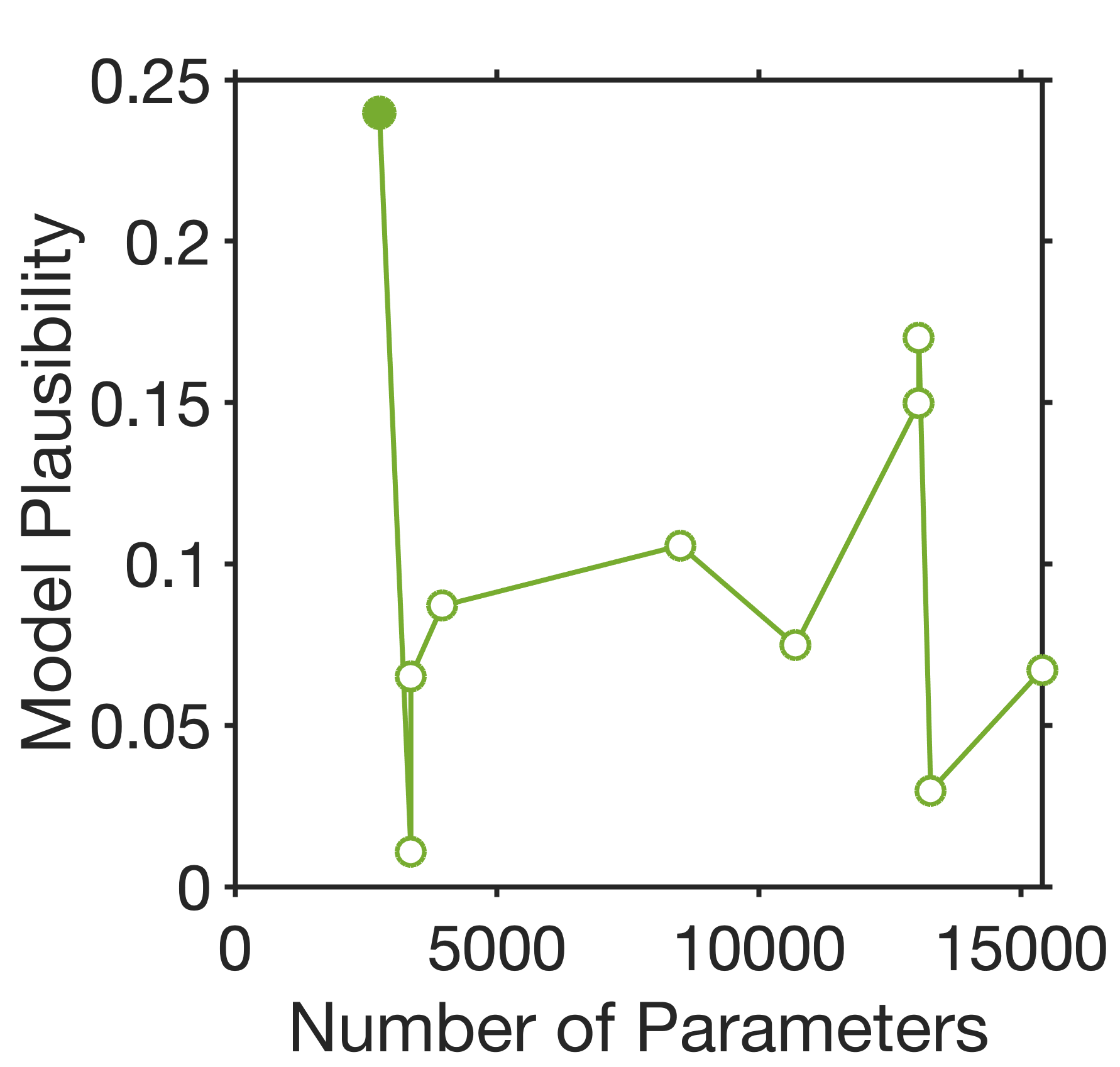}
\\(a)\\
%--- BG prediction - fale
\includegraphics[width=0.45\linewidth]{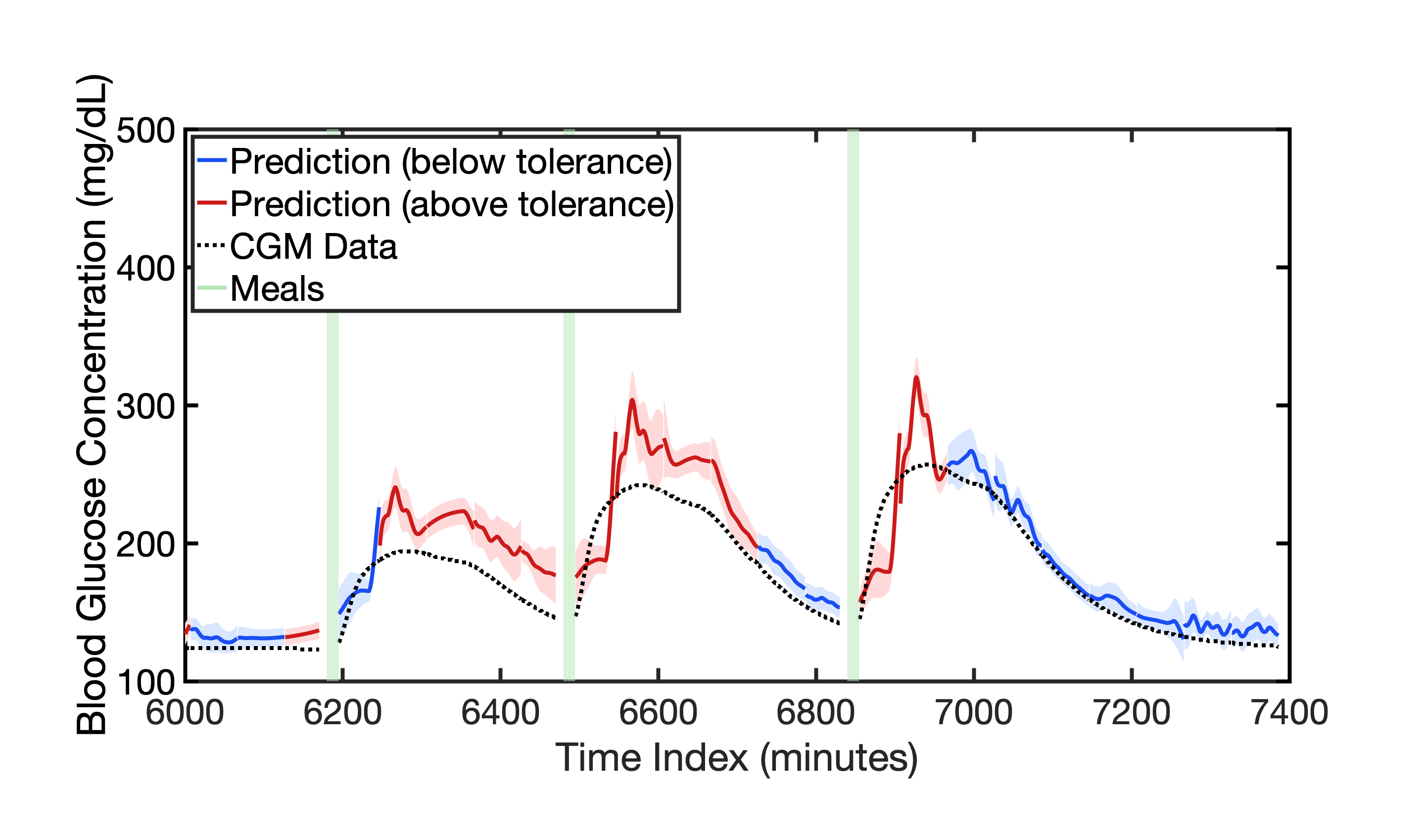}
\includegraphics[width=0.45\linewidth]{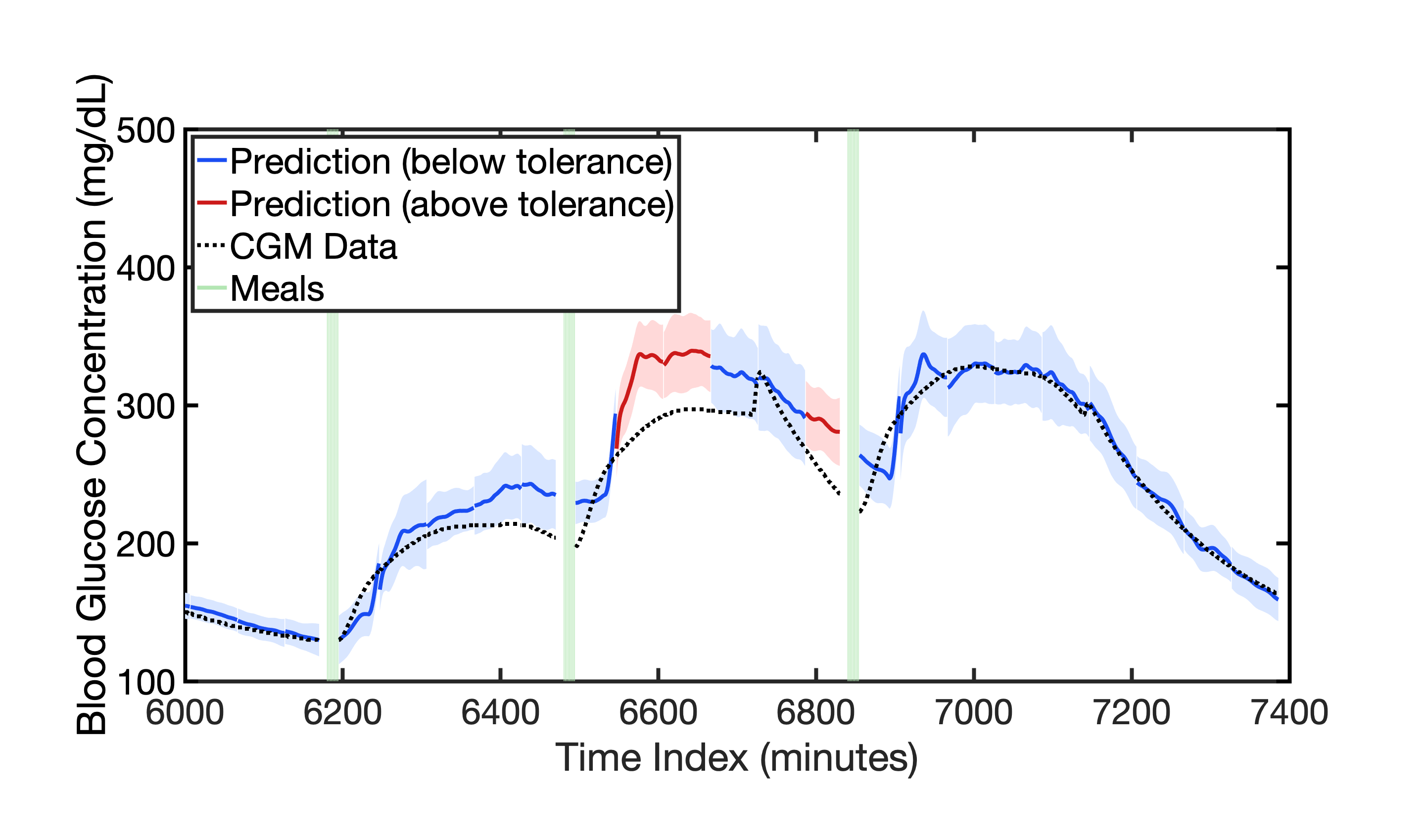}
\\(b) \hspace{2.5in} (c)\\
\caption{
Level~1 architecture ranking and validation outcome. 
(a) Posterior plausibility \(\rho\) versus model size (number of parameters) for Level~1 architectures across validation folds (left to right: folds 1–5). 
(b,c) Representative validation failure for most plausible architectures on held-out subjects: (b) Patient 2 from fold 3, Architecture: $Level_1Arch_4$, (c) Patient 3 from fold 4, Architecture: $Level_1Arch_8$.
Red segments indicate one-hour prediction windows that violates the acceptance criterion $NLPD \leq Tol$.
}
\label{fig:level1_plausibility_predict}
\end{figure}

\paragraph{Capacity Level 1 ($l=1$)}
The lowest-capacity TCN candidates with less than $3 \times 10^4$ parameters, that still satisfy the receptive-field feasibility constraints, are in Level~1.
As shown in Figure \ref{fig:level1_plausibility_predict}, model evidence is not uniformly distributed within this level and it concentrates on a small subset of architectures in each held-out validation fold. 
Despite this, \blue{the shortlisted architecture in Level~1 does not} satisfy the validation criterion. As illustrated in Figure~\ref{fig:level1_plausibility_predict}, predictions on held-out subjects frequently violate the NLPD tolerance across one-hour windows (shown in red). These failures indicate that, although some architectures are favored by the population-based training data, their representational capacity is insufficient to generalize to subject-specific glucose dynamics. Consequently, EVIDENT rejects Level~1 and proceeds to the next capacity level.

\begin{figure}[H]
\vspace{-0.1in}
\centering
\includegraphics[width=0.18\textwidth]{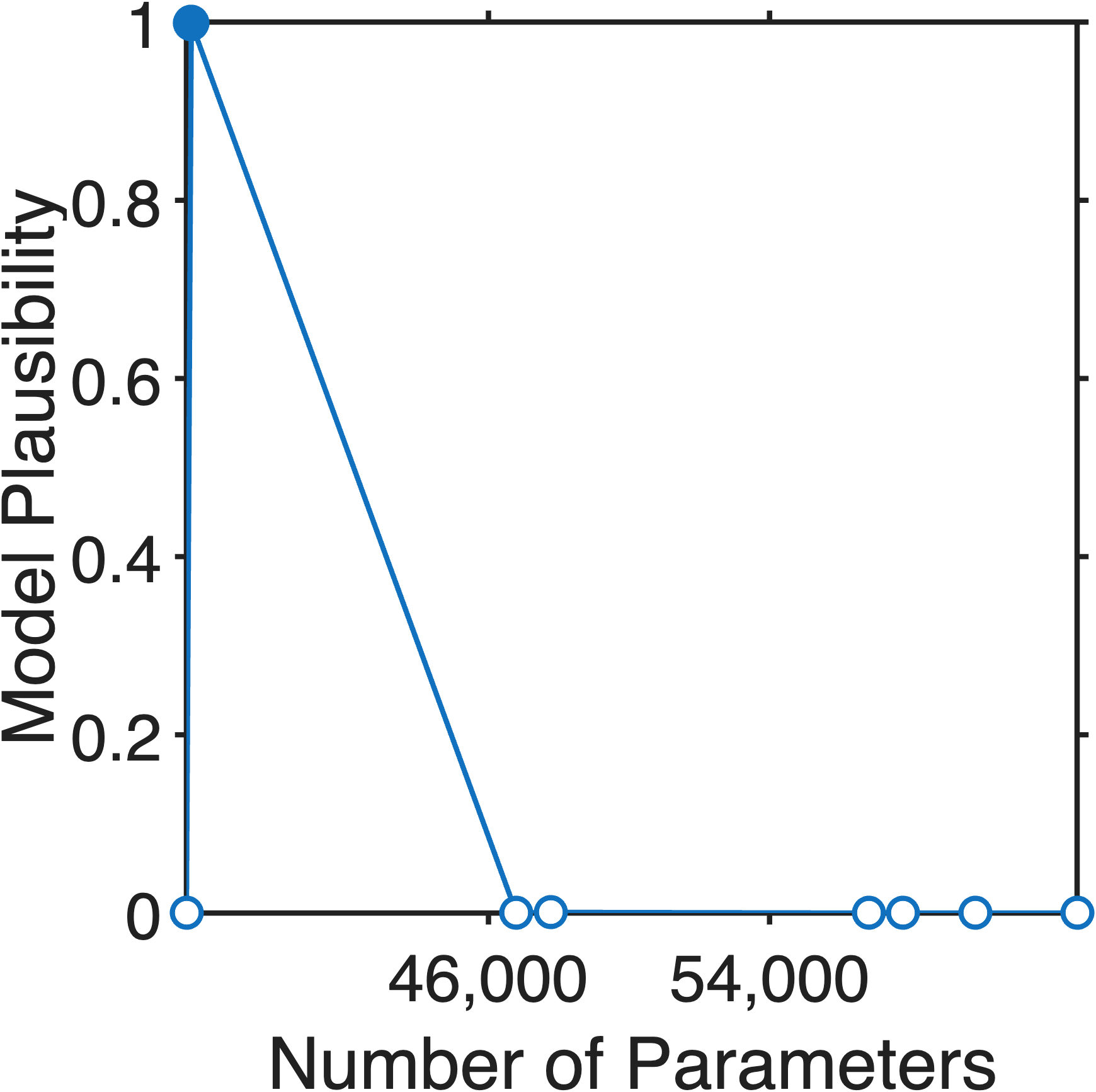}\hfill
\includegraphics[width=0.18\textwidth]{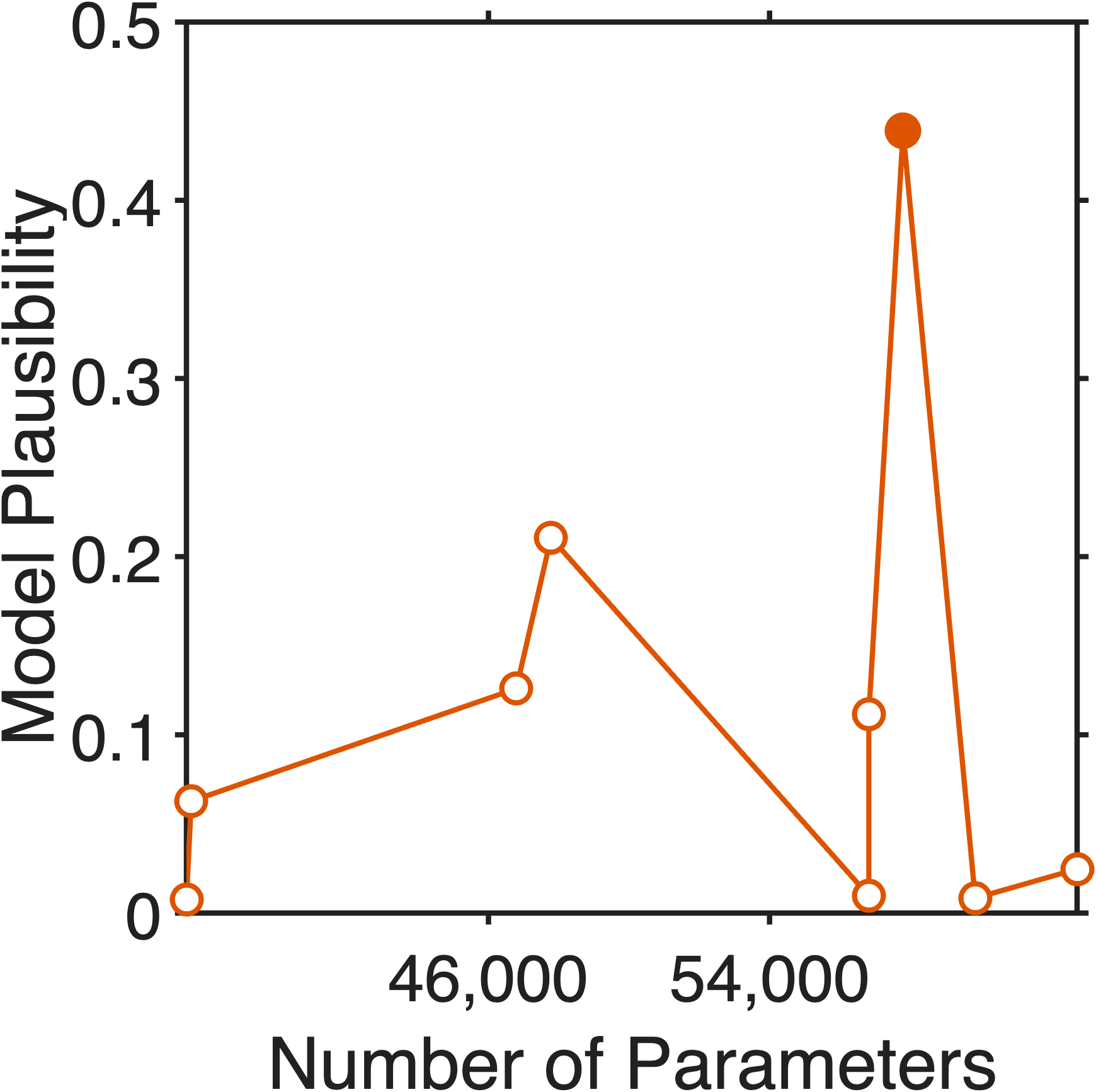}\hfill
\includegraphics[width=0.18\textwidth]{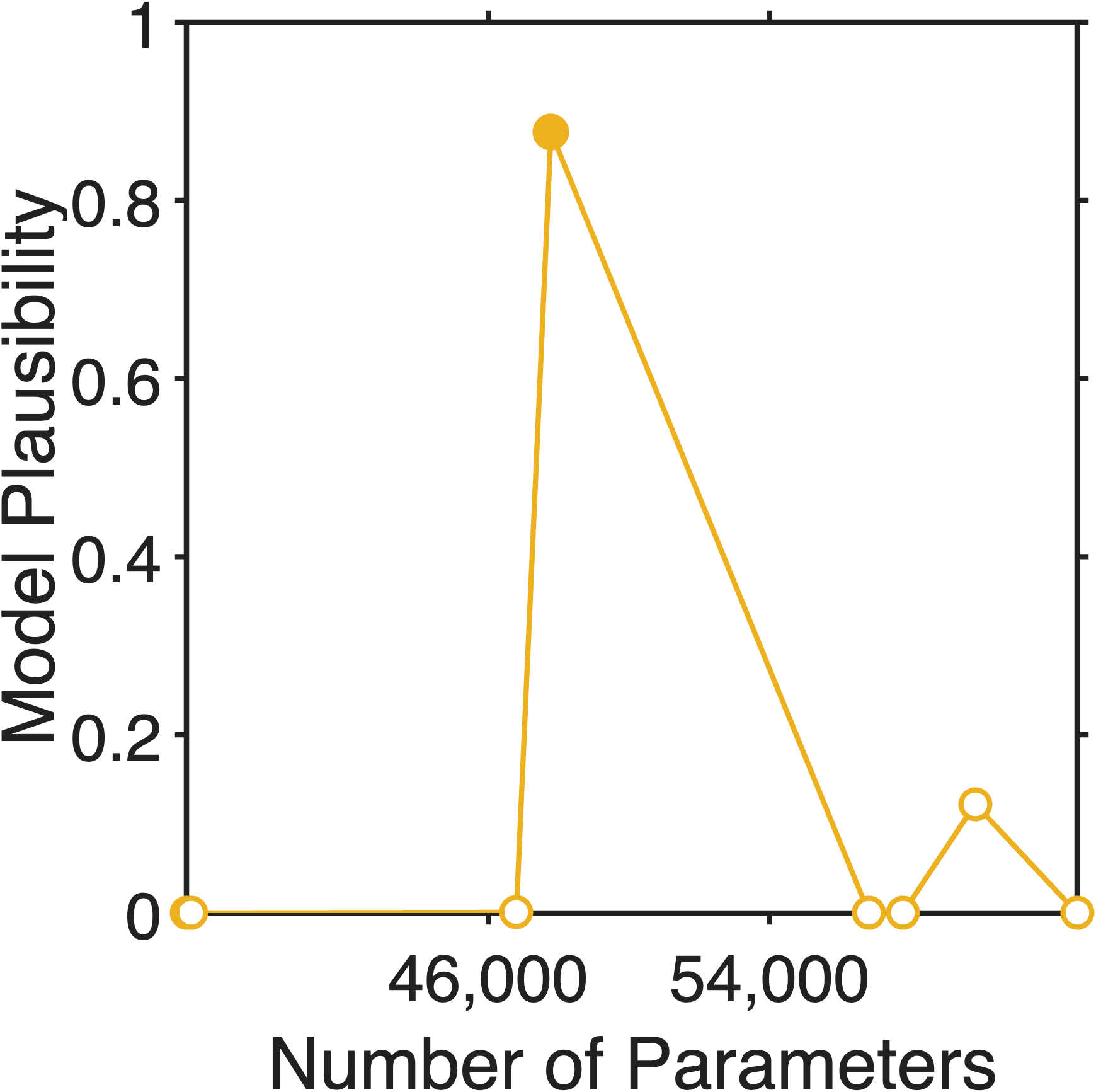}\hfill
\includegraphics[width=0.18\textwidth]{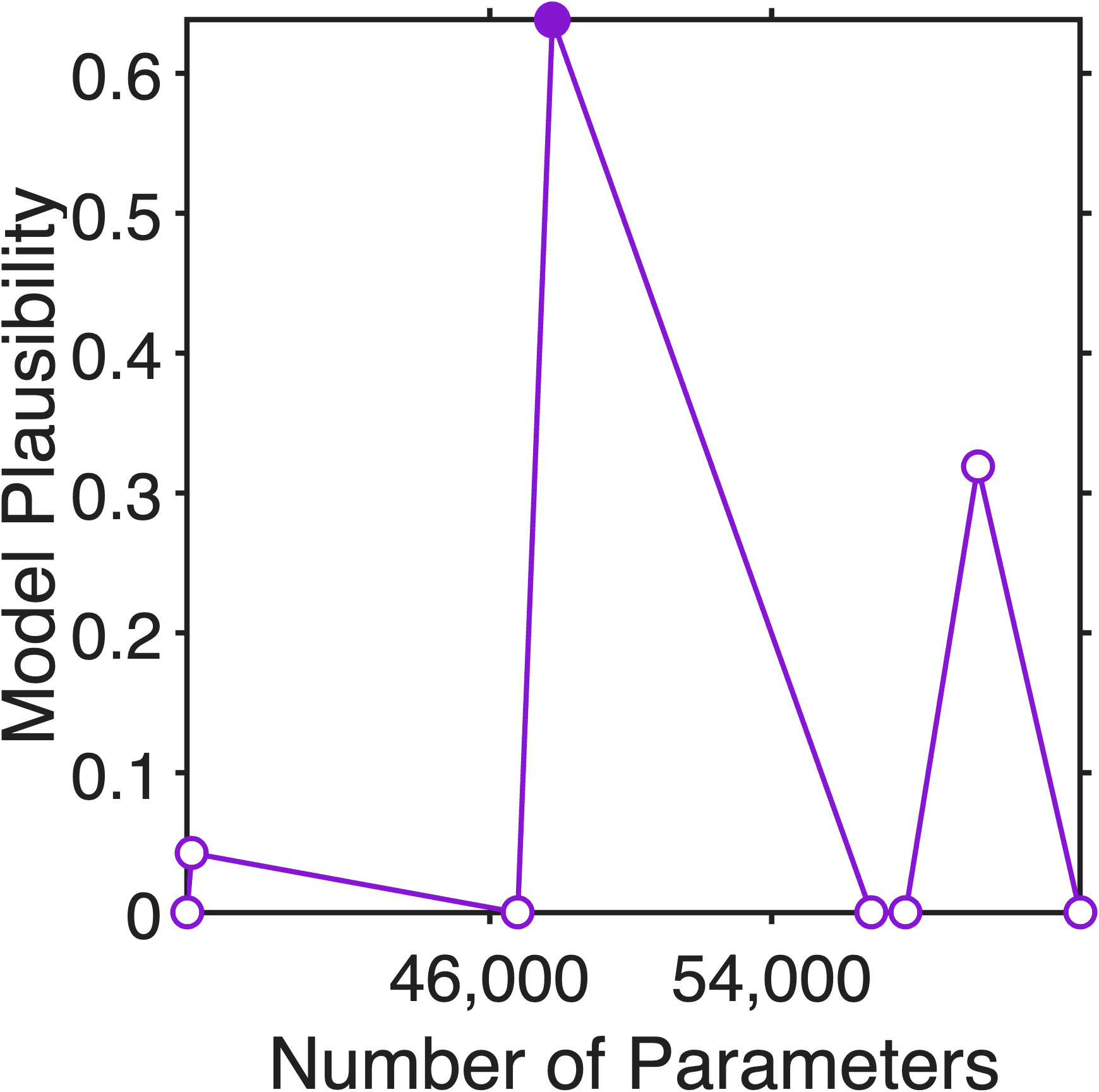}\hfill
\includegraphics[width=0.18\textwidth]{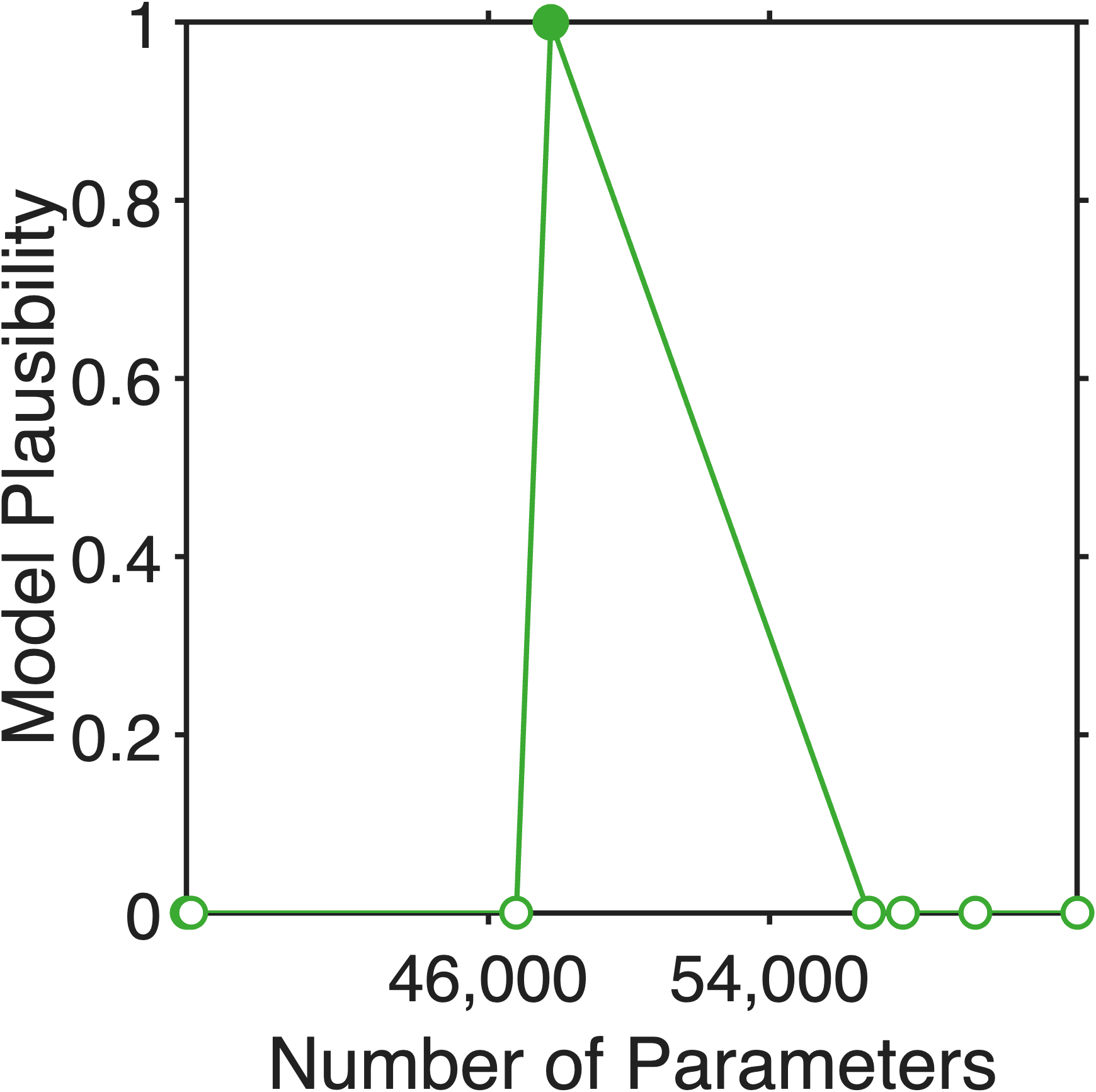}
\\
\includegraphics[width=0.20\textwidth]{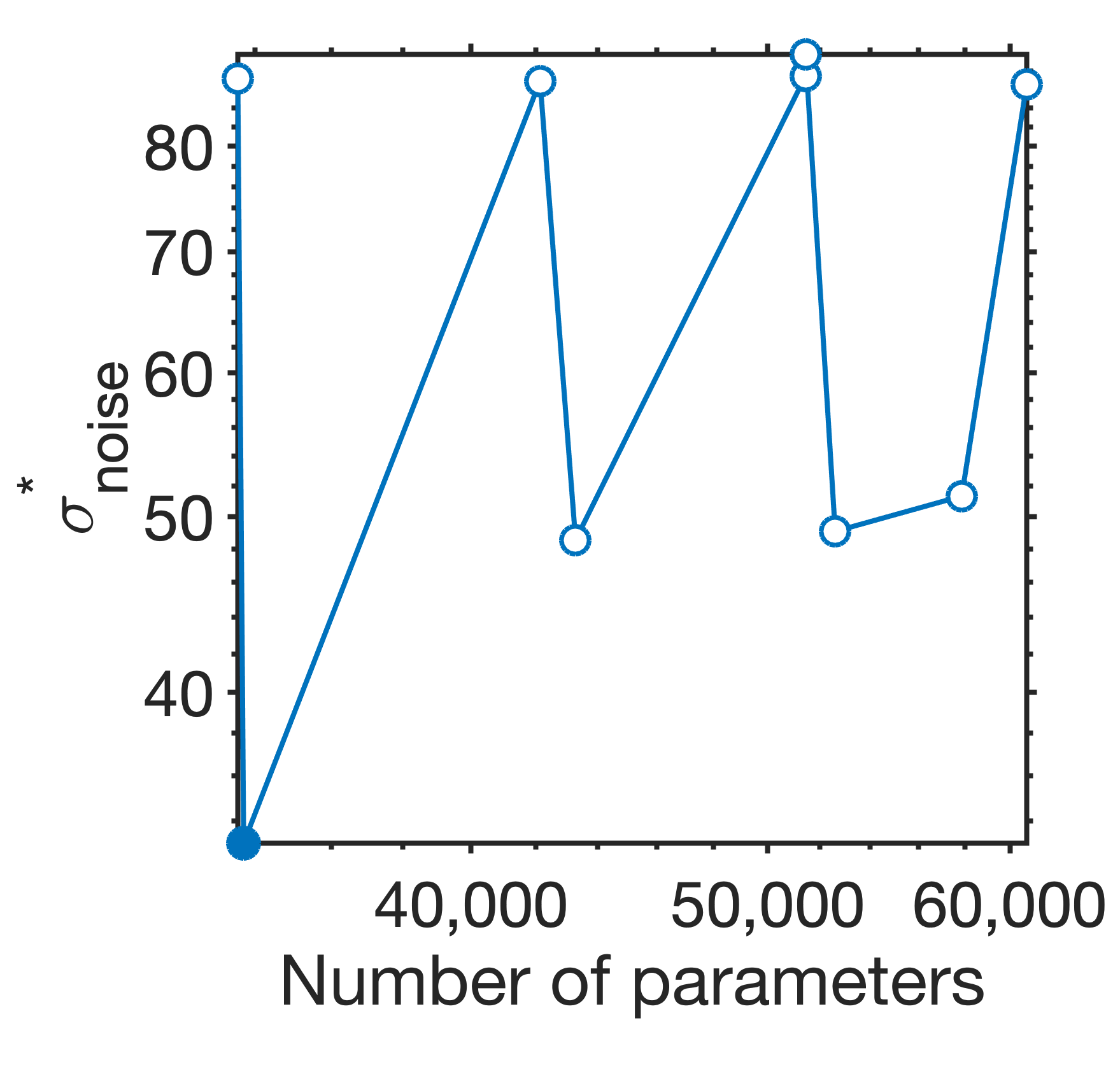}\hfill
\includegraphics[width=0.20\textwidth]{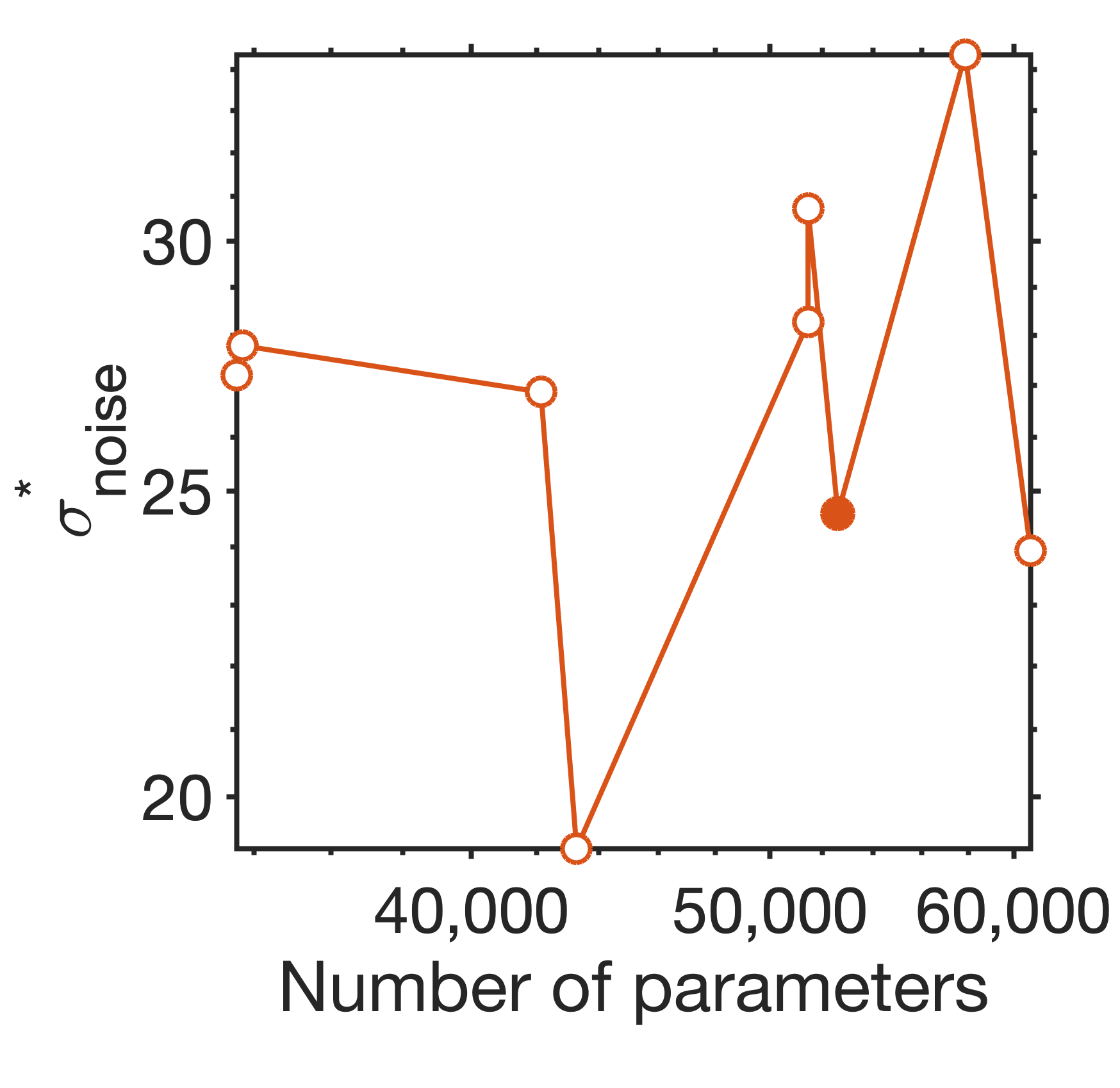}\hfill
\includegraphics[width=0.20\textwidth]{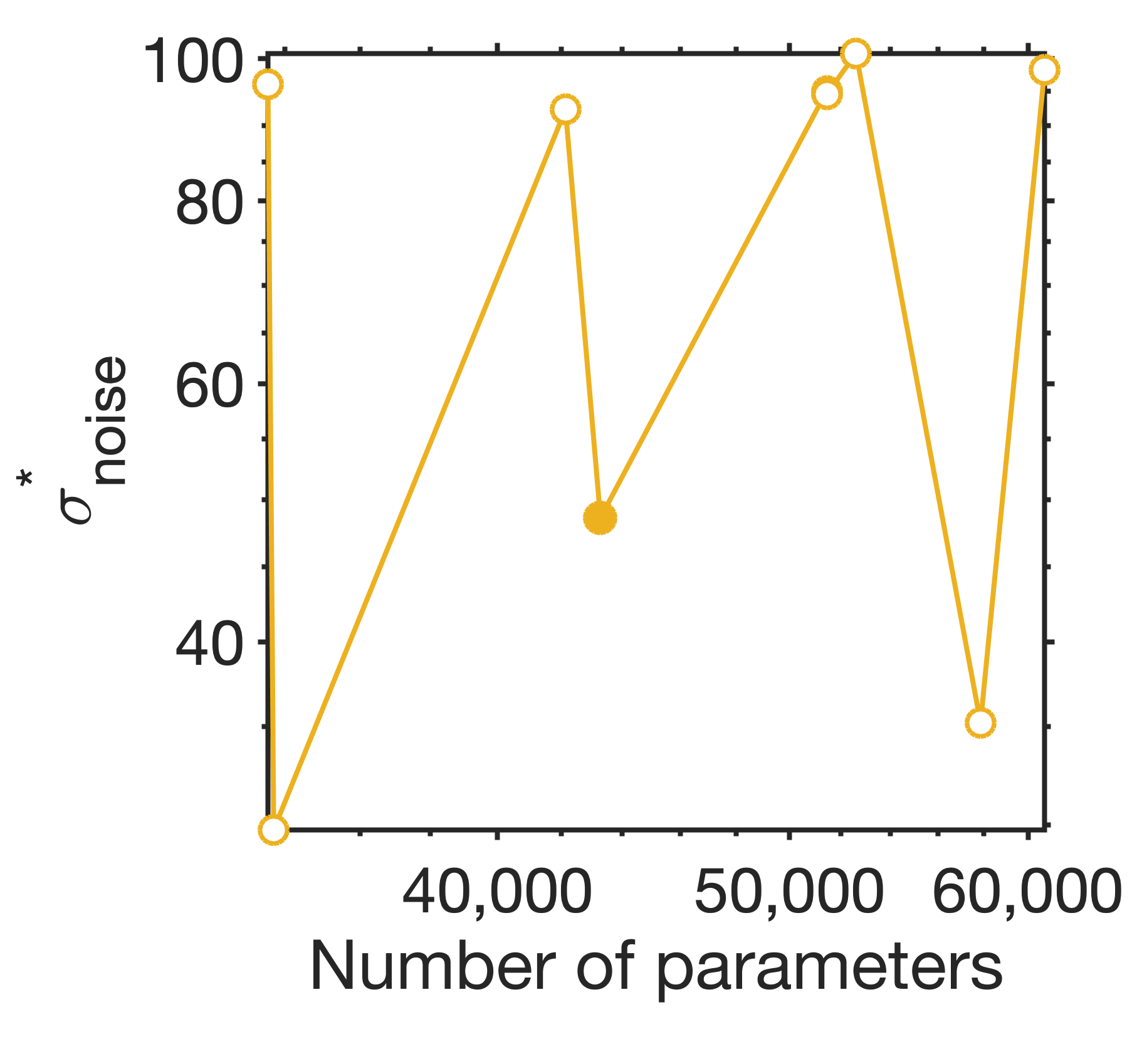}\hfill
\includegraphics[width=0.20\textwidth]{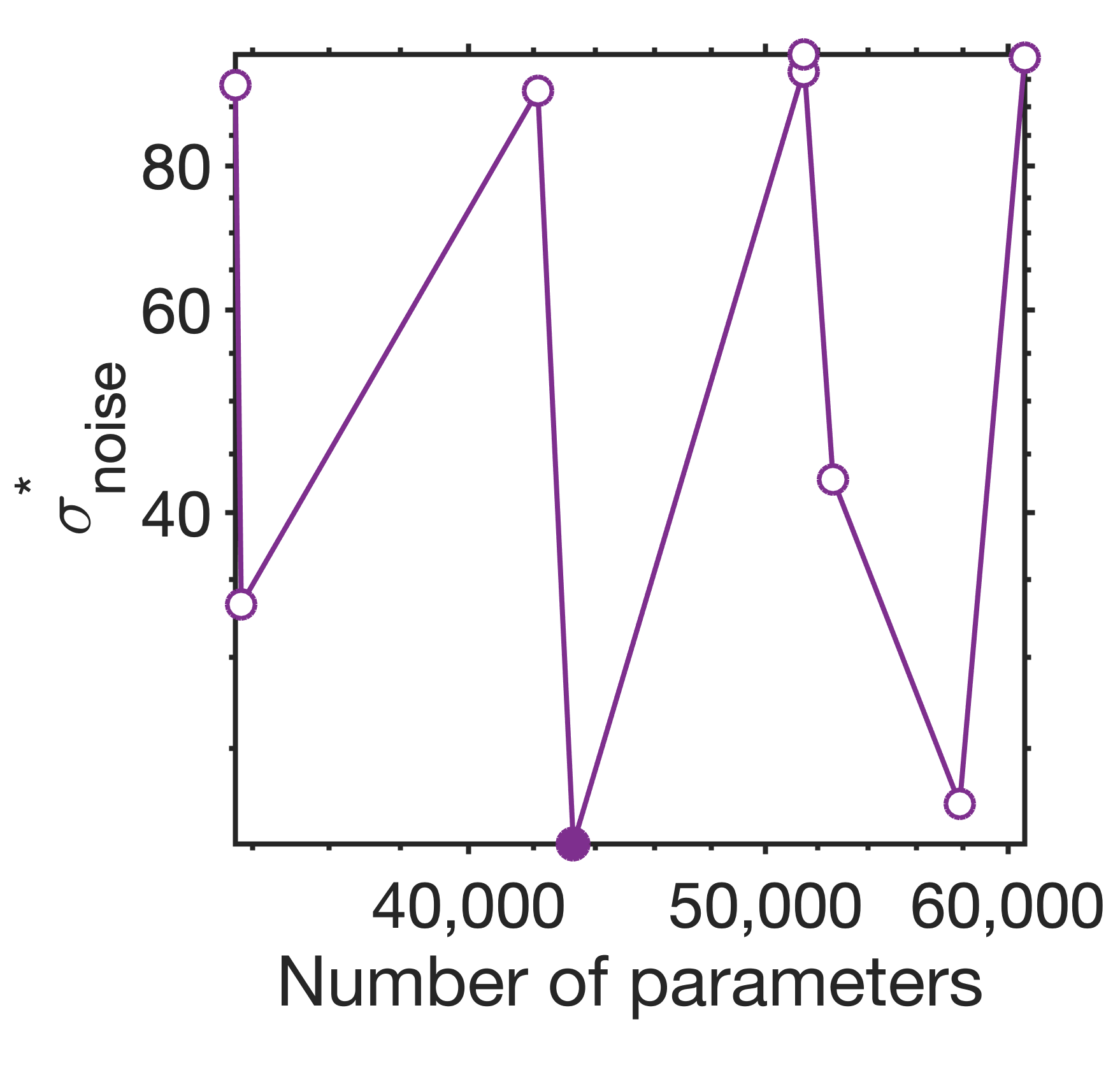}\hfill
\includegraphics[width=0.20\textwidth]{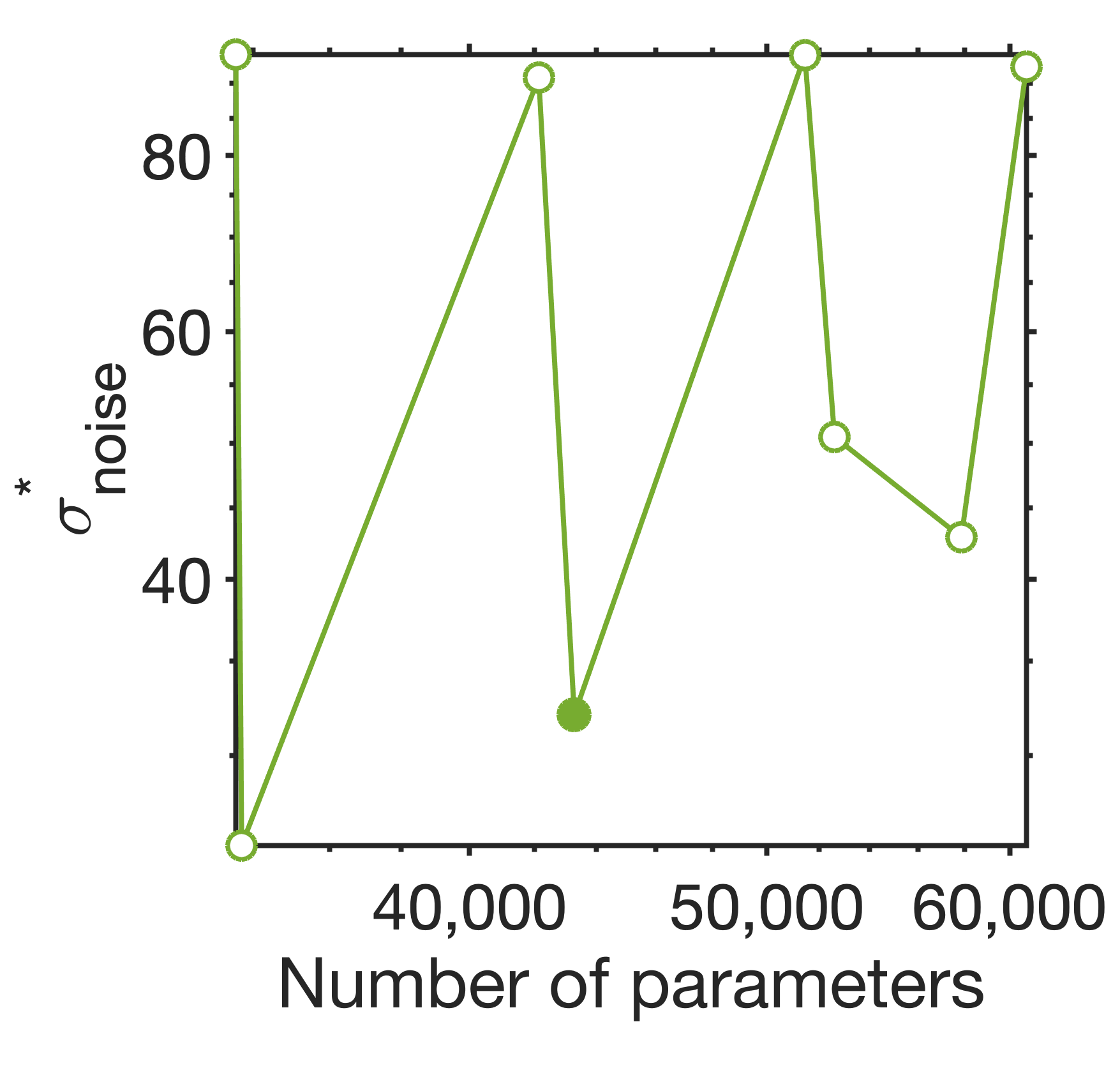}
\vspace{-0.1in}
\\ (a)\\
  \includegraphics[width=0.43\linewidth]{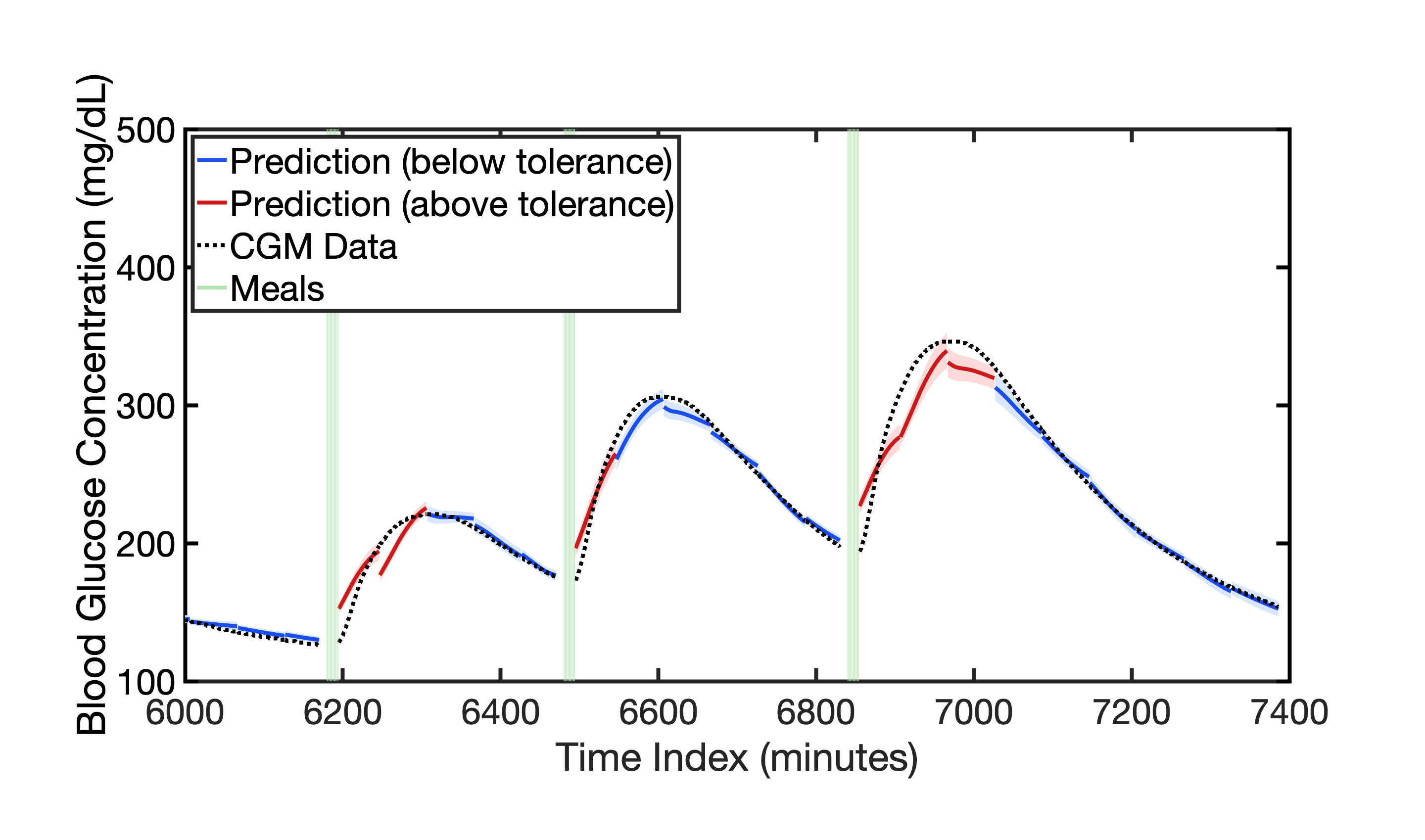}
~
  \includegraphics[width=0.43\linewidth]{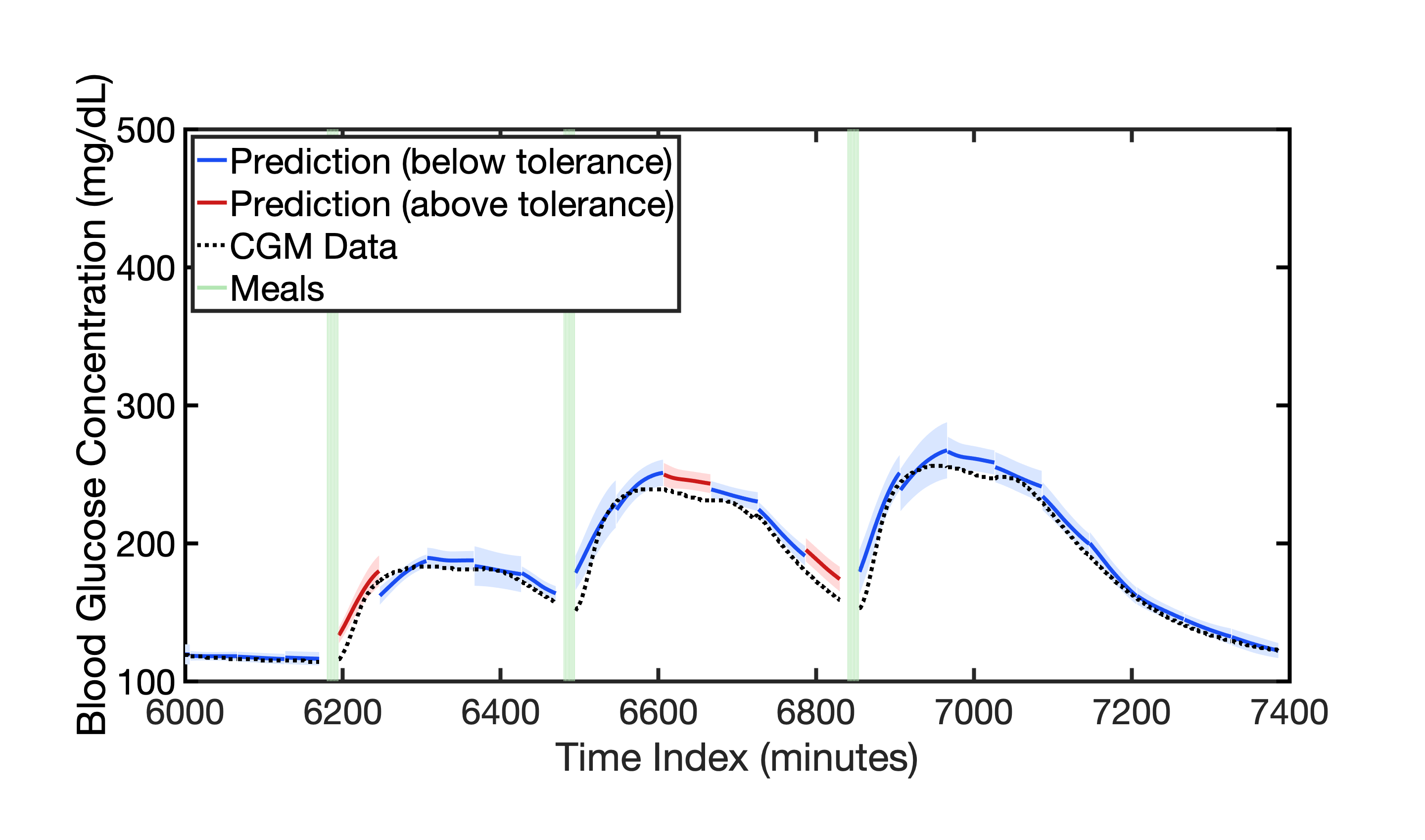}
  \vspace{-0.1in}
\\ (b)  \hspace{2.5in}  (c)\\
  \includegraphics[width=0.43\linewidth]{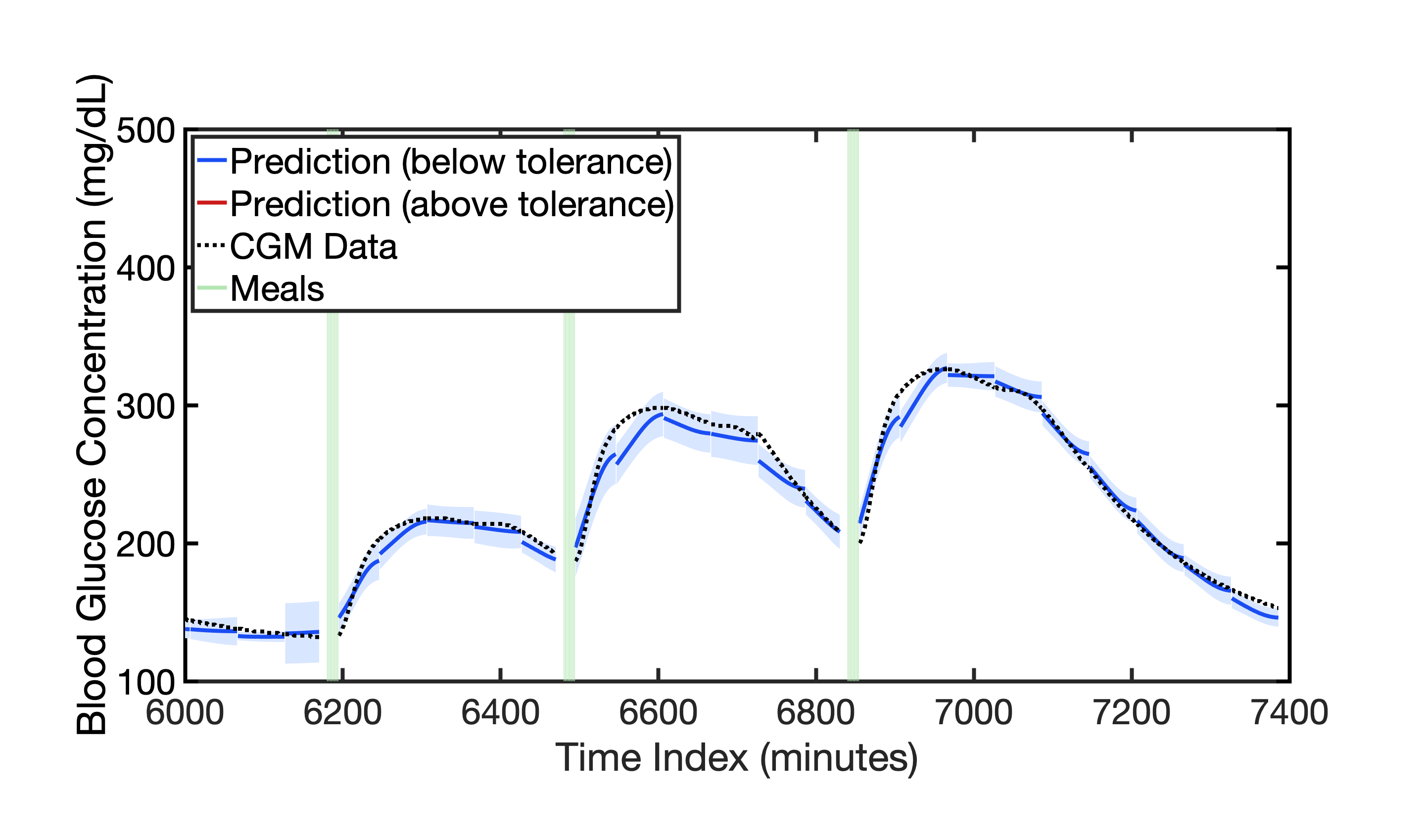}
  ~
  \includegraphics[width=0.43\linewidth]{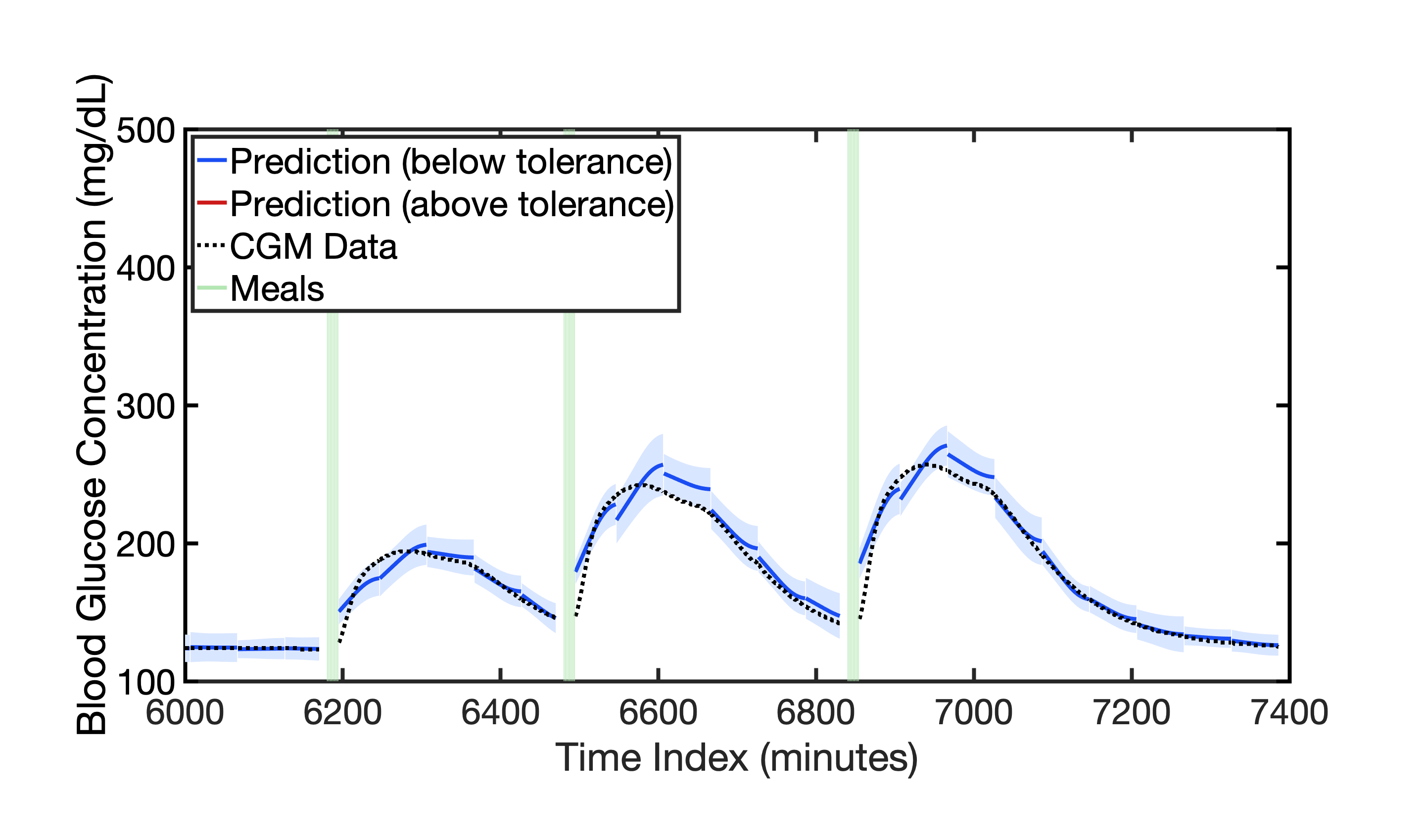}
  \vspace{-0.1in}
\\ (d)  \hspace{2.5in}  (e)\\
\vspace{-0.05in}
  \includegraphics[width=0.43\linewidth]{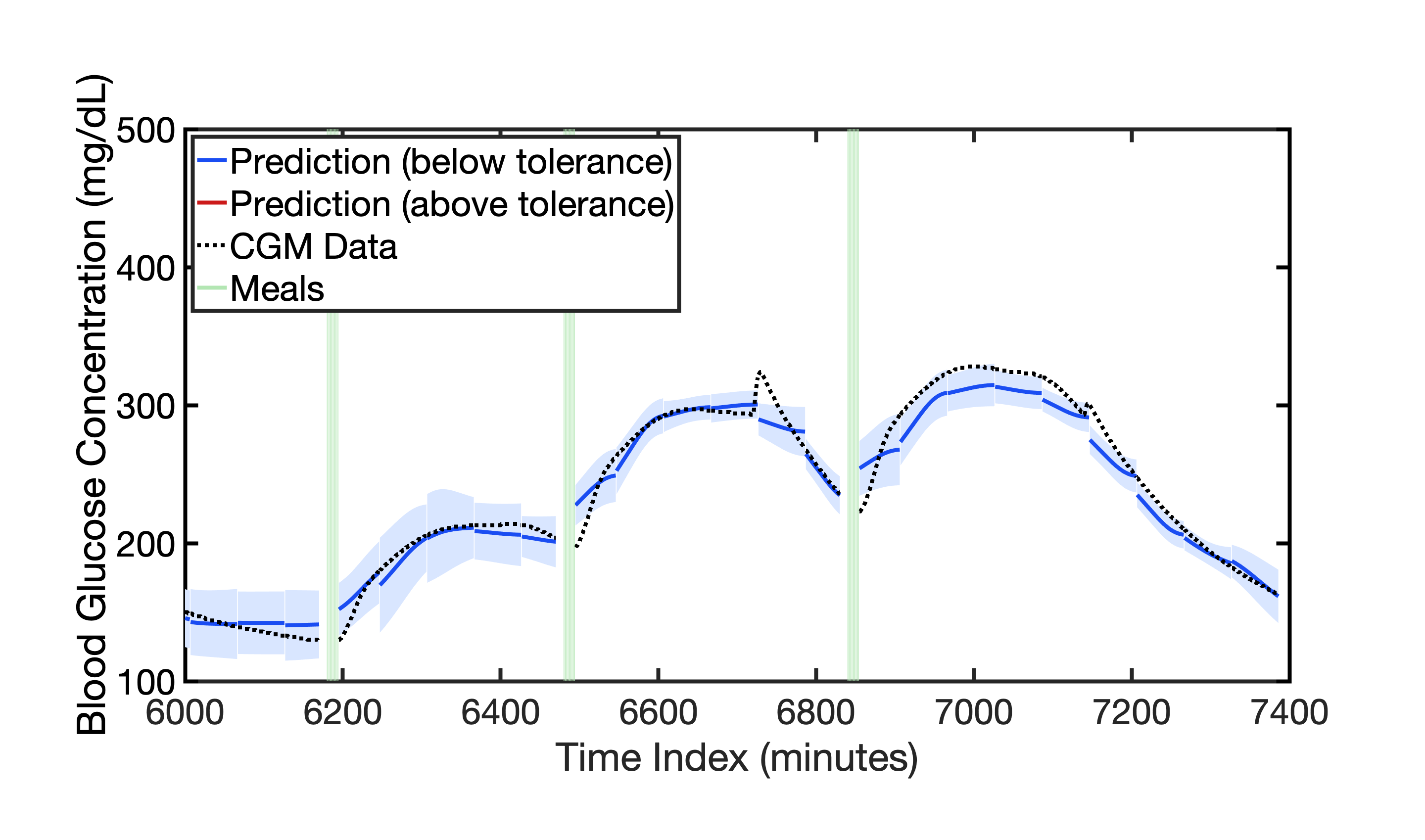}
  ~
  \includegraphics[width=0.43\linewidth]{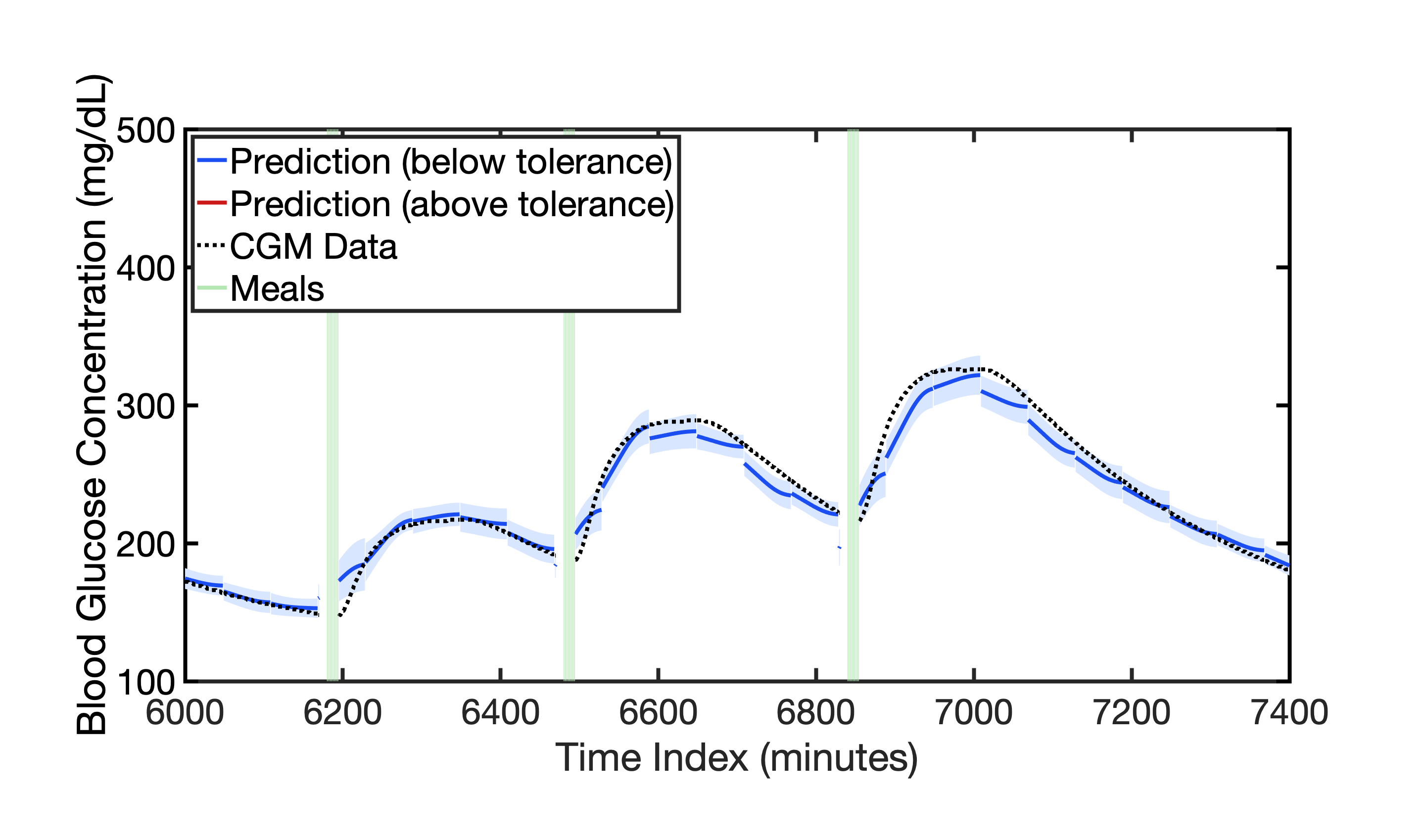}
  \vspace{-0.1in}
\\ (f)  \hspace{2.5in}  (g)
\vspace{-0.1in}
\caption{
Level~2 architecture ranking and validation outcomes. 
(a) Posterior plausibility \(\rho\) and MAP estimate of \(\sigma_{\text{noise}}\) as functions of model size (number of trainable parameters) for Level~2 architectures across validation folds (left to right: folds 1–5).
(b, c) Representative validation failures for most plausible architecture (\(Level_2Arch_2\)) on held-out patients 9 and 10 from Fold~1. Red segments indicate one-hour prediction windows that violate the acceptance criterion \(\mathrm{NLPD} \leq Tol\).
(d--g) Validation results for the EVIDENT-selected architecture (\(Level_2Arch_4\)) across multiple folds and held-out patients: (d,e) patients 1--2 and (f,g) patients 3--4. The one-hour forecasts of (\(Level_2Arch_4\)) passes all the validation tests, demonstrating EVIDENT \blue{accepted architecture}.
}
\label{fig:level2_plausibility_failure_sucess}
\end{figure}

\paragraph{Capacity Level~2 ($l=2$)}
Level~2 contains the intermediate-capacity regime ($3\times10^4 - 1\times10^5$ parameters) in which EVIDENT identifies architectures that are both plausible and validation-acceptable. 
%-- Level 2: plausibility/sigma-noise
Figure \ref{fig:level2_plausibility_failure_sucess}(a) summarizes, for each held-out validation fold, the variation of posterior plausibility $\rho$ and the corresponding MAP estimate of $\sigma_{\text{noise}}$ across the candidate architectures in this level. It is observed that a small subset of architectures receives substantial posterior model plausibility, while the most plausible architecture varies across folds, reflecting the inter-patient heterogeneity of the population-based training data in Figure \ref{fig:uva}. While most subjects exhibit comparable glucose ranges and excursion patterns, Patient 7 and Patient 5 display substantially larger amplitudes, which changes the temporal structure seen during training. As a result, EVIDENT favors different architectures across folds. Nevertheless, $Level_2Arch_4$ with 47,779 parameters, which consists of 5 temporal blocks with dilation base $b=3$, filter size $f=9$, and encoder channels $[2,4,8,16,32]$, emerges as a competitive architecture in three validation folds. This architecture achieves a one-day receptive field through gradual dilation growth across depth, while retaining sufficient local temporal resolution to capture meal-driven glucose excursions.
The $\sigma_{{\rm noise}_{MAP}}$ trends provide a diagnostic such that low-plausibility architectures require inflated noise levels, indicating that unresolved temporal structure is absorbed as noise, whereas competitive models achieve high plausibility without such inflation.

% Level 2 plot -- plausibility, representative failur, sucess
%-- Level 2: validation
The top-ranked architectures in each fold are subsequently evaluated on the held-out subjects using the validation protocol. Although Level~2 architectures exhibit improved predictive performance relative to Level~1, most candidates still fail the validation criterion by violating the NLPD tolerance in a subset of one-hour prediction windows (representative failure cases are shown in Figure~\ref{fig:level2_plausibility_failure_sucess}(b,c)).
EVIDENT, however, identifies a single architecture, \(Level_2Arch_4\), that satisfies the validation criterion across all folds and held-out subjects. As shown in Figure~\ref{fig:level2_plausibility_failure_sucess}(d--g), this model produces consistent one-hour forecasts that remain within the prescribed tolerance across all validation windows. This architecture is therefore accepted candidate architecture by EVIDENT.

\begin{figure}[!ht]
\centering
  \includegraphics[width=0.45\linewidth]{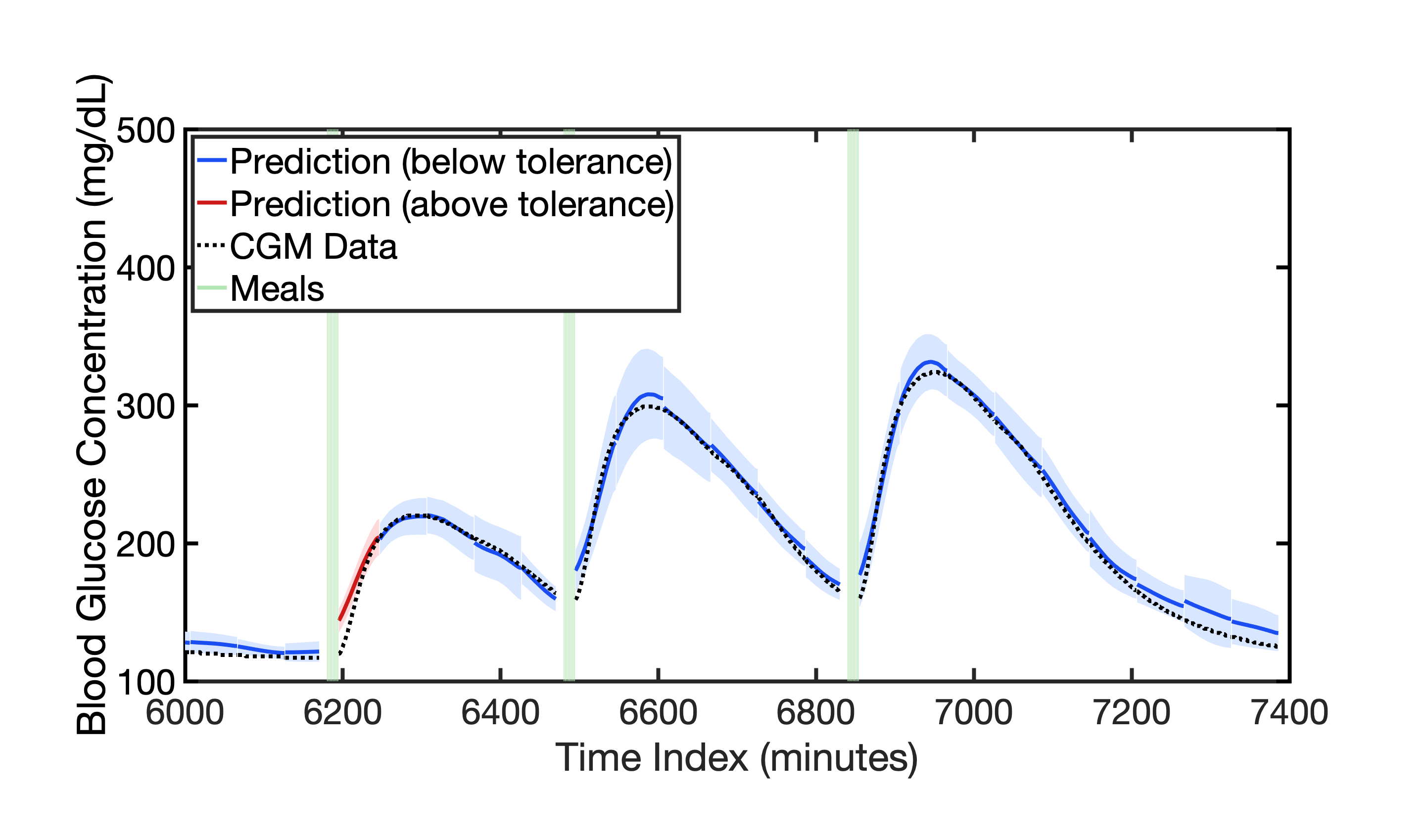}
    ~
  \includegraphics[width=0.45\linewidth]{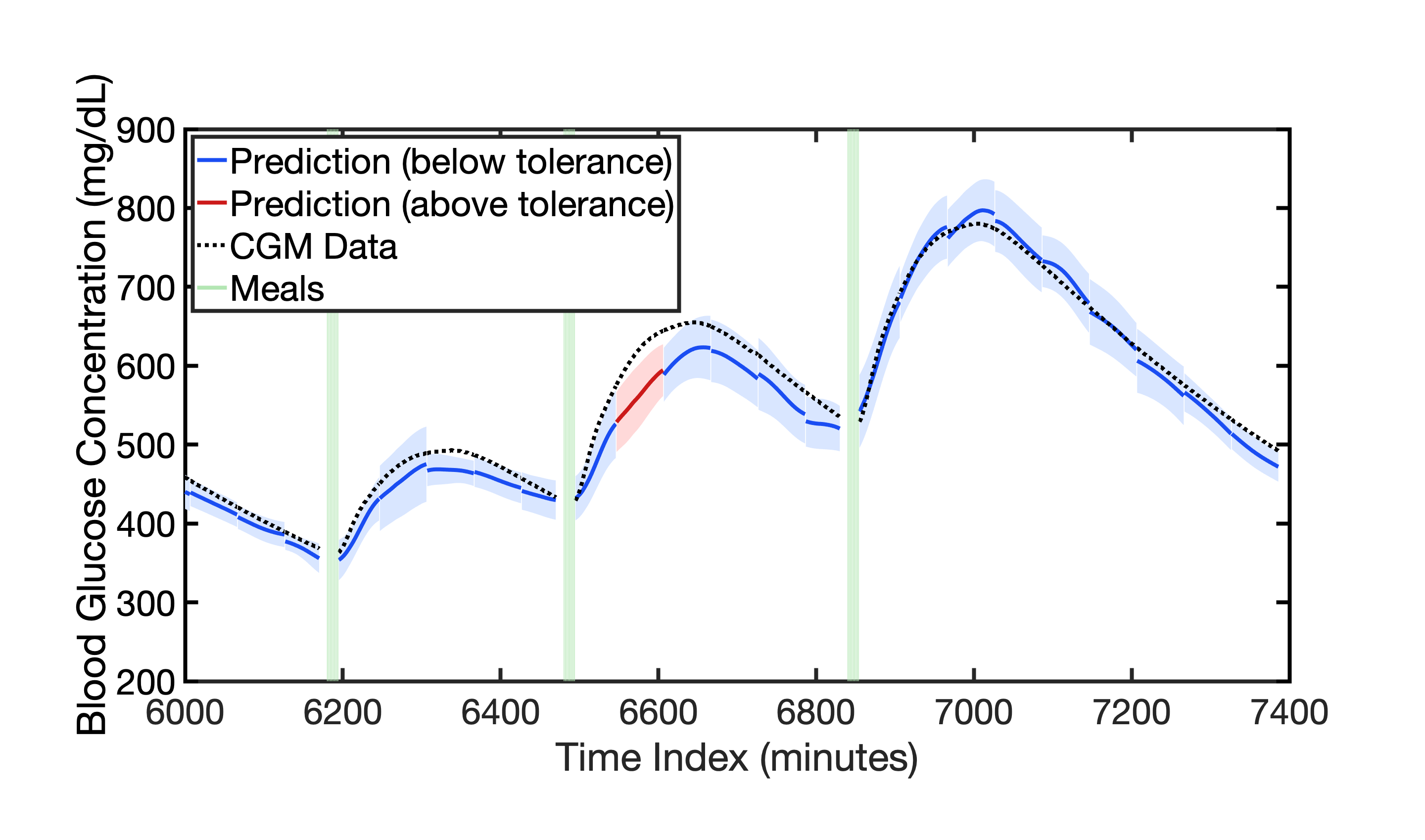}
  \vspace{-0.1in}
  \\ (a) \hspace{2.5in} (b) \\
  \includegraphics[width=0.45\linewidth]{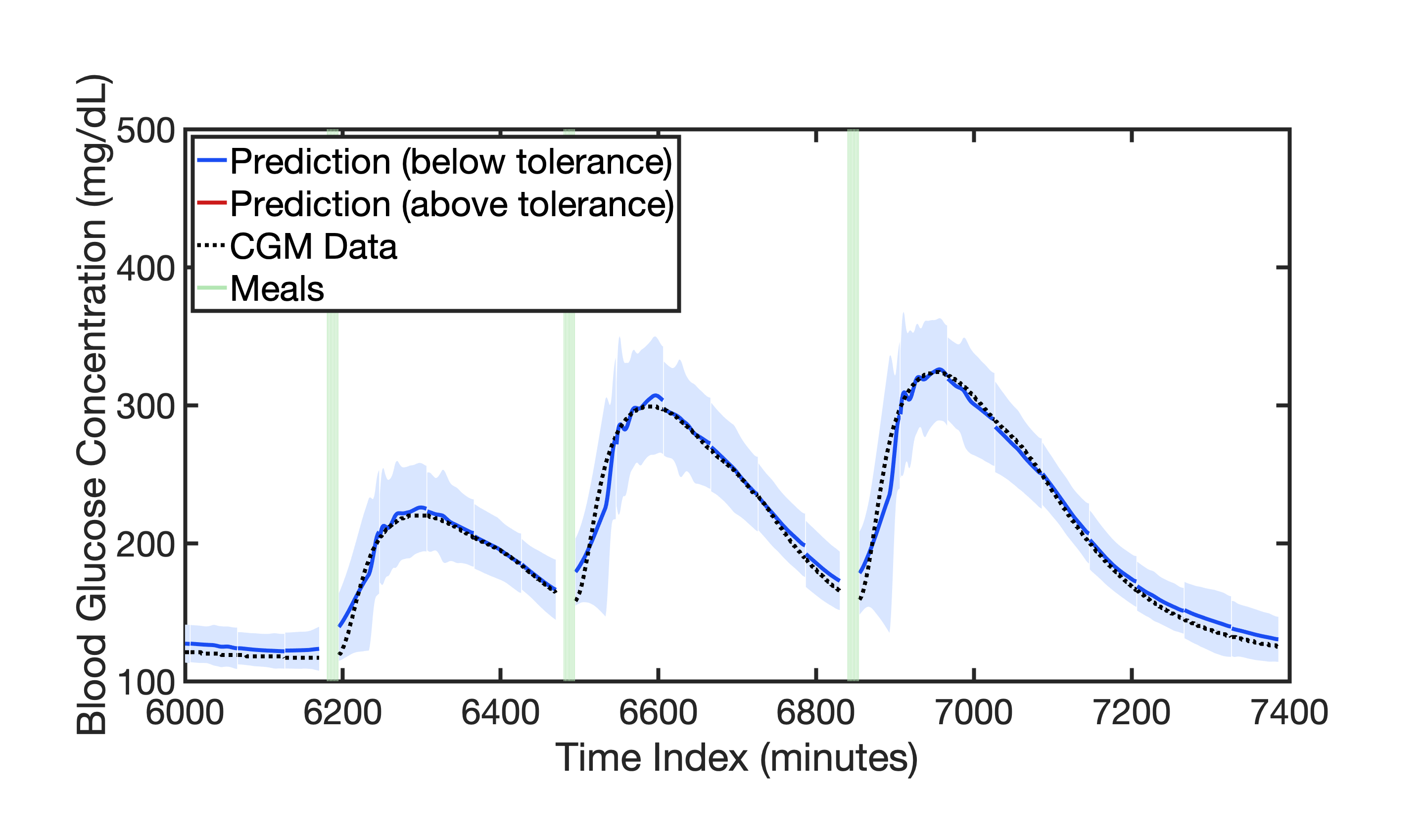}
  ~
  \includegraphics[width=0.45\linewidth]{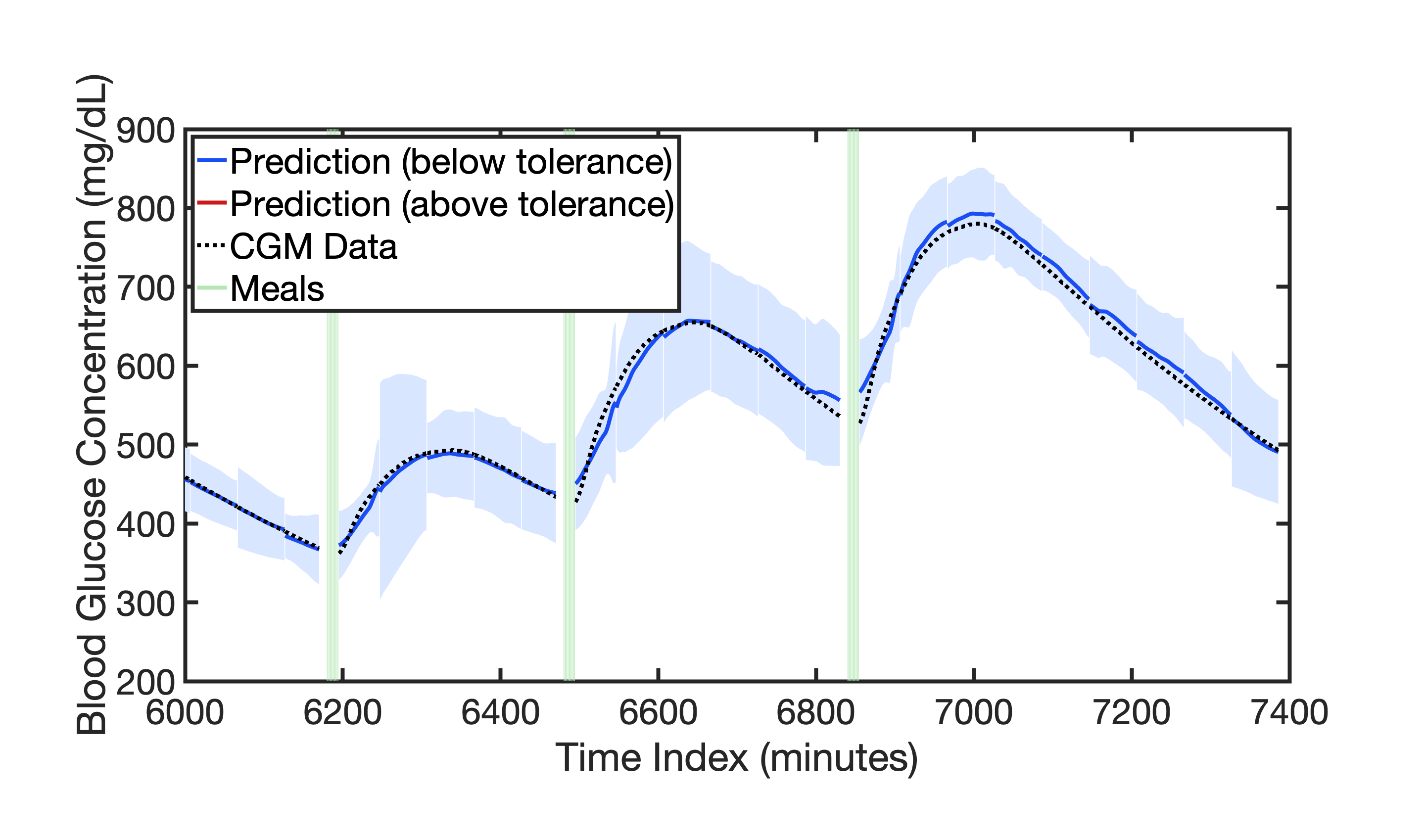}
  \vspace{-0.1in}
  \\ (c) \hspace{2.5in} (d)
  \vspace{-0.1in}
\caption{Comparison of single-architecture and plausibility-weighted ensemble predictions for representative Level~2 candidates in fold~2. 
(a–-b) Predictions from a single architecture ($Level_2Arch_7$) for Patients 6–7 in fold~2, and 
(c–-d) corresponding ensemble predictions from all architectures in Level~2.
Red segments indicate one-hour prediction windows that violates the acceptance criterion $NLPD \leq Tol$.}
\label{fig:bma}
\end{figure}

% -- Level 2: BMA
In addition to single-architecture selection, EVIDENT also supports a plausibility-weighted ensemble predictor when multiple architectures remain competitive. To assess this, we compare the predictions of the most plausible architecture with the ensemble prediction defined in \eqref{eq:bma} over all Level~2 architectures for fold~2 (Figure~\ref{fig:bma}).
\blue{In this fold, the plausibility-weighted ensemble achieves
lower prediction error and NLPD than the individual predictors
considered. Its wider uncertainty bands also reflect variation
between architecture specific predictions.}
These results indicate that, although a single architecture can satisfy the validation criterion, the plausibility-weighted ensemble provides a mechanism for improving predictive performance while capturing model uncertainty in the presence of multiple candidates.

\begin{figure}[!ht]
\centering
  \includegraphics[width=0.45\linewidth]{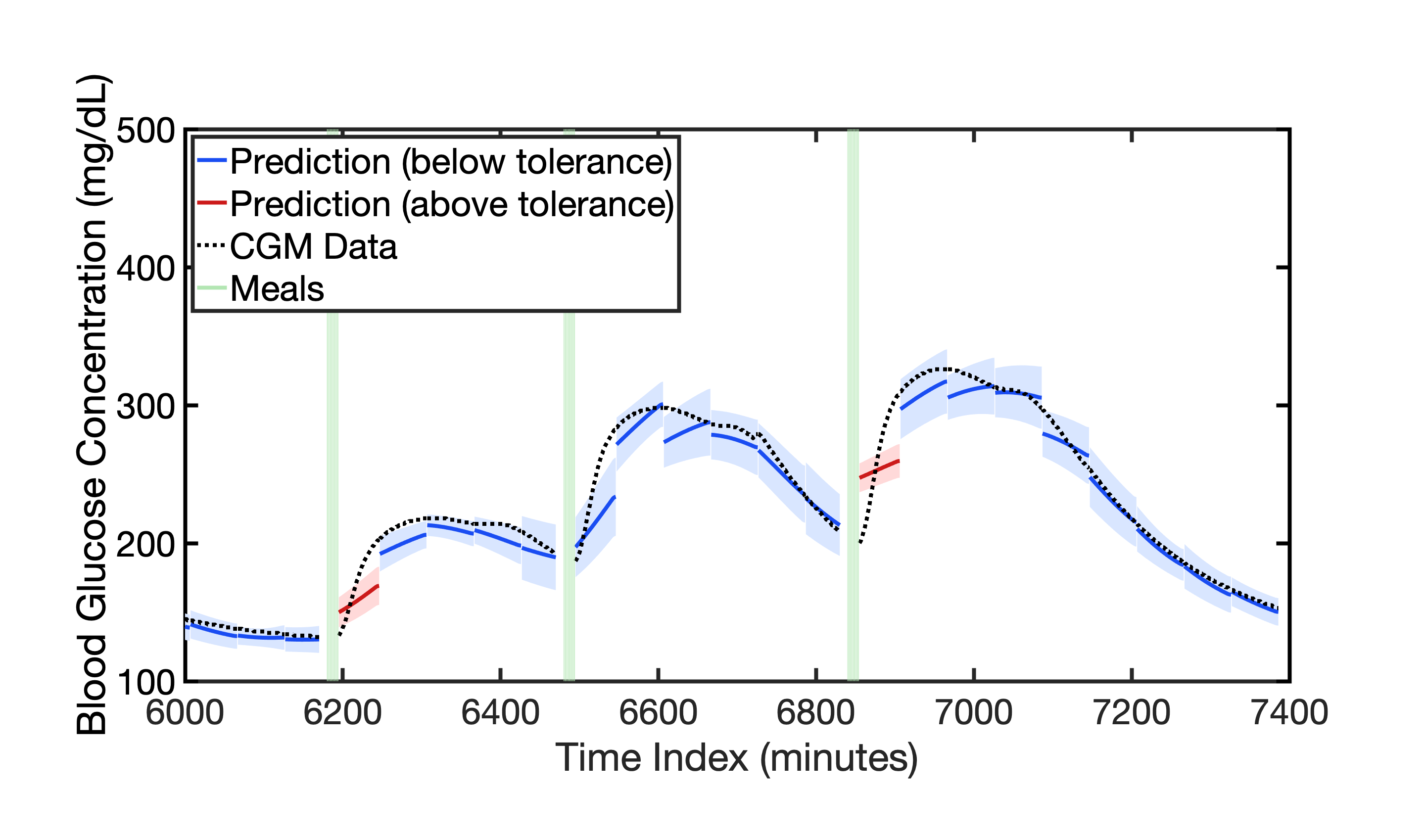}
  ~
  \includegraphics[width=0.45\linewidth]{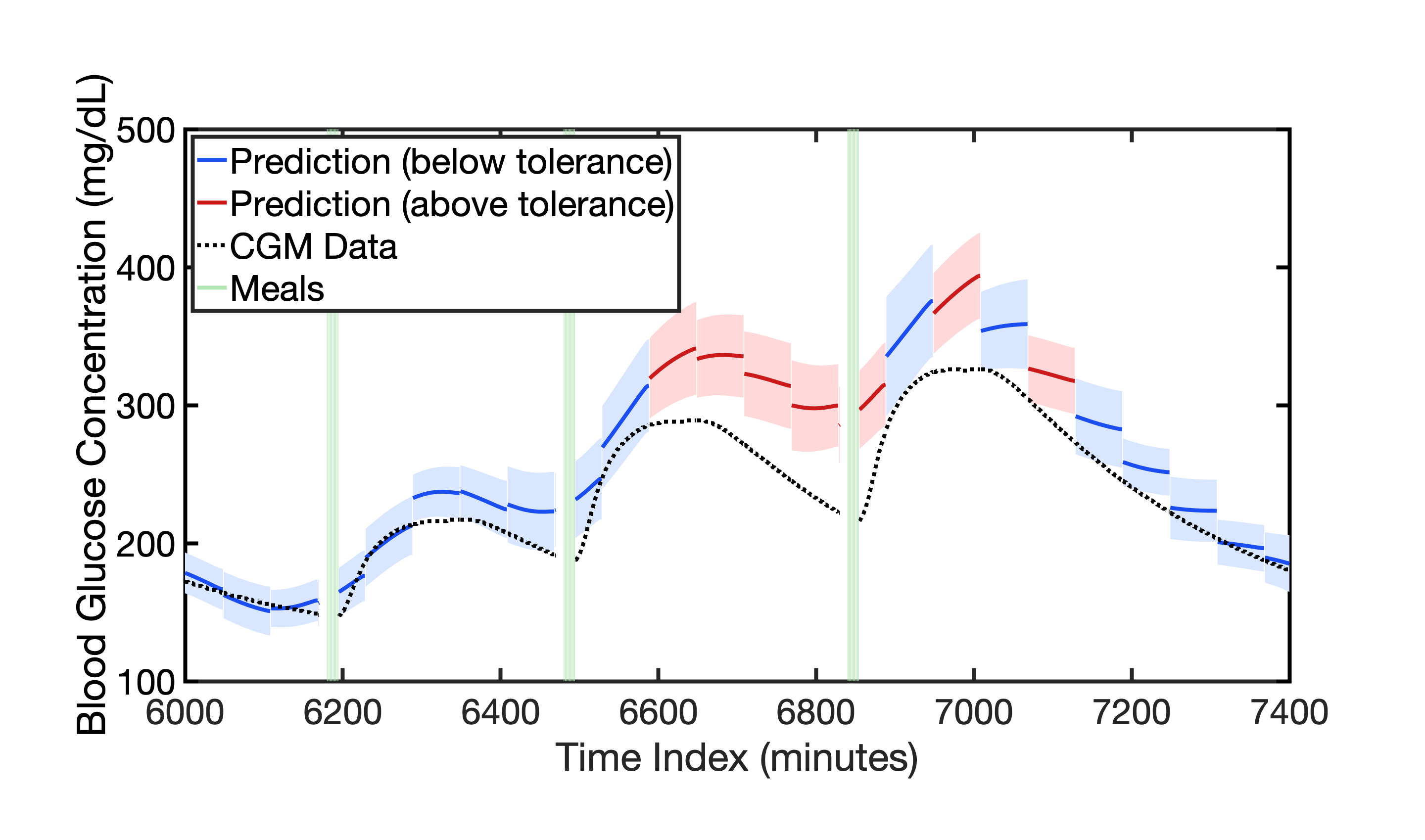}
  \vspace{-0.1in}
  \\ (a) \hspace{2.5in} (b)
  \vspace{-0.1in}
\caption{Representative validation failures for Level~3 architectures on held-out patients. (a) Patient 1 from fold~3 using $Level_3Arch_{10}$ and (b) Patient 4 from fold~4 using $Level_3Arch_4$. Red segments indicate one-hour prediction windows that violate the acceptance criterion $NLPD \leq Tol$.}
\label{fig:pred_cat3}
\end{figure}

\paragraph{Capacity level 3 ($l=3$)\blue{: additional diagnostic evaluations beyond the stopping point}}
Level~3 contains higher-capacity architectures ($10^5$–$4\times10^5$ parameters). Despite their increased representational capacity, these models do not improve validation performance and frequently violate the NLPD criterion across multiple prediction windows (Figure~\ref{fig:pred_cat3}).
The additional parameters at this level are not sufficiently learned by the available data, leading to degraded generalization under inter-patient variability. In contrast to Level~2, where intermediate-capacity architectures satisfy the validation criterion, increasing capacity beyond this regime yields no consistent improvement and introduces  prediction errors.

\begin{figure}[!ht]
\centering
\begin{minipage}[t]{0.5\textwidth}
  \centering
  \includegraphics[width=0.75\linewidth]{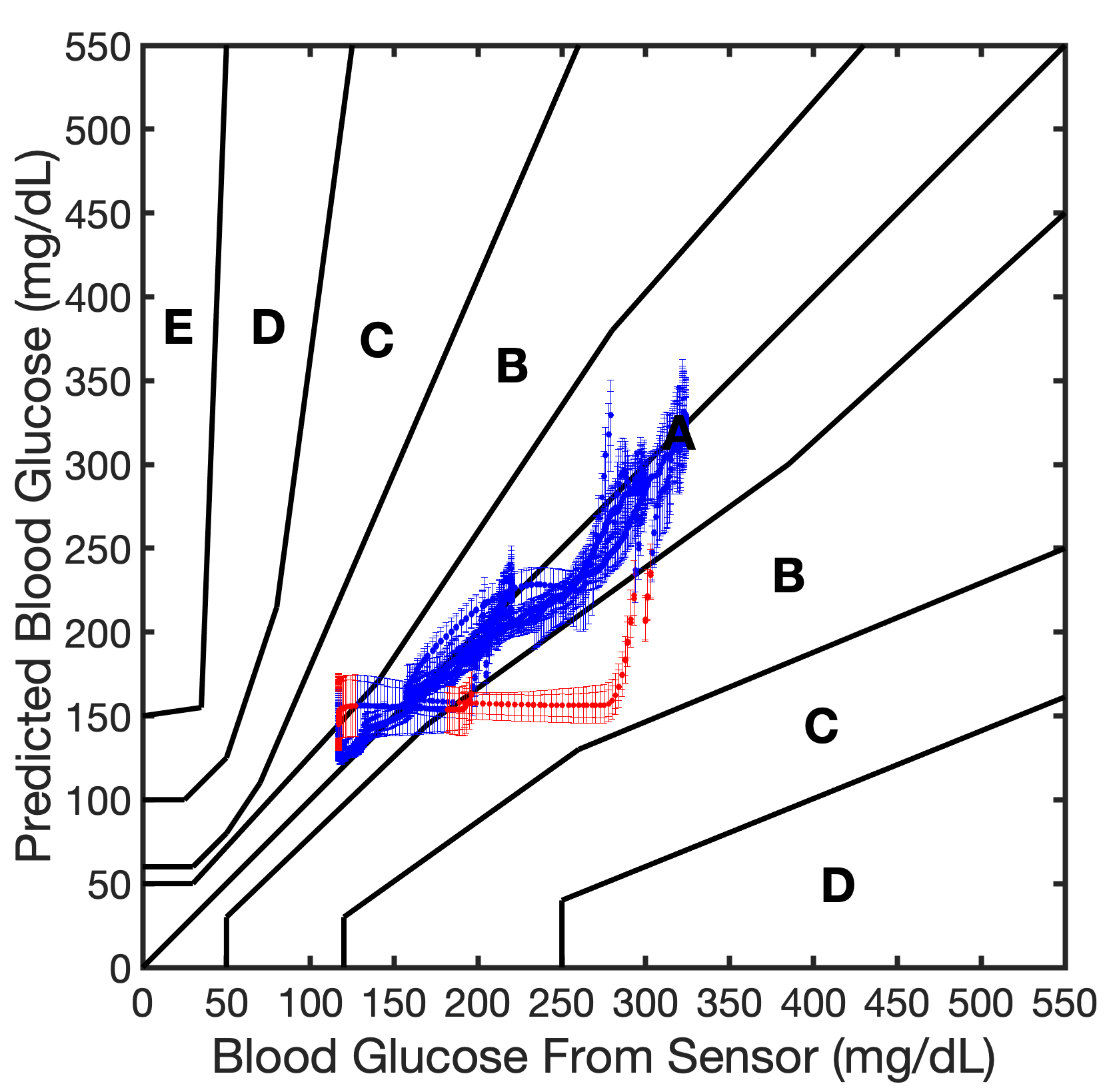}
  \caption*{(a)}
\end{minipage}\hfill
\begin{minipage}[t]{0.5\textwidth}
  \centering
 \includegraphics[width=0.75\linewidth]{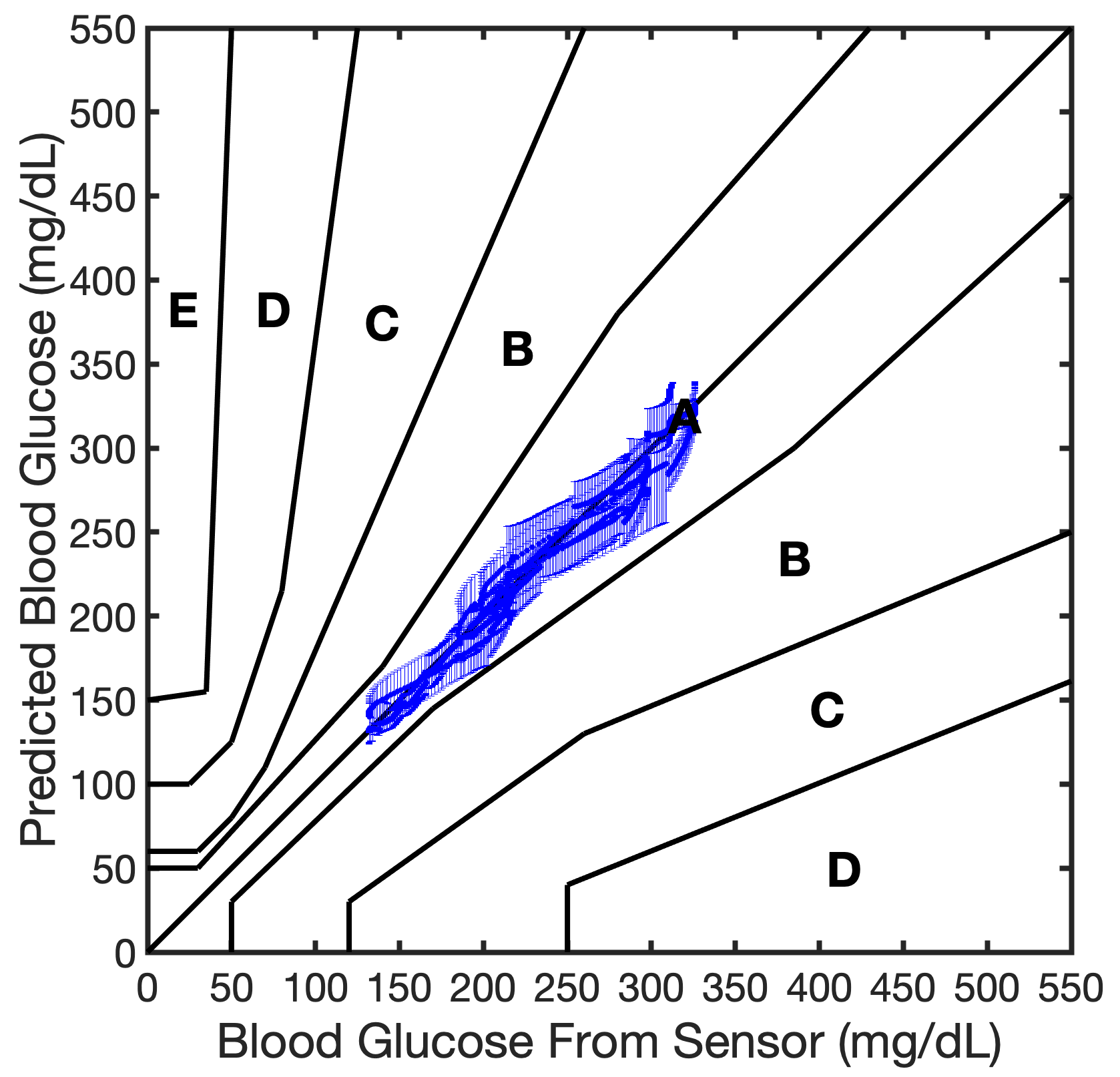}
  \caption*{(b)}
\end{minipage}
\caption{Parkes error grid diagnostics for representative rejected and accepted architectures by EVIDENT. (a) Rejected Level~1 architecture ($Level_1Arch_1$) shows uncertainty extending into higher-risk Zone B. 
(b) Selected Level~2 architecture ($Level_2Arch_4$) concentrates predictions within Zone A.}
\label{fig:parkes}
\vspace{-0.3in}
\end{figure}

%----Parks error map
\paragraph{\blue{Parkes error-grid diagnostics}}
To complement NLPD-based validation, we evaluate representative rejected and accepted architectures using the Parkes error grid \cite{parkes2000new, pfutzner2013technical}, a clinically established mapping for assessing the impact of glucose prediction errors on treatment decisions. The grid partitions errors into risk zones including no clinical impact (Zone A), minimal clinical consequence (Zone B), incorrect action affecting
clinical outcome (Zone C), significant medical risk (Zone D), and 
to dangerous treatment decisions (Zone E).
Figure~\ref{fig:parkes} compares a rejected Level~1 architecture with the selected Level~2 model. To account for uncertainty, we overlay uncertainty bands from both CGM measurements noise and TCN predictions. 
\blue{For these examples, the rejected Level~1 representative has
predictive intervals extending into Zone~B, whereas the accepted
Level~2 representative concentrates predictions and associated
intervals within Zone~A. These plots provide complementary
error-grid diagnostics for forecasting and do not establish clinical safety or
suitability for insulin-treatment decisions.}

%++++++++++++++++++++++++++++++++++++++++++++++++++++++++++++++++++++++++
\subsubsection{\blue{Comparison with predictive performance-based architecture selection}}
\label{sec:baseline}

\begin{figure}
\vspace{-0.9in} 
\centering
\begin{minipage}[t]{0.5\textwidth}
  \centering
  \includegraphics[width=\linewidth]{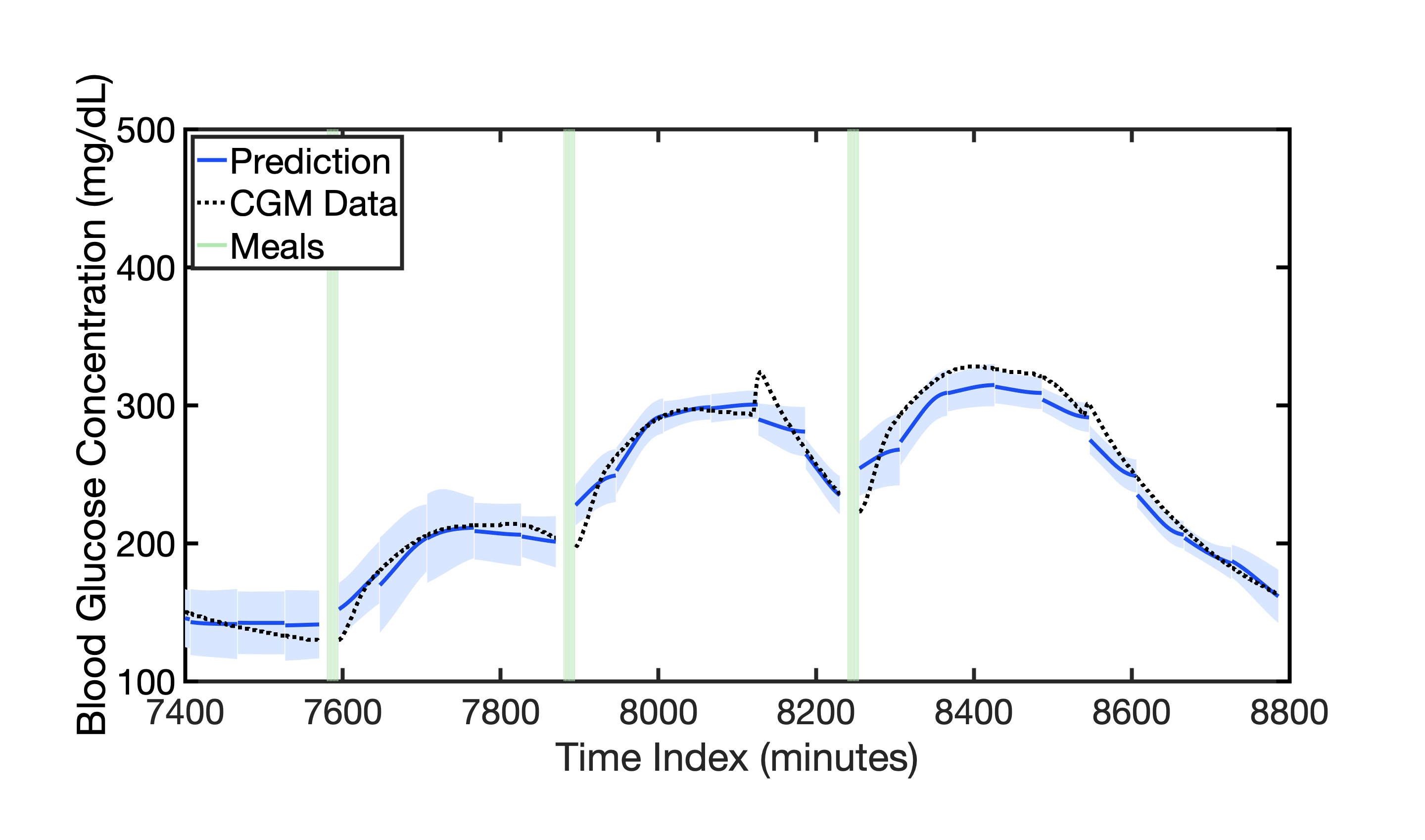}
  \caption*{(a)}
\end{minipage}\hfill
\begin{minipage}[t]{0.5\textwidth}
  \centering
  \includegraphics[width=\linewidth]{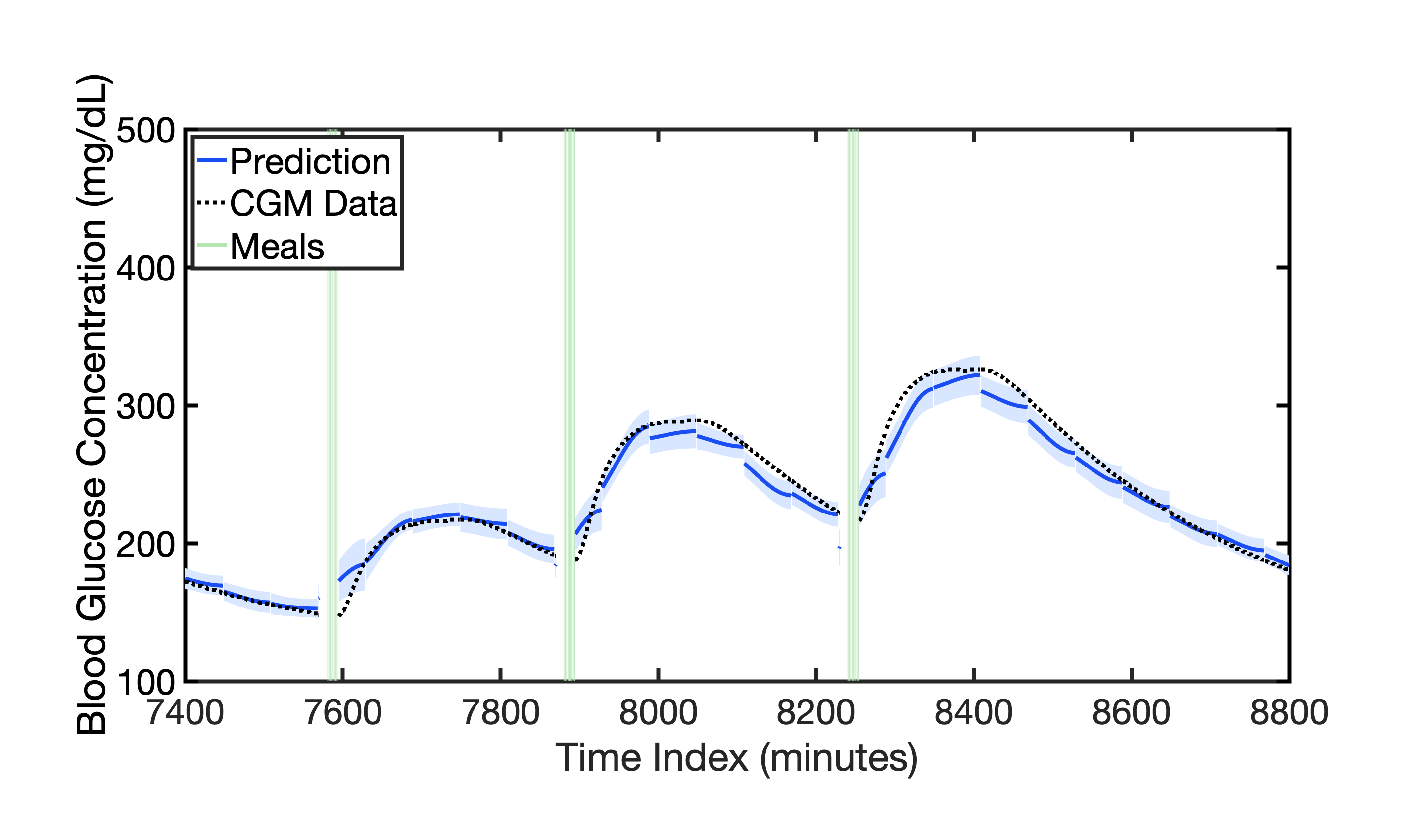}
  \caption*{(b)}
\end{minipage}
\begin{minipage}[t]{0.5\textwidth}
  \centering
  \includegraphics[width=\linewidth]{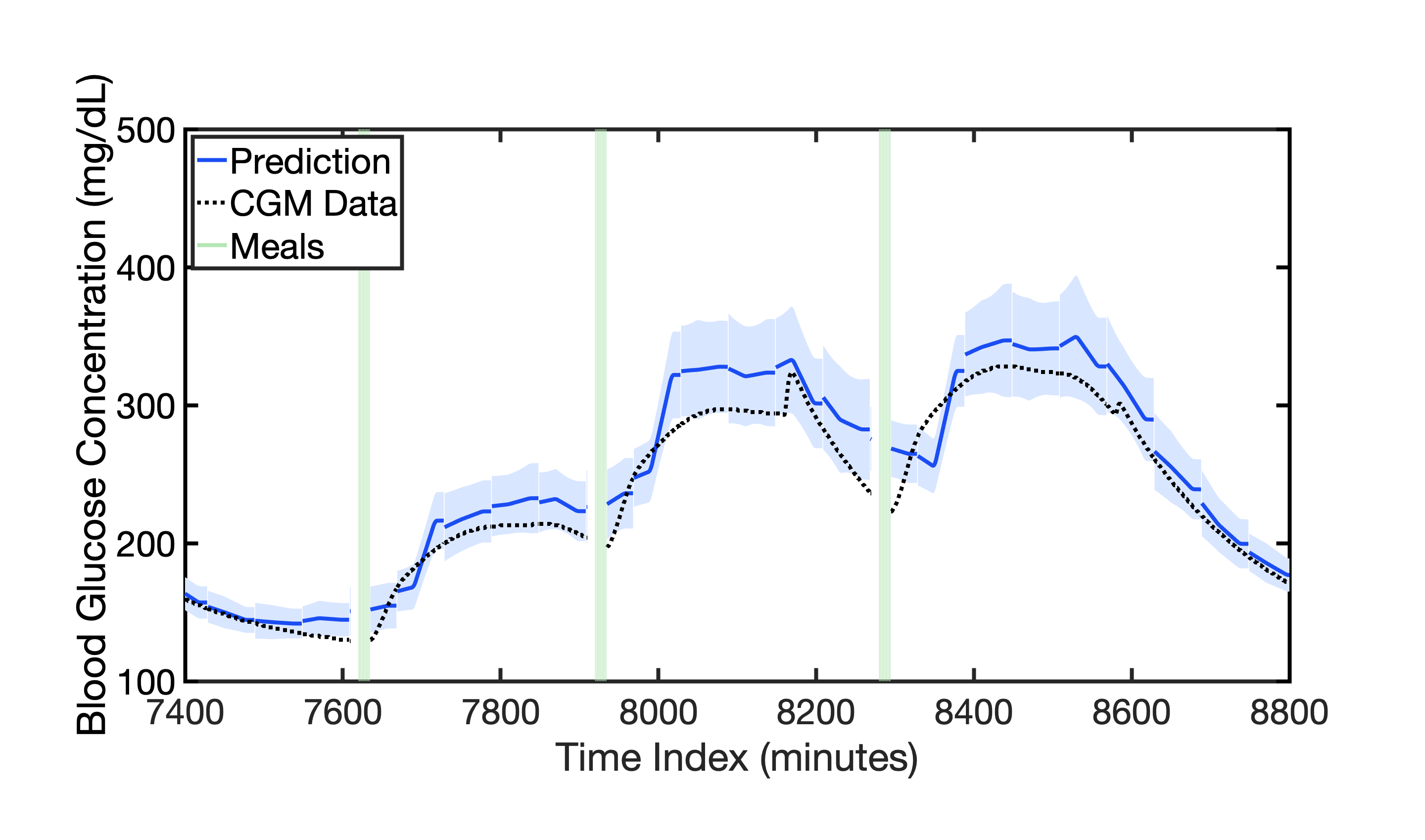}
  \caption*{(c)}
\end{minipage}\hfill
\begin{minipage}[t]{0.5\textwidth}
  \centering
  \includegraphics[width=\linewidth]{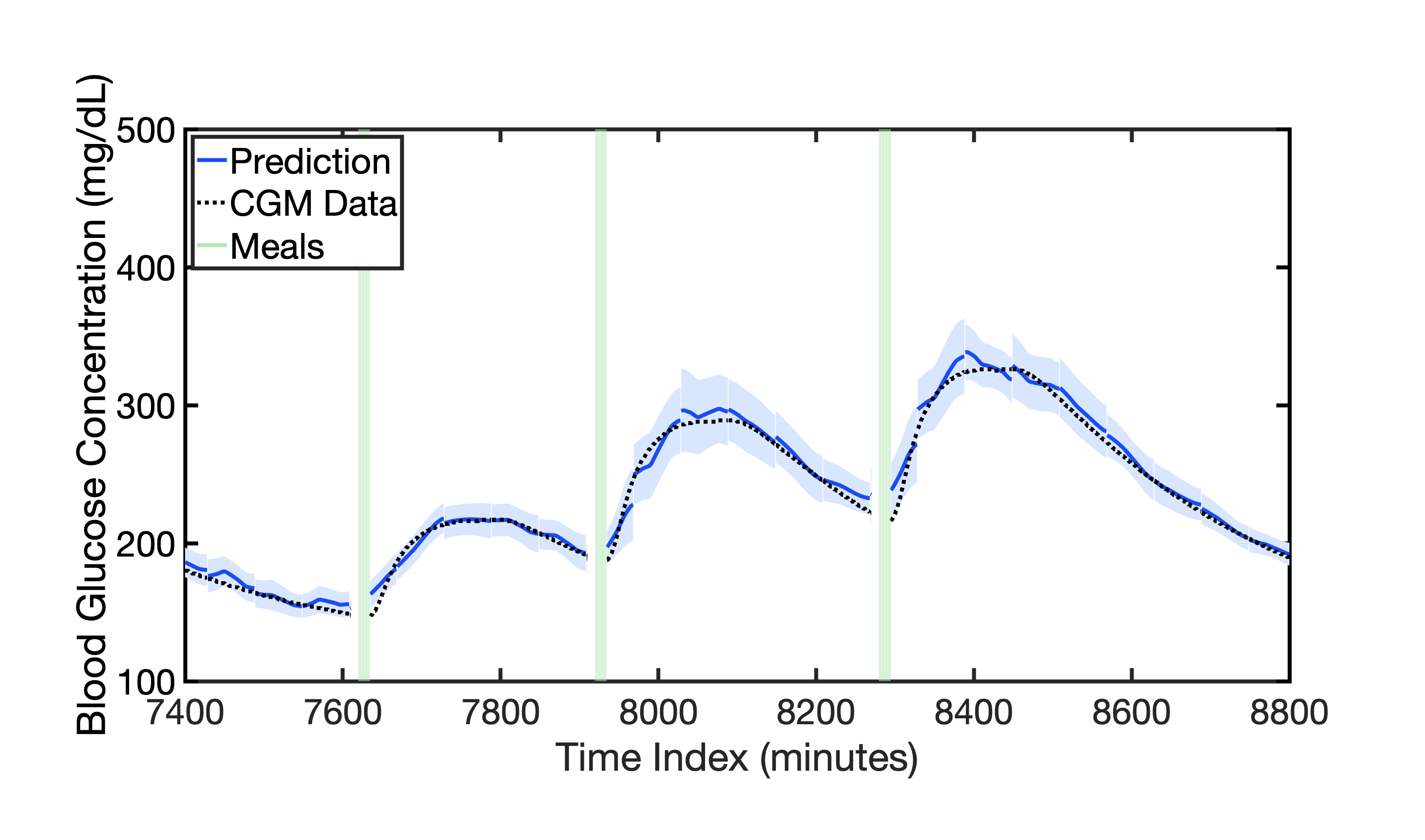}
  \caption*{(d)}
\end{minipage}
\begin{minipage}[t]{0.5\textwidth}
  \centering
  \includegraphics[width=\linewidth]{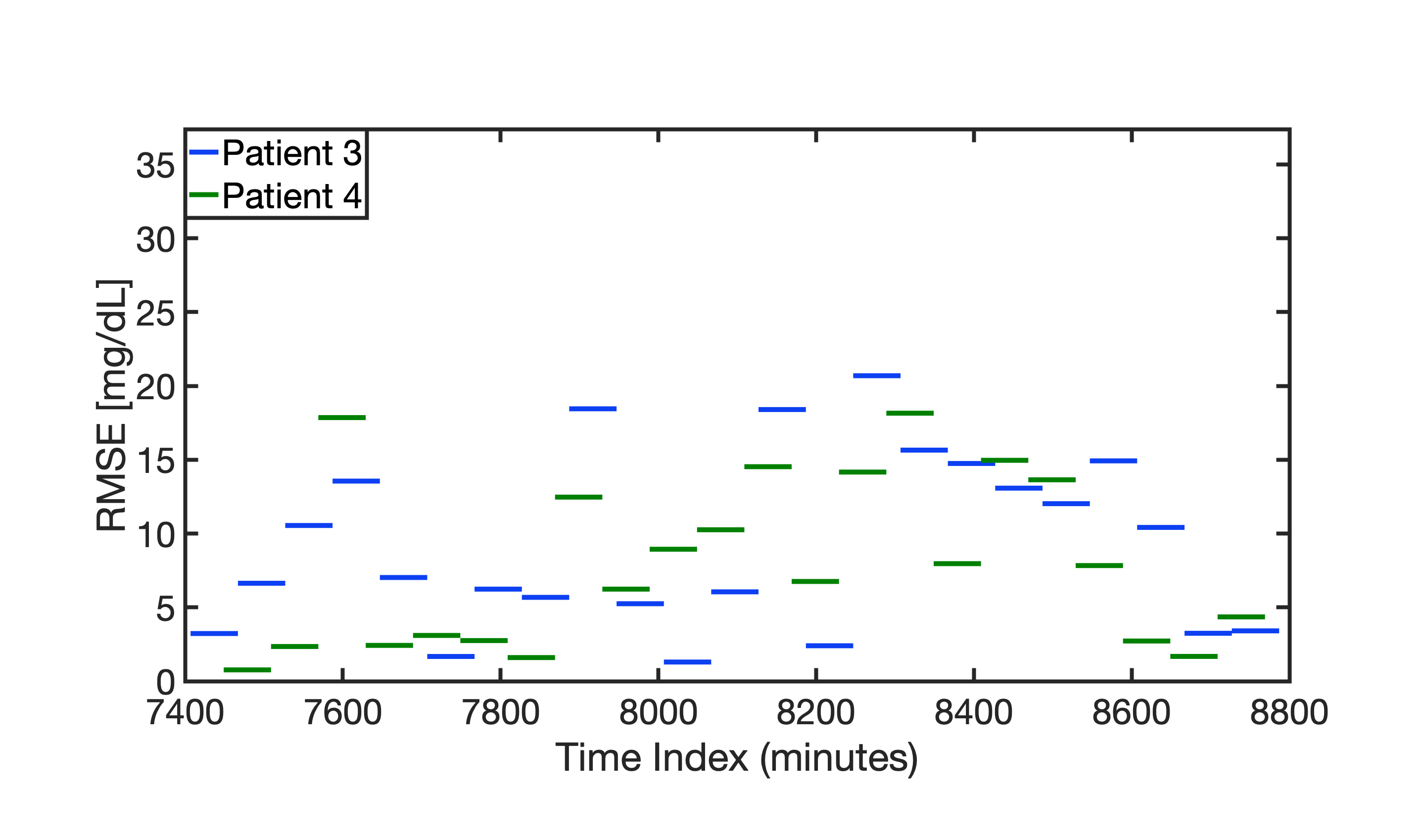}
  \caption*{(e)}
\end{minipage}\hfill
\begin{minipage}[t]{0.5\textwidth}
  \centering
  \includegraphics[width=\linewidth]{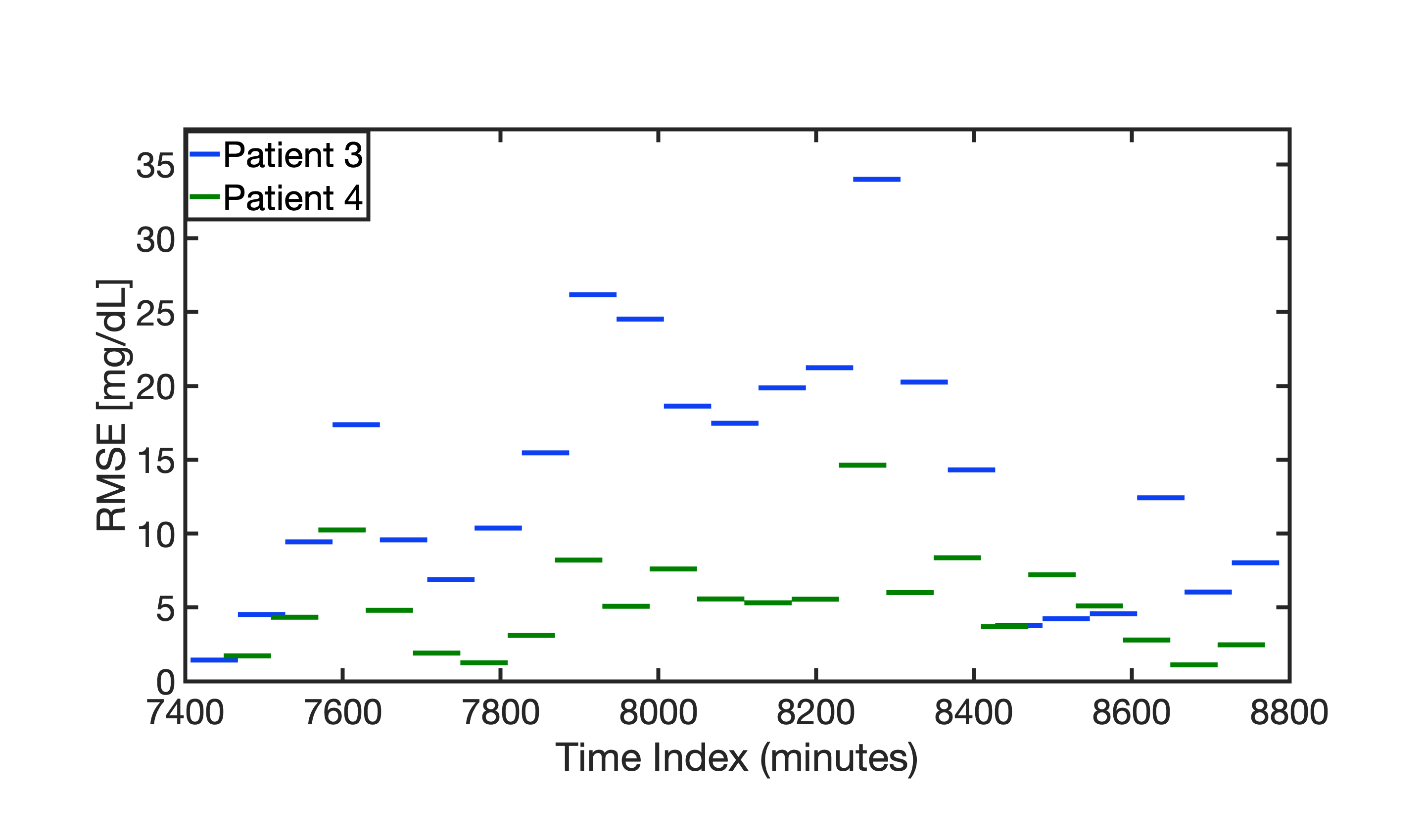}
  \caption*{(f)}
\end{minipage}
\begin{minipage}[t]{0.5\textwidth}
  \centering
  \includegraphics[width=\linewidth]{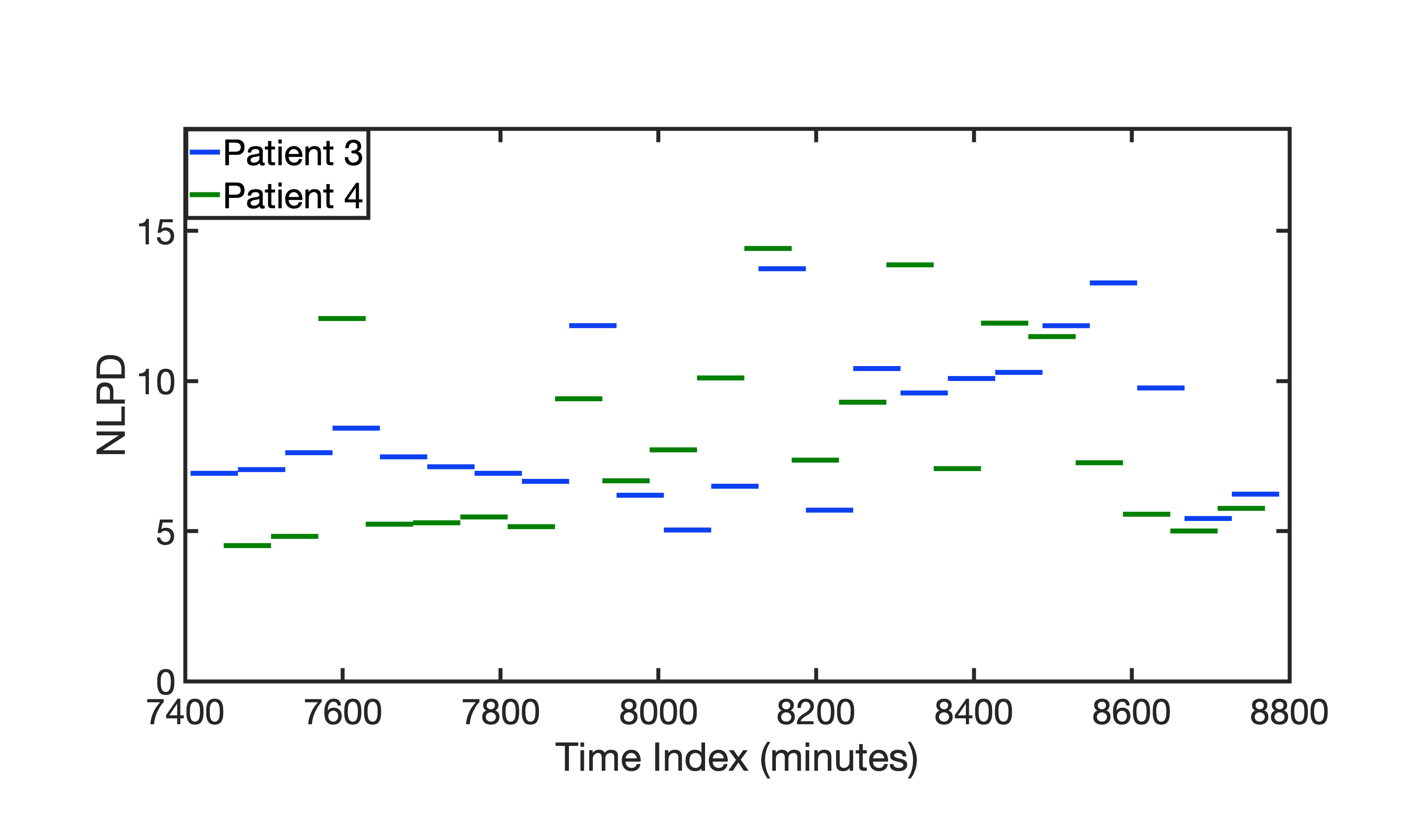}
  \caption*{(g)}
\end{minipage}\hfill
\begin{minipage}[t]{0.5\textwidth}
  \centering
  \includegraphics[width=\linewidth]{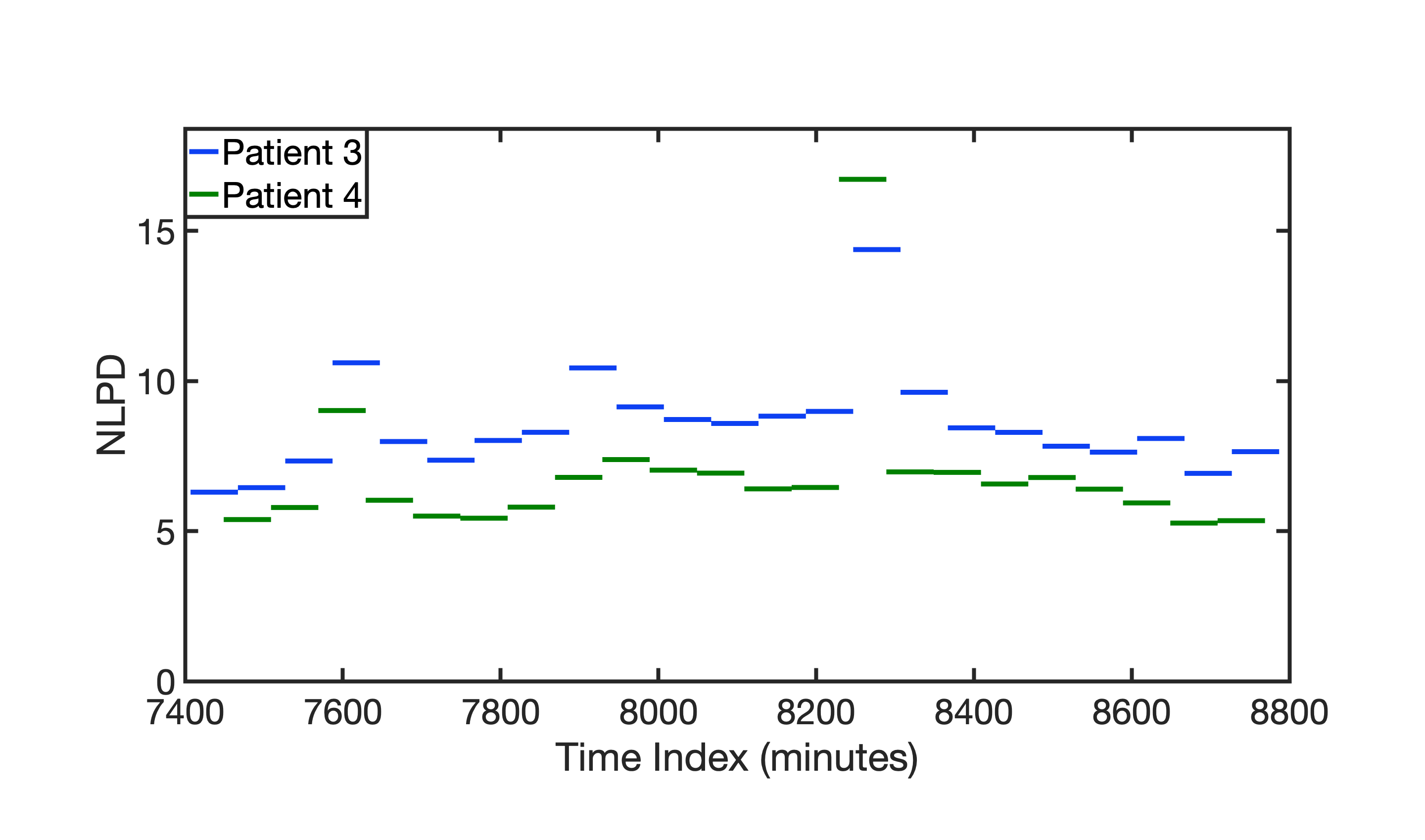}
  \caption*{(h)}
\end{minipage}
\caption{
\blue{
Performance comparison between the predictive performance-selected architecture $RandArch_8$ and the EVIDENT-selected architecture $Level_2Arch_4$ for the held-out patients in Fold 4 and independent data of day~6 excluded from model construction and validation. Panels (a,b) show the 1-hour rolling glucose predictions of $Level_2Arch_4$ for patients 3 and 4, respectively. Panels (c,d) 1-hour rolling glucose predictions of $RandArch_8$ for patients 3 and 4, respectively. 
Panels (e,f) show rolling-window RMSE for
$Level_2Arch_4$ and $RandArch_8$, respectively, with both patients
shown in each panel. Panels (g,h) show the corresponding
rolling-window NLPD scores.
Both models retain their parameters from adaptation on days~1--4.}}
\label{fig:random_search}
\end{figure}

\blue{We conduct an illustrative comparison with predictive
performance-based architecture selection within a fixed subset of
19 architectures from the 61-candidate feasible pool.
Random search is a standard baseline for hyperparameter and architecture selection because it provides a non-adaptive sampling strategy and has been shown to be competitive with grid search under fixed computational budgets \cite{bergstra2012random}.
The subset, listed in ~\ref{app:architecture_pool}, includes
the EVIDENT-selected $RandArch_6$$(=Level_2Arch_4)$ by construction.
Its size equals the number of candidates in the first two EVIDENT
capacity levels, matching candidate counts, rather than
establishing equal computational cost.}
\blue{The comparison is restricted to Fold~4, with patients~3 and~4
excluded from population training. Each of the 19 population-trained
candidates is adapted separately to each patient using days~1--4.
One shared comparator architecture is selected by minimizing day~5
mean window NLPD, first averaging over windows within each patient
and then averaging the two patient means equally.
EVIDENT retains its evidence-based ranking and day~5 acceptance
procedure. Both selected architectures are evaluated on independent day~6 data,
which was excluded from model development, using identical forecasting
and scoring conventions. Their four-day-adapted parameters remain
frozen, with no additional adaptation on day~5.}
\blue{Predictive performance-based selection identifies
$RandArch_8$ ($=Level_3Arch_1$), with 148,729 trainable parameters,
whereas EVIDENT selects $Level_2Arch_4$, with 47,779 parameters.
Figure~\ref{fig:random_search} presents their day~6 forecasts and
rolling-window NLPD and root mean-squared error (RMSE) results
for patients~3 and~4. For each forecast window, RMSE is defined as,}
\vspace{-0.1in}
\begin{equation}
\mathrm{RMSE}
=
\left[
\frac{1}{H}
\left\|
G_{\mathrm{data}}^{(k+1:k+H)}
-
\overline{\mathcal{T}}_{w,\xi}\!\left(
G_{\mathrm{data}}^{(k-d_G:k)},
M_{\mathrm{data}}^{(k-d_M:k)},
I_{\mathrm{data}}^{(k-d_I:k)}
\right)
\right\|_2^2
\right]^{1/2}.
\label{eq:rmse}
\end{equation}
\vspace{-0.25in}

\blue{Table~\ref{tab:day6_comparison} summarizes the independent
day 6 results shown in Figure~\ref{fig:random_search}, including
mean and maximum window NLPD, mean window RMSE, and rejected-window
counts at $Tol=14.5$. The two procedures select different
architectures, with EVIDENT selecting the smaller model in this
comparison. Its forecasts have no rejected windows for either
patient, whereas the comparator architecture has one rejected window for
patient~4 despite achieving lower mean NLPD and mean window RMSE
for that patient. These results illustrate the distinction between
average predictive performance and satisfaction of a window-wise
forecasting requirement. The comparison uses the same adaptation
protocol and is limited to the fixed 19-architecture subset and
the two subjects in Fold~4 and thus it does not establish general
superiority over random search.}

\clearpage
%%%%%%%%%%%%%%%%%%%%%%%%%%%%%%%%%%%%%
{
\color{blue}

\begin{table}[h]
\centering
\small
\caption{\blue{Day~6 evaluation of the EVIDENT-selected architecture
$Level_2Arch_4$ and the predictive performance-selected architecture $RandArch_8$ for the two held-out patients in Fold~4. Both models retain their parameters from adaptation on days~1--4 and
rejected window counts use $Tol=14.5$.}}
\label{tab:day6_comparison}
\setlength{\tabcolsep}{4pt}
\begin{tabular}{lcrrrc}
\toprule
\textbf{\shortstack{Architecture\\ID}} &
\textbf{Patient} &
\textbf{\shortstack{Mean window\\NLPD}} &
\textbf{\shortstack{Max.\ window\\NLPD}} &
\textbf{\shortstack{Mean window\\RMSE (mg/dL)}} &
\textbf{\shortstack{Windows\\rejected/total}} \\
\midrule
$Level_2Arch_4$ & 3 & 8.44 & 13.73 & 9.33 & 0/24 \\
                & 4 & 7.98 & 14.41 & 7.98 & 0/24 \\
\midrule
$RandArch_8$    & 3 & 8.52 & 14.38 & 13.51 & 0/24 \\
                & 4 & 6.86 & 16.71 & 5.28 & 1/24 \\
\bottomrule
\end{tabular}
\end{table}
%++++++++++++++++++++++++++++++++++++++++++++++++++++++++++++++++++++++++
%++++++++++++++++++++++++++++++++++++++++++++++++++++++++++++++++++++++++
\subsection{EVIDENT for glucose forecasting using recorded patient data}
\label{sec:ohio_results}

To evaluate EVIDENT using measured real-world patient data, we consider blood glucose forecasting with the OhioT1DM dataset \cite{marling2020ohiot1dm}. The dataset contains approximately eight weeks of data from each of twelve individuals with type~1 diabetes receiving insulin-pump therapy. 
The dataset contains records collected under free-living conditions,
including continuous glucose monitoring (CGM) measurements at five-minute intervals, insulin-pump
basal and bolus records, and self-reported meal times and carbohydrate
estimates. The records were collected using Medtronic Enlite CGM sensors.
For the predictive calculations, we use an observation-noise
variance of $\sigma_{\mathrm{data}}^2=6^2\,(\mathrm{mg/dL})^2$
\cite{biagi2017modeling}. These records provide an observational setting for
evaluating EVIDENT beyond the controlled in silico study in Section \ref{sec:evident_results}. 
Five patients are used in this experiment. As shown in Figure~\ref{fig:data_ohio}(a--d), patients 559, 563, 570, and 591 each contribute three days of data for population-based training and architecture selection.
Patient 588 is excluded from population training and evidence-based
ranking. Six days of this patient's data are used for subject-specific
adaptation, and a separate, later one-day interval is used for validation
(Figure~\ref{fig:data_ohio}(e,f)). This validation interval governs
architecture acceptance and stopping and therefore forms part of
architecture selection rather than an independent test.

\begin{figure}[!h]
\centering

\begin{minipage}{0.45\linewidth}
  \centering
  \includegraphics[width=\linewidth]{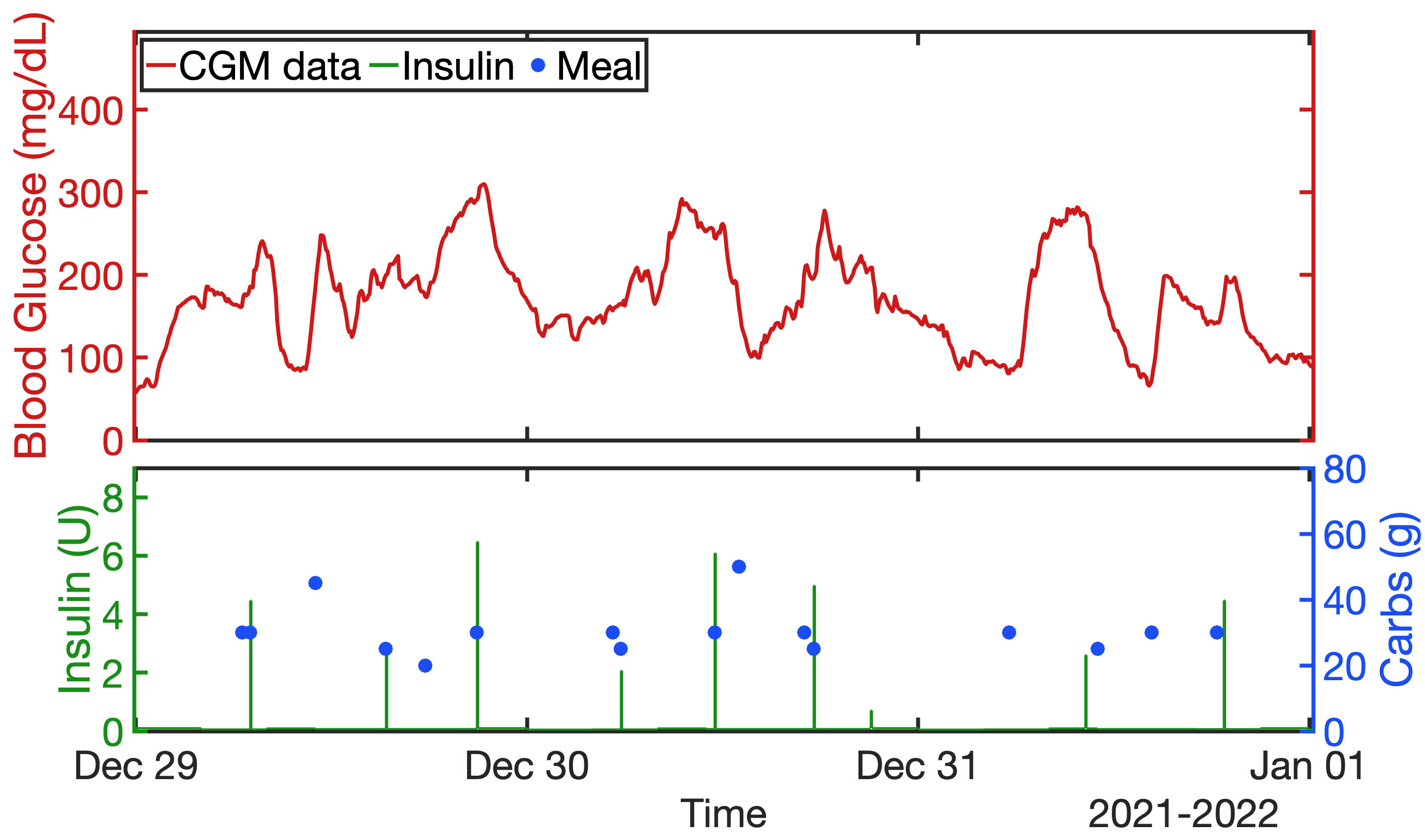}
  \par\smallskip (a) Patient 559 
\end{minipage}
\hfill
\begin{minipage}{0.45\linewidth}
  \centering
  \includegraphics[width=\linewidth]{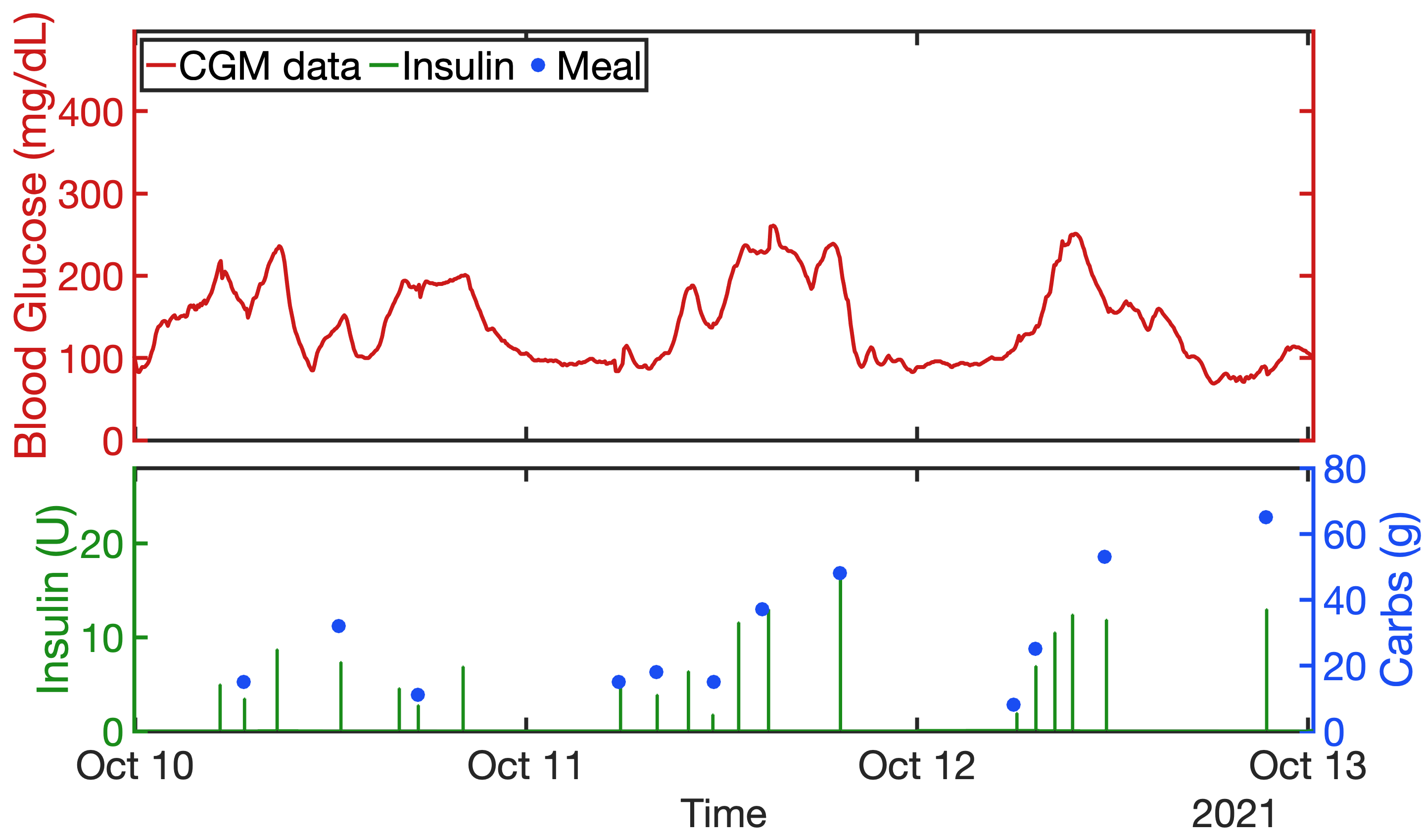}
  \par\smallskip (b) Patient 563
\end{minipage}

\begin{minipage}{0.45\linewidth}
  \centering
  \includegraphics[width=\linewidth]{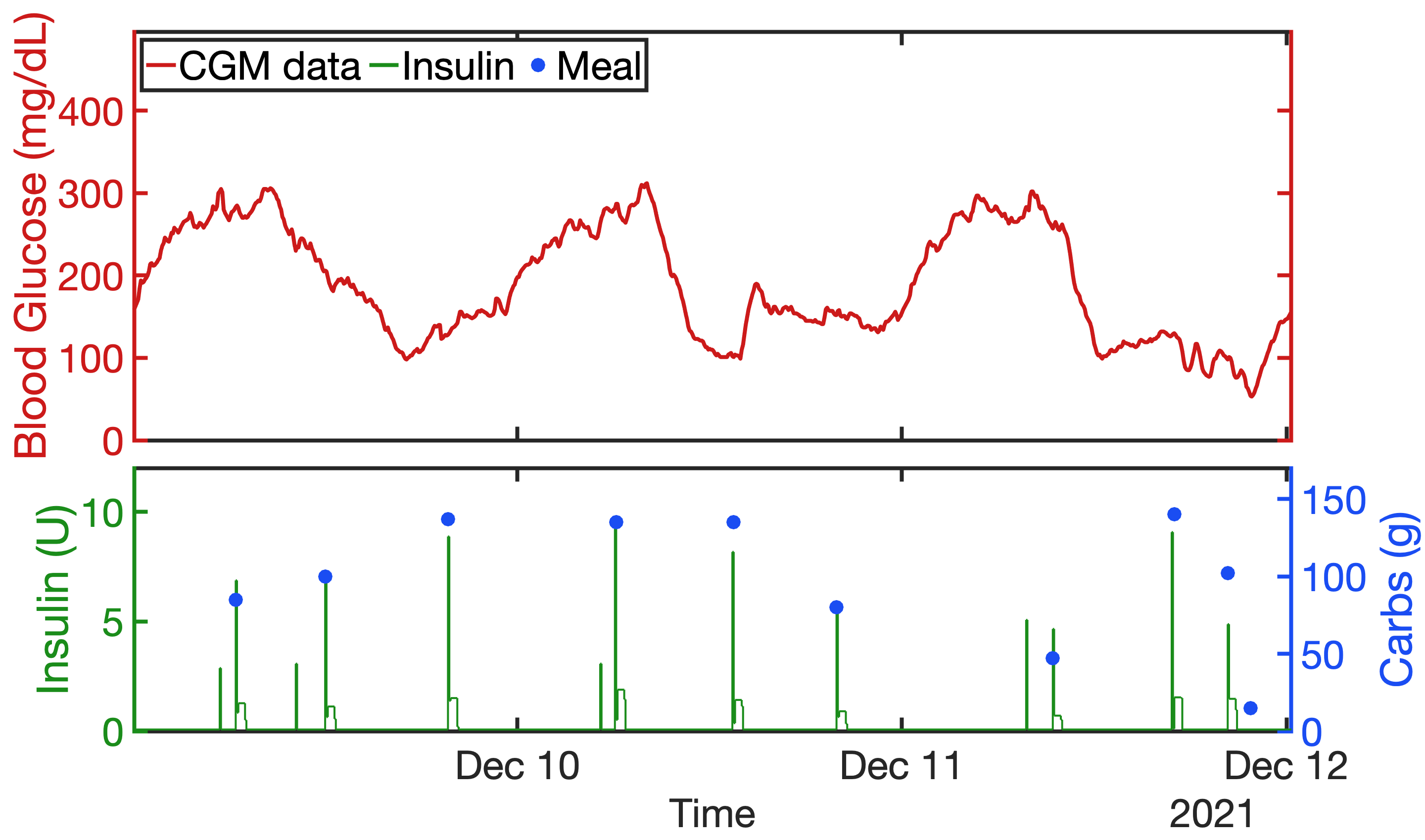}
  \par\smallskip (c) Patient 570
\end{minipage}
\hfill
\begin{minipage}{0.45\linewidth}
  \centering
  \includegraphics[width=\linewidth]{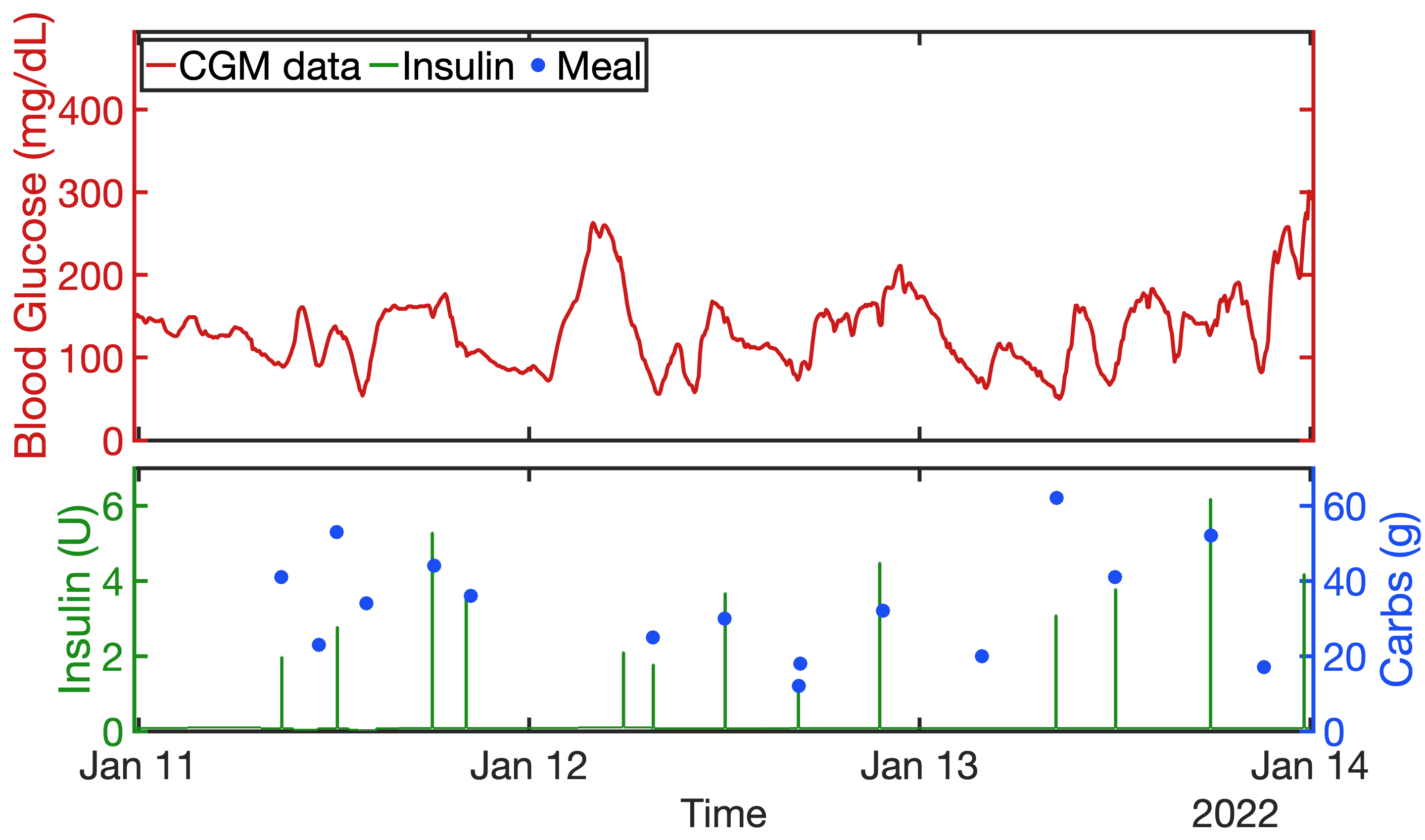}
  \par\smallskip (d) Patient 591
\end{minipage}

\begin{minipage}{0.45\linewidth}
  \centering
  \includegraphics[width=\linewidth]{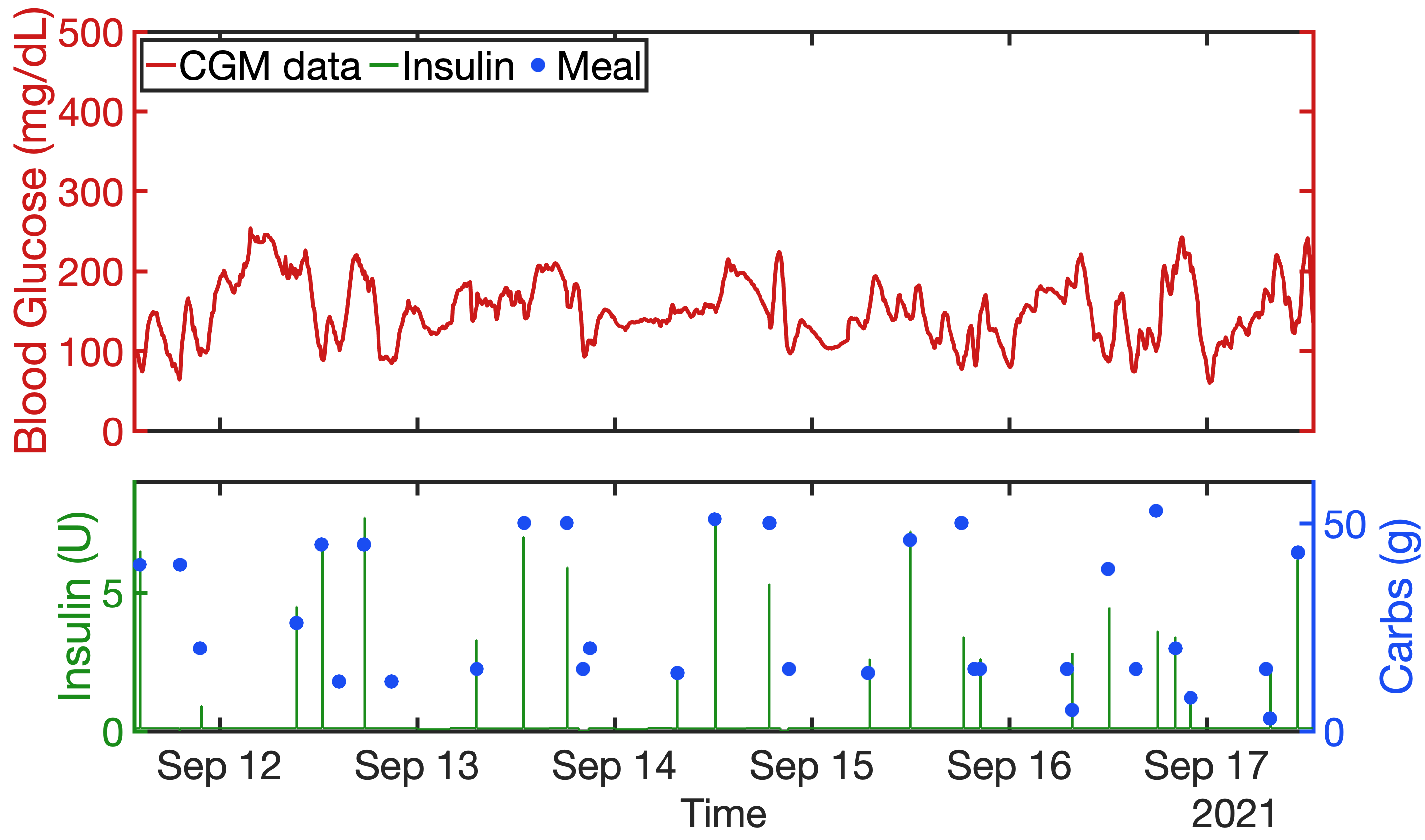}
  \par\smallskip (e) Patient 588
\end{minipage}
\hfill
\begin{minipage}{0.45\linewidth}
  \centering
  \includegraphics[width=\linewidth]{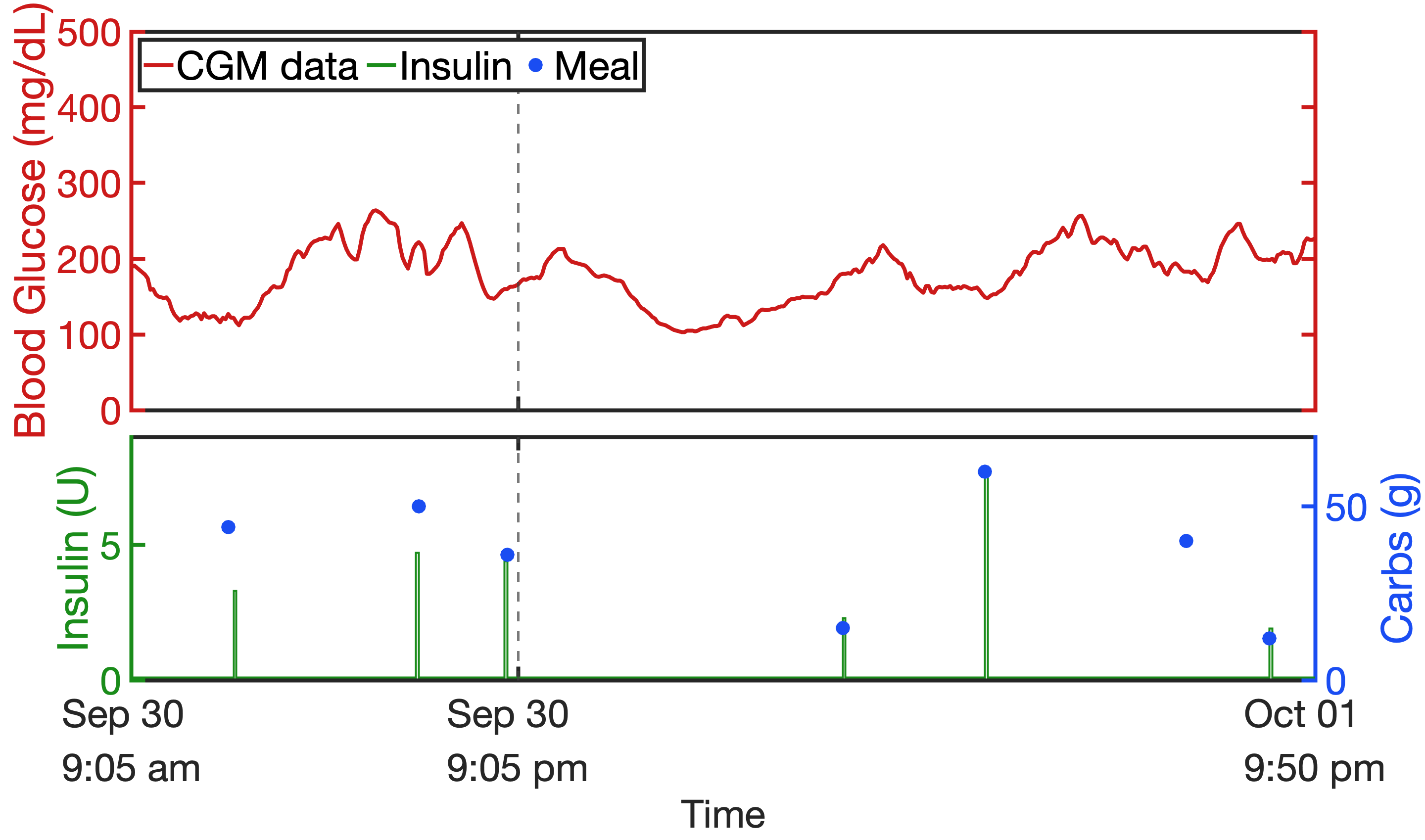}
  \par\smallskip (f) Patient 588
\end{minipage}

\caption{\blue{OhioT1DM data used for population-based training, patient adaptation, and validation. (a--d) Three days of data from each of four patients used for population-based training. (e) Patient adaptation using six days of data from an unseen patient 588, excluded from population training. (f) One day of data from the same patient reserved for validation; the left of the dotted black line shows the input for the first prediction window. Red curves show CGM glucose measurements, green bars indicate insulin delivery, and blue markers represent carbohydrate intake.}}
\label{fig:data_ohio}
\end{figure}

\subsubsection{Population training and architecture ranking}
Three days of data from each of the four patients are combined to form the population-based dataset. The candidate pool follows the same TCN architecture family and structural constraints described in Section~\ref{sec:evident_results}, including two convolutions per temporal block, geometrically increasing encoder dilations, doubling encoder channels, mirrored decoder channels, and the condition $f\geq b$. For this experiment, the receptive-field bounds are removed, allowing additional temporal configurations to be considered. An architecture search algorithm, Optuna \cite{optuna_2019}, is used to explore candidate TCN architecture within each
capacity level, with approximate log-evidence as the search objective. The candidate pool is partitioned into six ordered
capacity levels according to the number of trainable parameters $W$,
as summarized in Table~\ref{tab:ohio_arch_categories}. Evaluated candidates
within each level are ranked by approximate log-evidence computed
using the architecture-level approximation in ~\ref{app:evidence}. The
highest evidence candidate found within a level is shortlisted for
subject-specific adaptation and validation under uncertainty. Validation determines
whether EVIDENT accepts the candidate and stops or proceeds to the
next capacity level.
Table~\ref{tab:ohio_selected} reports the highest-evidence
candidate identified in each of capacity levels 1-4.

\begin{footnotesize}
\setlength{\tabcolsep}{4pt}
\renewcommand{\arraystretch}{1.3}
\begin{longtable}{
    p{1.5cm}
    p{2.5cm}
    >{\centering\arraybackslash}p{2.4cm}
    p{2.0cm}
    p{1.5cm}
    p{1.5cm}
    p{1.5cm}
}
\caption{Expanded TCN architecture pool for the OhioT1DM experiment.}
\label{tab:ohio_arch_categories}\\

\toprule
\textbf{\shortstack{Capacity \\ level $l$}} &
\textbf{\shortstack{Parameter\\range}} &
\textbf{\shortstack{Number of\\architectures}} &
\textbf{\shortstack{Temporal\\blocks $N$}} &
\textbf{\shortstack{Filter\\size $f$}} &
\textbf{\shortstack{Dilation\\base $b$}} &
\textbf{\shortstack{Starting\\channels $c_1$}} \\
\midrule
\endfirsthead

\toprule
\textbf{Capacity level $l$} &
\textbf{\shortstack{Parameter\\range}} &
\textbf{\shortstack{Number of\\architectures}} &
\textbf{\shortstack{Temporal\\blocks $N$}} &
\textbf{\shortstack{Filter\\size $f$}} &
\textbf{\shortstack{Dilation\\base $b$}} &
\textbf{\shortstack{Starting\\channels $c_1$}} \\
\midrule
\endhead

\endfoot
\bottomrule
\endlastfoot

1 & $<2{,}500$
   & 46 & $[2,3]$ & $[3,15]$ & $[2,5]$
   & $\{2,4\}$ \\

2 & $2{,}500 - 5{,}000$
   & 38 & $[2,4]$ & $[3,15]$ & $[2,5]$
   & $\{2,4,8\}$ \\

3 & $5{,}000 - 15{,}000$
   & 48 & $[2,4]$ & $[5,11]$ & $[2,5]$
   & $\{2,4,8\}$ \\

4 & $15{,}000 - 40{,}000$
   & 64 & $[2,5]$ & $[3,15]$ & $[2,5]$
   & $\{2,4,8,16\}$ \\

5 & $40{,}000 - 65{,}000$
   & 38 & $[2,5]$ & $[3,13]$ & $[2,5]$
   & $\{2,4,8,16,32\}$ \\

6 & $65{,}000 - 100{,}000$
   & 40 & $[2,6]$ & $[3,15]$ & $[2,5]$
   & $\{2,4,8,16,32\}$ \\

\end{longtable}
\end{footnotesize}
 
Within each capacity level, candidate architectures were trained using
the population-based dataset and ranked by approximate Bayesian evidence.
Table~\ref{tab:ohio_selected} summarizes the highest-evidence candidate among those evaluated within the capacity level.

\begin{table}[htbp]
\centering
\small
\setlength{\tabcolsep}{4pt}
\renewcommand{\arraystretch}{1.2}
\caption{Highest-evidence Bayesian TCN architecture in each capacity
level for the OhioT1DM experiment.}
\label{tab:ohio_selected}
\begin{tabular}{lrrrrcc}
\toprule
\textbf{\shortstack{Architecture\\ID}} &
\textbf{\shortstack{Parameter\\count $W$}} &
\textbf{\shortstack{Temporal\\blocks $N$}} &
\textbf{\shortstack{Filter\\size $f$}} &
\textbf{\shortstack{Dilation\\base $b$}} &
\textbf{\shortstack{Encoder\\channels}} &
\textbf{\shortstack{Decoder\\channels}} \\
\midrule
$Level_1Arch_{20}$ & 1,061  & 2 & 3 & 3 & $[4,8]$   & $[8,4]$   \\
$Level_2Arch_{18}$ & 3,977  & 2 & 3 & 3 & $[8,16]$  & $[16,8]$  \\
$Level_3Arch_{3}$ & 6,329  & 2 & 5 & 4 & $[8,16]$  & $[16,8]$  \\
$Level_4Arch_{39}$ & 26,729 & 3 & 5 & 4 & $[8,16,32]$ & $[32,16,8]$ \\ 
\bottomrule
\end{tabular}
\end{table}

\subsubsection{Subject-specific validation results}
For each selected architecture, the posterior distribution of weights obtained from population-based training is used as the prior for the patient 588. 
Subject-specific adaptation uses six days of data, and validation is
performed on the separate one-day interval shown in
Figure~\ref{fig:data_ohio}(f).
Forecasts are evaluated against recorded CGM measurements using the
NLPD criterion described in Section~\ref{sec:evident}, with a
study-specific tolerance $\mathrm{Tol}=15$.
Acceptance requires every scored validation window to satisfy
this tolerance.
Figure~\ref{fig:pred_ohio}(a--d) compares forecasts from the architecture with highest log-evidence in capacity levels 1--4 including
$Level_1Arch_{20}$,
$Level_2Arch_{18}$,
$Level_3Arch_3$, and
$Level_4Arch_{39}$, respectively.
The dotted black curves show recorded CGM measurements,
the colored curves show predictive means, and the shaded bands
represent 95\% predictive intervals.
The representatives of capacity levels 1--3
each violate the acceptance criterion in one or more forecast
windows. In contrast, $Level_4Arch_{39}$, with 26,729 trainable
parameters, satisfies the criterion in all scored windows over the
one-day validation interval. EVIDENT therefore accepts this candidate
at capacity level 4, the first level whose evaluated representative
satisfies the prescribed criterion.
% coverage plot 
\blue{Figure~\ref{fig:pred_ohio}(e) examines whether the predictive
intervals are consistent with observed forecast errors.
For each nominal coverage level, empirical coverage is the fraction
of observed CGM values contained within the corresponding central
predictive intervals, pooled across validation windows and forecast
lead times. The intervals use the same predictive variance as NLPD,
including CGM noise.
The representatives of levels~1--3 exhibit substantial undercoverage,
indicating overconfident predictions relative to their observed errors.
The level~4 representative follows the nominal coverage reference
more closely. Its broader intervals therefore provide a better
match to the observed errors on this validation sequence.
This diagnostic complements NLPD-based acceptance by assessing
pointwise coverage.}
% appendix results
\blue{Additionally, a tolerance sensitivity analysis for the 
$Level_4Arch_{39}$ is reported in
~\ref{app:tolerance}, illustrating how the prescribed
tolerance affects acceptance of this architecture. With predictions and validation windows
held unchanged, the predictor violates the criterion in 17, 5,
and 2 of the 24 windows at tolerances of 8, 10, and 12,
respectively, and satisfies it in all windows at tolerances
of 15 and 20.}
\blue{We note that these findings are limited
to the evaluated validation day for patient~588 and does not establish suitability
for insulin-control decisions.}

\begin{figure}[!h]
\centering
%\vspace{-1in}
\begin{minipage}{0.49\linewidth}
  \centering
  \includegraphics[width=\linewidth]{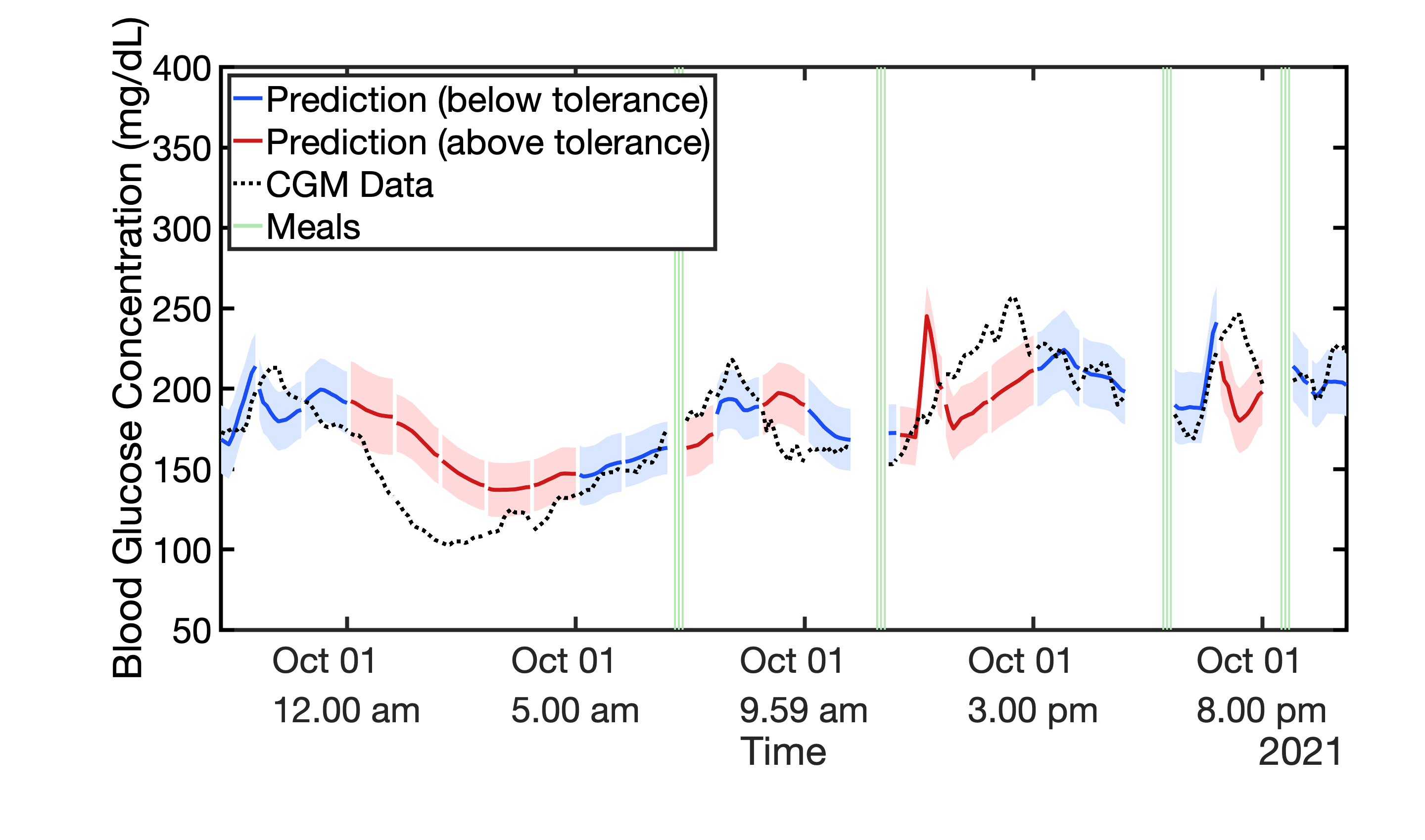}
  \par\smallskip (a) 
\end{minipage}
\hfill
\begin{minipage}{0.49\linewidth}
  \centering
  \includegraphics[width=\linewidth]{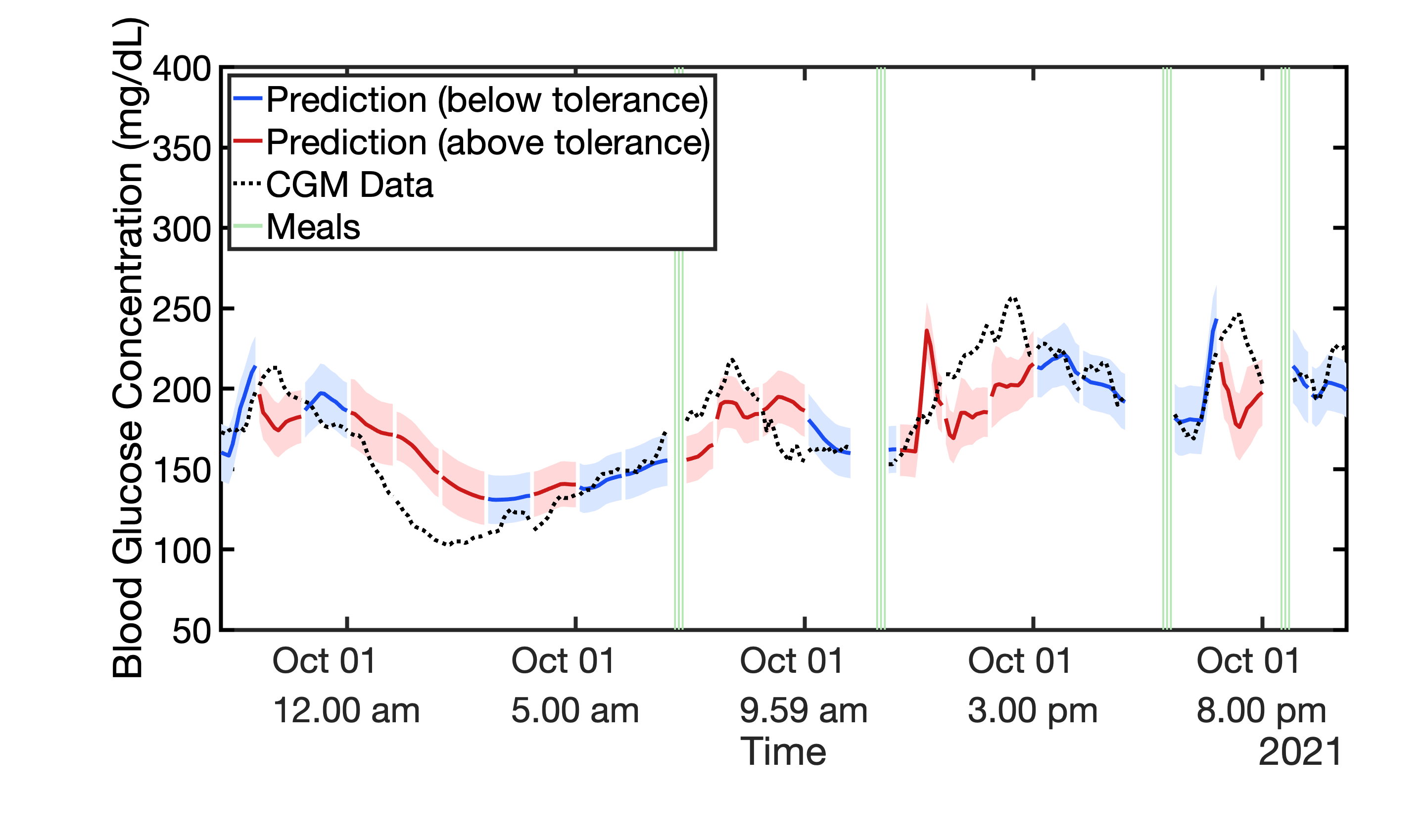}
  \par\smallskip (b)
\end{minipage}

\begin{minipage}{0.49\linewidth}
  \centering
  \includegraphics[width=\linewidth]{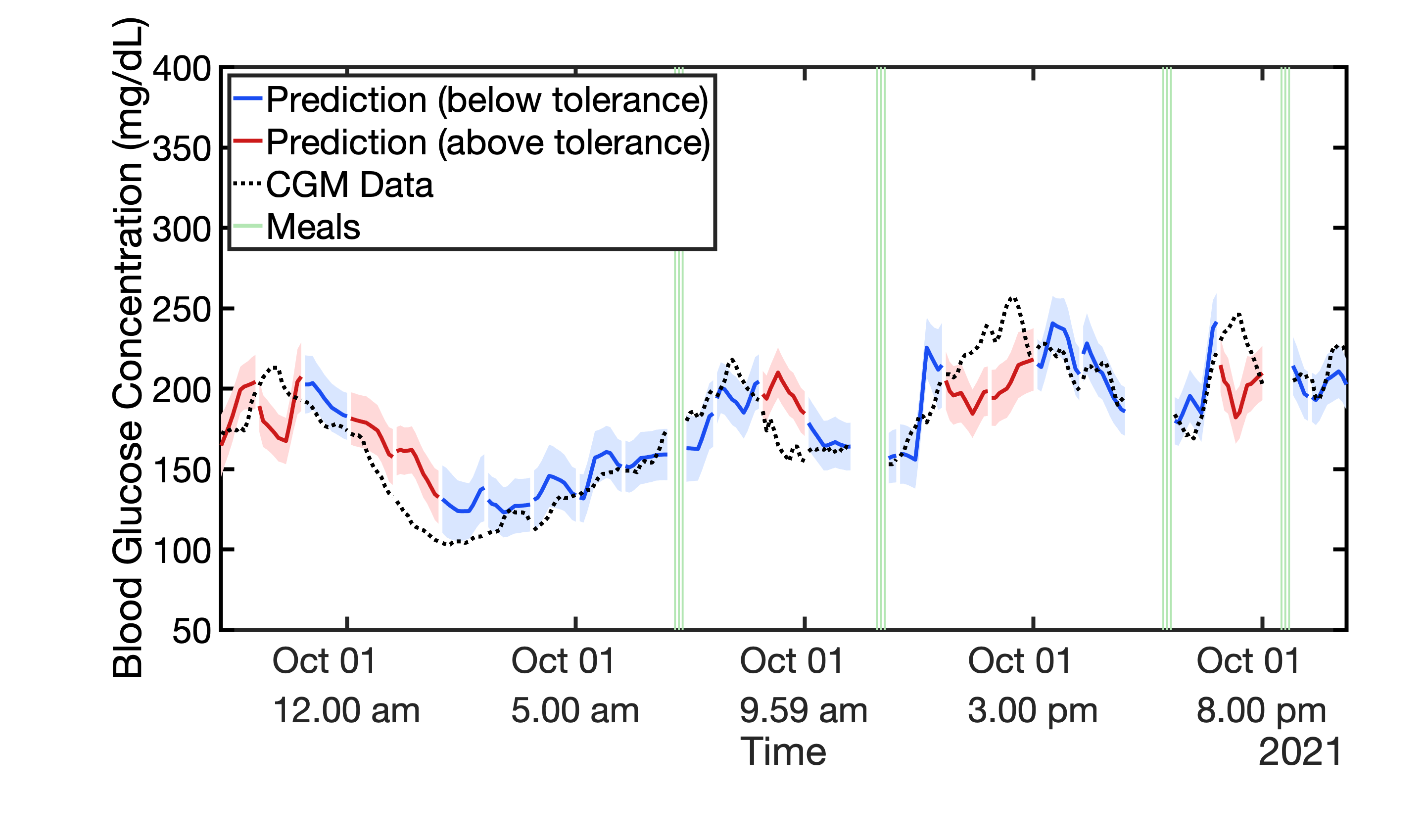}
  \par\smallskip (c)
\end{minipage}
\hfill
\begin{minipage}{0.5\linewidth}
  \centering
  \includegraphics[width=\linewidth]{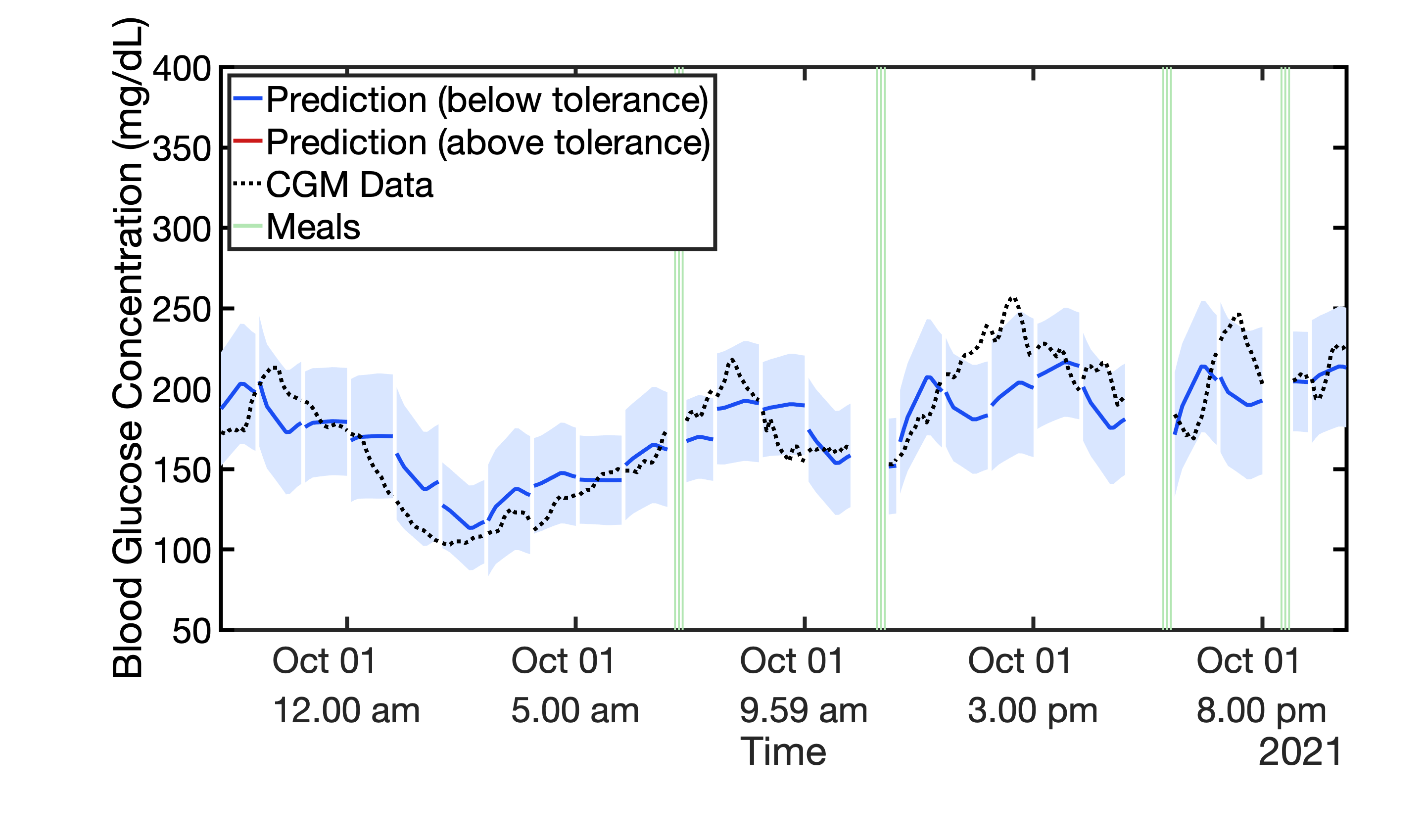}
  \par\smallskip (d)
\end{minipage}

\begin{minipage}{0.6\linewidth}
  \centering
  \includegraphics[width=\linewidth]{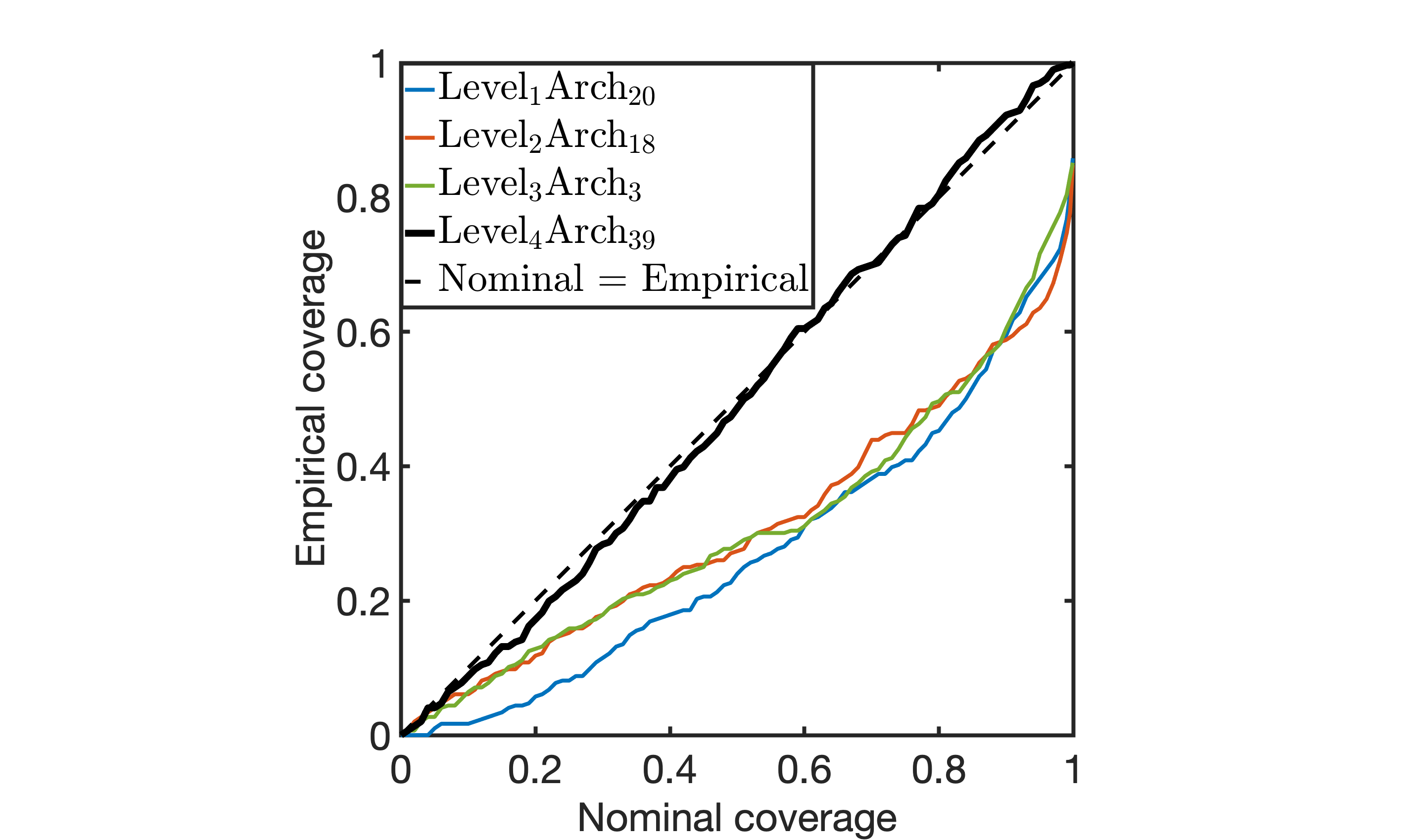}
  \par\smallskip (e)
\end{minipage}

\caption{
\blue{Subject-specific validation on recorded CGM data from patient 588
using the evidence-selected representatives of
(a) capacity level 1, (b) capacity level 2,
(c) capacity level 3, and (d) capacity level 4.
Dotted black curves show recorded CGM measurements, solid curves show
predictive means, and the shaded bands represent 95\% predictive intervals.
Blue and red forecast segments indicate window NLPD values at or
below, and above, $\mathrm{Tol}=15$, respectively.
The capacity level 4 representative satisfies the criterion in all
scored windows.
(e) Nominal versus empirical pointwise coverage, pooled across
validation windows and forecast lead times, for nominal levels up
to 0.99. Predictive intervals include observation-noise variance.
The dashed diagonal denotes agreement between nominal and empirical
coverage.
}
}
\label{fig:pred_ohio}
\end{figure}

%========================================================
%========================================================
\subsection{EVIDENT for synthetic ECG forecasting with Bayesian LSTMs}
\label{sec:ecg_results}

To examine whether the EVIDENT procedure extends beyond glucose forecasting and TCN predictors, we consider short-horizon electrocardiogram (ECG) forecasting using Bayesian long short-term memory (LSTM) networks. 
Six synthetic subject records are used in this experiment including five contribute data for population training and evidence-based architecture ranking, while
the sixth is excluded from these stages and used for subject-specific
adaptation and subsequent validation. 
Two architecture levels were evaluated using Optuna \cite{optuna_2019} and the architecture-level evidence approximation in ~\ref{app:evidence}. The highest-evidence Level~1 architecture failed the prescribed ECG validation criterion, whereas the highest-evidence Level~2 architecture passed. This study therefore examines the transfer of the EVIDENT procedure to a different signal, sampling rate, forecasting horizon, and network family.

\subsubsection{Synthetic ECG data}
\label{sec:synthetic_ecg_data}
The ECG experiment uses six synthetic subject records generated using a
customized implementation of the dynamical model of McSharry et al.\
\cite{mcsharry2003dynamical}. Each record spans 300~s at a sampling frequency
of 256~Hz. Inter-patient variability is introduced through independently
sampled heart-rate, heart-rate-variability, respiratory, morphology,
voltage-gain, and measurement-noise parameters. The forecasting inputs are
the synthetic ECG voltage $E$, elapsed time since the most recently available
detected R peak $\tau_R$, the last available detected RR interval
$RR_{\mathrm{last}}$, and a simulated respiratory-belt signal $B$ with added noise. A one-second
history of these four channels is used to predict the subsequent 0.25~s of
ECG. The data-generation equations and parameter distributions are provided in
~\ref{app:ecg_generation}.

\subsubsection{Bayesian LSTM predictor}
\label{sec:bayesian_lstm_ecg}

The ECG predictor is implemented using an LSTM network \cite{graves2012long, gers2000learning}. At recurrent step $j$, the first layer receives
$\boldsymbol{x}^{(j)}
=[E^{(j)},\tau_R^{(j)},RR_{\mathrm{last}}^{(j)},B^{(j)}]^\top$ containing four input features at sample index $j$.
The subsequent LSTM updates are
\begin{equation}
\begin{aligned}
\boldsymbol{i}^{(j)}
&=\operatorname{sigm}\!\left(
\boldsymbol{W}_{i}\boldsymbol{x}^{(j)}
+\boldsymbol{U}_{i}\boldsymbol{h}^{(j-1)}
+\boldsymbol{b}_{i}\right),\\
\boldsymbol{f}^{(j)}
&=\operatorname{sigm}\!\left(
\boldsymbol{W}_{f}\boldsymbol{x}^{(j)}
+\boldsymbol{U}_{f}\boldsymbol{h}^{(j-1)}
+\boldsymbol{b}_{f}\right),\\
\boldsymbol{o}^{(j)}
&=\operatorname{sigm}\!\left(
\boldsymbol{W}_{o}\boldsymbol{x}^{(j)}
+\boldsymbol{U}_{o}\boldsymbol{h}^{(j-1)}
+\boldsymbol{b}_{o}\right),\\
\widetilde{\boldsymbol{c}}^{(j)}
&=\tanh\!\left(
\boldsymbol{W}_{c}\boldsymbol{x}^{(j)}
+\boldsymbol{U}_{c}\boldsymbol{h}^{(j-1)}
+\boldsymbol{b}_{c}\right),\\
\boldsymbol{c}^{(j)}
&=\boldsymbol{f}^{(j)}\odot\boldsymbol{c}^{(j-1)}
+\boldsymbol{i}^{(j)}\odot\widetilde{\boldsymbol{c}}^{(j)},\\
\boldsymbol{h}^{(j)}
&=\boldsymbol{o}^{(j)}
\odot\tanh\!\left(\boldsymbol{c}^{(j)}\right).
\end{aligned}
\label{eq:ecg_lstm_updates}
\end{equation}
Here, $\operatorname{sigm}$ denotes the logistic sigmoid function,
$\odot$ denotes element-wise multiplication,
$\boldsymbol{i}^{(j)}$, $\boldsymbol{f}^{(j)}$, and
$\boldsymbol{o}^{(j)}$ are the input, forget, and output gates;
$\boldsymbol{c}^{(j)}$ and $\boldsymbol{h}^{(j)}$ are the cell and hidden
states. For $q\in\{i,f,o,c\}$, $\boldsymbol{W}_q$ denotes the
input-to-hidden weight matrix, $\boldsymbol{U}_q$ denotes the
hidden-to-hidden recurrent weight matrix, and $\boldsymbol{b}_q$
denotes the corresponding bias vector. With $h$ hidden units and
input dimension $d$, their dimensions are
$\boldsymbol{W}_q\in\mathbb{R}^{h\times d}$,
$\boldsymbol{U}_q\in\mathbb{R}^{h\times h}$, and
$\boldsymbol{b}_q\in\mathbb{R}^{h}$.

At forecast origin $k$, a Bayesian linear output layer maps the final hidden
state of layer $N_{LSTM}$ to the complete future ECG trajectory:
\begin{equation}
\begin{aligned}
\widehat{E}^{(k+1:k+64)}
&=\mathcal{L}_{w,\xi}\!\left(
E^{(k-255:k)},\tau_R^{(k-255:k)},
RR_{\mathrm{last}}^{(k-255:k)},B^{(k-255:k)}\right)\\
\end{aligned}
\label{eq:ecg_operator}
\end{equation}
The input has dimensions $256\times4$, and the output contains 64 future
ECG values. The LSTM weights are assigned independent
Gaussian variational posteriors and trained using the Bayesian learning in Section~\ref{sec:methods}. 

\subsubsection{Population training and architecture ranking}

Candidate Bayesian LSTM architectures are organized into two ordered
capacity levels according to their number of trainable parameters,
as summarized in Table~\ref{tab:ecg_capacity}. Here, $W$ counts each network weight and bias once, including those in the recurrent layers and the output layer. Every recurrent and output-layer weight is assigned an independent Gaussian variational posterior. 

\begin{footnotesize}
\begin{longtable}{ccccc}
\caption{Capacity levels and parameter ranges for the Bayesian LSTM architecture.
}
\label{tab:ecg_capacity}\\
\toprule
\textbf{\shortstack{Capacity\\level $l$}} &
\textbf{\shortstack{Parameter\\range}} &
\textbf{\shortstack{Number of\\architectures}} &
\textbf{\shortstack{Number of\\layers $N_{LSTM}$}} &
\textbf{\shortstack{Hidden units\\per layer $h$}} \\
\midrule
\endfirsthead

\toprule
\textbf{\shortstack{Capacity\\level $l$}} &
\textbf{\shortstack{Parameter\\range}} &
\textbf{\shortstack{Number of\\architectures}} &
\textbf{\shortstack{Number of\\LSTM layers $L$}} &
\textbf{\shortstack{Hidden units\\per layer $h$}} \\
\midrule
\endhead

\endfoot
\bottomrule
\endlastfoot

1 & $\le 1{,}500$   & 19 & $[1,3]$ & $[1,8]$  \\
2 & $1{,}500$--$5{,}000$ & 22 & $[1,3]$ & $[6,17]$ \\
\end{longtable}
\end{footnotesize}

Five synthetic ECG patients, each providing 120 seconds (2 minutes) of data, were used to construct the population-based dataset. 
Within each capacity level, Optuna is used to explore candidate
architectures. Evaluated candidates are trained on the population
data and ranked by approximate log-evidence using the
architecture-level approximation.
The highest-evidence candidate found within each level is
shortlisted for subject-specific adaptation and validation.
The posterior distribution of weights obtained from population-based training was then used as the prior for the sixth patient. Subject-specific adaptation used approximately 4 minutes of data, while 3 seconds of data was reserved for validation using the prescribed NLPD criterion.
Table~\ref{tab:ecg_selected} summarizes the most plausible architectures for both levels. 

\begin{table}[htbp]
\centering
\small
\caption{Highest-evidence Bayesian LSTM architecture in each capacity level.}
\label{tab:ecg_selected}
\begin{tabular}{lrrr}
\toprule
\textbf{\shortstack{Architecture\\ID}} &
\textbf{\shortstack{Parameter\\count $W$}} &
\textbf{\shortstack{LSTM\\layers $N_{LSTM}$}} &
\textbf{\shortstack{Hidden units\\per layer $h$}} \\
\midrule
$Level_1Arch_{19}$   & 1,408 & 1 & 8  \\
$Level_2Arch_{21}$ & 3,964 & 2 & 13 \\
\bottomrule
\end{tabular}
\end{table}

\subsubsection{Subject-specific validation results}
The study-specific NLPD tolerance is set to
$\mathrm{Tol}=-0.5$.
A forecast window passed when its mean NLPD was at or below this threshold and failed otherwise; architecture acceptance required every evaluated window to pass. Over the selected three-second validation interval, thirteen forecast windows were evaluated. As shown in Figure \ref{fig:pred_ecg}, the selected Level 1 architecture produced eight passing windows and five failing windows, marked in blue and red, respectively, and therefore failed validation. Figure \ref{fig:pred_ecg} shows that the selected Level 2 architecture passed all thirteen windows, with no tolerance violations, and consequently satisfied the validation criterion over the evaluated interval. 
EVIDENT therefore accepts \textit{Level2\_Arch21}, with 3,964
trainable parameters, at capacity level 2, the first level whose
evaluated representative satisfies the prescribed criterion.
%
% To examine dependence on the capacity partition,
% Table~\ref{tab:ecg_category_comparison} compares the original two-level
% organization with a single level containing the same 41 evaluated
% architectures. Ranking uses raw approximate log-evidence computed
% on the same population data, while adaptation, validation, and the
% tolerance are held fixed. Both partitions select
% \textit{Level2\_Arch21} indicating that in this particular problem the levels changes the shortlisting
% sequence but not the accepted architecture in this evaluated architecture pool.
%
This result demonstrates implementation of the selection procedure
for synthetic ECG forecasting with Bayesian LSTMs; it does not
establish performance on recorded ECG or superiority over other
forecasting architectures.

\begin{figure}[!h]
\centering
\begin{minipage}{0.49\linewidth}
  \centering
  \includegraphics[width=\linewidth]{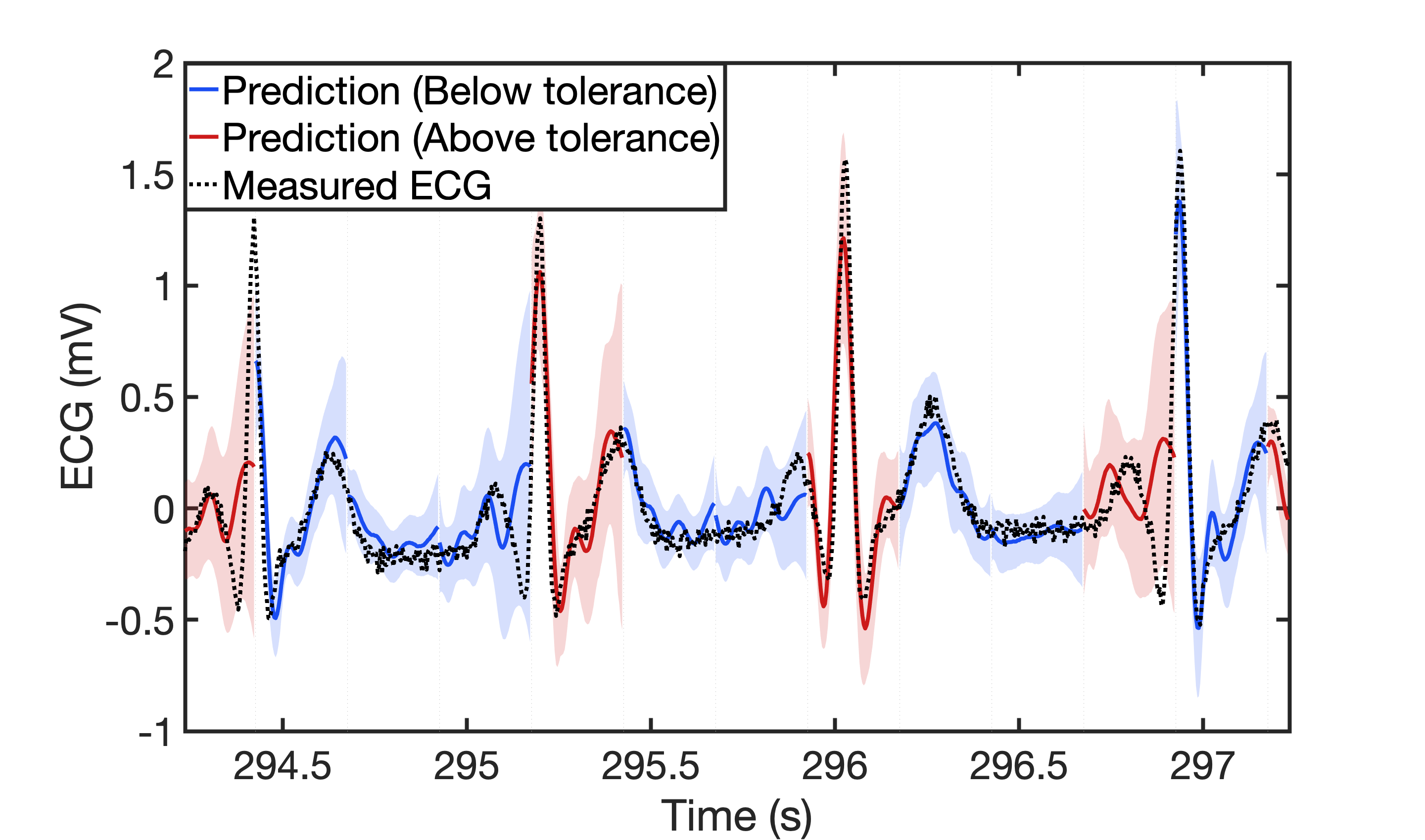}
  \par\smallskip (a)
\end{minipage}
\hfill
\begin{minipage}{0.49\linewidth}
  \centering
  \includegraphics[width=\linewidth]{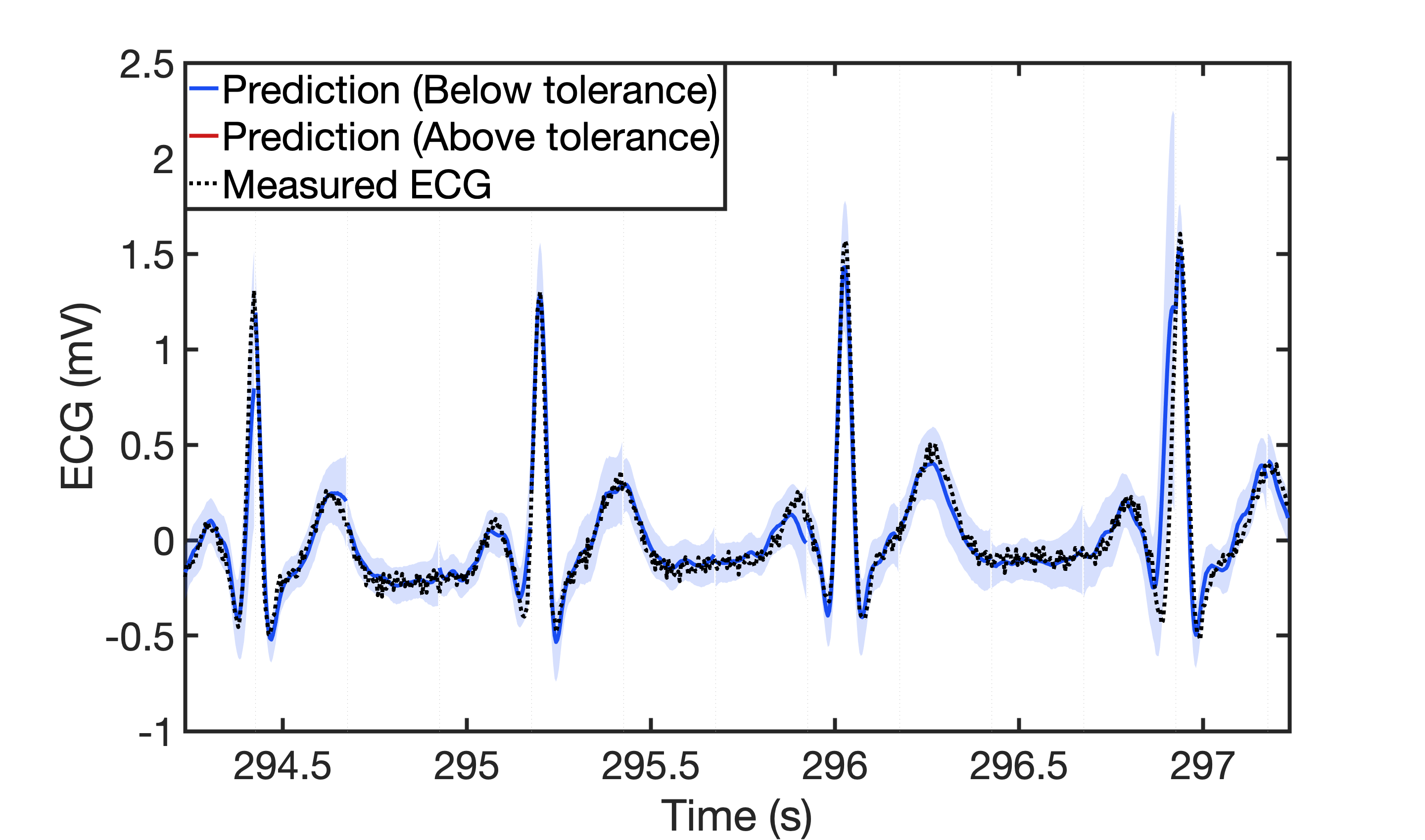}
  \par\smallskip (b)
\end{minipage}
\caption{ECG validation on held-out synthetic subject~6, excluded from population training, over a three-second validation interval using the evidence-selected architectures: (a) $Level_1Arch_{19}$ and (b) $Level_2Arch_{21}$. The dotted black curve represents the synthetic ECG observations, solid curves represent predictive means. Blue and red segments denote forecast windows with mean NLPD that is at or below, or above $Tol=-0.5$, respectively.}
\label{fig:pred_ecg}
\end{figure}

}

\paragraph{\blue{Sensitivity and component analyses}}
\blue{Three analyses examine the selection and validation decisions
in this experiment. First, merging the two capacity levels into
a single level containing the same 41 evaluated architectures
leaves the accepted architecture, $Level_2Arch_{21}$, unchanged.
Ranking by raw approximate log-evidence shortlists this candidate
directly, changing the selection sequence
(\ref{app:ecg_partition}).
Second, replacing evidence-based ranking with population-training
RMSE selects the same representative in each capacity level and
therefore yields the same acceptance outcome
(\ref{app:ecg_ranking}).
Third, omitting model predictive variance from the validation score,
while retaining predictive means and observation-noise variance,
changes the NLPD values but leaves the rejected-window counts
unchanged at five and zero for the level~1 and level~2
representatives, respectively
(\ref{app:ecg_variance}).
The latter diagnostic holds the fitted models, validation windows,
and tolerance fixed and examines only the contribution of model
variance to scoring.}

%========================================================
% Discussion
%========================================================
\section{Discussion}\label{sec:discussion}

%-------------------------------------------------------
% 1. Summarize the key methods and results (one-two paragraphs)
%-------------------------------------------------------
\blue{The primary methodological contribution of this work is a
temporal neural architecture-selection procedure that seeks a balance
between representational capacity, support from available data, and
forecasting adequacy. Ordered capacity levels structure exploration
from less to more expressive architectures, while approximate Bayesian
evidence ranks evaluated candidates within each level under the assumed
prior and likelihood. Task-specific validation separately assesses
forecasting adequacy under predictive uncertainty and governs
acceptance, progression, and stopping. These decisions serve distinct
purposes, that is an evidence-favored candidate can fail the forecasting
requirement, prompting progression when no shortlisted candidate
is accepted at the current level. Building on established Bayesian
model comparison and validation principles \cite{mackay1992practical, oden2017predictive}, EVIDENT makes these
decisions operational within a common procedure for temporal
forecasting. Architecture choice therefore reflects both data supported
flexibility and performance under the prescribed forecasting protocol.}
\blue{The numerical studies provide complementary evidence for this framework. The preliminary analysis illustrates the non-uniform dependence of approximate evidence on TCN architecture and temporal forecasting configuration. In the in silico glucose study, accepted TCN predictors occur within an intermediate-capacity regime, while selected candidates at lower and higher capacities fail the prescribed validation criterion. 
This behavior is consistent with the balance motivating
EVIDENT, such that insufficient capacity can restrict the representation of
temporal dynamics, whereas additional flexibility need not improve
forecasting adequacy under the available data. We note that the observed
intermediate-capacity regime is specific to this candidate pool
and forecasting scenario, rather than a general property of
intermediate-sized networks.
The recorded patient glucose study using the OhioT1DM dataset
extends the evaluation to measured patient records, with architecture
acceptance assessed after adaptation to a subject excluded from
population training. The synthetic ECG study demonstrates the
procedure with Bayesian LSTMs, extending its implementation to
another signal and network family.}

%-------------------------------------------------------
% 2. Extension and broader interpretation of EVIDENT as a framework (one paragraph)
%-------------------------------------------------------
\blue{EVIDENT's modular structure accommodates different candidate
search procedures, predictor families, and validation protocols.
Within-level exploration uses enumeration in the in silico glucose
study and neural architecture search in the recorded patient glucose
and synthetic ECG studies. Ordered capacity levels permit stopping
once a shortlisted candidate is accepted, without exploring higher
levels. The selected architecture depends on the capacity partition
and the candidates evaluated.
The plausibility-weighted ensemble in the in silico glucose study
illustrates a further extension, combining predictions from competitive
architectures while accounting for variation across them.
In the evaluated fold, it improves predictive accuracy and NLPD
relative to the individual predictors considered.
Whether predictions come from a single architecture or an ensemble,
acceptance remains governed by the task-specific validation requirement.}

%-------------------------------------------------------
% 3. Limitations of the study and possible extensions (one paragraph)
%-------------------------------------------------------
\blue{Several limitations motivate further evaluation and methodological
development. All studies explore finite architecture spaces with
prescribed capacity levels and search settings. The integration of
neural architecture search provides a methodological foundation for
extending EVIDENT to substantially larger spaces. Further evaluation
should examine computational scaling, variability across search
realizations, and the effects of alternative capacity partitions
\cite{optuna_2019,elsken2019neural}.
Acceptance also depends on the problem specific validation tolerance, whose numerical
meaning is specific to the signal scaling, scoring convention,
and forecasting scenario.}
Moreover, the Bayesian formulation employed in this work relies on several approximations. The posterior over network weights is approximated using mean-field variational inference, while the model evidence is estimated through a Laplace approximation based on the \blue{variational posterior}. Although these approximations provide a computationally tractable alternative to exact Bayesian inference, they can influence the resulting plausibility landscape and, consequently, the ranking of candidate architectures. 
Future work will examine richer variational families and
sampling-based inference, including Markov chain Monte Carlo
\cite{psaros2023uncertainty}, together with alternative evidence
estimators, to assess the sensitivity of architecture ranking
to these approximations.
\blue{Finally, the empirical evaluation remains limited in cohort size and
temporal coverage. The recorded patient glucose study using OhioT1D assesses
adaptation and acceptance for one target patient over one validation
day, while the synthetic ECG study uses a short validation interval
from one synthetic subject. Further studies should include larger
cohorts, longer observation periods, recorded ECG data, and systematic
assessment of missing observations, temporal misalignment, and
changing forecasting conditions.
One direction is reconstruction of incomplete or
irregularly sampled inputs on a common time grid using probabilistic models, with reconstruction
uncertainty propagated through the forecaster to 
validation under uncertainty. 
These extensions require separate evaluation,
particularly for unrecorded events that cannot be reliably
recovered from the available observations.
The independent day 6 evaluation in in silico glucose study provides a separate assessment
with selected architectures and adapted parameters held fixed,
but is limited to the two subjects in Fold~4.
Broader independent evaluation and comparisons with alternative
NAS methods and established forecasters under common data protocols
would further establish the framework's practical scope.
Clinical safety and suitability for treatment decisions remain
unaddressed in this work, requiring multiple validation scenarios and metrics for different temporal regimes of the forecaster.}

%========================================================
% Conclusions
%========================================================
\section{Conclusions}\label{sec:conclusions}
% Summarize the main take-home points from the study
\blue{This paper introduced EVIDENT, an architecture selection
framework that seeks a balance between representational capacity,
support from available data, and forecasting adequacy and reliability.
Approximate Bayesian evidence prioritizes candidates within capacity
levels, while task-specific validation under predictive uncertainty
governs acceptance, progression, and stopping.
Glucose and synthetic ECG studies demonstrate this procedure with
Bayesian TCNs and LSTMs, illustrating that relative model support
and forecasting acceptance are distinct assessments.
Future work will evaluate broader architecture
exploration and missing data in other time-series applications.}

%++++++++++++++++++++++++++++++++++++++++++++++++++++++++++++++++++++++++
\section*{Acknowledgments}
Md Azharul Islam, Tarunraj Singh, and Danial Faghihi acknowledge support from the National Science Foundation (NSF) under award number DMS-2533946. 
\blue{The authors thank Karim Soliman of the Control, Dynamics and Estimation Lab at the University at Buffalo, for providing the curated OhioT1DM continuous glucose monitoring data used in the recorded-patient glucose study.}
The authors also acknowledge computational support provided by the Center for Computational Research at the University at Buffalo.

\section*{Funding}
This work was supported by the NSF under award number DMS-2533946. NSF had no role in the analysis, interpretation of results, writing of the manuscript, or decision to submit the article for publication.

\section*{Declaration of competing interest}
The authors declare that they have no known competing financial interests or personal relationships that could have appeared to influence the work reported in this paper.

\section*{Data and code availability}
\blue{This study uses synthetic glucose data generated with the
Bergman minimal model and the UVA/Padova Type 1 Diabetes simulator,
recorded patient data from the OhioT1DM dataset, and synthetic ECG
data generated using the model described in this manuscript.
The implementation code, synthetic Bergman glucose data, and
synthetic ECG data are publicly available at
\url{https://github.com/PCELab/EVIDENT-Neural-Architecture-Selection}.
The repository does not include the UVA/Padova simulator or its
generated data, or the OhioT1DM records. Access to the UVA/Padova
simulator and OhioT1DM dataset must be obtained from their respective
providers under the applicable licensing and data use requirements.}

\section*{Declaration of generative AI in the manuscript preparation process}
During the preparation of this work, the authors used Grammarly and ChatGPT to assist with language editing, clarity, and readability. After using this tool, the authors reviewed and edited the manuscript as needed and take full responsibility for the content of the published article.

%========================================================================
% Appendix
% \section{Appendices}\label{sec:appendix}
\appendix

\section{Methodological comparison}
\begin{table}[H]
\centering
\caption{Selection objective, uncertainty treatment, and acceptance rule of representative approaches.}
\label{tab:related_work_comparison}
\small
\setlength{\tabcolsep}{3pt}
\renewcommand{\arraystretch}{1.15}
\begin{tabularx}{\linewidth}{@{}>{\raggedright\arraybackslash}p{0.20\linewidth}*{3}{>{\raggedright\arraybackslash}X}@{}}
\toprule
Approach & Objective & Uncertainty & \shortstack{Acceptance/\\termination} \\
\midrule
ENAS, DARTS~\cite{pham2018efficient,hanxiao2019darts}
& Validation reward/loss & Forecast uncertainty not required & Search budget/convergence \\
Bayesian optimization, BOHB~\cite{snoek2012practical,falkner2018bohb}
& User-specified score & Configuration-performance surrogate & Budget; multi-fidelity pruning \\
Evidence selection~\cite{mackay1992practical,immer2021scalable}
& Marginal likelihood & Bayesian predictions & Evidence-based comparison \\
N-BEATS, Autoformer, TimesNet~\cite{oreshkin2019n,wu2021autoformer,wu2022timesnet}
& Forecast error & Point forecasts in original formulations & Benchmark evaluation; no capacity-level rule \\
DeepAR, Gluformer~\cite{salinas2020deepar,sergazinov2023gluformer}
& Forecast likelihood & Predictive distributions & No prescribed capacity-level acceptance \\
CRNN, DRNN~\cite{li2020crnn,zhu2020dilated}
& Absolute or squared forecast error & Point forecasts & Fixed design or grid search on validation RMSE \\
GluNet~\cite{li2020glunet}
& Log-likelihood of quantized glucose change & Categorical distribution; point evaluation metrics & Validation-tuned; baseline comparison on RMSE, MARD \\
EVIDENT
& Approximate log-evidence & Posterior predictive distribution & First capacity level passing NLPD tolerance \\
\bottomrule
\end{tabularx}
\end{table}

\section{Evidence approximation details}
\label{app:evidence}

This appendix summarizes the Laplace approximations used to evaluate the evidence terms introduced in Section~\ref{sec:plausibility}. 
For a fixed architecture \(\xi\) and inference hyperparameters
\(\sigma=(\sigma_{\mathrm{pr}},\sigma_{\mathrm{noise}})\), the evidence distribution in \eqref{eq:bayes_w} is approximated by
\begin{align}
\log \pi_{\mathrm{evid}}(\mathcal{D}\mid \sigma,\xi)
\approx\;&
-\frac{1}{2\sigma_{\mathrm{noise}}^{2}}
\sum_{k=1}^{N_D}
\left\|
G_{\mathrm{data}}^{(k+1:k+H)}
-
\mathcal{T}_{w_{\mathrm{MAP}},\xi}
\!\big(
G_{\mathrm{data}}^{(k-d_G:k)},
M_{\mathrm{data}}^{(k-d_M:k)},
I_{\mathrm{data}}^{(k-d_I:k)}
\big)
\right\|_2^{2}
\nonumber\\
&-\frac{N_D H}{2}\log\!\big(2\pi\sigma_{\mathrm{noise}}^{2}\big)
-\frac{1}{2\sigma_{\mathrm{pr}}^{2}}\|w_{\mathrm{MAP}}\|_2^{2}
-\frac{W}{2}\log\!\big(2\pi\sigma_{\mathrm{pr}}^{2}\big)
\nonumber\\
&-\frac{1}{2}\log\det\!\left(\frac{P}{2\pi}\right).
\label{eq:log_evid}
\end{align}
Here, \(N_D\) denotes the number of training input--output windows and \(H\) the prediction horizon, so that \(N_DH\) is the total number of blood glucose observations. The data-misfit term is evaluated at \(w_{\mathrm{MAP}}\), approximated by the mean of the variational posterior.
The matrix \(P\) denotes the Hessian of the negative log-posterior with respect to the trainable weights. Rather than computing \(P\) explicitly, we approximate \(P^{-1}\) using the diagonal covariance of the mean-field variational posterior, yielding \(\log\det(P)\) as the negative sum of the logarithms of the posterior variances.

The evidence distribution in \eqref{eq:sigma_post} for architecture comparison under Laplace approximation is obtained by marginalizing the inference hyperparameters around their MAP estimate:
\begin{equation}
\pi_{\mathrm{evid}}(\mathcal{D}\mid \xi)
\approx
\pi_{\mathrm{evid}}(\mathcal{D}\mid \sigma_{\mathrm{MAP}},\xi)\,
\pi_{\mathrm{pr}}(\sigma_{\mathrm{MAP}}\mid \xi)\,
\Big[
2\pi\sqrt{
\mathrm{Var}(\sigma_{\mathrm{pr}}^{2})\,
\mathrm{Var}(\sigma_{\mathrm{noise}}^{2})
}
\Big].
\label{eq:evid3}
\end{equation}
The variances in \eqref{eq:evid3} correspond to the Gaussian approximation of the posterior over the inference hyperparameters.

\section{Single-patient \blue{synthetic glucose} data generation model}
\label{app:bergman}
This appendix provides the simulator equations and parameter values used to generate the single-patient dataset in Figure~\ref{fig:bergman}.
The Bergman states are plasma glucose $G(t)$ [mg/dL], remote insulin action $X(t)$ [min$^{-1}$], and plasma insulin $I(t)$ [mU/L], with basal values $G_b$ and $I_b$. For a Type-1 diabetic (T1D) subject, endogenous pancreatic secretion is removed and insulin is administered externally. The stabilized T1D model is
\begin{align}
\dot G(t) &= -\big(X(t)+p_1\big)\,G(t) + p_1 G_b + \frac{R_{ag}(t)}{V_g}, \label{eq:t1d_bergman_G}\\
\dot X(t) &= -p_2 X(t) + p_3\big(I(t)-I_b\big), \label{eq:t1d_bergman_X}\\
\dot I(t) &= -p_4\big(I(t)-I_b\big) + U(t), \label{eq:t1d_bergman_I}
\end{align}
where $V_g$ [dL] is the glucose distribution volume and $U(t)$ is the insulin control input around basal. This form is obtained by writing the physical infusion as
\begin{equation}
U_0(t) = U(t) + p_4 I_b,
\end{equation}
so that $p_4 I_b$ mimics the basal insulin dose required to prevent the open-loop glucose drift in T1D.
Meal absorption is modeled with stomach--gut compartments whose output is $R_{ag}(t)$. Let $q_{\mathrm{sto}1}(t)$ and $q_{\mathrm{sto}2}(t)$ [mg] denote stomach glucose in solid and liquid phases, and $q_{\mathrm{gut}}(t)$ [mg] denote intestinal glucose. A meal of size $D$ [mg] consumed at time $t_m$ is injected using a Dirac impulse:
\begin{align}
\dot q_{\mathrm{sto}1}(t) &= -k_{21}\,q_{\mathrm{sto}1}(t) + D\,\delta(t-t_m), \label{eq:gut1}\\
\dot q_{\mathrm{sto}2}(t) &= -k_{\mathrm{empt}}\!\big(q_{\mathrm{sto}}(t)\big)\,q_{\mathrm{sto}2}(t) + k_{21}\,q_{\mathrm{sto}1}(t), \label{eq:gut2}\\
\dot q_{\mathrm{gut}}(t) &= -k_{\mathrm{abs}}\,q_{\mathrm{gut}}(t) + k_{\mathrm{empt}}\!\big(q_{\mathrm{sto}}(t)\big)\,q_{\mathrm{sto}2}(t), \label{eq:gut3}\\
R_{ag}(t) &= f\,k_{\mathrm{abs}}\,q_{\mathrm{gut}}(t), \qquad q_{\mathrm{sto}}(t)=q_{\mathrm{sto}1}(t)+q_{\mathrm{sto}2}(t), \label{eq:Ra}
\end{align}
where $k_{21}$ [min$^{-1}$] governs transfer from solid to liquid stomach phase, $k_{\mathrm{abs}}$ [min$^{-1}$] governs intestinal absorption, and $f$ is a dimensionless bioavailability factor. Gastric emptying is modeled as a smooth nonlinearity between $k_{\min}$ and $k_{\max}$:
\begin{align}
k_{\mathrm{empt}}(q_{\mathrm{sto}}) &=
k_{\min} + \tfrac12\big(k_{\max}-k_{\min}\big)\Big(
\tanh\!\big[\alpha(q_{\mathrm{sto}}-bD)\big]
-\tanh\!\big[\beta(q_{\mathrm{sto}}-cD)\big] + 2\Big), \label{eq:kempt}\\
\alpha &= \frac{5}{2D(1-b)}, \qquad
\beta  = \frac{5}{2Dc}, \label{eq:alpha_beta}
\end{align}
with $b,c\in(0,1)$ dimensionless shape parameters. 

\begin{table}[t]
\footnotesize
\centering
\caption{Parameter values for the Type-1 diabetic patient used in simulation.}
\label{tab:t1d_params}
\begin{tabular}{lll}
\hline
\textbf{Parameter} & \textbf{Value} & \textbf{Units} \\
\hline
$p_1$ & $0.028735$ & min$^{-1}$ \\
$p_2$ & $0.028344$ & min$^{-1}$ \\
$p_3$ & $5.035\times 10^{-5}$ & min$^{-2}\cdot$L/mU \\
$p_4$ & $5/54$ & min$^{-1}$ \\
$G_b$ & $119.1858$ & mg/dL \\
$I_b$ & $15.3872$ & mU/L \\
$V_g$ & $128.8237$ & dL \\
$k_{\max}$ & $0.0429$ & min$^{-1}$ \\
$k_{\min}$ & $0.0141$ & min$^{-1}$ \\
$k_{\mathrm{abs}}$ & $0.2062$ & min$^{-1}$ \\
$k_{21}$ & $0.0558$ & min$^{-1}$ \\
$b$ & $0.7612$ & -- \\
$c$ & $0.1372$ & -- \\
$f$ & $0.9$ & -- \\
$\gamma$ & N/A (T1D) & -- \\
$h$ & N/A (T1D) & mg/dL \\
\hline
\end{tabular}
\end{table}

For creating the training data in Figure~\ref{fig:bergman}, 
meal variability is introduced by perturbing both meal size and meal timing. For each meal, the carbohydrate amount is sampled as
\(D \sim \mathcal{U}(0.7D_{\mathrm{nom}},1.3D_{\mathrm{nom}})\),
with \(D_{\mathrm{nom}}\in\{20,40,60\}\) g for breakfast, lunch, and dinner, respectively. Meal timing is independently jittered by an offset sampled from
\(\mathcal{U}(-50,50)\) minutes.
{
\color{blue}
\section{Synthetic ECG data generation}
\label{app:ecg_generation}

This appendix describes the dynamica ECG model and parameter values for generating synthetic
patient records in \ref{sec:synthetic_ecg_data}.

Let $x(t)$ and $y(t)$ describe motion in the limit-cycle plane, and let
$z(t)$ describe the ECG waveform coordinate. The governing equations are
\begin{align}
\dot{x}(t)
&=\alpha(t)x(t)-\omega(t)y(t),
\label{eq:ecg_dx}\\
\dot{y}(t)
&=\alpha(t)y(t)+\omega(t)x(t),
\label{eq:ecg_dy}\\
\dot{z}(t)
&=-\sum_{r\in\{P,Q,R,S,T\}}
a_r\Delta\theta_r(t)
\exp\!\left[-\frac{\Delta\theta_r(t)^2}{2b_r^2}\right]
-\bigl[z(t)-z_0(t)\bigr],
\label{eq:ecg_dz}
\end{align}
where
\begin{equation}
\alpha(t)=1-\sqrt{x(t)^2+y(t)^2},
\qquad
\theta(t)=\operatorname{atan2}\!\bigl(y(t),x(t)\bigr).
\label{eq:ecg_phase}
\end{equation}
The quantity $\Delta\theta_r(t)$ is the phase difference
$\theta(t)-\theta_r$, wrapped to $[-\pi,\pi)$. For each waveform component
$r$, $\theta_r$ specifies its phase location, $a_r$ its signed strength,
and $b_r$ its phase width. Patient-specific morphology is represented
through variations in these parameters. Respiratory baseline variation
is modeled as
\begin{equation}
z_0(t)=A_{\mathrm{resp}}\sin(2\pi f_{\mathrm{resp}}t),
\label{eq:ecg_baseline}
\end{equation}
where $A_{\mathrm{resp}}$ and $f_{\mathrm{resp}}$ are the baseline amplitude
and respiratory frequency, respectively.

The instantaneous angular frequency is determined by the RR process:
\begin{equation}
\omega(t)=\frac{2\pi}{RR(t)}.
\label{eq:ecg_omega}
\end{equation}
In the underlying McSharry formulation, the RR process has a bimodal
power spectrum with low-frequency (LF) and high-frequency (HF) components:
\begin{equation}
S_{RR}(f)=
\sum_{\nu\in\{\mathrm{LF},\mathrm{HF}\}}
\frac{P_{\nu}}{\sqrt{2\pi}\,d_{\nu}}
\exp\!\left[-\frac{(f-f_{\nu})^2}{2d_{\nu}^{2}}\right],
\qquad
\rho_{\mathrm{LF/HF}}=\frac{P_{\mathrm{LF}}}{P_{\mathrm{HF}}}.
\label{eq:ecg_rr_spectrum}
\end{equation}
Here, $P_{\nu}$, $f_{\nu}$, and $d_{\nu}$ denote the power, central
frequency, and spectral width of component $\nu$. A real-valued process
is constructed by inverse Fourier transformation using random phases and
conjugate symmetry. Its mean and standard deviation are adjusted to the
patient-specific RR statistics. The nominal mean RR interval is
$\overline{RR}=60/\overline{HR}$ seconds when the nominal heart rate
$\overline{HR}$ is expressed in beats/min. The HF component is associated
with respiration, with $f_{\mathrm{HF}}=f_{\mathrm{resp}}$.

Patient heterogeneity is introduced by independently drawing the
heart-rate, heart-rate-variability, respiratory, morphology, voltage-gain,
and measurement-noise parameters.
Table~\ref{tab:ecg_generation_parameters} summarizes the reported numerical
distributions.

\begin{table}[htbp]
\centering
\caption{Reported parameter distributions for the synthetic ECG patients.
In $\mathcal{N}(\mu,\sigma^2)$, the second argument is the variance;
$\mathcal{U}$ denotes a uniform distribution.}
\label{tab:ecg_generation_parameters}
\begin{tabular}{lll}
\toprule
Parameter & Distribution & Units\\
\midrule
Mean heart rate $\overline{HR}$
& $\mathcal{U}(55,85)$ & beats/min\\
Normal-RR standard deviation
& $\mathcal{N}(50,30^2)$ & ms\\
LF/HF power ratio $\rho_{\mathrm{LF/HF}}$
& $\mathcal{N}(2,1.5^2)$ & --\\
Respiratory frequency $f_{\mathrm{resp}}$
& $\mathcal{U}(0.20,0.33)$ & Hz\\
LF frequency $f_{\mathrm{LF}}$
& $\mathcal{U}(0.08,0.12)$ & Hz\\
Respiratory baseline amplitude $A_{\mathrm{resp}}$
& $\mathcal{U}(0.05,0.20)$ & mV\\
Voltage gain $g$
& $\mathcal{U}(0.75,1.30)$ & --\\
ECG noise standard deviation $\sigma_E$
& $\mathcal{U}(0.015,0.035)$ & mV\\
\bottomrule
\end{tabular}
\end{table}
}

%=================================
\section{\blue{Sensitivity and component analyses}}
\label{app:analyses}
\blue{This appendix examines the effects of validation tolerance,
capacity partitioning, ranking criterion, and model predictive
variance on architecture selection and acceptance. Each analysis
retains the other elements of its corresponding forecasting protocol.}

\subsection{\blue{Sensitivity of glucose forecasting acceptance to tolerance}}
\label{app:tolerance}
% new table
\blue{We first examine tolerance-dependent stopping for the
evaluated representatives of capacity levels~1 and~2 in Fold~4
of the in silico glucose study. Table~\ref{tab:uva_max_nlpd}
reports their maximum day-5 window NLPD for each held-out patient
and across both patients. Because acceptance requires every
window for both patients to satisfy the tolerance, these maxima
determine the acceptance boundary for each representative.}
%%%

\begin{table}[h]
\centering
\small
\caption{Maximum day-5 windows NLPD of the most plausible architectures in capacity levels 1 and 2, separately
for each held-out patient in Fold~4 and across both patients.}
\label{tab:uva_max_nlpd}
\setlength{\tabcolsep}{4pt}
\begin{tabular}{llccc}
\toprule
\textbf{\shortstack{Level\\$l$}} &
\textbf{\shortstack{Architecture\\ID}} &
\textbf{\shortstack{Max.\ NLPD\\patient~3}} &
\textbf{\shortstack{Max.\ NLPD\\patient~4}} &
\textbf{\shortstack{Max.\ NLPD\\both patients}} \\
\midrule
1 & $Level_1Arch_8$ & 19.84 & 50.7  & 50.7  \\
2 & $Level_2Arch_4$ & 13.74 & 14.41 & 14.41 \\
\bottomrule
\end{tabular}
\end{table}

\blue{Based on the reported scores, EVIDENT stops at capacity
level~1 for $Tol\ge50.70$. For $14.41\le Tol<50.70$, the level~1
representative is rejected and the level~2 representative is
accepted, so the procedure stops at level~2. The reference
tolerance $Tol=14.5$ lies within this latter range.}

%%%
\blue{We additionally examine acceptance sensitivity for two fixed
predictors: $Level_2Arch_4$ for patient~1 in the in silico glucose
study and $Level_4Arch_{39}$ for patient~588 in the recorded-patient
glucose study, as reported in
Tables~\ref{tab:uva_tol_sensitivity} and~\ref{tab:ohio_tol_sensitivity},
respectively. Predictions and validation windows are held fixed
while the tolerance is varied, with acceptance requiring all
24 windows to satisfy the criterion. These two tables report
predictor acceptance and rejected-window counts, rather than
a rerun of architecture selection.}
\begin{table}[h]
\centering
\small
\caption{Sensitivity of UVA/Padova architecture acceptance to
the NLPD tolerance $Tol$. The originally validated architecture
$Level_2Arch_4$, with patient 1 was used in this study.}
\label{tab:uva_tol_sensitivity}
\begin{tabular}{crlcl}
\toprule
\textbf{Tolerance} &
\textbf{\shortstack{Windows\\ Rejected/total}} &
\textbf{\shortstack{Acceptance}} &
\textbf{\shortstack{Capacity\\level $l$}} &
\textbf{\shortstack{Evaluated\\architecture}} \\
\midrule
6.0           & 17/24 & Rejected & 2 & $Level_2Arch_4$ \\
8.0           &  6/24 & Rejected & 2 & $Level_2Arch_4$ \\
12.5          &  0/24 & Accepted & 2    & $Level_2Arch_4$ \\
\textbf{14.5} &  \textbf{0/24} & \textbf{Accepted} & \textbf{2}    & \textbf{$Level_2Arch_4$} \\
16.5          &  0/24 & Accepted & 2    & $Level_2Arch_4$ \\
\bottomrule
\end{tabular}
\end{table}

\begin{table}[h]
\centering
\small
\caption{Sensitivity of OhioT1DM architecture acceptance to
the NLPD tolerance $Tol$. The originally
validated architecture $Level_4Arch_{39}$ is evaluated on
patient~588.}
\label{tab:ohio_tol_sensitivity}
\begin{tabular}{crlcl}
\toprule
\textbf{Tolerance} &
\textbf{\shortstack{Windows\\ Rejected/total}} &
\textbf{Acceptance} &
\textbf{\shortstack{Capacity\\level $l$}} &
\textbf{\shortstack{Evaluated\\architecture}} \\
\midrule
8.0         & 17/24 & Rejected & 4 & $Level_4Arch_{39}$ \\
10.0        &  5/24 & Rejected & 4 & $Level_4Arch_{39}$ \\
12.0        &  2/24 & Rejected & 4 & $Level_4Arch_{39}$ \\
\textbf{15} &  \textbf{0/24} & \textbf{Accepted} & \textbf{4}    & \textbf{$Level_4Arch_{39}$} \\
20.0        &  0/24 & Accepted & 4    & $Level_4Arch_{39}$ \\
\bottomrule
\end{tabular}
\end{table}

\blue{For patient~1, $Level_2Arch_4$ satisfies the criterion at
tolerances of 12.5, 14.5, and 16.5
(Table~\ref{tab:uva_tol_sensitivity}). For patient~588,
$Level_4Arch_{39}$ satisfies it at 15 and 20
(Table~\ref{tab:ohio_tol_sensitivity}). Both predictors are rejected
at the smaller tolerances examined. These patient-specific
acceptance results complement the Fold~4 stopping analysis above.
For $Tol<14.41$, neither representative is accepted and
further exploration would be required. These boundaries are
approximate to the displayed precision and conditional on the
evaluated representatives and Fold~4 validation data.}

\subsection{\blue{Sensitivity to the ECG capacity partition}}
\label{app:ecg_partition}
\blue{We compare two partitions of the same 41 evaluated Bayesian
LSTM architectures, consisting of the original two capacity levels containing
19 and 22 candidates, and a single level containing all candidates.
The comparison reuses the population-training results and ranks
candidates by approximate log-evidence, rather than plausibility
weights normalized separately within the original levels.
Subject-specific adaptation, validation windows, 
and the tolerance remain unchanged.}

\begin{table}[H]
\centering
\small
\caption{Comparison of two capacity partitions for the
synthetic ECG study using the same evaluated architecture pool. Validation outcomes reuse the
original protocol with $Tol=-0.5$.}
\label{tab:ecg_category_comparison}
\begin{tabular}{lcrlrl}
\toprule
\textbf{\shortstack{Capacity\\ partition}} &
\textbf{\shortstack{level\\$l$}} &
\textbf{\shortstack{Evaluated\\architectures}} &
\textbf{\shortstack{Highest-evidence\\architecture}} &
\textbf{Log-evidence} &
\textbf{Validation} \\
\midrule
Two capacity levels & 1 & 19 &
$Level_1Arch_{19}$ & 143,005.354 & Rejected \\

Two capacity levels & 2 & 22 &
$Level_2Arch_{21}$ & 210,660.035 & Accepted \\
\midrule
One capacity level & 1 & 41 &
$Level_2Arch_{21}$ & 210,660.035 & Accepted \\
\bottomrule
\end{tabular}
\end{table}

\blue{Under the original partition, $Level_1Arch_{19}$ fails validation
and the procedure advances to the second level, where
$Level_2Arch_{21}$ is accepted. Under the merged partition,
$Level_2Arch_{21}$ has the largest approximate log-evidence among
the 41 evaluated candidates and is shortlisted directly.
Thus, the partition changes the shortlisting sequence but leaves
the accepted architecture unchanged for this candidate pool
and validation scenario.}

\subsection{\blue{Evidence-based versus population-training RMSE architecture ranking}}
\label{app:ecg_ranking}
\blue{To examine the role of the ranking criterion, we replace
evidence-based ranking with population-training RMSE computed
from each candidate's predictive mean. The evaluated candidate pool,
capacity levels, subject specific adaptation, validation protocol,
and stopping rule remain unchanged. Within each level, the candidate
with the smallest population-training RMSE is shortlisted.
Both ranking criteria identify $Level_1Arch_{19}$ and
$Level_2Arch_{21}$ as their respective representatives, yielding
the same validation outcomes and accepted architecture.
This agreement is specific to the evaluated pool and does not
establish general equivalence between the ranking criteria.}

\begin{table}[h]
\centering
\small
\caption{Architecture ranking with population RMSE for ECG dataset. Max NLPD represent highest average NLPD for a prediction window.}
\label{tab:ecg-ranking-ablation}
\begin{tabular}{lclrrcl}
\toprule
\textbf{Ranking} &
\textbf{\shortstack{Level\\$l$}} &
\textbf{\shortstack{Architecture\\ID}} &
\textbf{\shortstack{Parameters\\$W$}} &
\textbf{\shortstack{Max.\\NLPD}} &
\textbf{\shortstack{Windows\\Rejected/total}} &
\textbf{Acceptance} \\
\midrule
Evidence        & 1 & $Level_1Arch_{19}$   & 1,408 & $0.336$  & 5/13 & Rejected \\
Evidence        & 2 & $Level_2Arch_{21}$ & 3,964 & $-0.770$ & 0/13 & Accepted \\
RMSE & 1 & $Level_1Arch_{19}$  & 1,408 & $0.336$  & 5/13 & Rejected \\
RMSE & 2 & $Level_2Arch_{21}$ & 3,964 & $-0.770$ & 0/13 & Accepted \\
\bottomrule
\end{tabular}
\end{table}

\subsection{Contribution of model predictive variance to ECG validation}
\label{app:ecg_variance}
\blue{For the two evidence-selected representatives, we recompute
window-level NLPD after omitting model predictive variance while
retaining the predictive means and observation-noise variance.
The fitted models, validation windows, scoring convention, and
tolerance remain unchanged.
Although the NLPD values change, the numbers of rejected windows
remain five and zero for the level~1 and level~2 representatives,
respectively. This diagnostic isolates the contribution of model
variance to the validation score, but it does not remove Bayesian
training.}

\begin{table}[h]
\centering
\small
\caption{Contribution of model predictive variance to ECG validation.
Window-level NLPD is computed using either data noise alone $\sigma^2_{\mathrm{data}}$ or data noise plus
model predictive variance $\sigma_{i,h}^{2}$ defined
in \eqref{eq:sigma_nlpd},
with predictive means and tolerance held fixed.
}
\label{tab:ecg-variance-ablation}
\begin{tabular}{clcrcl}
\toprule
\textbf{\shortstack{Level\\$l$}} &
\textbf{\shortstack{Architecture\\ID}} &
\textbf{\shortstack{NLPD\\variance}} &
\textbf{\shortstack{Max.\\NLPD}} &
\textbf{\shortstack{Windows\\Rejected/total}} &
\textbf{Acceptance} \\
\midrule
1 & $Level_1Arch_{19}$  & $\sigma_{i,h}^{2}$ & $0.336$  & 5/13 & Rejected \\
1 & $Level_1Arch_{19}$  & $\sigma^2_{\mathrm{data}}$                    & $0.384$  & 5/13 & Rejected \\
2 & $Level_2Arch_{21}$ & $\sigma_{i,h}^{2}$ & $-0.770$ & 0/13 & Accepted \\
2 & $Level_2Arch_{21}$ & $\sigma^2_{\mathrm{data}}$                    & $-0.778$ & 0/13 & Accepted \\
\bottomrule
\end{tabular}
\end{table}

%==========================================
\section{Computational cost}
\label{app:cost}
\blue{
Table~\ref{tab:evident_walltime} distinguishes population training,
evidence estimation, subject-specific adaptation, and validation
for the first two capacity levels of the UVA/Padova experiment.
Computations were performed on the NCSA Delta system using
32 allocated CPU cores on an AMD EPYC 7763 (Milan) processor,
16~GB of allocated host memory, and one NVIDIA A100 GPU with
40~GB of GPU memory. The software environment used the
\texttt{pytorch-conda/2.8} and \texttt{cuda/12.9} modules.
The accepted architecture, $Level_2Arch_4$, has a reported parameter
count of $W=47{,}779$, five temporal blocks per encoder/decoder,
filter size $f=9$, dilation base $b=3$, and encoder channels
$[2,4,8,16,32]$, with reversed decoder channels.
For an input history of 1400 time steps with three input channels
and a one-hour forecast horizon of 60 time steps, posterior
predictive inference uses $50$ samples.
At batch size one, a single forward pass requires approximately
24.519~MFLOPs, corresponding to approximately 1.226~GFLOPs for
50 forward passes. A complete probabilistic forecast, including
the 50 sampled-weight forward passes and computation of the
predictive mean and variance, takes 1.12~s per input window.
}
\begin{table}[h]
\centering
\small
\caption{Computational costs for EVIDENT in the in silico glaucous forecasting study, restricted to capacity Levels~1 and~2.
Population training and evidence estimation use the Fold~4
training set, excluding patients~3 and~4; adaptation and
validation are performed separately for these held-out patients.
}
\label{tab:evident_walltime}
\setlength{\tabcolsep}{5pt}
\renewcommand{\arraystretch}{1.15}
\begin{tabular}{llr}
\toprule
\textbf{Stage} &
\textbf{Scope} &
\textbf{\shortstack{Wall time\\(hh:mm:ss)}} \\
\midrule
Population training
& Levels~1--2, 19 candidates & 15:56:21 \\
Evidence estimation
& Levels~1--2, 19 candidates &  00:40:28\\
\midrule
Subject-specific adaptation
& Level~1, patient~3 & 01:32:16 \\
& Level~1, patient~4 & 01:40:21 \\
& Level~2, patient~3 & 02:24:14 \\
& Level~2, patient~4 & 02:31:45 \\
\midrule
Validation total
& Levels~1--2, patients~3 and~4 & 00:04:00 \\
\cmidrule(lr){2-3}
\textbf{Total EVIDENT time}
& \textbf{Levels~1--2; Patients~3 and~4} & \textbf{24:49:25} \\
\bottomrule
\end{tabular}

\end{table}

\newpage
%==========================================
\section{Architecture specifications and candidate pools}
\label{app:architecture_pool}

\blue{Architecture specifications are grouped by study.
Architecture identifiers are local to each study and should
be interpreted together with the corresponding candidate pool.}

\subsection{In silico glucose study}
This section reports the implementation constraints and detailed architecture specifications used to construct the 61 candidate TCNs summarized in Table~\ref{tab:arch_level}.
Kernel sizes are restricted to ($f \le 15$), which is sufficient to represent local glucose excursions without excessively large convolutional filters. 
Encoder channels double across successive temporal blocks
from the architecture-specific starting width listed in the tables;
decoder channels follow the mirrored schedule.
Tables~\ref{tab:level1}--\ref{tab:level3} summarize the Level~1--3 architectures explored by EVIDENT in the numerical experiments, and Table~\ref{tab:random} lists the 19 architectures used in the \blue{predictive performance-based comparison}.

\begin{footnotesize}
\setlength{\tabcolsep}{5pt}
\renewcommand{\arraystretch}{1.3}
\begin{longtable}{p{2.2cm}p{1.8cm}p{1.6cm}p{1.4cm}p{1.6cm}p{2.0cm}}
\caption{TCN architectures for Level 1. The dilation factor in encoder block $j$ is defined as $\delta_j = b^{j-1}$, where $b$ is the dilation base. The decoder channels $\{\tilde{c}_j\}_{j=1}^{N}$ mirror the encoder channels $\{c_j\}_{j=1}^{N}$ in reverse order.}
\label{tab:level1}\\
\toprule
\textbf{Architecture ID} & \textbf{Parameter count} & \textbf{Temporal blocks $N$} & \textbf{Filter size $f$} & \textbf{Dilation base $b$} & \textbf{Encoder channels $\{c_j\}_{j=1}^{N}$} \\
\midrule
\endfirsthead

\toprule
\textbf{Architecture ID} & \textbf{Parameter count} & \textbf{Temporal blocks $N$} & \textbf{Filter size $f$} & \textbf{Dilation base $b$} & \textbf{Encoder channels $\{c_j\}_{j=1}^{N}$} \\
\midrule
\endhead

\endfoot

\bottomrule
\endlastfoot

$Level_1Arch_1$    & 3035  & 3 & 9  & 9 & $[2, 4, 8]$     \\
$Level_1Arch_2$    & 3671  & 3 & 11 & 8 & $[2, 4, 8]$     \\
$Level_1Arch_3$    & 3671  & 3 & 11 & 9 & $[2, 4, 8]$     \\
$Level_1Arch_4$    & 4307  & 3 & 13 & 8 & $[2, 4, 8]$     \\
$Level_1Arch_5$    & 9463  & 4 & 7  & 5 & $[2, 4, 8, 16]$ \\
$Level_1Arch_6$    & 11845 & 3 & 9  & 9 & $[4, 8, 16]$    \\
$Level_1Arch_7$    & 14365 & 3 & 11 & 8 & $[4, 8, 16]$    \\
$Level_1Arch_8$    & 14365 & 3 & 11 & 9 & $[4, 8, 16]$    \\
$Level_1Arch_9$    & 14575 & 4 & 11 & 4 & $[2, 4, 8, 16]$ \\
$Level_1Arch_{10}$ & 16885 & 3 & 13 & 8 & $[4, 8, 16]$    \\
\end{longtable}
\end{footnotesize}

\begin{footnotesize}
\setlength{\tabcolsep}{5pt}
\renewcommand{\arraystretch}{1.3}
\begin{longtable}{p{2.2cm}p{1.8cm}p{1.6cm}p{1.4cm}p{1.6cm}p{2.0cm}}
\caption{TCN architectures for Level 2. The dilation factor in encoder block $j$ is defined as $\delta_j = b^{j-1}$, where $b$ is the dilation base. The decoder channels $\{\tilde{c}_j\}_{j=1}^{N}$ mirror the encoder channels $\{c_j\}_{j=1}^{N}$ in reverse order.
EVIDENT identifies $Level_2Arch_4$ \blue{as the accepted architecture}.
}
\label{tab:level2}\\
\toprule
\textbf{Architecture ID} & \textbf{Parameter count} & \textbf{Temporal blocks $N$} & \textbf{Filter size $f$} & \textbf{Dilation base $b$} & \textbf{Encoder channels $\{c_j\}_{j=1}^{N}$} \\
\midrule
\endfirsthead

\toprule
\textbf{Architecture ID} & \textbf{Parameter count} & \textbf{Temporal blocks $N$} & \textbf{Filter size $f$} & \textbf{Dilation base $b$} & \textbf{Encoder channels $\{c_j\}_{j=1}^{N}$} \\
\midrule
\endhead

\endfoot

\bottomrule
\endlastfoot

$Level_2Arch_1$ & 37405 & 4 & 7  & 5 & $[4, 8, 16, 32]$        \\
$Level_2Arch_2$ & 37543 & 5 & 7  & 3 & $[2, 4, 8, 16, 32]$     \\
$Level_2Arch_3$ & 46793 & 3 & 9  & 9 & $[8, 16, 32]$           \\
$\mathbf{Level_2Arch_4}$ & \textbf{47779} & \textbf{5} & \textbf{9}  & \textbf{3} & $\textbf{[2, 4, 8, 16, 32]}$     \\
$Level_2Arch_5$ & 56825 & 3 & 11 & 8 & $[8, 16, 32]$           \\
$Level_2Arch_6$ & 56825 & 3 & 11 & 9 & $[8, 16, 32]$           \\
$Level_2Arch_7$ & 57805 & 4 & 11 & 4 & $[4, 8, 16, 32]$        \\
$Level_2Arch_8$ & 59599 & 6 & 3  & 3 & $[2, 4, 8, 16, 32, 64]$ \\
$Level_2Arch_9$ & 62857 & 3 & 13 & 8 & $[8, 16, 32]$           \\
\end{longtable}
\end{footnotesize}

\begin{footnotesize}
\setlength{\tabcolsep}{5pt}
\renewcommand{\arraystretch}{1.3}
\begin{longtable}{p{2.2cm}p{1.8cm}p{1.6cm}p{1.4cm}p{1.6cm}p{2.0cm}}
\caption{TCN architectures for additional diagnostic exploration in Level 3. The dilation factor in encoder block $j$ is defined as $\delta_j = b^{j-1}$, where $b$ is the dilation base. The decoder channels $\{\tilde{c}_j\}_{j=1}^{N}$ mirror the encoder channels $\{c_j\}_{j=1}^{N}$ in reverse order.}
\label{tab:level3}\\
\toprule
\textbf{Architecture ID} & \textbf{Parameter count} & \textbf{Temporal blocks $N$} & \textbf{Filter size $f$} & \textbf{Dilation base $b$} & \textbf{Encoder channels $\{c_j\}_{j=1}^{N}$} \\
\midrule
\endfirsthead
\toprule
\textbf{Architecture ID} & \textbf{Parameter count} & \textbf{Temporal blocks $N$} & \textbf{Filter size $f$} & \textbf{Dilation base $b$} & \textbf{Encoder channels $\{c_j\}_{j=1}^{N}$} \\
\midrule
\endhead
\endfoot
\bottomrule
\endlastfoot
$Level_3Arch_1$    & 148729 & 4 & 7  & 5 & $[8, 16, 32, 64]$         \\
$Level_3Arch_2$    & 149373 & 5 & 7  & 3 & $[4, 8, 16, 32, 64]$      \\
$Level_3Arch_3$    & 186001 & 3 & 9  & 9 & $[16, 32, 64]$            \\
$Level_3Arch_4$    & 190293 & 5 & 9  & 3 & $[4, 8, 16, 32, 64]$      \\
$Level_3Arch_5$    & 226033 & 3 & 11 & 8 & $[16, 32, 64]$            \\
$Level_3Arch_6$    & 226033 & 3 & 11 & 9 & $[16, 32, 64]$            \\
$Level_3Arch_7$    & 230233 & 4 & 11 & 4 & $[8, 16, 32, 64]$         \\
$Level_3Arch_8$    & 268941 & 6 & 3  & 3 & $[4, 8, 16, 32, 64, 128]$ \\
$Level_3Arch_9$    & 266065 & 3 & 13 & 8 & $[16, 32, 64]$            \\
$Level_3Arch_{10}$ & 272379 & 6 & 13 & 2 & $[2, 4, 8, 16, 32, 64]$   \\
\end{longtable}
\end{footnotesize}

\subsection{Fixed 19-architecture comparison subset}

\begin{footnotesize}
\setlength{\tabcolsep}{4pt}
\renewcommand{\arraystretch}{1.3}
\begin{longtable}{p{3.0cm}p{2.0cm}p{1.8cm}p{1.7cm}p{1.3cm}p{1.4cm}p{2.2cm}}
\caption{\blue{Fixed 19-architecture subset used for the illustrative
comparison with predictive performance-based selection.
The subset is drawn from the 61-candidate feasible pool and includes
$RandArch_6=Level_2Arch_4$ by construction.} }
\label{tab:random}\\
\toprule
\textbf{Architecture ID} &
\textbf{EVIDENT ID} &
\textbf{Parameter count} &
\textbf{Temporal blocks $N$} &
\textbf{Filter size $f$} &
\textbf{Dilation base $b$} &
\textbf{Encoder channels $\{c_j\}_{j=1}^{N}$} \\
\midrule
\endfirsthead

\toprule
\textbf{Random-search Architecture ID} &
\textbf{EVIDENT ID} &
\textbf{Parameter count} &
\textbf{Temporal blocks $N$} &
\textbf{Filter size $f$} &
\textbf{Dilation base $b$} &
\textbf{Encoder channels $\{c_j\}_{j=1}^{N}$} \\
\midrule
\endhead

\endfoot

\bottomrule
\endlastfoot

$RandArch_1$     & $Level_1Arch_1$     & 3035     & 3 & 9  & 9 & $[2, 4, 8]$ \\
$RandArch_2$     & $Level_1Arch_3$     & 3671     & 3 & 11 & 9 & $[2, 4, 8]$ \\
$RandArch_3$     & $Level_1Arch_4$     & 4307     & 3 & 13 & 8 & $[2, 4, 8]$ \\
$RandArch_4$     & $Level_1Arch_9$     & 14575    & 4 & 11 & 4 & $[2, 4, 8, 16]$ \\

$RandArch_5$     & $Level_2Arch_1$     & 37405    & 4 & 7  & 5 & $[4, 8, 16, 32]$ \\
$RandArch_6$     & $Level_2Arch_4$     & 47779    & 5 & 9  & 3 & $[2, 4, 8, 16, 32]$ \\
$RandArch_7$     & $Level_2Arch_8$     & 59599    & 6 & 3  & 3 & $[2, 4, 8, 16, 32, 64]$ \\

$\mathbf{RandArch_8}$ & $\mathbf{Level_3Arch_1}$ & \textbf{148729} & \textbf{4} & \textbf{7} & \textbf{5} & \textbf{[8,16,32,64]} \\
$RandArch_9$     & $Level_3Arch_4$     & 190293   & 5 & 9  & 3 & $[4, 8, 16, 32, 64]$ \\
$RandArch_{10}$  & $Level_3Arch_5$     & 226033   & 3 & 11 & 8 & $[16, 32, 64]$ \\
$RandArch_{11}$  & $Level_3Arch_8$     & 268941   & 6 & 3  & 3 & $[4, 8, 16, 32, 64, 128]$ \\

$RandArch_{12}$  & $Level_4Arch_1$     & 593137   & 4 & 7  & 5 & $[16, 32, 64, 128]$ \\
$RandArch_{13}$  & $Level_4Arch_4$     & 741665   & 3 & 9  & 9 & $[32, 64, 128]$ \\
$RandArch_{14}$  & $Level_4Arch_5$     & 759529   & 5 & 9  & 3 & $[8, 16, 32, 64, 128]$ \\

$RandArch_{15}$  & $Level_5Arch_1$     & 2368993  & 4 & 7  & 5 & $[32, 64, 128, 256]$ \\
$RandArch_{16}$  & $Level_5Arch_5$     & 3034833  & 5 & 9  & 3 & $[16, 32, 64, 128, 256]$ \\
$RandArch_{17}$  & $Level_5Arch_6$     & 3601345  & 3 & 11 & 8 & $[64, 128, 256]$ \\

$RandArch_{18}$  & $Level_6Arch_4$     & 11838593 & 3 & 9  & 9 & $[128, 256, 512]$ \\
$RandArch_{19}$  & $Level_6Arch_{10}$  & 17388177 & 6 & 13 & 2 & $[16, 32, 64, 128, 256, 512]$ \\
\end{longtable}
\end{footnotesize}

\subsection{Recorded-patient glucose study using OhioT1DM dataset}
\blue{The following tables list the TCN configurations in the
prescribed capacity levels for the recorded patient glucose study.
The reported selection procedure accepts a candidate at level~4;
the inclusion of higher-level configurations here documents the
candidate space and does not imply that they were required
for acceptance.}

{
\color{blue}
\begingroup
\footnotesize
\setlength{\tabcolsep}{3pt}
\renewcommand{\arraystretch}{1.15}
\setlength{\LTleft}{\fill}
\setlength{\LTright}{\fill}

\begin{longtable}{@{}
>{\raggedright\arraybackslash}p{0.21\textwidth}
>{\raggedleft\arraybackslash}p{0.10\textwidth}
>{\centering\arraybackslash}p{0.10\textwidth}
>{\centering\arraybackslash}p{0.083\textwidth}
>{\centering\arraybackslash}p{0.075\textwidth}
>{\centering\arraybackslash}p{0.156\textwidth}
>{\centering\arraybackslash}p{0.156\textwidth}
@{}}

\caption{TCN architectures for OhioT1DM capacity Level~1}
\label{tab:ohio_level1}\\

\toprule
\textbf{\shortstack{Architecture\\ID}} &
\textbf{\shortstack{Parameter\\ Count}}  &
\textbf{\shortstack{Blocks\\$N$}} &
\textbf{\shortstack{Filter\\$f$}} &
\textbf{\shortstack{Base\\$b$}} &
\textbf{\shortstack{Encoder\\channels}} &
\textbf{\shortstack{Decoder\\channels}} \\
\midrule
\endfirsthead

\toprule
\textbf{\shortstack{Architecture\\ID}} &
\textbf{\shortstack{Parameter\\ Count}} &
\textbf{\shortstack{Blocks\\$N$}} &
\textbf{\shortstack{Filter\\$f$}} &
\textbf{\shortstack{Base\\$b$}} &
\textbf{\shortstack{Encoder\\channels}} &
\textbf{\shortstack{Decoder\\channels}} \\
\midrule
\endhead

\midrule
\multicolumn{7}{r}{Continued on next page}\\
\endfoot

\bottomrule
\endlastfoot

$\mathrm{Level}_{1}\mathrm{Arch}_{1}$  & 299   & 2 & 3  & 2 & [2, 4] & [4, 2] \\
$\mathrm{Level}_{1}\mathrm{Arch}_{2}$  & 299   & 2 & 3  & 3 & [2, 4] & [4, 2] \\
$\mathrm{Level}_{1}\mathrm{Arch}_{3}$  & 455   & 2 & 5  & 2 & [2, 4] & [4, 2] \\
$\mathrm{Level}_{1}\mathrm{Arch}_{4}$  & 455   & 2 & 5  & 3 & [2, 4] & [4, 2] \\
$\mathrm{Level}_{1}\mathrm{Arch}_{5}$  & 455   & 2 & 5  & 4 & [2, 4] & [4, 2] \\
$\mathrm{Level}_{1}\mathrm{Arch}_{6}$  & 455   & 2 & 5  & 5 & [2, 4] & [4, 2] \\
$\mathrm{Level}_{1}\mathrm{Arch}_{7}$  & 611   & 2 & 7  & 2 & [2, 4] & [4, 2] \\
$\mathrm{Level}_{1}\mathrm{Arch}_{8}$  & 611   & 2 & 7  & 3 & [2, 4] & [4, 2] \\
$\mathrm{Level}_{1}\mathrm{Arch}_{9}$  & 611   & 2 & 7  & 4 & [2, 4] & [4, 2] \\
$\mathrm{Level}_{1}\mathrm{Arch}_{10}$ & 611   & 2 & 7  & 5 & [2, 4] & [4, 2] \\
$\mathrm{Level}_{1}\mathrm{Arch}_{11}$ & 767   & 2 & 9  & 2 & [2, 4] & [4, 2] \\
$\mathrm{Level}_{1}\mathrm{Arch}_{12}$ & 767   & 2 & 9  & 3 & [2, 4] & [4, 2] \\
$\mathrm{Level}_{1}\mathrm{Arch}_{13}$ & 767   & 2 & 9  & 4 & [2, 4] & [4, 2] \\
$\mathrm{Level}_{1}\mathrm{Arch}_{14}$ & 767   & 2 & 9  & 5 & [2, 4] & [4, 2] \\
$\mathrm{Level}_{1}\mathrm{Arch}_{15}$ & 923   & 2 & 11 & 2 & [2, 4] & [4, 2] \\
$\mathrm{Level}_{1}\mathrm{Arch}_{16}$ & 923   & 2 & 11 & 3 & [2, 4] & [4, 2] \\
$\mathrm{Level}_{1}\mathrm{Arch}_{17}$ & 923   & 2 & 11 & 4 & [2, 4] & [4, 2] \\
$\mathrm{Level}_{1}\mathrm{Arch}_{18}$ & 923   & 2 & 11 & 5 & [2, 4] & [4, 2] \\
$\mathrm{Level}_{1}\mathrm{Arch}_{19}$ & 1,061 & 2 & 3  & 2 & [4, 8] & [8, 4] \\
$\mathrm{Level}_{1}\mathrm{Arch}_{20}$ & 1,061 & 2 & 3  & 3 & [4, 8] & [8, 4] \\
$\mathrm{Level}_{1}\mathrm{Arch}_{21}$ & 1,079 & 2 & 13 & 2 & [2, 4] & [4, 2] \\
$\mathrm{Level}_{1}\mathrm{Arch}_{22}$ & 1,079 & 2 & 13 & 3 & [2, 4] & [4, 2] \\
$\mathrm{Level}_{1}\mathrm{Arch}_{23}$ & 1,079 & 2 & 13 & 4 & [2, 4] & [4, 2] \\
$\mathrm{Level}_{1}\mathrm{Arch}_{24}$ & 1,079 & 2 & 13 & 5 & [2, 4] & [4, 2] \\
$\mathrm{Level}_{1}\mathrm{Arch}_{25}$ & 1,127 & 3 & 3  & 2 & [2, 4, 8] & [8, 4, 2] \\
$\mathrm{Level}_{1}\mathrm{Arch}_{26}$ & 1,127 & 3 & 3  & 3 & [2, 4, 8] & [8, 4, 2] \\
$\mathrm{Level}_{1}\mathrm{Arch}_{27}$ & 1,235 & 2 & 15 & 2 & [2, 4] & [4, 2] \\
$\mathrm{Level}_{1}\mathrm{Arch}_{28}$ & 1,235 & 2 & 15 & 3 & [2, 4] & [4, 2] \\
$\mathrm{Level}_{1}\mathrm{Arch}_{29}$ & 1,235 & 2 & 15 & 4 & [2, 4] & [4, 2] \\
$\mathrm{Level}_{1}\mathrm{Arch}_{30}$ & 1,235 & 2 & 15 & 5 & [2, 4] & [4, 2] \\
$\mathrm{Level}_{1}\mathrm{Arch}_{31}$ & 1,661 & 2 & 5  & 2 & [4, 8] & [8, 4] \\
$\mathrm{Level}_{1}\mathrm{Arch}_{32}$ & 1,661 & 2 & 5  & 3 & [4, 8] & [8, 4] \\
$\mathrm{Level}_{1}\mathrm{Arch}_{33}$ & 1,661 & 2 & 5  & 4 & [4, 8] & [8, 4] \\
$\mathrm{Level}_{1}\mathrm{Arch}_{34}$ & 1,661 & 2 & 5  & 5 & [4, 8] & [8, 4] \\
$\mathrm{Level}_{1}\mathrm{Arch}_{35}$ & 1,763 & 3 & 5  & 2 & [2, 4, 8] & [8, 4, 2] \\
$\mathrm{Level}_{1}\mathrm{Arch}_{36}$ & 1,763 & 3 & 5  & 3 & [2, 4, 8] & [8, 4, 2] \\
$\mathrm{Level}_{1}\mathrm{Arch}_{37}$ & 1,763 & 3 & 5  & 4 & [2, 4, 8] & [8, 4, 2] \\
$\mathrm{Level}_{1}\mathrm{Arch}_{38}$ & 1,763 & 3 & 5  & 5 & [2, 4, 8] & [8, 4, 2] \\
$\mathrm{Level}_{1}\mathrm{Arch}_{39}$ & 2,261 & 2 & 7  & 2 & [4, 8] & [8, 4] \\
$\mathrm{Level}_{1}\mathrm{Arch}_{40}$ & 2,261 & 2 & 7  & 3 & [4, 8] & [8, 4] \\
$\mathrm{Level}_{1}\mathrm{Arch}_{41}$ & 2,261 & 2 & 7  & 4 & [4, 8] & [8, 4] \\
$\mathrm{Level}_{1}\mathrm{Arch}_{42}$ & 2,261 & 2 & 7  & 5 & [4, 8] & [8, 4] \\
$\mathrm{Level}_{1}\mathrm{Arch}_{43}$ & 2,399 & 3 & 7  & 2 & [2, 4, 8] & [8, 4, 2] \\
$\mathrm{Level}_{1}\mathrm{Arch}_{44}$ & 2,399 & 3 & 7  & 3 & [2, 4, 8] & [8, 4, 2] \\
$\mathrm{Level}_{1}\mathrm{Arch}_{45}$ & 2,399 & 3 & 7  & 4 & [2, 4, 8] & [8, 4, 2] \\
$\mathrm{Level}_{1}\mathrm{Arch}_{46}$ & 2,399 & 3 & 7  & 5 & [2, 4, 8] & [8, 4, 2] \\

\end{longtable}
\endgroup

\begingroup
\footnotesize
\setlength{\tabcolsep}{3pt}
\renewcommand{\arraystretch}{1.15}
\setlength{\LTleft}{\fill}
\setlength{\LTright}{\fill}

\begin{longtable}{@{}
>{\raggedright\arraybackslash}p{0.21\textwidth}
>{\raggedleft\arraybackslash}p{0.10\textwidth}
>{\centering\arraybackslash}p{0.10\textwidth}
>{\centering\arraybackslash}p{0.083\textwidth}
>{\centering\arraybackslash}p{0.075\textwidth}
>{\centering\arraybackslash}p{0.156\textwidth}
>{\centering\arraybackslash}p{0.156\textwidth}
@{}}

\caption{TCN architectures for OhioT1DM capacity Level~2}
\label{tab:ohio_level2}\\

\toprule
\textbf{\shortstack{Architecture\\ID}} &
\textbf{\shortstack{Parameter\\ Count}} &
\textbf{\shortstack{Blocks\\$N$}} &
\textbf{\shortstack{Filter\\$f$}} &
\textbf{\shortstack{Base\\$b$}} &
\textbf{\shortstack{Encoder\\channels}} &
\textbf{\shortstack{Decoder\\channels}} \\
\midrule
\endfirsthead

\toprule
\textbf{\shortstack{Architecture\\ID}} &
\textbf{\shortstack{Parameter\\ Count}} &
\textbf{\shortstack{Blocks\\$N$}} &
\textbf{\shortstack{Filter\\$f$}} &
\textbf{\shortstack{Base\\$b$}} &
\textbf{\shortstack{Encoder\\channels}} &
\textbf{\shortstack{Decoder\\channels}} \\
\midrule
\endhead

\midrule
\multicolumn{7}{r}{Continued on next page}\\
\endfoot

\bottomrule
\endlastfoot

$\mathrm{Level}_{2}\mathrm{Arch}_{1}$  & 2,861 & 2 & 9  & 2 & [4, 8] & [8, 4] \\
$\mathrm{Level}_{2}\mathrm{Arch}_{2}$  & 2,861 & 2 & 9  & 3 & [4, 8] & [8, 4] \\
$\mathrm{Level}_{2}\mathrm{Arch}_{3}$  & 2,861 & 2 & 9  & 4 & [4, 8] & [8, 4] \\
$\mathrm{Level}_{2}\mathrm{Arch}_{4}$  & 2,861 & 2 & 9  & 5 & [4, 8] & [8, 4] \\
$\mathrm{Level}_{2}\mathrm{Arch}_{5}$  & 3,035 & 3 & 9  & 2 & [2, 4, 8] & [8, 4, 2] \\
$\mathrm{Level}_{2}\mathrm{Arch}_{6}$  & 3,035 & 3 & 9  & 3 & [2, 4, 8] & [8, 4, 2] \\
$\mathrm{Level}_{2}\mathrm{Arch}_{7}$  & 3,035 & 3 & 9  & 4 & [2, 4, 8] & [8, 4, 2] \\
$\mathrm{Level}_{2}\mathrm{Arch}_{8}$  & 3,035 & 3 & 9  & 5 & [2, 4, 8] & [8, 4, 2] \\
$\mathrm{Level}_{2}\mathrm{Arch}_{9}$  & 3,461 & 2 & 11 & 2 & [4, 8] & [8, 4] \\
$\mathrm{Level}_{2}\mathrm{Arch}_{10}$ & 3,461 & 2 & 11 & 3 & [4, 8] & [8, 4] \\
$\mathrm{Level}_{2}\mathrm{Arch}_{11}$ & 3,461 & 2 & 11 & 4 & [4, 8] & [8, 4] \\
$\mathrm{Level}_{2}\mathrm{Arch}_{12}$ & 3,461 & 2 & 11 & 5 & [4, 8] & [8, 4] \\
$\mathrm{Level}_{2}\mathrm{Arch}_{13}$ & 3,671 & 3 & 11 & 2 & [2, 4, 8] & [8, 4, 2] \\
$\mathrm{Level}_{2}\mathrm{Arch}_{14}$ & 3,671 & 3 & 11 & 3 & [2, 4, 8] & [8, 4, 2] \\
$\mathrm{Level}_{2}\mathrm{Arch}_{15}$ & 3,671 & 3 & 11 & 4 & [2, 4, 8] & [8, 4, 2] \\
$\mathrm{Level}_{2}\mathrm{Arch}_{16}$ & 3,671 & 3 & 11 & 5 & [2, 4, 8] & [8, 4, 2] \\
$\mathrm{Level}_{2}\mathrm{Arch}_{17}$ & 3,977 & 2 & 3  & 2 & [8, 16] & [16, 8] \\
$\mathrm{Level}_{2}\mathrm{Arch}_{18}$ & 3,977 & 2 & 3  & 3 & [8, 16] & [16, 8] \\
$\mathrm{Level}_{2}\mathrm{Arch}_{19}$ & 4,061 & 2 & 13 & 2 & [4, 8] & [8, 4] \\
$\mathrm{Level}_{2}\mathrm{Arch}_{20}$ & 4,061 & 2 & 13 & 3 & [4, 8] & [8, 4] \\
$\mathrm{Level}_{2}\mathrm{Arch}_{21}$ & 4,061 & 2 & 13 & 4 & [4, 8] & [8, 4] \\
$\mathrm{Level}_{2}\mathrm{Arch}_{22}$ & 4,061 & 2 & 13 & 5 & [4, 8] & [8, 4] \\
$\mathrm{Level}_{2}\mathrm{Arch}_{23}$ & 4,285 & 3 & 3  & 2 & [4, 8, 16] & [16, 8, 4] \\
$\mathrm{Level}_{2}\mathrm{Arch}_{24}$ & 4,285 & 3 & 3  & 3 & [4, 8, 16] & [16, 8, 4] \\
$\mathrm{Level}_{2}\mathrm{Arch}_{25}$ & 4,307 & 3 & 13 & 2 & [2, 4, 8] & [8, 4, 2] \\
$\mathrm{Level}_{2}\mathrm{Arch}_{26}$ & 4,307 & 3 & 13 & 3 & [2, 4, 8] & [8, 4, 2] \\
$\mathrm{Level}_{2}\mathrm{Arch}_{27}$ & 4,307 & 3 & 13 & 4 & [2, 4, 8] & [8, 4, 2] \\
$\mathrm{Level}_{2}\mathrm{Arch}_{28}$ & 4,307 & 3 & 13 & 5 & [2, 4, 8] & [8, 4, 2] \\
$\mathrm{Level}_{2}\mathrm{Arch}_{29}$ & 4,351 & 4 & 3  & 2 & [2, 4, 8, 16] & [16, 8, 4, 2] \\
$\mathrm{Level}_{2}\mathrm{Arch}_{30}$ & 4,351 & 4 & 3  & 3 & [2, 4, 8, 16] & [16, 8, 4, 2] \\
$\mathrm{Level}_{2}\mathrm{Arch}_{31}$ & 4,661 & 2 & 15 & 2 & [4, 8] & [8, 4] \\
$\mathrm{Level}_{2}\mathrm{Arch}_{32}$ & 4,661 & 2 & 15 & 3 & [4, 8] & [8, 4] \\
$\mathrm{Level}_{2}\mathrm{Arch}_{33}$ & 4,661 & 2 & 15 & 4 & [4, 8] & [8, 4] \\
$\mathrm{Level}_{2}\mathrm{Arch}_{34}$ & 4,661 & 2 & 15 & 5 & [4, 8] & [8, 4] \\
$\mathrm{Level}_{2}\mathrm{Arch}_{35}$ & 4,943 & 3 & 15 & 2 & [2, 4, 8] & [8, 4, 2] \\
$\mathrm{Level}_{2}\mathrm{Arch}_{36}$ & 4,943 & 3 & 15 & 3 & [2, 4, 8] & [8, 4, 2] \\
$\mathrm{Level}_{2}\mathrm{Arch}_{37}$ & 4,943 & 3 & 15 & 4 & [2, 4, 8] & [8, 4, 2] \\
$\mathrm{Level}_{2}\mathrm{Arch}_{38}$ & 4,943 & 3 & 15 & 5 & [2, 4, 8] & [8, 4, 2] \\

\end{longtable}
\endgroup

\begingroup
\footnotesize
\setlength{\tabcolsep}{3pt}
\renewcommand{\arraystretch}{1.15}
\setlength{\LTleft}{\fill}
\setlength{\LTright}{\fill}

\begin{longtable}{@{}
>{\raggedright\arraybackslash}p{0.21\textwidth}
>{\raggedleft\arraybackslash}p{0.10\textwidth}
>{\centering\arraybackslash}p{0.10\textwidth}
>{\centering\arraybackslash}p{0.083\textwidth}
>{\centering\arraybackslash}p{0.075\textwidth}
>{\centering\arraybackslash}p{0.156\textwidth}
>{\centering\arraybackslash}p{0.156\textwidth}
@{}}

\caption{TCN architectures for OhioT1DM capacity Level~3}
\label{tab:ohio_level3}\\

\toprule
\textbf{\shortstack{Architecture\\ID}} &
\textbf{\shortstack{Parameter\\ Count}}  &
\textbf{\shortstack{Blocks\\$N$}} &
\textbf{\shortstack{Filter\\$f$}} &
\textbf{\shortstack{Base\\$b$}} &
\textbf{\shortstack{Encoder\\channels}} &
\textbf{\shortstack{Decoder\\channels}} \\
\midrule
\endfirsthead

\multicolumn{7}{c}{\tablename\ \thetable\ (continued)}\\
\toprule
\textbf{\shortstack{Architecture\\ID}} &
\textbf{\shortstack{Parameter\\ Count}}  &
\textbf{\shortstack{Blocks\\$N$}} &
\textbf{\shortstack{Filter\\$f$}} &
\textbf{\shortstack{Base\\$b$}} &
\textbf{\shortstack{Encoder\\channels}} &
\textbf{\shortstack{Decoder\\channels}} \\
\midrule
\endhead

\midrule
\multicolumn{7}{r}{Continued on next page}\\
\endfoot

\bottomrule
\endlastfoot

$\mathrm{Level}_{3}\mathrm{Arch}_{1}$  & 6,329  & 2 & 5  & 2 & [8, 16] & [16, 8] \\
$\mathrm{Level}_{3}\mathrm{Arch}_{2}$  & 6,329  & 2 & 5  & 3 & [8, 16] & [16, 8] \\
$\mathrm{Level}_{3}\mathrm{Arch}_{3}$  & 6,329  & 2 & 5  & 4 & [8, 16] & [16, 8] \\
$\mathrm{Level}_{3}\mathrm{Arch}_{4}$  & 6,329  & 2 & 5  & 5 & [8, 16] & [16, 8] \\
$\mathrm{Level}_{3}\mathrm{Arch}_{5}$  & 6,805  & 3 & 5  & 2 & [4, 8, 16] & [16, 8, 4] \\
$\mathrm{Level}_{3}\mathrm{Arch}_{6}$  & 6,805  & 3 & 5  & 3 & [4, 8, 16] & [16, 8, 4] \\
$\mathrm{Level}_{3}\mathrm{Arch}_{7}$  & 6,805  & 3 & 5  & 4 & [4, 8, 16] & [16, 8, 4] \\
$\mathrm{Level}_{3}\mathrm{Arch}_{8}$  & 6,805  & 3 & 5  & 5 & [4, 8, 16] & [16, 8, 4] \\
$\mathrm{Level}_{3}\mathrm{Arch}_{9}$  & 6,907  & 4 & 5  & 2 & [2, 4, 8, 16] & [16, 8, 4, 2] \\
$\mathrm{Level}_{3}\mathrm{Arch}_{10}$ & 6,907  & 4 & 5  & 3 & [2, 4, 8, 16] & [16, 8, 4, 2] \\
$\mathrm{Level}_{3}\mathrm{Arch}_{11}$ & 6,907  & 4 & 5  & 4 & [2, 4, 8, 16] & [16, 8, 4, 2] \\
$\mathrm{Level}_{3}\mathrm{Arch}_{12}$ & 6,907  & 4 & 5  & 5 & [2, 4, 8, 16] & [16, 8, 4, 2] \\
$\mathrm{Level}_{3}\mathrm{Arch}_{13}$ & 8,681  & 2 & 7  & 2 & [8, 16] & [16, 8] \\
$\mathrm{Level}_{3}\mathrm{Arch}_{14}$ & 8,681  & 2 & 7  & 3 & [8, 16] & [16, 8] \\
$\mathrm{Level}_{3}\mathrm{Arch}_{15}$ & 8,681  & 2 & 7  & 4 & [8, 16] & [16, 8] \\
$\mathrm{Level}_{3}\mathrm{Arch}_{16}$ & 8,681  & 2 & 7  & 5 & [8, 16] & [16, 8] \\
$\mathrm{Level}_{3}\mathrm{Arch}_{17}$ & 9,325  & 3 & 7  & 2 & [4, 8, 16] & [16, 8, 4] \\
$\mathrm{Level}_{3}\mathrm{Arch}_{18}$ & 9,325  & 3 & 7  & 3 & [4, 8, 16] & [16, 8, 4] \\
$\mathrm{Level}_{3}\mathrm{Arch}_{19}$ & 9,325  & 3 & 7  & 4 & [4, 8, 16] & [16, 8, 4] \\
$\mathrm{Level}_{3}\mathrm{Arch}_{20}$ & 9,325  & 3 & 7  & 5 & [4, 8, 16] & [16, 8, 4] \\
$\mathrm{Level}_{3}\mathrm{Arch}_{21}$ & 9,463  & 4 & 7  & 2 & [2, 4, 8, 16] & [16, 8, 4, 2] \\
$\mathrm{Level}_{3}\mathrm{Arch}_{22}$ & 9,463  & 4 & 7  & 3 & [2, 4, 8, 16] & [16, 8, 4, 2] \\
$\mathrm{Level}_{3}\mathrm{Arch}_{23}$ & 9,463  & 4 & 7  & 4 & [2, 4, 8, 16] & [16, 8, 4, 2] \\
$\mathrm{Level}_{3}\mathrm{Arch}_{24}$ & 9,463  & 4 & 7  & 5 & [2, 4, 8, 16] & [16, 8, 4, 2] \\
$\mathrm{Level}_{3}\mathrm{Arch}_{25}$ & 11,033 & 2 & 9  & 2 & [8, 16] & [16, 8] \\
$\mathrm{Level}_{3}\mathrm{Arch}_{26}$ & 11,033 & 2 & 9  & 3 & [8, 16] & [16, 8] \\
$\mathrm{Level}_{3}\mathrm{Arch}_{27}$ & 11,033 & 2 & 9  & 4 & [8, 16] & [16, 8] \\
$\mathrm{Level}_{3}\mathrm{Arch}_{28}$ & 11,033 & 2 & 9  & 5 & [8, 16] & [16, 8] \\
$\mathrm{Level}_{3}\mathrm{Arch}_{29}$ & 11,845 & 3 & 9  & 2 & [4, 8, 16] & [16, 8, 4] \\
$\mathrm{Level}_{3}\mathrm{Arch}_{30}$ & 11,845 & 3 & 9  & 3 & [4, 8, 16] & [16, 8, 4] \\
$\mathrm{Level}_{3}\mathrm{Arch}_{31}$ & 11,845 & 3 & 9  & 4 & [4, 8, 16] & [16, 8, 4] \\
$\mathrm{Level}_{3}\mathrm{Arch}_{32}$ & 11,845 & 3 & 9  & 5 & [4, 8, 16] & [16, 8, 4] \\
$\mathrm{Level}_{3}\mathrm{Arch}_{33}$ & 12,019 & 4 & 9  & 2 & [2, 4, 8, 16] & [16, 8, 4, 2] \\
$\mathrm{Level}_{3}\mathrm{Arch}_{34}$ & 12,019 & 4 & 9  & 3 & [2, 4, 8, 16] & [16, 8, 4, 2] \\
$\mathrm{Level}_{3}\mathrm{Arch}_{35}$ & 12,019 & 4 & 9  & 4 & [2, 4, 8, 16] & [16, 8, 4, 2] \\
$\mathrm{Level}_{3}\mathrm{Arch}_{36}$ & 12,019 & 4 & 9  & 5 & [2, 4, 8, 16] & [16, 8, 4, 2] \\
$\mathrm{Level}_{3}\mathrm{Arch}_{37}$ & 13,385 & 2 & 11 & 2 & [8, 16] & [16, 8] \\
$\mathrm{Level}_{3}\mathrm{Arch}_{38}$ & 13,385 & 2 & 11 & 3 & [8, 16] & [16, 8] \\
$\mathrm{Level}_{3}\mathrm{Arch}_{39}$ & 13,385 & 2 & 11 & 4 & [8, 16] & [16, 8] \\
$\mathrm{Level}_{3}\mathrm{Arch}_{40}$ & 13,385 & 2 & 11 & 5 & [8, 16] & [16, 8] \\
$\mathrm{Level}_{3}\mathrm{Arch}_{41}$ & 14,365 & 3 & 11 & 2 & [4, 8, 16] & [16, 8, 4] \\
$\mathrm{Level}_{3}\mathrm{Arch}_{42}$ & 14,365 & 3 & 11 & 3 & [4, 8, 16] & [16, 8, 4] \\
$\mathrm{Level}_{3}\mathrm{Arch}_{43}$ & 14,365 & 3 & 11 & 4 & [4, 8, 16] & [16, 8, 4] \\
$\mathrm{Level}_{3}\mathrm{Arch}_{44}$ & 14,365 & 3 & 11 & 5 & [4, 8, 16] & [16, 8, 4] \\
$\mathrm{Level}_{3}\mathrm{Arch}_{45}$ & 14,575 & 4 & 11 & 2 & [2, 4, 8, 16] & [16, 8, 4, 2] \\
$\mathrm{Level}_{3}\mathrm{Arch}_{46}$ & 14,575 & 4 & 11 & 3 & [2, 4, 8, 16] & [16, 8, 4, 2] \\
$\mathrm{Level}_{3}\mathrm{Arch}_{47}$ & 14,575 & 4 & 11 & 4 & [2, 4, 8, 16] & [16, 8, 4, 2] \\
$\mathrm{Level}_{3}\mathrm{Arch}_{48}$ & 14,575 & 4 & 11 & 5 & [2, 4, 8, 16] & [16, 8, 4, 2] \\

\end{longtable}
\endgroup

\begingroup
\footnotesize
\setlength{\tabcolsep}{3pt}
\renewcommand{\arraystretch}{1.15}
\setlength{\LTleft}{\fill}
\setlength{\LTright}{\fill}

\begin{longtable}{@{}
>{\raggedright\arraybackslash}p{0.21\textwidth}
>{\raggedleft\arraybackslash}p{0.10\textwidth}
>{\centering\arraybackslash}p{0.10\textwidth}
>{\centering\arraybackslash}p{0.083\textwidth}
>{\centering\arraybackslash}p{0.075\textwidth}
>{\centering\arraybackslash}p{0.156\textwidth}
>{\centering\arraybackslash}p{0.156\textwidth}
@{}}

\caption{TCN architectures for OhioT1DM capacity Level~4}
\label{tab:ohio_level4}\\

\toprule
\textbf{\shortstack{Architecture\\ID}} &
\textbf{\shortstack{Parameter\\ Count}}  &
\textbf{\shortstack{Blocks\\$N$}} &
\textbf{\shortstack{Filter\\$f$}} &
\textbf{\shortstack{Base\\$b$}} &
\textbf{\shortstack{Encoder\\channels}} &
\textbf{\shortstack{Decoder\\channels}} \\
\midrule
\endfirsthead

\toprule
\textbf{\shortstack{Architecture\\ID}} &
\textbf{\shortstack{Parameter\\ Count}}  &
\textbf{\shortstack{Blocks\\$N$}} &
\textbf{\shortstack{Filter\\$f$}} &
\textbf{\shortstack{Base\\$b$}} &
\textbf{\shortstack{Encoder\\channels}} &
\textbf{\shortstack{Decoder\\channels}} \\
\midrule
\endhead

\midrule
\multicolumn{7}{r}{Continued on next page}\\
\endfoot

\bottomrule
\endlastfoot

$\mathrm{Level}_{4}\mathrm{Arch}_{1}$  & 15,377 & 2 & 3  & 2 & [16, 32] & [32, 16] \\
$\mathrm{Level}_{4}\mathrm{Arch}_{2}$  & 15,377 & 2 & 3  & 3 & [16, 32] & [32, 16] \\
$\mathrm{Level}_{4}\mathrm{Arch}_{3}$  & 15,737 & 2 & 13 & 2 & [8, 16] & [16, 8] \\
$\mathrm{Level}_{4}\mathrm{Arch}_{4}$  & 15,737 & 2 & 13 & 3 & [8, 16] & [16, 8] \\
$\mathrm{Level}_{4}\mathrm{Arch}_{5}$  & 15,737 & 2 & 13 & 4 & [8, 16] & [16, 8] \\
$\mathrm{Level}_{4}\mathrm{Arch}_{6}$  & 15,737 & 2 & 13 & 5 & [8, 16] & [16, 8] \\
$\mathrm{Level}_{4}\mathrm{Arch}_{7}$  & 16,697 & 3 & 3  & 2 & [8, 16, 32] & [32, 16, 8] \\
$\mathrm{Level}_{4}\mathrm{Arch}_{8}$  & 16,697 & 3 & 3  & 3 & [8, 16, 32] & [32, 16, 8] \\
$\mathrm{Level}_{4}\mathrm{Arch}_{9}$  & 16,885 & 3 & 13 & 2 & [4, 8, 16] & [16, 8, 4] \\
$\mathrm{Level}_{4}\mathrm{Arch}_{10}$ & 16,885 & 3 & 13 & 3 & [4, 8, 16] & [16, 8, 4] \\
$\mathrm{Level}_{4}\mathrm{Arch}_{11}$ & 16,885 & 3 & 13 & 4 & [4, 8, 16] & [16, 8, 4] \\
$\mathrm{Level}_{4}\mathrm{Arch}_{12}$ & 16,885 & 3 & 13 & 5 & [4, 8, 16] & [16, 8, 4] \\
$\mathrm{Level}_{4}\mathrm{Arch}_{13}$ & 17,005 & 4 & 3  & 2 & [4, 8, 16, 32] & [32, 16, 8, 4] \\
$\mathrm{Level}_{4}\mathrm{Arch}_{14}$ & 17,005 & 4 & 3  & 3 & [4, 8, 16, 32] & [32, 16, 8, 4] \\
$\mathrm{Level}_{4}\mathrm{Arch}_{15}$ & 17,071 & 5 & 3  & 2 & [2, 4, 8, 16, 32] & [32, 16, 8, 4, 2] \\
$\mathrm{Level}_{4}\mathrm{Arch}_{16}$ & 17,071 & 5 & 3  & 3 & [2, 4, 8, 16, 32] & [32, 16, 8, 4, 2] \\
$\mathrm{Level}_{4}\mathrm{Arch}_{17}$ & 17,131 & 4 & 13 & 2 & [2, 4, 8, 16] & [16, 8, 4, 2] \\
$\mathrm{Level}_{4}\mathrm{Arch}_{18}$ & 17,131 & 4 & 13 & 3 & [2, 4, 8, 16] & [16, 8, 4, 2] \\
$\mathrm{Level}_{4}\mathrm{Arch}_{19}$ & 17,131 & 4 & 13 & 4 & [2, 4, 8, 16] & [16, 8, 4, 2] \\
$\mathrm{Level}_{4}\mathrm{Arch}_{20}$ & 17,131 & 4 & 13 & 5 & [2, 4, 8, 16] & [16, 8, 4, 2] \\
$\mathrm{Level}_{4}\mathrm{Arch}_{21}$ & 18,089 & 2 & 15 & 2 & [8, 16] & [16, 8] \\
$\mathrm{Level}_{4}\mathrm{Arch}_{22}$ & 18,089 & 2 & 15 & 3 & [8, 16] & [16, 8] \\
$\mathrm{Level}_{4}\mathrm{Arch}_{23}$ & 18,089 & 2 & 15 & 4 & [8, 16] & [16, 8] \\
$\mathrm{Level}_{4}\mathrm{Arch}_{24}$ & 18,089 & 2 & 15 & 5 & [8, 16] & [16, 8] \\
$\mathrm{Level}_{4}\mathrm{Arch}_{25}$ & 19,405 & 3 & 15 & 2 & [4, 8, 16] & [16, 8, 4] \\
$\mathrm{Level}_{4}\mathrm{Arch}_{26}$ & 19,405 & 3 & 15 & 3 & [4, 8, 16] & [16, 8, 4] \\
$\mathrm{Level}_{4}\mathrm{Arch}_{27}$ & 19,405 & 3 & 15 & 4 & [4, 8, 16] & [16, 8, 4] \\
$\mathrm{Level}_{4}\mathrm{Arch}_{28}$ & 19,405 & 3 & 15 & 5 & [4, 8, 16] & [16, 8, 4] \\
$\mathrm{Level}_{4}\mathrm{Arch}_{29}$ & 19,687 & 4 & 15 & 2 & [2, 4, 8, 16] & [16, 8, 4, 2] \\
$\mathrm{Level}_{4}\mathrm{Arch}_{30}$ & 19,687 & 4 & 15 & 3 & [2, 4, 8, 16] & [16, 8, 4, 2] \\
$\mathrm{Level}_{4}\mathrm{Arch}_{31}$ & 19,687 & 4 & 15 & 4 & [2, 4, 8, 16] & [16, 8, 4, 2] \\
$\mathrm{Level}_{4}\mathrm{Arch}_{32}$ & 19,687 & 4 & 15 & 5 & [2, 4, 8, 16] & [16, 8, 4, 2] \\
$\mathrm{Level}_{4}\mathrm{Arch}_{33}$ & 24,689 & 2 & 5  & 2 & [16, 32] & [32, 16] \\
$\mathrm{Level}_{4}\mathrm{Arch}_{34}$ & 24,689 & 2 & 5  & 3 & [16, 32] & [32, 16] \\
$\mathrm{Level}_{4}\mathrm{Arch}_{35}$ & 24,689 & 2 & 5  & 4 & [16, 32] & [32, 16] \\
$\mathrm{Level}_{4}\mathrm{Arch}_{36}$ & 24,689 & 2 & 5  & 5 & [16, 32] & [32, 16] \\
$\mathrm{Level}_{4}\mathrm{Arch}_{37}$ & 26,729 & 3 & 5  & 2 & [8, 16, 32] & [32, 16, 8] \\
$\mathrm{Level}_{4}\mathrm{Arch}_{38}$ & 26,729 & 3 & 5  & 3 & [8, 16, 32] & [32, 16, 8] \\
$\mathrm{Level}_{4}\mathrm{Arch}_{39}$ & 26,729 & 3 & 5  & 4 & [8, 16, 32] & [32, 16, 8] \\
$\mathrm{Level}_{4}\mathrm{Arch}_{40}$ & 26,729 & 3 & 5  & 5 & [8, 16, 32] & [32, 16, 8] \\
$\mathrm{Level}_{4}\mathrm{Arch}_{41}$ & 27,205 & 4 & 5  & 2 & [4, 8, 16, 32] & [32, 16, 8, 4] \\
$\mathrm{Level}_{4}\mathrm{Arch}_{42}$ & 27,205 & 4 & 5  & 3 & [4, 8, 16, 32] & [32, 16, 8, 4] \\
$\mathrm{Level}_{4}\mathrm{Arch}_{43}$ & 27,205 & 4 & 5  & 4 & [4, 8, 16, 32] & [32, 16, 8, 4] \\
$\mathrm{Level}_{4}\mathrm{Arch}_{44}$ & 27,205 & 4 & 5  & 5 & [4, 8, 16, 32] & [32, 16, 8, 4] \\
$\mathrm{Level}_{4}\mathrm{Arch}_{45}$ & 27,307 & 5 & 5  & 2 & [2, 4, 8, 16, 32] & [32, 16, 8, 4, 2] \\
$\mathrm{Level}_{4}\mathrm{Arch}_{46}$ & 27,307 & 5 & 5  & 3 & [2, 4, 8, 16, 32] & [32, 16, 8, 4, 2] \\
$\mathrm{Level}_{4}\mathrm{Arch}_{47}$ & 27,307 & 5 & 5  & 4 & [2, 4, 8, 16, 32] & [32, 16, 8, 4, 2] \\
$\mathrm{Level}_{4}\mathrm{Arch}_{48}$ & 27,307 & 5 & 5  & 5 & [2, 4, 8, 16, 32] & [32, 16, 8, 4, 2] \\
$\mathrm{Level}_{4}\mathrm{Arch}_{49}$ & 34,001 & 2 & 7  & 2 & [16, 32] & [32, 16] \\
$\mathrm{Level}_{4}\mathrm{Arch}_{50}$ & 34,001 & 2 & 7  & 3 & [16, 32] & [32, 16] \\
$\mathrm{Level}_{4}\mathrm{Arch}_{51}$ & 34,001 & 2 & 7  & 4 & [16, 32] & [32, 16] \\
$\mathrm{Level}_{4}\mathrm{Arch}_{52}$ & 34,001 & 2 & 7  & 5 & [16, 32] & [32, 16] \\
$\mathrm{Level}_{4}\mathrm{Arch}_{53}$ & 36,761 & 3 & 7  & 2 & [8, 16, 32] & [32, 16, 8] \\
$\mathrm{Level}_{4}\mathrm{Arch}_{54}$ & 36,761 & 3 & 7  & 3 & [8, 16, 32] & [32, 16, 8] \\
$\mathrm{Level}_{4}\mathrm{Arch}_{55}$ & 36,761 & 3 & 7  & 4 & [8, 16, 32] & [32, 16, 8] \\
$\mathrm{Level}_{4}\mathrm{Arch}_{56}$ & 36,761 & 3 & 7  & 5 & [8, 16, 32] & [32, 16, 8] \\
$\mathrm{Level}_{4}\mathrm{Arch}_{57}$ & 37,405 & 4 & 7  & 2 & [4, 8, 16, 32] & [32, 16, 8, 4] \\
$\mathrm{Level}_{4}\mathrm{Arch}_{58}$ & 37,405 & 4 & 7  & 3 & [4, 8, 16, 32] & [32, 16, 8, 4] \\
$\mathrm{Level}_{4}\mathrm{Arch}_{59}$ & 37,405 & 4 & 7  & 4 & [4, 8, 16, 32] & [32, 16, 8, 4] \\
$\mathrm{Level}_{4}\mathrm{Arch}_{60}$ & 37,405 & 4 & 7  & 5 & [4, 8, 16, 32] & [32, 16, 8, 4] \\
$\mathrm{Level}_{4}\mathrm{Arch}_{61}$ & 37,543 & 5 & 7  & 2 & [2, 4, 8, 16, 32] & [32, 16, 8, 4, 2] \\
$\mathrm{Level}_{4}\mathrm{Arch}_{62}$ & 37,543 & 5 & 7  & 3 & [2, 4, 8, 16, 32] & [32, 16, 8, 4, 2] \\
$\mathrm{Level}_{4}\mathrm{Arch}_{63}$ & 37,543 & 5 & 7  & 4 & [2, 4, 8, 16, 32] & [32, 16, 8, 4, 2] \\
$\mathrm{Level}_{4}\mathrm{Arch}_{64}$ & 37,543 & 5 & 7  & 5 & [2, 4, 8, 16, 32] & [32, 16, 8, 4, 2] \\

\end{longtable}
\endgroup

\begingroup
\footnotesize
\setlength{\tabcolsep}{3pt}
\renewcommand{\arraystretch}{1.15}
\setlength{\LTleft}{\fill}
\setlength{\LTright}{\fill}

\begin{longtable}{@{}
>{\raggedright\arraybackslash}p{0.21\textwidth}
>{\raggedleft\arraybackslash}p{0.10\textwidth}
>{\centering\arraybackslash}p{0.10\textwidth}
>{\centering\arraybackslash}p{0.083\textwidth}
>{\centering\arraybackslash}p{0.075\textwidth}
>{\centering\arraybackslash}p{0.156\textwidth}
>{\centering\arraybackslash}p{0.156\textwidth}
@{}}

\caption{TCN architectures for OhioT1DM capacity Level~5
($40{,}000\leq W<65{,}000$).}
\label{tab:ohio_level5}\\

\toprule
\textbf{\shortstack{Architecture\\ID}} &
\textbf{\shortstack{Parameter\\ Count}}  &
\textbf{\shortstack{Blocks\\$N$}} &
\textbf{\shortstack{Filter\\$f$}} &
\textbf{\shortstack{Base\\$b$}} &
\textbf{\shortstack{Encoder\\channels}} &
\textbf{\shortstack{Decoder\\channels}} \\
\midrule
\endfirsthead

\toprule
\textbf{\shortstack{Architecture\\ID}} &
\textbf{\shortstack{Parameter\\ Count}} &
\textbf{\shortstack{Blocks\\$N$}} &
\textbf{\shortstack{Filter\\$f$}} &
\textbf{\shortstack{Base\\$b$}} &
\textbf{\shortstack{Encoder\\channels}} &
\textbf{\shortstack{Decoder\\channels}} \\
\midrule
\endhead

\midrule
\multicolumn{7}{r}{Continued on next page}\\
\endfoot

\bottomrule
\endlastfoot

$\mathrm{Level}_{5}\mathrm{Arch}_{1}$  & 43,313 & 2 & 9  & 2 & [16, 32] & [32, 16] \\
$\mathrm{Level}_{5}\mathrm{Arch}_{2}$  & 43,313 & 2 & 9  & 3 & [16, 32] & [32, 16] \\
$\mathrm{Level}_{5}\mathrm{Arch}_{3}$  & 43,313 & 2 & 9  & 4 & [16, 32] & [32, 16] \\
$\mathrm{Level}_{5}\mathrm{Arch}_{4}$  & 43,313 & 2 & 9  & 5 & [16, 32] & [32, 16] \\
$\mathrm{Level}_{5}\mathrm{Arch}_{5}$  & 46,793 & 3 & 9  & 2 & [8, 16, 32] & [32, 16, 8] \\
$\mathrm{Level}_{5}\mathrm{Arch}_{6}$  & 46,793 & 3 & 9  & 3 & [8, 16, 32] & [32, 16, 8] \\
$\mathrm{Level}_{5}\mathrm{Arch}_{7}$  & 46,793 & 3 & 9  & 4 & [8, 16, 32] & [32, 16, 8] \\
$\mathrm{Level}_{5}\mathrm{Arch}_{8}$  & 46,793 & 3 & 9  & 5 & [8, 16, 32] & [32, 16, 8] \\
$\mathrm{Level}_{5}\mathrm{Arch}_{9}$  & 47,605 & 4 & 9  & 2 & [4, 8, 16, 32] & [32, 16, 8, 4] \\
$\mathrm{Level}_{5}\mathrm{Arch}_{10}$ & 47,605 & 4 & 9  & 3 & [4, 8, 16, 32] & [32, 16, 8, 4] \\
$\mathrm{Level}_{5}\mathrm{Arch}_{11}$ & 47,605 & 4 & 9  & 4 & [4, 8, 16, 32] & [32, 16, 8, 4] \\
$\mathrm{Level}_{5}\mathrm{Arch}_{12}$ & 47,605 & 4 & 9  & 5 & [4, 8, 16, 32] & [32, 16, 8, 4] \\
$\mathrm{Level}_{5}\mathrm{Arch}_{13}$ & 47,779 & 5 & 9  & 2 & [2, 4, 8, 16, 32] & [32, 16, 8, 4, 2] \\
$\mathrm{Level}_{5}\mathrm{Arch}_{14}$ & 47,779 & 5 & 9  & 3 & [2, 4, 8, 16, 32] & [32, 16, 8, 4, 2] \\
$\mathrm{Level}_{5}\mathrm{Arch}_{15}$ & 47,779 & 5 & 9  & 4 & [2, 4, 8, 16, 32] & [32, 16, 8, 4, 2] \\
$\mathrm{Level}_{5}\mathrm{Arch}_{16}$ & 47,779 & 5 & 9  & 5 & [2, 4, 8, 16, 32] & [32, 16, 8, 4, 2] \\
$\mathrm{Level}_{5}\mathrm{Arch}_{17}$ & 52,625 & 2 & 11 & 2 & [16, 32] & [32, 16] \\
$\mathrm{Level}_{5}\mathrm{Arch}_{18}$ & 52,625 & 2 & 11 & 3 & [16, 32] & [32, 16] \\
$\mathrm{Level}_{5}\mathrm{Arch}_{19}$ & 52,625 & 2 & 11 & 4 & [16, 32] & [32, 16] \\
$\mathrm{Level}_{5}\mathrm{Arch}_{20}$ & 52,625 & 2 & 11 & 5 & [16, 32] & [32, 16] \\
$\mathrm{Level}_{5}\mathrm{Arch}_{21}$ & 56,825 & 3 & 11 & 2 & [8, 16, 32] & [32, 16, 8] \\
$\mathrm{Level}_{5}\mathrm{Arch}_{22}$ & 56,825 & 3 & 11 & 3 & [8, 16, 32] & [32, 16, 8] \\
$\mathrm{Level}_{5}\mathrm{Arch}_{23}$ & 56,825 & 3 & 11 & 4 & [8, 16, 32] & [32, 16, 8] \\
$\mathrm{Level}_{5}\mathrm{Arch}_{24}$ & 56,825 & 3 & 11 & 5 & [8, 16, 32] & [32, 16, 8] \\
$\mathrm{Level}_{5}\mathrm{Arch}_{25}$ & 57,805 & 4 & 11 & 2 & [4, 8, 16, 32] & [32, 16, 8, 4] \\
$\mathrm{Level}_{5}\mathrm{Arch}_{26}$ & 57,805 & 4 & 11 & 3 & [4, 8, 16, 32] & [32, 16, 8, 4] \\
$\mathrm{Level}_{5}\mathrm{Arch}_{27}$ & 57,805 & 4 & 11 & 4 & [4, 8, 16, 32] & [32, 16, 8, 4] \\
$\mathrm{Level}_{5}\mathrm{Arch}_{28}$ & 57,805 & 4 & 11 & 5 & [4, 8, 16, 32] & [32, 16, 8, 4] \\
$\mathrm{Level}_{5}\mathrm{Arch}_{29}$ & 58,015 & 5 & 11 & 2 & [2, 4, 8, 16, 32] & [32, 16, 8, 4, 2] \\
$\mathrm{Level}_{5}\mathrm{Arch}_{30}$ & 58,015 & 5 & 11 & 3 & [2, 4, 8, 16, 32] & [32, 16, 8, 4, 2] \\
$\mathrm{Level}_{5}\mathrm{Arch}_{31}$ & 58,015 & 5 & 11 & 4 & [2, 4, 8, 16, 32] & [32, 16, 8, 4, 2] \\
$\mathrm{Level}_{5}\mathrm{Arch}_{32}$ & 58,015 & 5 & 11 & 5 & [2, 4, 8, 16, 32] & [32, 16, 8, 4, 2] \\
$\mathrm{Level}_{5}\mathrm{Arch}_{33}$ & 60,449 & 2 & 3  & 2 & [32, 64] & [64, 32] \\
$\mathrm{Level}_{5}\mathrm{Arch}_{34}$ & 60,449 & 2 & 3  & 3 & [32, 64] & [64, 32] \\
$\mathrm{Level}_{5}\mathrm{Arch}_{35}$ & 61,937 & 2 & 13 & 2 & [16, 32] & [32, 16] \\
$\mathrm{Level}_{5}\mathrm{Arch}_{36}$ & 61,937 & 2 & 13 & 3 & [16, 32] & [32, 16] \\
$\mathrm{Level}_{5}\mathrm{Arch}_{37}$ & 61,937 & 2 & 13 & 4 & [16, 32] & [32, 16] \\
$\mathrm{Level}_{5}\mathrm{Arch}_{38}$ & 61,937 & 2 & 13 & 5 & [16, 32] & [32, 16] \\

\end{longtable}
\endgroup

\begingroup
\footnotesize
\setlength{\tabcolsep}{3pt}
\renewcommand{\arraystretch}{1.15}
\setlength{\LTleft}{\fill}
\setlength{\LTright}{\fill}

\begin{longtable}{@{}
>{\raggedright\arraybackslash}p{0.21\textwidth}
>{\raggedleft\arraybackslash}p{0.10\textwidth}
>{\centering\arraybackslash}p{0.10\textwidth}
>{\centering\arraybackslash}p{0.083\textwidth}
>{\centering\arraybackslash}p{0.075\textwidth}
>{\centering\arraybackslash}p{0.156\textwidth}
>{\centering\arraybackslash}p{0.156\textwidth}
@{}}

\caption{TCN architectures for OhioT1DM capacity Level~6}
\label{tab:ohio_level6}\\

\toprule
\textbf{\shortstack{Architecture\\ID}} &
\textbf{\shortstack{parameter\\ Count}} &
\textbf{\shortstack{Blocks\\$N$}} &
\textbf{\shortstack{Filter\\$f$}} &
\textbf{\shortstack{Base\\$b$}} &
\textbf{\shortstack{Encoder\\channels}} &
\textbf{\shortstack{Decoder\\channels}} \\
\midrule
\endfirsthead

\toprule
\textbf{\shortstack{Architecture\\ID}} &
\textbf{\shortstack{Parameter\\ Count}}  &
\textbf{\shortstack{Blocks\\$N$}} &
\textbf{\shortstack{Filter\\$f$}} &
\textbf{\shortstack{Base\\$b$}} &
\textbf{\shortstack{Encoder\\channels}} &
\textbf{\shortstack{Decoder\\channels}} \\
\midrule
\endhead

\midrule

\endfoot

\bottomrule
\endlastfoot

$\mathrm{Level}_{6}\mathrm{Arch}_{1}$  & 65,905 & 3 & 3  & 2 & [16, 32, 64] & [64, 32, 16] \\
$\mathrm{Level}_{6}\mathrm{Arch}_{2}$  & 65,905 & 3 & 3  & 3 & [16, 32, 64] & [64, 32, 16] \\
$\mathrm{Level}_{6}\mathrm{Arch}_{3}$  & 66,857 & 3 & 13 & 2 & [8, 16, 32] & [32, 16, 8] \\
$\mathrm{Level}_{6}\mathrm{Arch}_{4}$  & 66,857 & 3 & 13 & 3 & [8, 16, 32] & [32, 16, 8] \\
$\mathrm{Level}_{6}\mathrm{Arch}_{5}$  & 66,857 & 3 & 13 & 4 & [8, 16, 32] & [32, 16, 8] \\
$\mathrm{Level}_{6}\mathrm{Arch}_{6}$  & 66,857 & 3 & 13 & 5 & [8, 16, 32] & [32, 16, 8] \\
$\mathrm{Level}_{6}\mathrm{Arch}_{7}$  & 67,225 & 4 & 3  & 2 & [8, 16, 32, 64] & [64, 32, 16, 8] \\
$\mathrm{Level}_{6}\mathrm{Arch}_{8}$  & 67,225 & 4 & 3  & 3 & [8, 16, 32, 64] & [64, 32, 16, 8] \\
$\mathrm{Level}_{6}\mathrm{Arch}_{9}$  & 67,533 & 5 & 3  & 2 & [4, 8, 16, 32, 64] & [64, 32, 16, 8, 4] \\
$\mathrm{Level}_{6}\mathrm{Arch}_{10}$ & 67,533 & 5 & 3  & 3 & [4, 8, 16, 32, 64] & [64, 32, 16, 8, 4] \\
$\mathrm{Level}_{6}\mathrm{Arch}_{11}$ & 67,599 & 6 & 3  & 2 & [2, 4, 8, 16, 32, 64] & [64, 32, 16, 8, 4, 2] \\
$\mathrm{Level}_{6}\mathrm{Arch}_{12}$ & 67,599 & 6 & 3  & 3 & [2, 4, 8, 16, 32, 64] & [64, 32, 16, 8, 4, 2] \\
$\mathrm{Level}_{6}\mathrm{Arch}_{13}$ & 68,005 & 4 & 13 & 2 & [4, 8, 16, 32] & [32, 16, 8, 4] \\
$\mathrm{Level}_{6}\mathrm{Arch}_{14}$ & 68,005 & 4 & 13 & 3 & [4, 8, 16, 32] & [32, 16, 8, 4] \\
$\mathrm{Level}_{6}\mathrm{Arch}_{15}$ & 68,005 & 4 & 13 & 4 & [4, 8, 16, 32] & [32, 16, 8, 4] \\
$\mathrm{Level}_{6}\mathrm{Arch}_{16}$ & 68,005 & 4 & 13 & 5 & [4, 8, 16, 32] & [32, 16, 8, 4] \\
$\mathrm{Level}_{6}\mathrm{Arch}_{17}$ & 68,251 & 5 & 13 & 2 & [2, 4, 8, 16, 32] & [32, 16, 8, 4, 2] \\
$\mathrm{Level}_{6}\mathrm{Arch}_{18}$ & 68,251 & 5 & 13 & 3 & [2, 4, 8, 16, 32] & [32, 16, 8, 4, 2] \\
$\mathrm{Level}_{6}\mathrm{Arch}_{19}$ & 68,251 & 5 & 13 & 4 & [2, 4, 8, 16, 32] & [32, 16, 8, 4, 2] \\
$\mathrm{Level}_{6}\mathrm{Arch}_{20}$ & 68,251 & 5 & 13 & 5 & [2, 4, 8, 16, 32] & [32, 16, 8, 4, 2] \\
$\mathrm{Level}_{6}\mathrm{Arch}_{21}$ & 71,249 & 2 & 15 & 2 & [16, 32] & [32, 16] \\
$\mathrm{Level}_{6}\mathrm{Arch}_{22}$ & 71,249 & 2 & 15 & 3 & [16, 32] & [32, 16] \\
$\mathrm{Level}_{6}\mathrm{Arch}_{23}$ & 71,249 & 2 & 15 & 4 & [16, 32] & [32, 16] \\
$\mathrm{Level}_{6}\mathrm{Arch}_{24}$ & 71,249 & 2 & 15 & 5 & [16, 32] & [32, 16] \\
$\mathrm{Level}_{6}\mathrm{Arch}_{25}$ & 76,889 & 3 & 15 & 2 & [8, 16, 32] & [32, 16, 8] \\
$\mathrm{Level}_{6}\mathrm{Arch}_{26}$ & 76,889 & 3 & 15 & 3 & [8, 16, 32] & [32, 16, 8] \\
$\mathrm{Level}_{6}\mathrm{Arch}_{27}$ & 76,889 & 3 & 15 & 4 & [8, 16, 32] & [32, 16, 8] \\
$\mathrm{Level}_{6}\mathrm{Arch}_{28}$ & 76,889 & 3 & 15 & 5 & [8, 16, 32] & [32, 16, 8] \\
$\mathrm{Level}_{6}\mathrm{Arch}_{29}$ & 78,205 & 4 & 15 & 2 & [4, 8, 16, 32] & [32, 16, 8, 4] \\
$\mathrm{Level}_{6}\mathrm{Arch}_{30}$ & 78,205 & 4 & 15 & 3 & [4, 8, 16, 32] & [32, 16, 8, 4] \\
$\mathrm{Level}_{6}\mathrm{Arch}_{31}$ & 78,205 & 4 & 15 & 4 & [4, 8, 16, 32] & [32, 16, 8, 4] \\
$\mathrm{Level}_{6}\mathrm{Arch}_{32}$ & 78,205 & 4 & 15 & 5 & [4, 8, 16, 32] & [32, 16, 8, 4] \\
$\mathrm{Level}_{6}\mathrm{Arch}_{33}$ & 78,487 & 5 & 15 & 2 & [2, 4, 8, 16, 32] & [32, 16, 8, 4, 2] \\
$\mathrm{Level}_{6}\mathrm{Arch}_{34}$ & 78,487 & 5 & 15 & 3 & [2, 4, 8, 16, 32] & [32, 16, 8, 4, 2] \\
$\mathrm{Level}_{6}\mathrm{Arch}_{35}$ & 78,487 & 5 & 15 & 4 & [2, 4, 8, 16, 32] & [32, 16, 8, 4, 2] \\
$\mathrm{Level}_{6}\mathrm{Arch}_{36}$ & 78,487 & 5 & 15 & 5 & [2, 4, 8, 16, 32] & [32, 16, 8, 4, 2] \\
$\mathrm{Level}_{6}\mathrm{Arch}_{37}$ & 97,505 & 2 & 5  & 2 & [32, 64] & [64, 32] \\
$\mathrm{Level}_{6}\mathrm{Arch}_{38}$ & 97,505 & 2 & 5  & 3 & [32, 64] & [64, 32] \\
$\mathrm{Level}_{6}\mathrm{Arch}_{39}$ & 97,505 & 2 & 5  & 4 & [32, 64] & [64, 32] \\
$\mathrm{Level}_{6}\mathrm{Arch}_{40}$ & 97,505 & 2 & 5  & 5 & [32, 64] & [64, 32] \\

\end{longtable}
\endgroup

}
%%%%%%%%%%%%%%%%%%%%%%%%%

%========================================================================
% Bibliography
%========================================================================
\clearpage
 \bibliographystyle{elsarticle-num} 
 \bibliography{references}

\end{document}